\documentclass[letterpaper, 10 pt, compsoc]{ieeeconf}  % Comment this line out if you need a4paper

\usepackage{times}
\usepackage{epsfig}
\usepackage{graphicx}
\usepackage{amsmath}
\usepackage{amssymb}
\usepackage{cite}
\usepackage{amsmath,amssymb,amsfonts}
\usepackage{textcomp}
\usepackage{subfigure}
\usepackage[linesnumbered,boxed,ruled,commentsnumbered]{algorithm2e}
\usepackage{algpseudocode}  
\usepackage{amsmath}  
\usepackage{multirow}
\usepackage{color}
\usepackage{algpseudocode}  
\UseRawInputEncoding

\newtheorem{proposition1}{Proposition}

\newtheorem{lemma1}{Lemma}

\newtheorem{proof1}{Proof}%[section]
\usepackage{caption}
\usepackage{url}

\IEEEoverridecommandlockouts                              % This command is only needed if 
\title{\LARGE \bf
Practical, Fast and Robust Point Cloud Registration for 3D Scene Stitching and Object Localization}

\author{Lei Sun$^{1,*}$% <-this % stops a space
\thanks{*Corresponding author. This work was not supported by any organization.}% <-this % stops a space
\thanks{$^{1}$Lei Sun is with School of Mechanical and Power Engineering, East China University of Science and Technology, Shanghai 200237, China;  
{\tt\small leisunjames@126.com}}%
}

\begin{document}

\maketitle
\thispagestyle{empty}
\pagestyle{empty}

%%%%%%%%%%%%%%%%%%%%%%%%%%%%%%%%%%%%%%%%%%%%%%%%%%%%%%%%%%%%%%%%%%%%%%%%%%%%%%%%
\begin{abstract}
3D point cloud registration ranks among the most fundamental problems in remote sensing, photogrammetry, robotics and geometric computer vision. Due to the limited accuracy of 3D feature matching techniques, \textit{outliers} may exist, sometimes even in very large numbers, among the correspondences. Since existing robust solvers may encounter high computational cost or restricted robustness, we propose a novel, fast and highly robust solution, named VOCRA (VOting with Cost function and Rotating Averaging), for the point cloud registration problem with extreme outlier rates. Our first contribution is to employ the Tukey's Biweight robust cost to introduce a new voting and correspondence sorting technique, which proves to be rather effective in distinguishing true inliers from outliers even with extreme (99\%) outlier rates. Our second contribution consists in designing a time-efficient consensus maximization paradigm based on robust rotation averaging, serving to seek inlier candidates among the correspondences. Finally, we apply Graduated Non-Convexity with Tukey's Biweight (GNC-TB) to estimate the correct transformation with the inlier candidates obtained, which is then used to find the complete inlier set. Both standard benchmarking and realistic experiments with application to two real-data problems are conducted, and we show that our solver VOCRA is robust against over 99\% outliers and more time-efficient than the state-of-the-art competitors.
\end{abstract}


%%%%%%%%%%%%%%%%%%%%%%%%%%%%%%%%%%%%%%%%%%%%%%%%%%%%%%%%%%%%%%%%%%%%%%%%%%%%%%%%
\section{Introduction}

With the development of the 3D measurement and scanning technologies (e.g. LiDAR scanners, 3D sensors), point cloud registration, which seeks to estimate the best rigid transformation (including rotation and translation) between multiple 3D point clouds or scans, becomes an increasingly important building block in remote sensing, photogrammetry, robotics perception and computer vision, and has found extensive applications in 3D reconstruction~\cite{henry2012rgb,choi2015robust,zhang2015visual}, object recognition and localization~\cite{drost2010model,zeng2017multi}, SLAM~\cite{zhang2014loam}, medical imaging~\cite{audette2000algorithmic}, etc.

To address the point cloud registration problem, Iterative Closest Point (ICP)~\cite{besl1992method} has been a well-known solver, but its downside lies in its high dependence on the initial guess of the rigid transformation. If the initial information given is poor, it is likely to converge to local minima and fail. Hence, correspondence-based registration methods that can be free of initial guess is growing increasingly popular. It consists in first matching keypoints between point clouds to construct putative correspondences and then estimating the best transformation using robust estimators.

It is known to all that the 3D keypoint matching technique (e.g. FPFH~\cite{rusu2009fast}, ISS~\cite{zhong2009intrinsic}), unlike its 2D counterparts (e.g. SIFT~\cite{lowe2004distinctive}, SURF~\cite{bay2006surf}), is very challenging in practice since we may encounter low texture, noisy environment, partiality, density variance, and cluttered or repetitive patterns in the 3D space, which may yield huge numbers of \textit{outliers} lurking in the correspondence set established. According to~\cite{bustos2017guaranteed}, having extreme outlier rates (e.g. over 95\%) is not uncommon in real-world scenes. Thus, various robust estimators~\cite{fischler1981random,chum2003locally,parra2014fast,yang2020graduated,zhou2016fast,bustos2017guaranteed,yang2019polynomial,yang2020teaser} are employed to reject outliers, but unfortunately, many of them suffer from issues like high computational cost or limited robustness.

RANSAC~\cite{fischler1981random} and Branch-and-Bound (BnB)~\cite{parra2014fast,horst2013global} are two well-known paradigms which, in essence, achieve robust estimation by maximizing the consensus set. However, both of the solvers have exponentially increasing runtime, where the former scales poorly with the outlier rate while the latter has the worst-case exponential time w.r.t. input size. Thus, the limitation in efficiency seems to be a fatal defect of them for practical use. More recently, Graduated Non-Convexity (GNC) is an effective global method for robust outlier rejection, typical examples including FGR~\cite{zhou2016fast} and GNC-TLS/GM~\cite{yang2020graduated}. Unfortunately, GNC solvers usually fail at no more than 90\% outliers, so it is too limited in robustness for high-outlier practical problems. In addition, GORE~\cite{bustos2017guaranteed}, is a outlier removal approach that guarantees to only remove true outliers, but it still requires long runtime in the low-outlier regime and is also likely to have exponential time due to the probable use of BnB within it. Though TEASER~\cite{yang2019polynomial,yang2020teaser} is a current highly robust solver, it still requires external maximal clique algorithms that may be slow to run without parallelism programming, especially when the putative correspondences are in abundance.

In this work, our goal is to design a new solver which can tolerate extreme outliers (e.g. over 99\%), has promising time-efficiency even with high outlier rates, and does not need any additional information (e.g. initial guess). To this end, we get inspirations from two aspects: (a) the line voting technique in~\cite{li2021practical} to sort the correspondences and (b) the fast robust single rotation averaging solver~\cite{lee2020robust}, and propose a novel robust registration solver named VOCRA (VOting with Cost function and Rotation Averaging).

The contributions of this paper include: 

(a) we introduce the Tukey's Biweight cost function in combination with the concept of GNC (GNC-TB) and propose a voting technique based on it, which proves to be more effective in extreme-outlier regime than simple 0-1 voting; 

(b) we present a novel consensus maximization framework using robust rotation averaging to rapidly seek the inlier consensus set; 

(c) we apply GNC-TB for further robust optimization of the consensus set and use its solution to find the complete inlier set. 

These three contributions lead to our robust solver VOCRA. Finally, we comparatively evaluate VOCRA in benchmarking and real-data experiments with applications to real-world problems against existing state-of-the-art robust registration solvers.

The rest of this paper is organized as: Section~\ref{sec-2} provides some concise reviews on related methods for robust registration, Section~\ref{pre-n} provides some preliminaries and mathematical notation for the registration problem and our method, Section~\ref{GNC-TB} introduces GNC-TB for both robust estimation and weighted voting, Section~\ref{sec-5} elaborates on the proposed method VOCRA, and Section~\ref{experiments} evaluates the performance of VOCRA through multiple experiments, followed by the conclusions in Section~\ref{sec-conclusion}.

\section{Related Work}\label{sec-2}

We review the related prior works w.r.t. the two key components of our solver: consensus maximization and M-estimation, as summarized in Table~\ref{review}.

\begin{figure}
\includegraphics[width=1\linewidth]{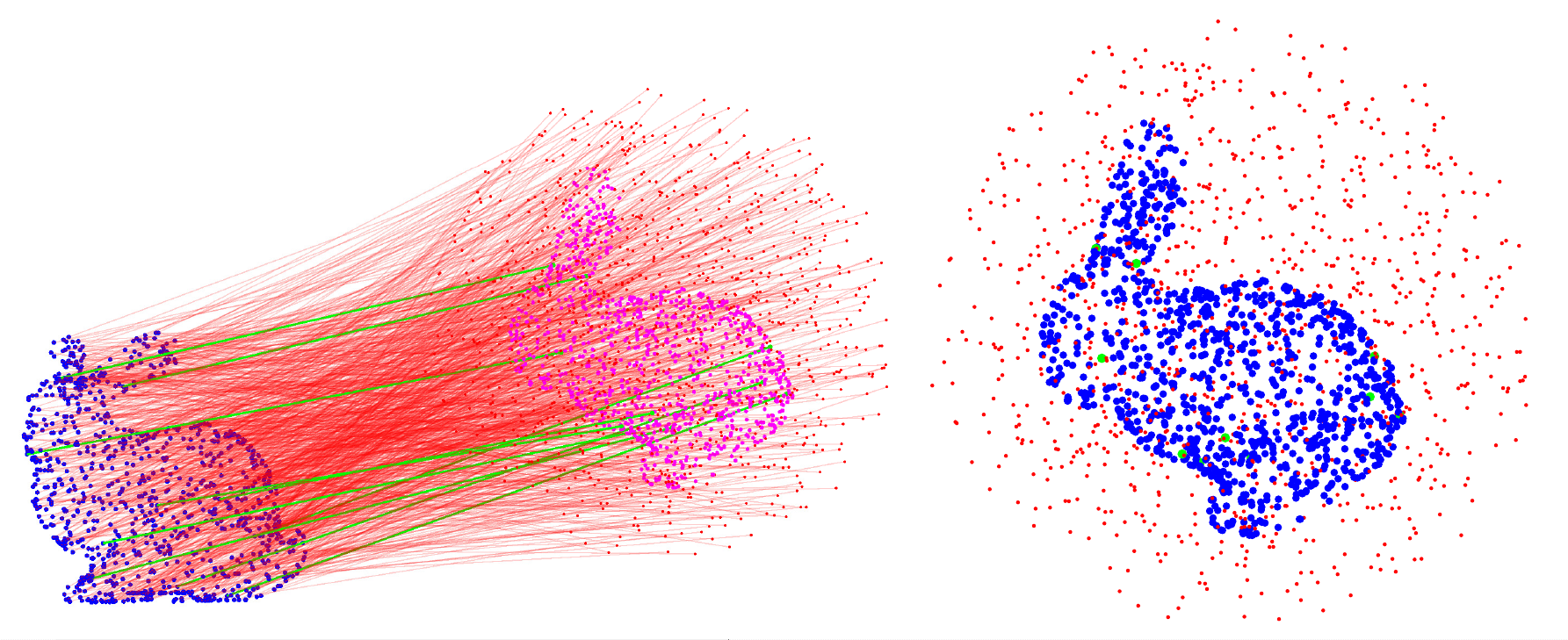}
\caption{Left: Example of a point cloud registration problem with 99\% outliers. Right: VOCRA can solve the registration problem within 3 seconds.}
\label{11}
\end{figure}

\subsection{Consensus Maximization}

RANSAC~\cite{fischler1981random} is probably the most popular robust estimator, which maximizes the consensus set using the iterative hypothesize-and-test framework: first making a hypothesis with minimal random samples and then testing the quality of its consensus. The variants of RANSAC have been proposed to enhance the performance (e.g. local optimization~\cite{chum2003locally,lebeda2012fixing}, correspondence sorting~\cite{chum2005matching}) against outliers. Nonetheless, the computational cost of the RANSAC-family solvers grows exponentially with the outlier rate, unsuitable for high-outlier problems. BnB~\cite{parra2014fast,horst2013global} is another consensus maximization approach, which searches in the parameter space to globally solve the optimization. Though with optimality guarantee, BnB's runtime is exponential to the input size (correspondence number), too slow for large problems. Moreover, GORE~\cite{bustos2017guaranteed} realizes consesnsus maximization via guaranteed outlier removal, but it may also encounter long runtime due to its possible use of BnB as a subroutine. Our solver shares small similarity with RANSAC since minimal models are estimated and best consensus set is iteratively updated in both solvers. But unlike RANSAC, VOCRA maximizes the consensus by fast rotation averaging over minimal rotations through enumeration, which is deterministic and tremendously faster for use since: (a) the scale-invariant constraint is introduced for subset filtering and saving much computational cost during consensus building, and (b) the enumeration process is based on the inlier reliability of correspondences rather than being random.

\begin{table*}[h]

\centering

\caption{Brief review on some typical registration methods (including our VOCRA).}
\label{review}

\setlength\tabcolsep{3pt}
%\addtolength{\tabcolsep}{-0pt}

\begin{tabular}{|c|c|c|c|}
\hline\rule{0pt}{6pt}
\textbf{\scriptsize{Solver Name}} & \textbf{\scriptsize{Category}}  & \textbf{\scriptsize{Brief Description}}  & \textbf{\scriptsize{Highlight \& Limitation}}  \\ \hline

\scriptsize{RANSAC~\cite{fischler1981random}} & \scriptsize{Consensus Maximization} & \scriptsize{Random sampling and model fitting} & \scriptsize{Fast with low outliers, but slow with high outliers} \\ \hline

\scriptsize{BnB~\cite{parra2014fast,horst2013global}} & \scriptsize{Consensus Maximization} & \scriptsize{Parameter-space searching} & \scriptsize{Globally optimal, but slow with large input size} \\ \hline

\scriptsize{GORE~\cite{bustos2017guaranteed}} & \scriptsize{Consensus Maximization} & \scriptsize{Guaranteed outlier removal} & \scriptsize{Globally optimal, but slow with large input size} \\ \hline

\scriptsize{GNC~\cite{yang2020graduated}, FGR~\cite{zhou2016fast}} & \scriptsize{M-Estimation} & \scriptsize{Non-minimal, alternating and iterative optimization} & \scriptsize{Fast, but limited in robustness} \\ \hline

\scriptsize{VOCRA} & \scriptsize{Consensus Maximization \& M-Estimation} & \scriptsize{Voting, consensus maximizing, and then M-estimation} & \scriptsize{Fast and highly robust} \\ \hline

\end{tabular}

\vspace{-1mm}

\end{table*}

\subsection{M-Estimation}

M-estimation consists in diminishing the effect of outlying data by optimizing over the \textit{cost functions} (or called the \textit{loss functions}). Iterative local solvers~\cite{agarwal2013robust,kummerle2011g,sunderhauf2012towards} have been applied to optimize M-estimation problems in earlier time, but they require initial guess and may get trapped into local minima. More currently, Graduated Non-Convexity (GNC)~\cite{yang2020graduated} is revisited and developed to work in conjunction with Black-Ragaranjan Duality~\cite{black1996unification} to reject outliers by adopting standard non-minimal solvers in multiple geometric vision problems. GNC does not require any initial guess but its downside is that it generally can only succeed with no more than 90\% outlier rates, hence brittle in harsh outlier situations. Our solver adopt the theory of GNC for two purposes: line voting for pre-proccessing and consensus refining (robustifying with only low outlier rates) for post-proccessing.

\section{Preliminaries and Notation}\label{pre-n}

\subsection{Problem Formulation}

Given two corresponding point sets: $\mathcal{P}=\{\boldsymbol{p}_i\}_{i=1}^{N}$ and $\mathcal{Q}=\{\boldsymbol{q}_i\}_{i=1}^{N}$ where $\boldsymbol{p}_i\leftrightarrow\boldsymbol{q}_i$ consitutes a point correspondence, the point cloud registration problem seeks to find the best rotation $\boldsymbol{R}\in SO(3)$ and translation $\boldsymbol{t}\in \mathbb{R}^{3}$ that align $\mathcal{P}$ and $\mathcal{Q}$. This problem can be formulated as a consensus maximization problem such that
\begin{equation}\label{CM}
\begin{gathered}
\underset{\mathcal{I}\subset \mathcal{N}}{\max}\, |\mathcal{I}| \\
s.t. \,\|\boldsymbol{R}^{\star}\boldsymbol{p}_i+\boldsymbol{t}^{\star}-\boldsymbol{q}_i\| \leq \xi,\, \forall i\in\mathcal{I}
\end{gathered}
\end{equation}
where $\xi$ denotes the inlier threshold or noise bound satisfying $\xi\geq\|\boldsymbol{\varepsilon}\|$, which is related to the standard deviation $\sigma$ of noise and used to differentiate inliers from outliers, and $\mathcal{I}$ is the consensus set of the optimal transformation $(\boldsymbol{R}^{\star},\boldsymbol{t}^{\star})$.

\subsection{Graduated Non-Convexity and Black-Ragaranjan Duality}

In~\cite{yang2020graduated}, Graudated Non-Convexity (GNC) is employed to develop a practically feasible and time-efficient solution for the robust estimation with outlier process (that is, Black-Ragaranjan Duality~\cite{black1996unification}). We summarize this technique as follows:

\begin{lemma1}[Black-Ragaranjan Duality with GNC]\label{Lem1}

As for the following correspondence-based robust estimation problem with outlier process~\cite{black1996unification}: 

\begin{equation}\label{robust-outlier}
\underset{\omega_i\in[0,1],\,\boldsymbol{x}\in\mathcal{K}}{\min} \sum_{i=1}^N \left[\omega_i \, {r}_i(\boldsymbol{p}_i,\boldsymbol{q}_i,\boldsymbol{x})+{\Psi}(\omega_i)\right],
\end{equation}

\noindent where $\boldsymbol{x}\in\mathcal{K}$ denotes the variables within domian $\mathcal{K}$, $\boldsymbol{p}_i\leftrightarrow\boldsymbol{q}_i$ ($i=1,2,\dots,N$) is the correspondence,  $\omega_i$ is the weight for the residual error ${r}_i(\boldsymbol{p}_i,\boldsymbol{q}_i,\boldsymbol{x})$ w.r.t. this correspondence that can also be abbreviated as ${r}_i$, and ${\Psi}(\omega_i)$ is the outlier process corresponding to a certian robust cost function $\rho({r}_i)$ which serves as a penalty function over weight $\omega_i$, we now introduce a surrogate function based on GNC, writable as $\rho_{\mu,\xi}({r}_i)$ that is jointly defined by a controlling parameter $\mu$ and the inlier threshold $\xi$, for function $\rho({r}_i)$. In this case, as $\mu$ gradually changes (usually monotonically increasing or decreasing), surrogate function $\rho_{\mu,\xi}({r}_i)$ starts from a approximately convex status and gradually approximates the original  $\rho({r}_i)$ with becoming increasingly non-convex. The optimization process can be operated alternately through iterations, and in each iteration, we first fix $\omega_i$ and optimize $\boldsymbol{x}$ with a non-minimal solver, and then fix $\boldsymbol{x}$ and solve (optimize) $\omega_i$ (usually in closed form). With $\mu$ continuously changing to push $\rho_{\mu,\xi}({r}_i)$ to recover $\rho({r}_i)$, the optimal solution $\boldsymbol{x}^{\star}$ and its corresponding weights $\omega_i$ can be approximated iteratively.
\end{lemma1}

\subsection{Robust Single Rotation Averaging}

\textit{Robust single rotation averaging} is a well-known problem in geometric vision, broadly applied to Structure-from-Motion~\cite{hartley2011l1}, attitude estimation~\cite{lam2007precision}, etc, which seeks to estimate the best (average) rotation from a group of rotations including possible outliers.

Lee et al~\cite{lee2020robust} provided an efficient chordal-distance-based robust approach for single rotation averaging, and its problem can be formulated as:

\begin{equation}
\boldsymbol{R}^{\star}=\underset{\boldsymbol{R}\in SO(3)}{\arg\min}\,\sum_{i=1}^{M} dis_{\text{chord}}(\boldsymbol{R}_i,\boldsymbol{R}),
\end{equation}

\noindent where $M$ is the total number of rotations (including outliers) given, and $dis_{\text{chord}}$ denotes the chordal distance between two rotations that can be given by:

\begin{equation}\label{chordal}
dis_{\text{chord}}(\boldsymbol{R}_1,\boldsymbol{R}_2)=\left\|\boldsymbol{R}_1-\boldsymbol{R}_2\right\|_{F}=2\sqrt{2}\sin (\frac{\angle_{\text{geo}}(\boldsymbol{R}_1,\boldsymbol{R}_2)}{2}),
\end{equation}

\noindent where $\|\,\,\|_F$ denotes the Frobenius norm and $\angle_\text{geo}(\,\,,\,\,)$ means the geodesic distance~\cite{hartley2013rotation} between rotations that is written as:

\begin{equation}
\angle_{\text{geo}}(\boldsymbol{R}_1,\boldsymbol{R}_2)=\left|\arccos\left(\frac{trace(\boldsymbol{R}_1^{\top}\boldsymbol{R}_2)-1}{2}\right)\right|.
\end{equation}

The explicit solver can be found in Algorithm 2 of~\cite{lee2020robust}, which runs in milliseconds with hundreds of rotations, rapid for practical use. In the following paper, we uniformly use function \textit{robustLeeChordal}($\mathcal{R}$) to represent this solver, where $\mathcal{R}=\{\boldsymbol{R}_i\}_{i=1}^M$ denotes the given group of rotations as the input.

\section{GNC-TB: From Robust Estimator to Weighted Voting Operator}
\label{GNC-TB}

We now introduce a GNC-based Black-Ragaranjan Duality formulation for robust estimation by adopting the Tukey's Biweight (TB) robust cost function as the outlier process, named GNC-TB, which is derived based on~\cite{yang2020graduated,black1996unification}. First, according to Figure 25 in~\cite{black1996unification}, the TB cost can be written as 

\begin{equation}\label{F25-p}
{\rho}(x)=\left\{\begin{array}{ll}
\frac{{x}^2}{c^2}-\frac{{x}^4}{c^4}+\frac{{x}^6}{3c^6} & \text { if } |{x}| \leq c, \\
\frac{1}{3} & \text { if } |{x}| > c,
\end{array}\right.
\end{equation}

\noindent and its robust objective with outlier process is writable as:

\begin{equation}\label{F25-E}
{\Psi}(z)=\frac{1}{3}-z+\frac{2}{3}z^{\frac{3}{2}},
\end{equation}

\noindent where the explicit definition of $x$ and $z$ can be found in~\cite{black1996unification}, and we replace `$\sigma$' in~\cite{black1996unification} with $c$ here to avoid confusion with the standard deviation of noise $\sigma$ in this paper.
Based on ${\rho}(x)$ and ${\Psi}(z)$, we can define GNC-TB as follows:

\begin{proposition1}[GNC-TB]\label{Prop1}
Following the robust problem~\eqref{robust-outlier}, we introduce a surrogate function for the TB cost by adopting $\mu$ as the controlling parameter and $\xi$ as the inlier threshold such that

\begin{equation}\label{rho-function}
{\rho}_{\mu,\xi}(r_i)=\left\{\begin{array}{ll}
\frac{{r_i}^2}{\mu\xi^2}-\frac{{r_i}^4}{\mu\xi^4}+\frac{{r_i}^6}{3\mu\xi^6} & \text { if } {r_{i}}^2 \leq \mu\xi^2, \\
\frac{1}{3} & \text { if } {r_{i}}^2 > \mu\xi^2.
\end{array}\right.
\end{equation}

\noindent We initiate $\mu$ with a sufficiently large value to make ${\rho}_{\mu,\xi}(r_i)$ approximately convex, indicating that the tolerance to residual errors is very lenient at first, and then we continuously diminish $\mu$ to gradually decrease the convexity and also make the filtration of residuals gradually stricter. When $\mu$ approximates 1, the original non-convex TB cost is recovered. Minimizing function~\eqref{rho-function} is equivalent to the following outlier process (with $\mu$ and $\xi$):

\begin{equation}\label{outlier-process}
{\Psi}_{\mu,\xi}(\omega_i)=\mu\xi^2\left(\frac{1}{3}-\omega_i+\frac{2}{3}\omega_i^{\frac{3}{2}}\right).
\end{equation}

\noindent Besides, we can update the weights $\omega_i$ in each iteration in closed-form such that

\begin{equation}\label{weight-updating}
\omega_i\leftarrow\left\{\begin{array}{ll}
\left(1-\frac{{r_i}^2}{\mu\xi^2}\right)^2 & \text { if } {r_{i}}^2 \leq \mu\xi^2, \\
0 & \text { if } {r_{i}}^2 > \mu\xi^2.
\end{array}\right.
\end{equation}

\end{proposition1}

\begin{proof1}

Following the traditional TB function as in~\eqref{F25-p}, in order to derive its GNC-based formulation, we can replace the fixed $c$ the varying $\mu\xi^2$ to build our surrogate function~\eqref{rho-function} controlled by $\mu$. Hence, the outlier process w.r.t. the surrogate function can be written as

\begin{equation}\label{robust-obj}
E(r_i,\mu,\xi,\omega_i)=\omega_i\frac{{r_i}^2}{\mu\xi^2}+{\Psi}_{\mu,\xi}(\omega_i),
\end{equation}

\noindent where ${\Psi}_{\mu,\xi}$ is given in~\eqref{outlier-process}. Then, we can derive its gradient w.r.t. $\omega_i$ such that

\begin{equation}\label{partial}
\frac{\partial E(r_i,\mu,\xi,\omega_i)}{\partial \omega_i}=\frac{{r_i}^2}{\mu\xi^2}-1+{\omega_i}^{\frac{1}{2}},
\end{equation}

\noindent and by letting the gradient $\frac{\partial E(r_i,\mu,\xi,\omega_i)}{\partial \omega_i}$ be zero, we can have:

\begin{equation}\label{w}
\omega_i=\left(1-\frac{{r_i}^2}{\mu\xi^2}\right)^2,
\end{equation}

\noindent which serves as the updating principle of the weights $\omega_i$.

\end{proof1}

We derive GNC-TB for two purposes: (a) the TB function underlies our voting process for correspondence sorting and is the cost function that yields the best results for weighted voting in Section~\ref{sorting-and-voting}, and (b) GNC-TB is taken as the robust outlier rejection approach in Section~\ref{prune-outliers}.

\begin{algorithm}[t]
\caption{\textit{votingTB}}
\label{Voting-TB}
\SetKwInOut{Input}{\textbf{Input}}
\SetKwInOut{Output}{\textbf{Output}}
\Input{putative correspondences $\{(\boldsymbol{p}_i,\boldsymbol{q}_i)\}_{i=1}^N$, threshold $\xi\leftarrow3\sigma$ and $\mu\leftarrow1.5$\;}
\Output{sorted correspondence indices $\boldsymbol{d}$\;}
%\BlankLine
Set empty votes $\boldsymbol{v}=\{v_i\}_{i=1}^N\leftarrow [0,0,\dots,0]$ for all $N$ correspondences and $eIn\leftarrow 0$\;
\For{$i=1:(N-1)$}{
\For{$j=(i+1):N$}{
Compute $S_{ij}$ for correspondence pair $(\boldsymbol{p}_i,\boldsymbol{q}_i)$ and $(\boldsymbol{p}_j,\boldsymbol{q}_j)$ using~\eqref{scale-invariant}\;
Vote for $v_{i}$ and $v_{j}$ with TB cost based on~\eqref{TB-voting}\;}
\If{$i\leq20$}{
\If{$v_i\geq 0.2N$}{
$eIn\leftarrow 1$\;
\textbf{break}
}
}
}
Sort votes $\boldsymbol{v}$ in descending order and also obtain its corresponding index vector $\boldsymbol{d}\in\mathbb{R}^N$\;
\Return sorted correspondence indices $\boldsymbol{d}$\;
\end{algorithm}

\section{The Proposed Methodology}\label{sec-5}

The proposed method VOCRA mainly consists of three key steps in the following three subsections: (a) line voting with the TB robust cost and sorting correspondences according to the probability to be inliers (Section~\ref{sorting-and-voting}) , (b) maximizing the consensus set based on robust rotation averaging (Section~\ref{consensus-maximize}), and (c) finally removing outliers and seeking true inliers with GNC-TB (Section~\ref{prune-outliers}).

\begin{figure*}[h]
\centering

\setlength\tabcolsep{0pt}
\addtolength{\tabcolsep}{0pt}
\begin{tabular}{ccc}

\footnotesize{(a) $N=1000$, 95\% outliers} & \footnotesize{(b) $N=1000$, 97\% outliers} &
\footnotesize{(c) $N=1000$, 98\% outliers} \\

\begin{minipage}[t]{0.33\linewidth}
\centering
\includegraphics[width=1\linewidth]{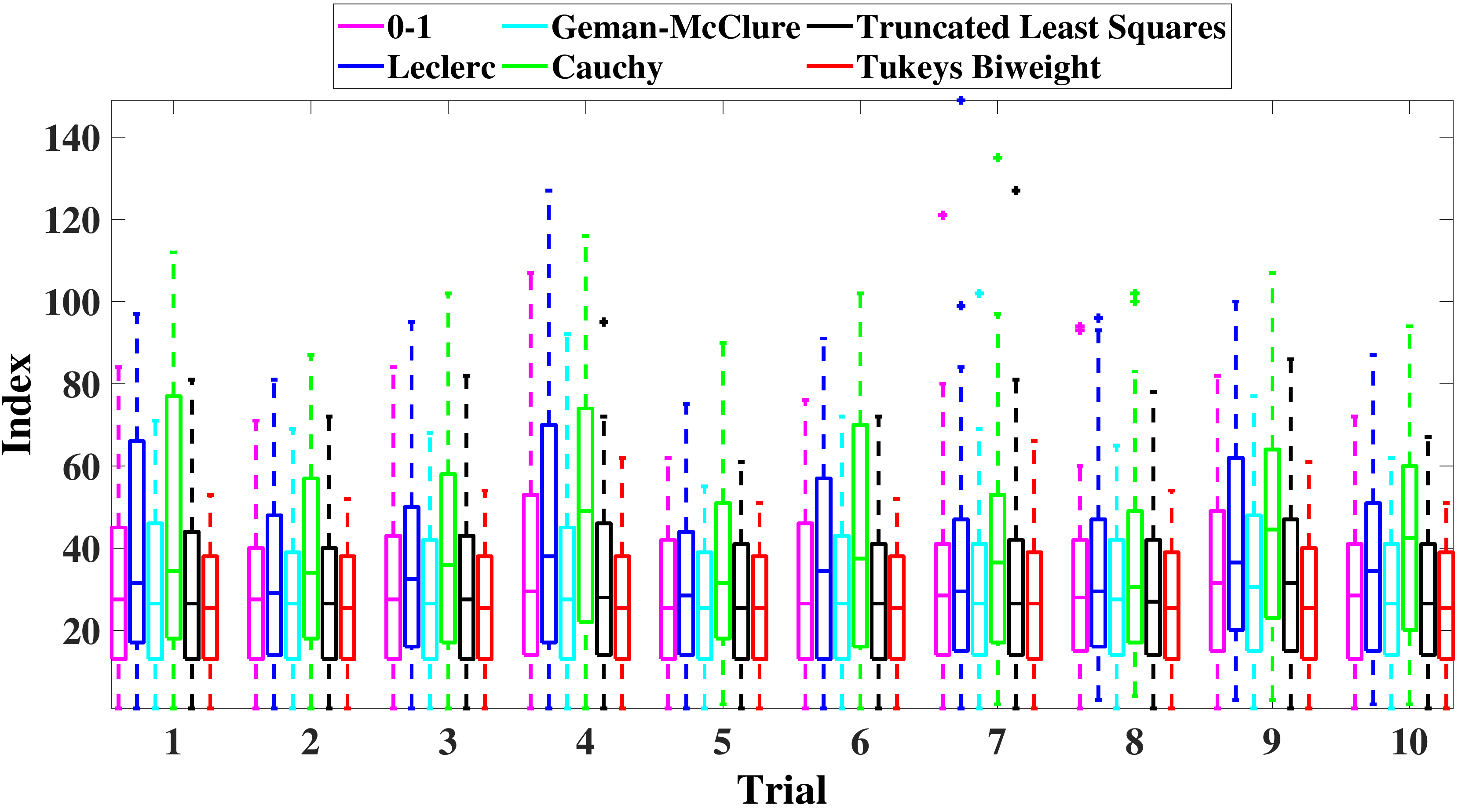}
\end{minipage}&

\begin{minipage}[t]{0.33\linewidth}
\centering
\includegraphics[width=1\linewidth]{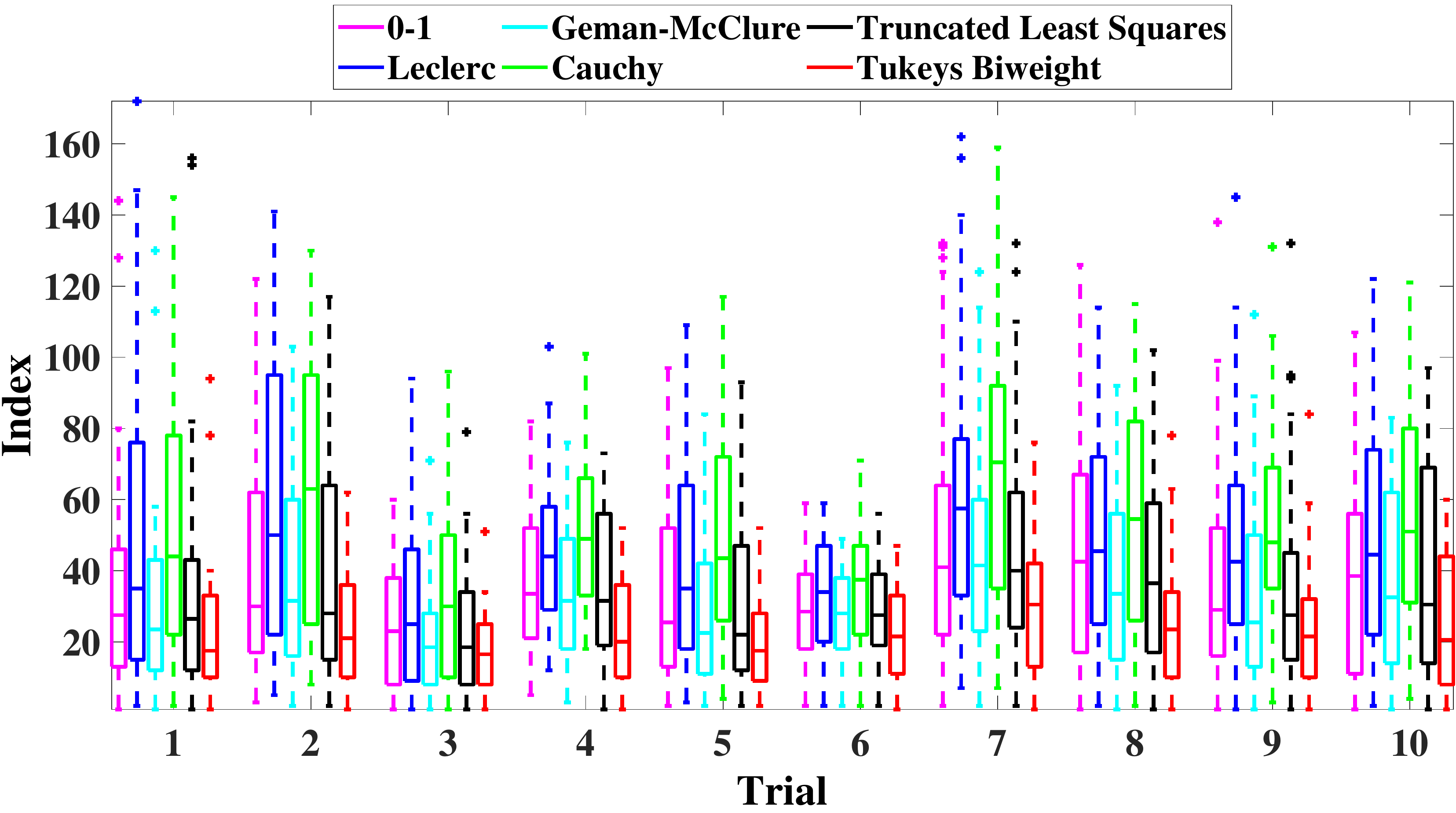}
\end{minipage}&

\begin{minipage}[t]{0.33\linewidth}
\centering
\includegraphics[width=1\linewidth]{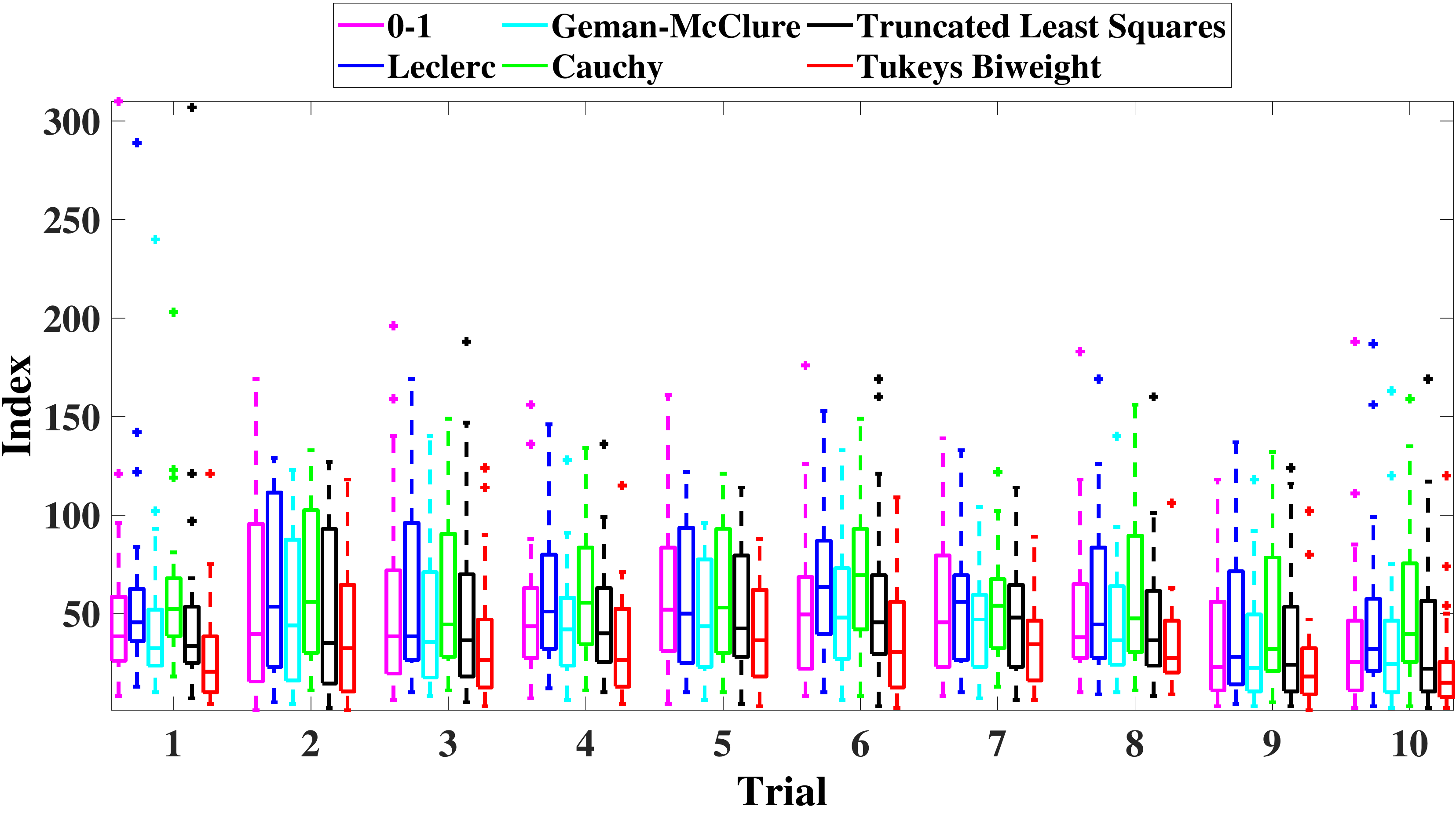}
\end{minipage}\\

\footnotesize{(d) $N=1000$, 99\% outliers}  & \footnotesize{(e) $N=100$, 95\% outliers} &  \footnotesize{(f) $N=500$, 98\% outliers} \\

\begin{minipage}[t]{0.33\linewidth}
\centering
\includegraphics[width=1\linewidth]{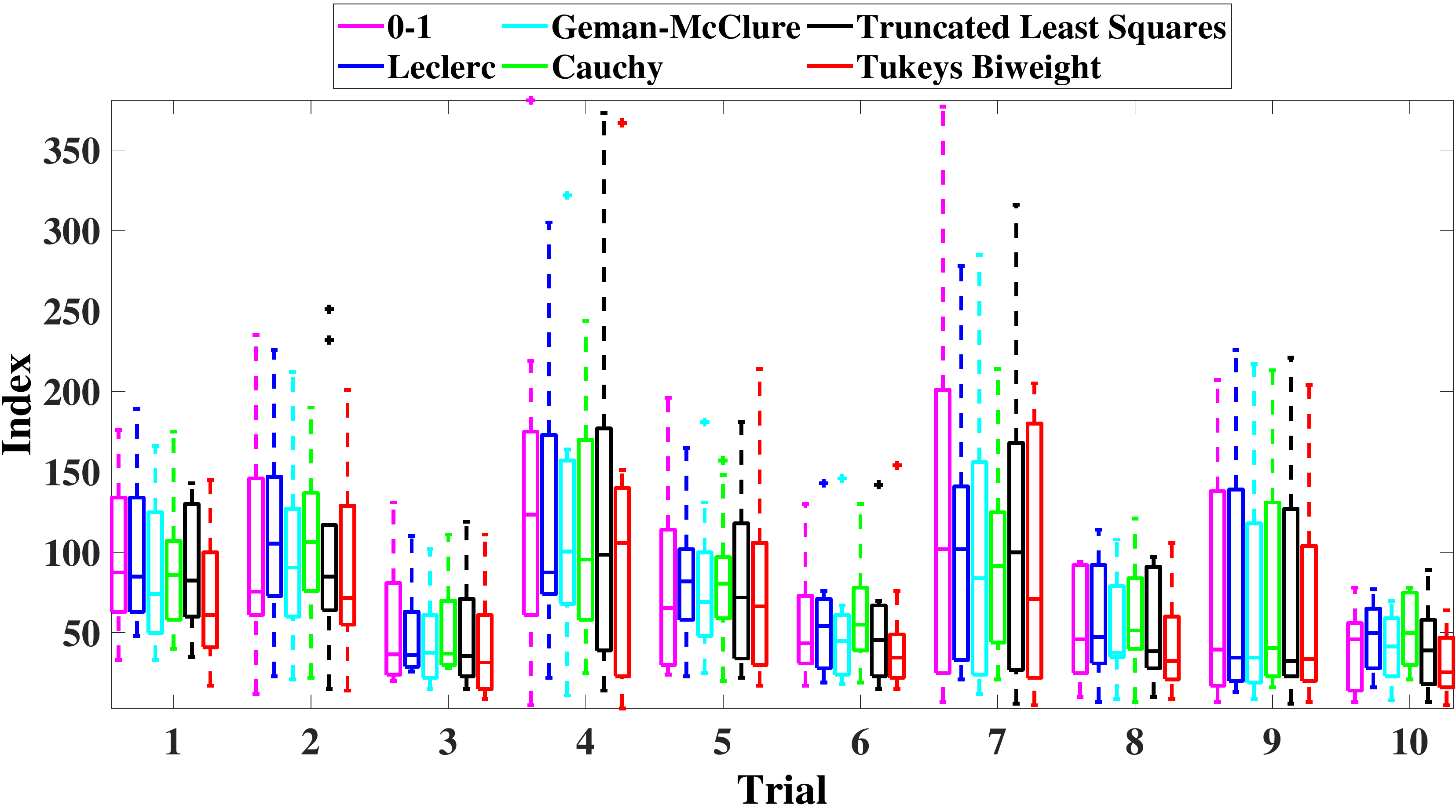}
\end{minipage}&

\begin{minipage}[t]{0.33\linewidth}
\centering
\includegraphics[width=1\linewidth]{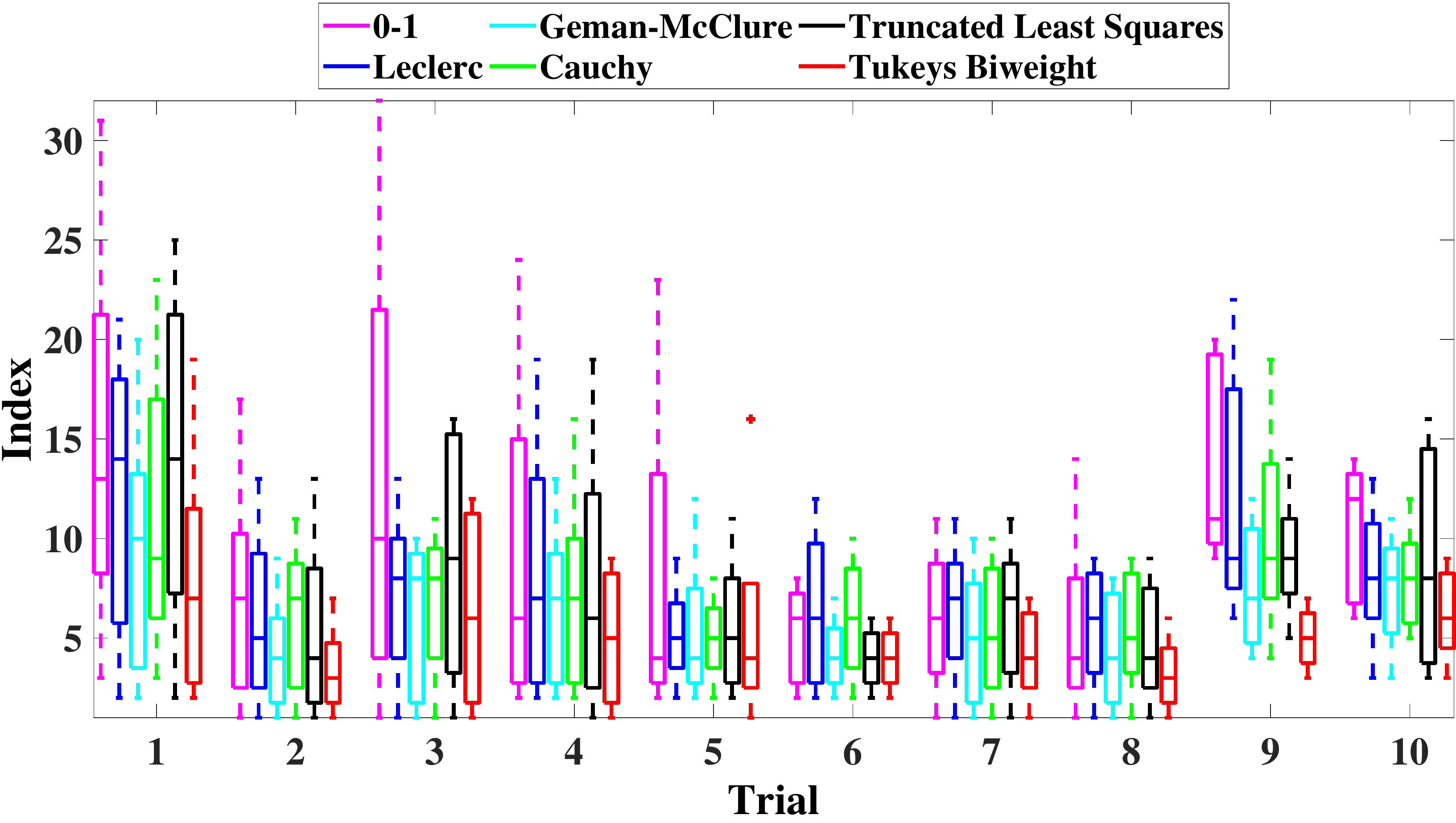}
\end{minipage}&

\begin{minipage}[t]{0.33\linewidth}
\centering
\includegraphics[width=1\linewidth]{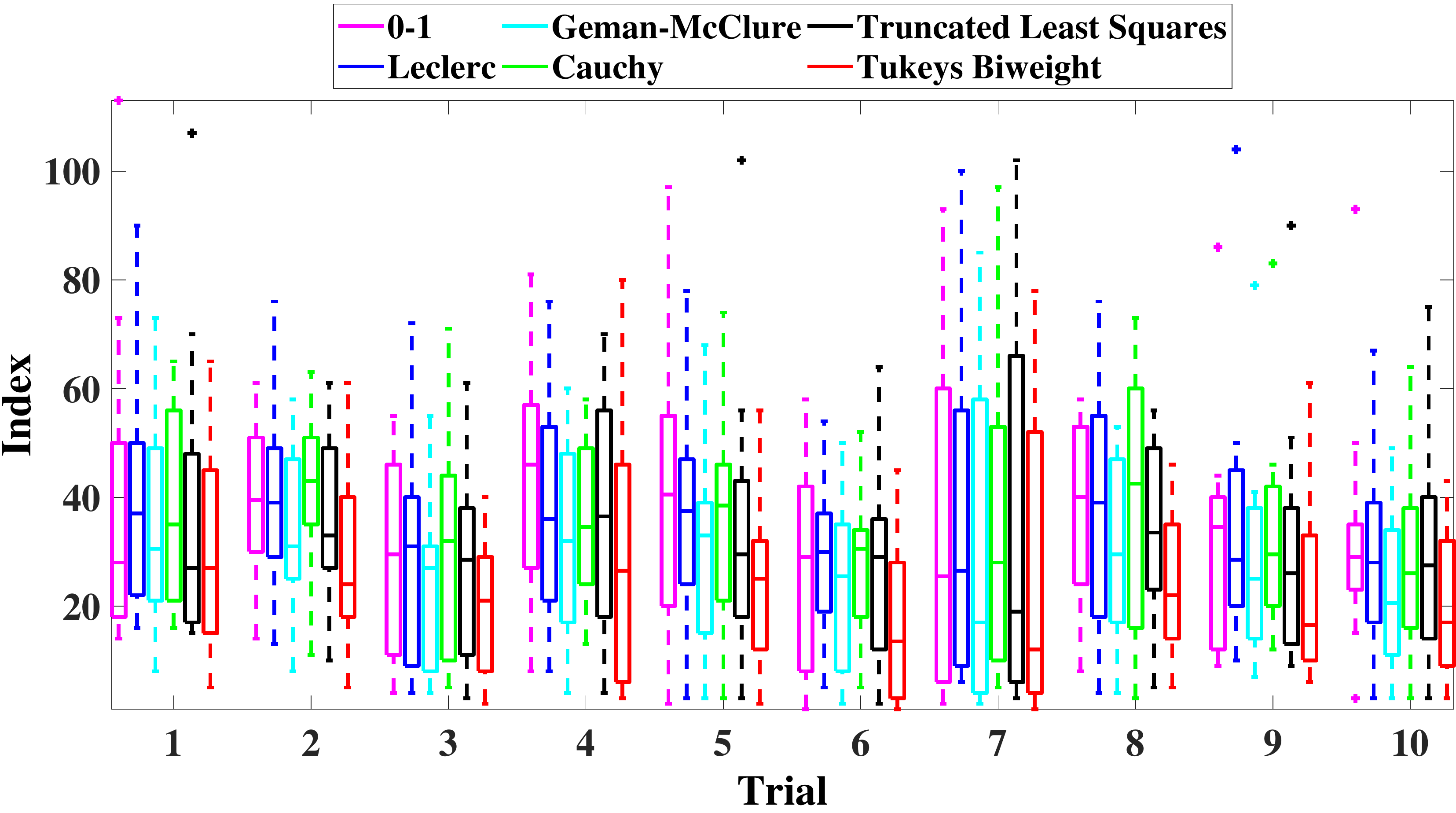}
\end{minipage}

\end{tabular}

\vspace{-3mm}
\caption{Voting results with different robust cost functions over 10 random trials. (a-d) Voting with $N=1000$ correspondences w.r.t. different outlier rates. (e-f) Challenging voting situations with $N=100$ and $N=500$, respectively. The superiority of TB function can be observed from the boxplots.}
\label{voting}
\vspace{-3mm}
\end{figure*}

\subsection{To Sort the Correspondences: Voting with Robust Cost Function}\label{sorting-and-voting}

After the 3D descriptors or feature matchers have established the putative correspondences between point clouds, we can start by raising a question: \textit{is it possible to sort these correspondences according to their respective quality (e.g. possibility to be true inliers) without any extra information required?} 

The answer must be \textit{yes}. Line voting~\cite{li2021practical} is a simple voting technique based on the scale-based invariant established over pairwise point correspondences that has been used to roughly reject outliers in various previous works~\cite{yang2019polynomial,yang2020graduated,quan2020compatibility,zach2015dynamic,michel2017global}. In line voting, each pair of correspondences $(\boldsymbol{p}_i,\boldsymbol{q}_i)$ and $(\boldsymbol{p}_j,\boldsymbol{q}_j)$ can be evaluated by condition: $\left| \|\boldsymbol{p}_i\|-\|\boldsymbol{q}_i\| \right| \leq 2 \xi $, and if this condition is satisfied, each of these two correspondences will get 1 vote. Through the voting process by every single correspondence pair, we are then able to sort the correspondences according to their respective total numbers of votes obtained because the more votes one correspondence can get, the more reliable (or likely) it is to be a true inlier.

However, this strategy only performs either 0 or 1 voting for the correspondences, which poorly reflects the realistic distribution of noise (usually assumed to be Gaussian~\cite{yang2019polynomial,yang2020teaser,yang2020graduated,zhou2016fast,shi2021robin}). As a result, though line voting works well with relatively lower outlier rates, it would become brittle once the outlier rate grows extreme (e.g. more than 95\%), since outliers that are in dominance are likely to get even more votes than the inliers that are only in the minority. Therefore, our goal is to design a more effective voting strategy to increase the votes of true inliers and meanwhile truncate that of outliers, so as to make the correspondence sorting process more reliable even with extreme outlier rates.

To this end, we render a new idea, that is, to vote with the \textit{robust cost functions}, typically including Leclerc (Welsh), Cauchy (Lorentzian), Geman-McClure (GM), Truncated Least Squares (TLS), Tukey's Biweight (TB), etc, which has been widely applied to reject outliers in conjunction with M-estimation or GNC~\cite{yang2020graduated}. (The explicit expressions of these cost functions can be found in~\cite{black1996unification,barron2019general}.) These functions can endow the correspondences with weights based on their input residual errors, where specifically, they set small or even zero weights for correspondences with high residual errors while providing large weights for those with low residual errors, which has empirically proved to suit well with the Gaussian distribution of noise in practice.

Before formally discussing the voting technique, we first start with introducing an inequality condition built upon the invariability of the scale (or say, the distance) between pairwise point correspondences as in~\cite{li2021practical}. 

\begin{proposition1}[Pairwise Scale-invariant Condition] \label{Prop1} For any pair of correspondences $(\boldsymbol{p}_i,\boldsymbol{q}_i)$ and $(\boldsymbol{p}_j,\boldsymbol{q}_j)$ that are both inliers, they must satisfy the condition such that

\begin{equation}\label{scale-invariant}
S_{ij}=\left|{\|\boldsymbol{q}_i-\boldsymbol{q}_j\|}-{\|\boldsymbol{p}_i-\boldsymbol{p}_j\|}\right| \leq 2\xi,
\end{equation}

\noindent where $\xi$ is still the noise bound: $\|\boldsymbol{\varepsilon}\| \leq \xi$ in~\eqref{CM}.

\end{proposition1}

\begin{proof1} Based on the triangular inequality and the fact that the norm is invariant to rotation $\boldsymbol{R}$, we can derive that

\begin{equation}\label{proof-of-scale}
\begin{gathered}
\left|{\|\boldsymbol{q}_i-\boldsymbol{q}_j\|}-{\|\boldsymbol{p}_i-\boldsymbol{p}_j\|}\right| \\
=\left|{\|\boldsymbol{R} \cdot (\boldsymbol{p}_i-\boldsymbol{p}_j+\boldsymbol{R}^{\top}\boldsymbol{\varepsilon}_i-\boldsymbol{R}^{\top}\boldsymbol{\varepsilon}_j)\|}-{\|\boldsymbol{p}_i-\boldsymbol{p}_j\|}\right| \\
\leq \left\| (\boldsymbol{p}_i-\boldsymbol{p}_j)-({\boldsymbol{p}_i-\boldsymbol{p}_j})+\boldsymbol{R}^{\top}\boldsymbol{\varepsilon}_i-\boldsymbol{R}^{\top}\boldsymbol{\varepsilon}_j \right\| \\ 
= {2\|\boldsymbol{R}^{\top} \cdot (\boldsymbol{\varepsilon}_i-\boldsymbol{\varepsilon}_j)\|}\leq 2\xi,
\end{gathered}
\end{equation}

\noindent where $\|\boldsymbol{\varepsilon}\|\leq\xi$, so it is apparent to have $S_{ij}\leq {2\xi}$ as in~\eqref{scale-invariant}.
\end{proof1}

The next step is to compute $S$ for every pair of correspondences. Now we let $v_i$ and $v_j$ denote the numbers of votes w.r.t. correspondence $i$ and $j$, respectively. When we obtain $S_{ij}$, $v_i$ and $v_j$ will be simultaneously updated based on the TB function such that

\begin{equation}\label{TB-voting}
\begin{gathered}
v_i\leftarrow v_i+\left\{\begin{array}{ll}
\left(1-\frac{{S_{ij}}^2}{4\mu\xi^2}\right)^2 & \text { if } {S_{ij}}^2 \leq 4\mu\xi^2, \\
0 & \text { if } {S_{ij}}^2 > 4\mu\xi^2, \end{array}\right.\\
v_j\leftarrow v_j+\left\{\begin{array}{ll}
\left(1-\frac{{S_{ij}}^2}{4\mu\xi^2}\right)^2 & \text { if } {S_{ij}}^2 \leq 4\mu\xi^2, \\
0 & \text { if } {S_{ij}}^2 > 4\mu\xi^2,
\end{array}\right.
\end{gathered}
\end{equation}

\noindent where we generally set $u=1.5$ for allowing a slightly more lenient inlier threshold here.

When all the correspondence pairs have been involved in the voting above, we can sort the correspondences according to their respective numbers of votes $v$. The implementation details of our voting technique is specified in Algorithm~\ref{Voting-TB}.=

\begin{algorithm*}[h]
\caption{\textit{maxRotConsensus}}
\label{MaximizeRotConsensus}

\SetKwInOut{Input}{\textbf{Input}}
\SetKwInOut{Output}{\textbf{Output}}
\Input{correspondences $\{(\boldsymbol{p}_i,\boldsymbol{q}_i)\}_{i=1}^N$, threshold $\xi\leftarrow5\sigma$, sorted indices $\boldsymbol{d}$, chordal threshold $\theta$, boolean value $eIn$, and the minimum inlier number $I$\;}
\Output{inlier set candidate $\mathcal{I}^{\star}$\;}
%\BlankLine
\uIf{$eIn=0$}{
$N^{\circ}\leftarrow 0.2N$, $K_{\text{max}}\leftarrow 0$, $numBreak\leftarrow I$\;}
\ElseIf{$eIn=1$}{
$N^{\circ}\leftarrow N$, $K_{\text{max}}\leftarrow 0$, $numBreak\leftarrow 1.5I$\;
}
\For{$i=\boldsymbol{d}[1:(N^{\circ}-2)]$}{
\For{$j=\boldsymbol{d}[(i+1):(N^{\circ}-1)]$}{
\If{$S_{ij}$ satisfies condition~\eqref{scale-invariant}}{
$\mathcal{R}\leftarrow\emptyset$, $\mathcal{K}\leftarrow\emptyset$, $c\leftarrow 0$\;
\For{$k=\boldsymbol{d}[(j+1):N^{\circ}]$}{
\If{$S_{ik}$ and $S_{jk}$ both satisfy~\eqref{scale-invariant}}{
Solve rotation $\boldsymbol{R}_{ijk}$ minimally\;
$\mathcal{R}\leftarrow \mathcal{R}\cup \{\boldsymbol{R}_{ijk}\}$, $\mathcal{K}\leftarrow \mathcal{K}\cup \{k\}$, $c\leftarrow c+1$\;
\If{$count\geq (I-3)$}{
$\boldsymbol{R}^{\circ}\leftarrow$ \textit{robustLeeChordal}($\mathcal{R}$)\;
Find the consensus set $\mathcal{R}^{\circ}$ (with threshold $\theta$) of $\boldsymbol{R}^{\circ}$ from set $\mathcal{R}$ and obtain its corresponding index set $\mathcal{K}^{\circ}$\;
\If{$|\mathcal{K}^{\circ}|\geq K_{\text{max}}$}{
$K_{\text{max}}\leftarrow |\mathcal{K}^{\circ}|$, $\mathcal{I}^{\star}\leftarrow \mathcal{K}^{\circ}\cup\{i,j\}$\;
\If{$|\mathcal{I}^{\star}|\geq numBreak$}{
$yes\leftarrow 1$\;
\textbf{break}
}
}
}
}
}
}
\If{$yes=1$}{
\textbf{break}
}
}
\If{$yes=1$}{
\textbf{break}
}
}
\Return Inlier set candidate $\mathcal{I}^{\star}$\;
\end{algorithm*}

Besides the voting process discussed above, there is one additional point in Algorithm~\ref{Voting-TB} worth discussing. We supplement a Boolean (0 or 1) parameter $eIn$ (abbreviation for `\textit{enough inliers}') to greatly shorten the runtime of our solver in the case of low outlier rates. Here, we define that when one correspondence is able to get over $0.2 N$ votes within the first 20 iterations, indicating that inliers are abundant, we could immediately stop voting since it seems no longer necessary. In this case, we can directly feed the correspondences to our consensus maximizer (Section~\ref{consensus-maximize}) because the inliers are not sparse and can be found easily even without the correspondences being sorted. This early-termination operation can save much time in practice, making our solver run in milliseconds with no higher than 80\% outliers (will show in Section~\ref{experiments}).

\subsubsection{Why Choosing Tukey's Biweight?}

Readers may wonder why we only choose the TB cost for voting. This is because TB generally displays the best voting and sorting results in relation to the other robust cost functions. Now we provide an empirical demonstration to display its superiority. 

Figure~\ref{voting} shows several examples of the indices of the true inliers among the sorted correspondences using various robust cost functions, including traditional 0-1 as in~\cite{li2021practical}, Geman-McClure (GM), Leclerc, Cauchy, Truncated Least Squares (TLS), and TB, respectively. For example, if a true inlier is the third element in the sorted correspondence set $\boldsymbol{d}$ after voting, then its index should be 3. The experimental setup is the same as the benchmarking in Section~\ref{exp-benchmark}, and we test different situations with varied correspondence numbers $N$ and outlier rates, where for each situation, we conduct 10 random trials.

We can clearly observe that with $N=1000$ and 95\% outlier rate, most of the cost functions yield promising voting results, whereas when the outlie rate increases to 97\% and even up to 99\%, some cost functions may fail to make the true inliers rank at the top of set $\boldsymbol{d}$. For instance, the 0-1 voting yields bad sorting results in the trial 9 of Figure~\ref{voting}(e) and the trial 4 of Figure~\ref{voting}(f) since the true inliers are distributed far from the top elements of set $\boldsymbol{d}$. However, through TB-based voting, true inliers generally are shown to have the greatest tendency to distribute in the forefront (top) of set $\boldsymbol{d}$. In other words, TB voting most often generates the lowest and shortest box in these trials, therefore capable of rendering the most reliable sorting results.

In sum, there is no doubt that 0-1 line voting is sufficient for low-outlier cases; however, when the outlier rate goes extreme as in Figure~\ref{voting}(d-f), 0-1 voting would become unable to sort the correspondences reliably since there exist so many outliers that some outliers may even get more votes than true inliers. Hence, we need to introduce the TB cost in voting so as to maximize the credibility of the voting and sorting results.

\subsection{To Maximize the Consensus Set: Robust Single Rotation Averaging}
\label{consensus-maximize}

Solving the point cloud registration problem with outliers, in essence, consists in maximizing the correspondence set in which all the correspondences can reach a consensus on the model (to be specific, the rigid transformation: $\boldsymbol{R}$ and $\boldsymbol{t}$), as discussed in~\eqref{CM}.

The popular consensus maximizer RANSAC usually suffers from two disadvantages: (a) the probability of selecting an all-inlier subset could be rather low when the outlier rate is high, and (b) the residual errors w.r.t. all the correspondences must be computed to build the consensus set in each iteration of sampling, taking much time with large problems. 

Nonetheless, in subsection~\ref{sorting-and-voting}, we managed to sort the correspondences according to their inlier-reliability with the TB cost function, so that the probability to obtain all-inlier subsets would be significantly increased if we sample in the order of $\boldsymbol{d}$.

Therefore, in this subsection, we propose a novel consensus maximization method for robust registration on the basis of robust \textit{rotation averaging} and the scale-invariant constraint~\eqref{scale-invariant}, which brings two benefits: (a) the times of computing minimal rotations will be greatly reduced and (b) building the consensus set will become much less time-consuming than traditional random sampling and model fitting.

We introduce our consensus maximization method with pseudocode given in Algorithm~\ref{MaximizeRotConsensus}. 

\subsubsection{Description of Algorithm~\ref{MaximizeRotConsensus}}

First of all, we evaluate the $eIn$ obtained from Algorithm~\ref{Voting-TB}. If $eIn=0$ which means the full correspondence set is sorted as $\boldsymbol{d}$, our consensus will be found and maximized within the top $0.2N$ correspondences in $\boldsymbol{d}$. And if $eIn=1$ which means we have enough inliers and the sorting is not fully conducted, our consensus will be sought directly from the full correspondence set of size $N$ (\textit{lines 1-5}). Note that since our consensus maximization method generally runs much faster than correspondence sorting in Section~\ref{sorting-and-voting}, sacrificing the probability of finding an all-inlier subset here to reduce the time for correspondence sorting is worthwhile.

We enumerate a pair of correspondences (say $i$ and $j$) following the order of $\boldsymbol{d}$ (\textit{lines 7-8}) and compute its $S_{ij}$. If $S_{ij}$ satisfies condition~\eqref{scale-invariant}, we continue to sample the third correspondence (say $k$, also in the order of $\boldsymbol{d}$) and check whether $S_{ik}$ and $S_{jk}$ can both fulfill condition~\eqref{scale-invariant}. If yes, we form a 3-point set with $i$, $j$ and $k$, based on which rotation $\boldsymbol{R}_{ijk}$ is estimated minimally using Horn's 3-point triad-based solver~\cite{horn1987closed} (\textit{lines 9-12}). Then, for correspondences $i$ and $j$, we consecutively sample $k$ from $\boldsymbol{d}$ and examine them with~\eqref{scale-invariant}; for all $k$ that pass the tests, we store their minimal rotations $\boldsymbol{R}_{ijk}$ and indices $k$ into $\mathcal{R}$ and $\mathcal{K}$, respectively (\textit{line 13}). Intuitively, this strategy uses the pairwise scale-invariant condition to examine the 3-point sets during enumeration, and only the eligible ones could be forwarded to consensus maximization, which, as a result, saves plenty of time in (a) minimal rotation estimation and (b) building of consensus set.

Note that the rotation-based consensus set is established by finding all the rotations in $\mathcal{R}^{\circ}$ that have chordal error (distance) lower than $\theta$ with the averaged rotation $\boldsymbol{R}^{\circ}$. As for the proper choice of $\theta$, we drive it based on the geodesic error. Assume a minimally solved rotation $\boldsymbol{R}$ whose correspondences in the minimal subset are perturbed by noise $\boldsymbol{\varepsilon}\in\mathbb{R}^3$, writable as $\boldsymbol{R}^*=\boldsymbol{{R}}\cdot{Exp}([\boldsymbol{\varepsilon}^{\boldsymbol{R}}]_{\times})$, where noise $\boldsymbol{\varepsilon}$ has standard deviation $\sigma$, $[\,\,]_{\times}$ denotes the skew-symmetric matrix of a size-3 vector, and ${Exp}$ is the exponential map of matrix. According to the properties of the geodesic distance in~\cite{hartley2013rotation}, we can derive that

\begin{equation}\label{geo-error}
\begin{gathered}
\angle_{geo}(\boldsymbol{R}^*, \boldsymbol{R})=\angle\left(\boldsymbol{{R}}\cdot {Exp}([\boldsymbol{\varepsilon}]_{\times}), \boldsymbol{{R}}\right)\\
=\left| arccos\left(\frac{tr({Exp}([\boldsymbol{\varepsilon}]_{\times})^{\top}\boldsymbol{R}^{\top}\boldsymbol{R}))-1}{2}\right)\right|\\
= \angle_{geo} \left({Exp}([\boldsymbol{\varepsilon}]_{\times}), \mathbf{I}_3 \right)\leq \theta_{geo},
\end{gathered}
\end{equation}

\noindent where $\theta_{geo}$ is the geodesic threshold. For an empirical choice of $\theta_{geo}$, we usually let $\boldsymbol{\varepsilon}=\delta\boldsymbol{y}$, where $\boldsymbol{y}\in\mathbb{R}^3$ is a random unit vector and usually $\delta=10\frac{\sigma}{D}$ (here $D$ is the maximum diameter of the largest surface of the point cloud). For example, if a point cloud have largest diameter 1 and $\sigma=0.01$, $\theta_{geo}=\angle_{geo}\left(10\frac{0.01}{1}\boldsymbol{y}, \mathbf{I}_3\right)=0.1\, rad$. Then we can obtain the equivalent threshold $\theta$ in chordal distance according to~\eqref{chordal} such that:

\begin{equation}\label{chord-error}
\theta=2\sqrt{2} \sin\left(\frac{\theta_{geo}}{2}\right).
\end{equation}

As for a certain correspondence pair $(i, j)$, when we obtain over $I-3$ eligible $k$, meaning that we have sufficient 3-point sets passing the scale-invariant condition~\eqref{scale-invariant}, we can then use Lee's chordal-distance solver~\cite{lee2020robust} to robustly average the rotation group $\mathcal{R}$ and find all rotations from $\mathcal{R}$ that are in consensus with the averaged rotation $\boldsymbol{R}^{\circ}$ (\textit{lines 14-16}). Subsequently, if the size of this consensus set exceeds the so-far-the-best consensus size, we update the best consensus set with the current one (\textit{lines 17-18}). Here, note that we do not always need to enumerate all the possible 3-point sets within the top $0.2N$ correspondences in $\boldsymbol{d}$. Usually, when we obtain the minimum inlier number expected, which is set as $I$ when $eIn=0$ and is set as $1.5I$ when $eIn=1$ (\textit{line 4}), we can directly stop the enumeration and return the currently best $\mathcal{K}^{\circ}$ plus $i$ and $j$ as the inlier set candidate $\mathcal{I}^{\star}$ (\textit{lines 19-22}). Typically, we set ${I}=\max(0.05N,5)$ for $N\in(0,200)$, ${I}=0.04N$ for $N\in[200,300)$, ${I}=0.03N$ for $N\in[300,500)$, ${I}=0.02N$ for $N\in[500,1000)$, ${I}=0.01N$ for $N\geq 1000$. This strategy is applied to terminate our solver earlier in case that the outlier rate is not extremely high, while if the actual inlier number is smaller than ${I}$, this solver will continue to return tha maximum consensus just as expected.

Generally, a great majority of $\mathcal{I}^{\star}$ should be true inliers, so that it can be further refined and robustified by a standard GNC framework (we directly use our GNC-TB in Proposition~\ref{Prop1}).

\begin{algorithm}[t]
\caption{\textit{solveGNCTB}}
\label{solveGNCTB}
\SetKwInOut{Input}{\textbf{Input}}
\SetKwInOut{Output}{\textbf{Output}}
\Input{correspondences $\{(\boldsymbol{p}_i,\boldsymbol{q}_i)\}_{i=1}^N$, threshold $\xi\leftarrow5\sigma$, and inlier set candidate $\mathcal{I}^{\star}$\;}
\Output{estimated rotation $\boldsymbol{\hat{R}}$\;}
%\BlankLine
$N^{\star}\leftarrow |\mathcal{I}^{\star}|$\;
Initiate weights $\boldsymbol{\omega}=\{\omega_i\}_{i=1}^{N^{\star}}\leftarrow [1,1,\dots,1]$, and set $\mu\leftarrow 100$\;
\While{true}{
Use SVD to solve $\boldsymbol{\hat{R}}$ from~\eqref{obj} with weights $\boldsymbol{\omega}$\;
Compute the residual errors $\{r_i\}_{i=1}^{N^{\star}}$\;
Update weights $\boldsymbol{\omega}$ based on~\eqref{weight-updating}\;
$\mu\leftarrow \mu/1.2$\;
\If{converge \,or\, $\mu<1$}{
\textbf{break}
}
}
\Return rotation $\boldsymbol{\hat{R}}$\;
\end{algorithm}

\begin{algorithm}[t]
\caption{\text{VOCRA}}
\label{VOCRA}
\SetKwInOut{Input}{\textbf{Input}}
\SetKwInOut{Output}{\textbf{Output}}
\Input{correspondences $\{(\boldsymbol{p}_i,\boldsymbol{q}_i)\}_{i=1}^N$, thresholds $\xi_1\leftarrow3\sigma$ and $\xi_2\leftarrow5\sigma$, chordal threshold $\theta$, and minimum inlier number $I$\;}
\Output{best rotation $\boldsymbol{R^{\star}}$ and inlier set $\mathcal{I}^{in}$\;}
%\BlankLine
Vote with TB cost function: $(\boldsymbol{d},\,eIn)\leftarrow$\textit{votingTB}$\left(\{(\boldsymbol{p}_i,\boldsymbol{q}_i)\}_{i=1}^N,1.5,\xi_1\right)$ \;
Maximize the consensus set using rotation averaging: $\mathcal{I}^{\star}\leftarrow$\textit{maxRotConsensus}$\left(\{(\boldsymbol{p}_i,\boldsymbol{q}_i)\}_{i=1}^N,\xi_2,\theta, eIn,\boldsymbol{d},I\right)$\;
Make a correct estimate for the rotation: $\boldsymbol{\hat{R}}\leftarrow$\textit{solveGNCTB}$\left(\{(\boldsymbol{p}_i,\boldsymbol{q}_i)\}_{i=1}^N,\xi_2,\mathcal{I}^{\star}\right)$ \;
Use $\boldsymbol{\hat{R}}$ to find the final consensus (inlier) set $\mathcal{I}^{in}$ from all $N$ correspondences and compute the best rotation $\boldsymbol{R^{\star}}$ with $\mathcal{I}^{in}$ using SVD\;
\Return best rotation $\boldsymbol{R^{\star}}$ and inlier set $\mathcal{I}^{in}$\;
\end{algorithm}

\begin{figure*}[t]
\centering

\setlength\tabcolsep{1pt}
\addtolength{\tabcolsep}{0pt}
\begin{tabular}{cc}

%\multicolumn{2}{c}{\footnotesize{(a) bunny}}\\

\begin{minipage}[t]{0.42\linewidth}
\centering
\includegraphics[width=1\linewidth]{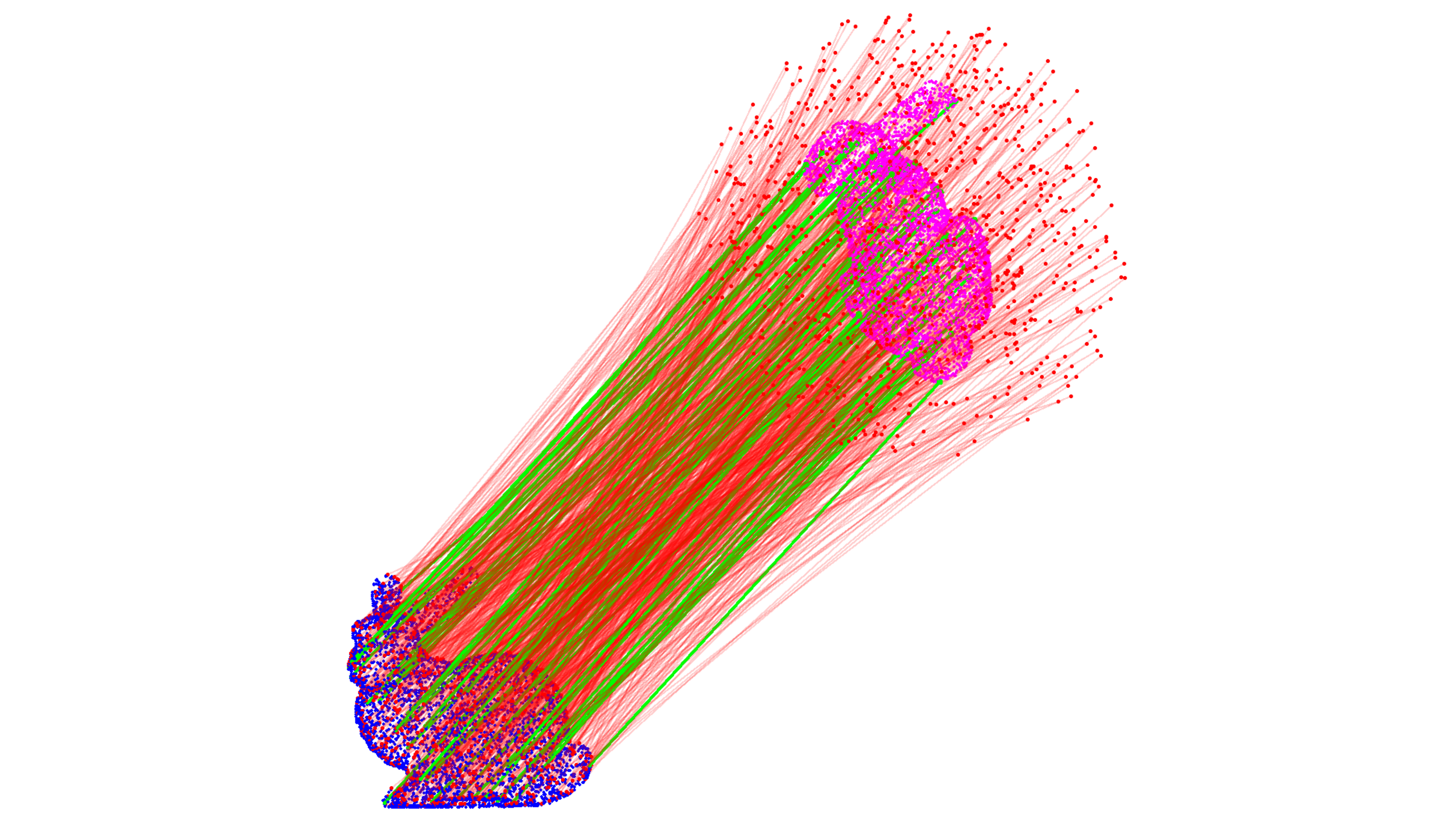}
\end{minipage}&

\begin{minipage}[t]{0.42\linewidth}
\centering
\includegraphics[width=1\linewidth]{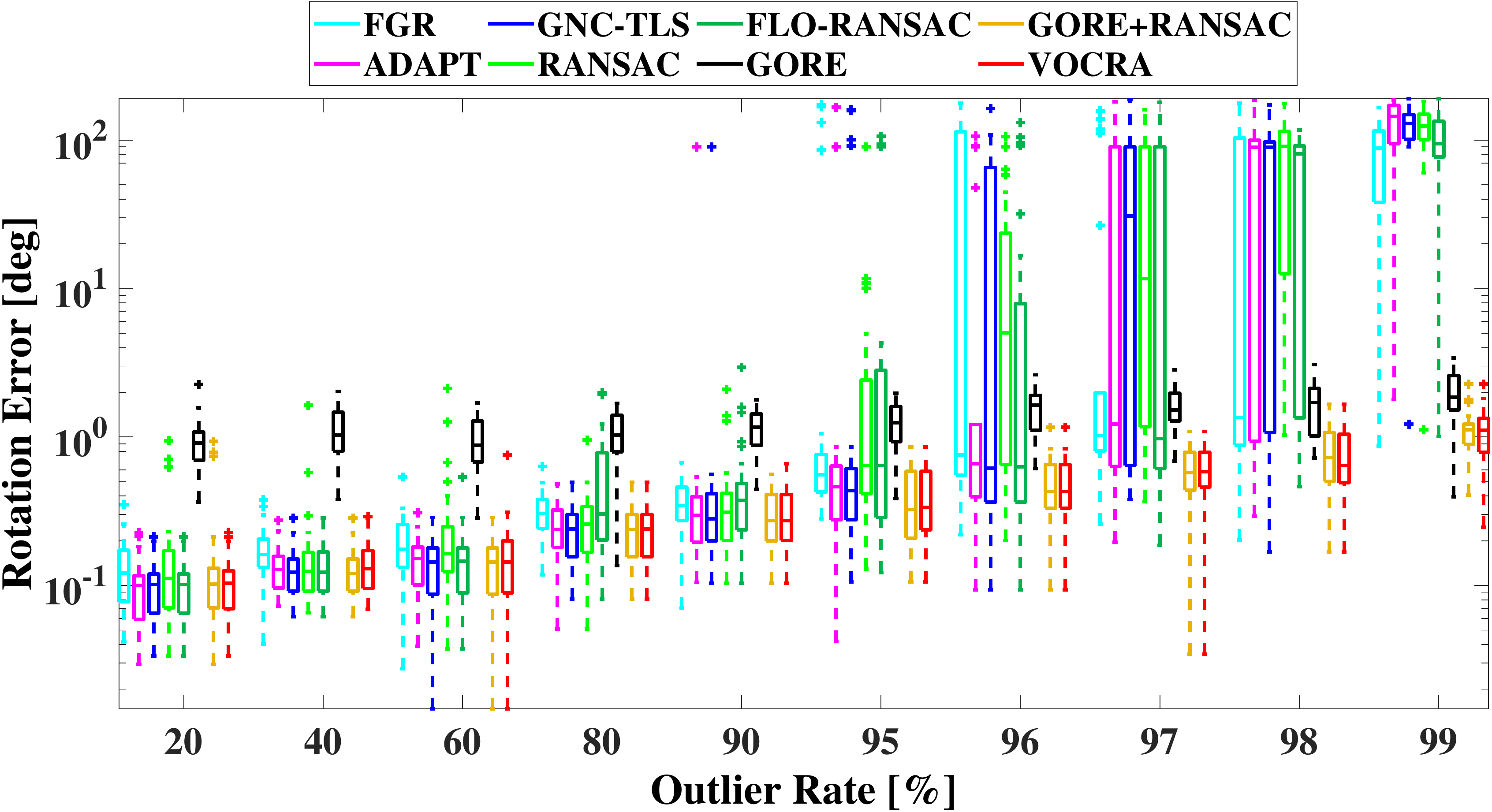}
\end{minipage}\\

\begin{minipage}[t]{0.42\linewidth}
\centering
\includegraphics[width=1\linewidth]{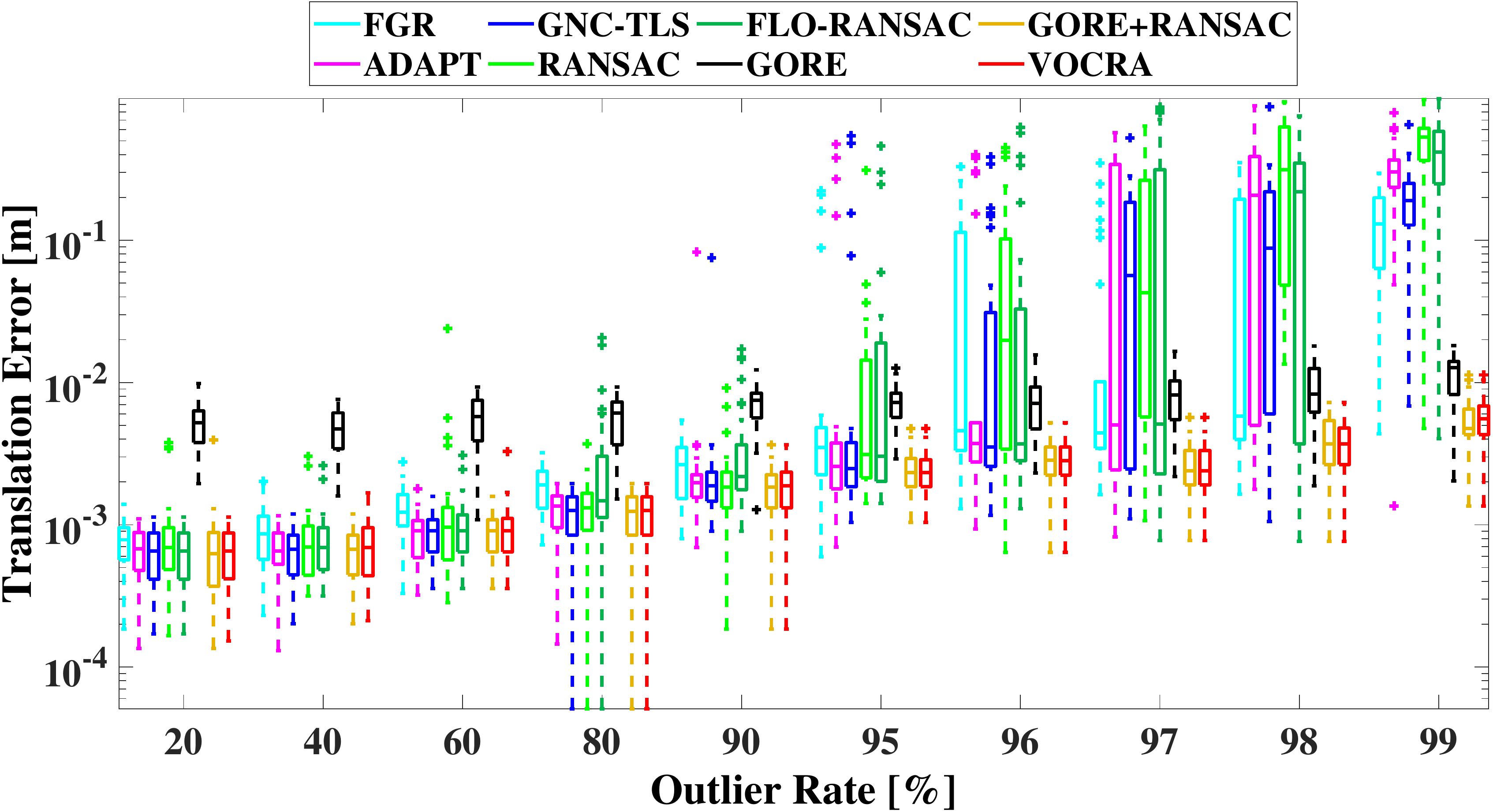}
\end{minipage}&

\begin{minipage}[t]{0.42\linewidth}
\centering
\includegraphics[width=1\linewidth]{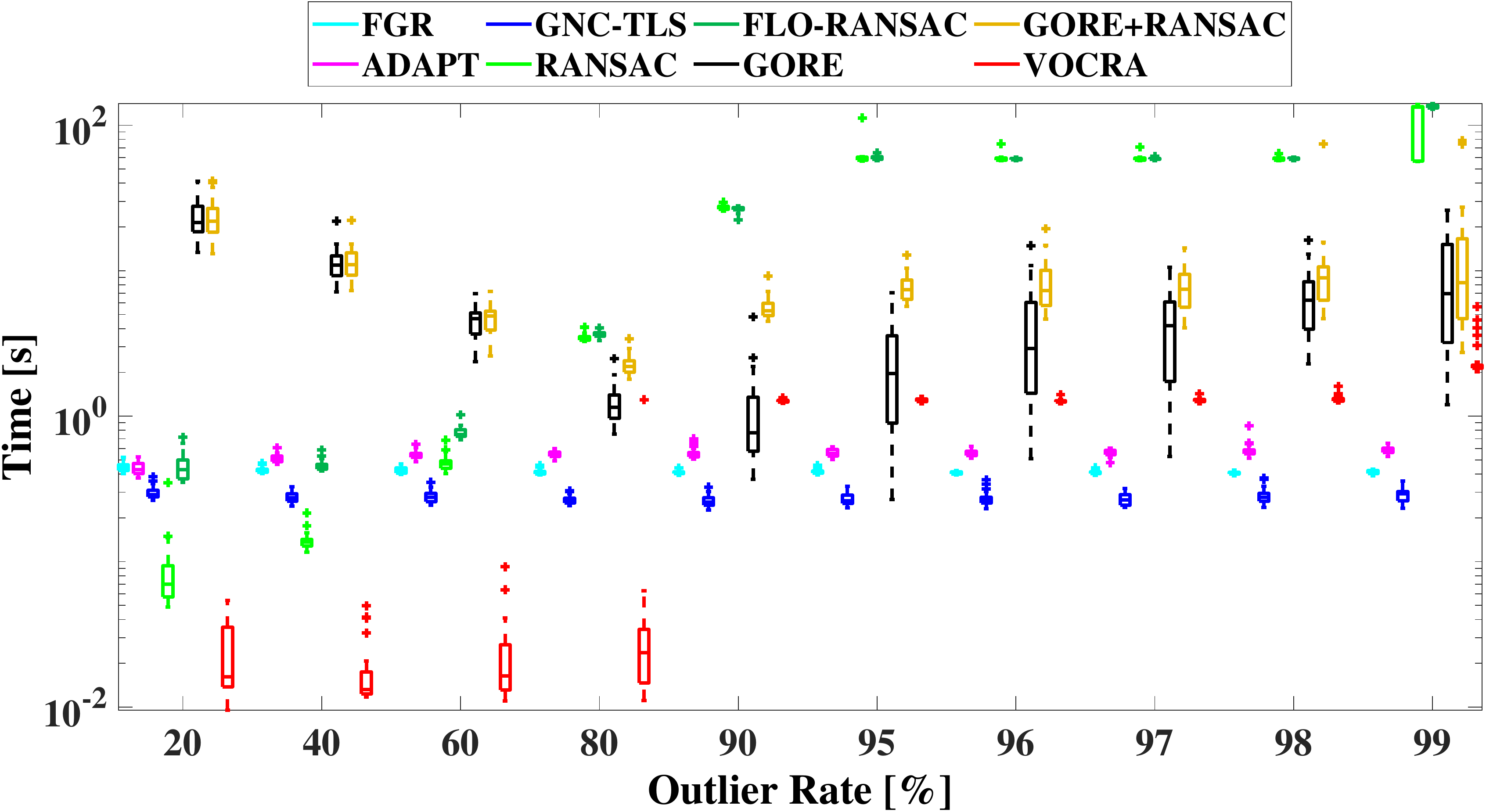}
\end{minipage} 

\end{tabular}
\vspace{-2mm}
\caption{Standard benchmarking over the \textit{bunny} point cloud~\cite{curless1996volumetric} using RANSAC~\cite{fischler1981random}, FLO-RANSAC~\cite{lebeda2012fixing}, FGR~\cite{zhou2016fast}, GNC-TLS~\cite{yang2020graduated}, ADAPT~\cite{tzoumas2019outlier}, GORE~\cite{bustos2017guaranteed}, GORE+RANSAC and VOCRA. The top-left image exemplifies correspondences with 95\% outliers in our setup where inliers are in green and outliers are in red.}
\label{bench-result1}
\vspace{-2mm}
\end{figure*}

\begin{figure*}[t]
\centering

\setlength\tabcolsep{1pt}
\addtolength{\tabcolsep}{0pt}
\begin{tabular}{cc}

%\multicolumn{2}{c}{\footnotesize{(b) armadillo}}\\

\begin{minipage}[t]{0.42\linewidth}
\centering
\includegraphics[width=1\linewidth]{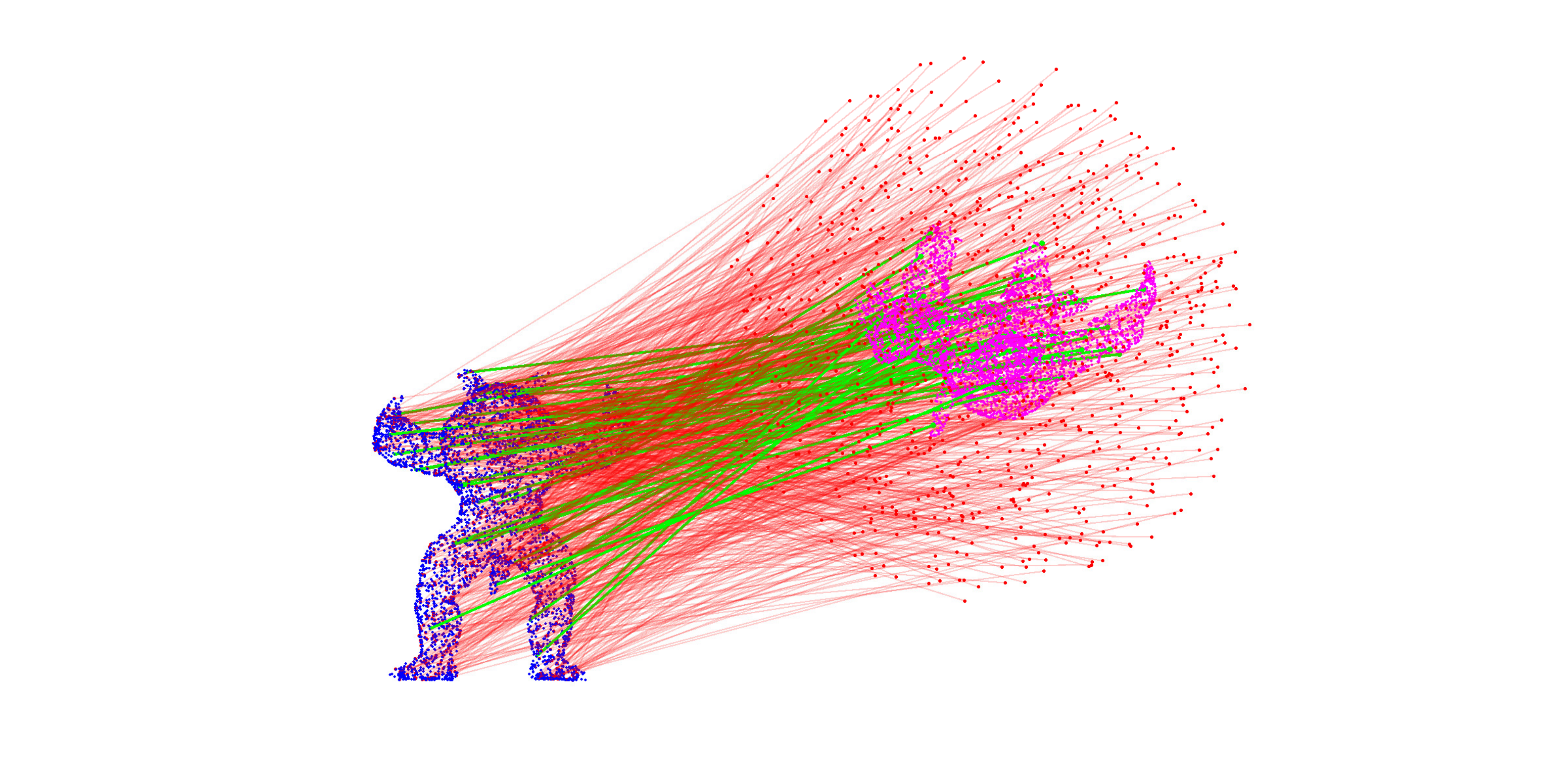}
\end{minipage}&

\begin{minipage}[t]{0.42\linewidth}
\centering
\includegraphics[width=1\linewidth]{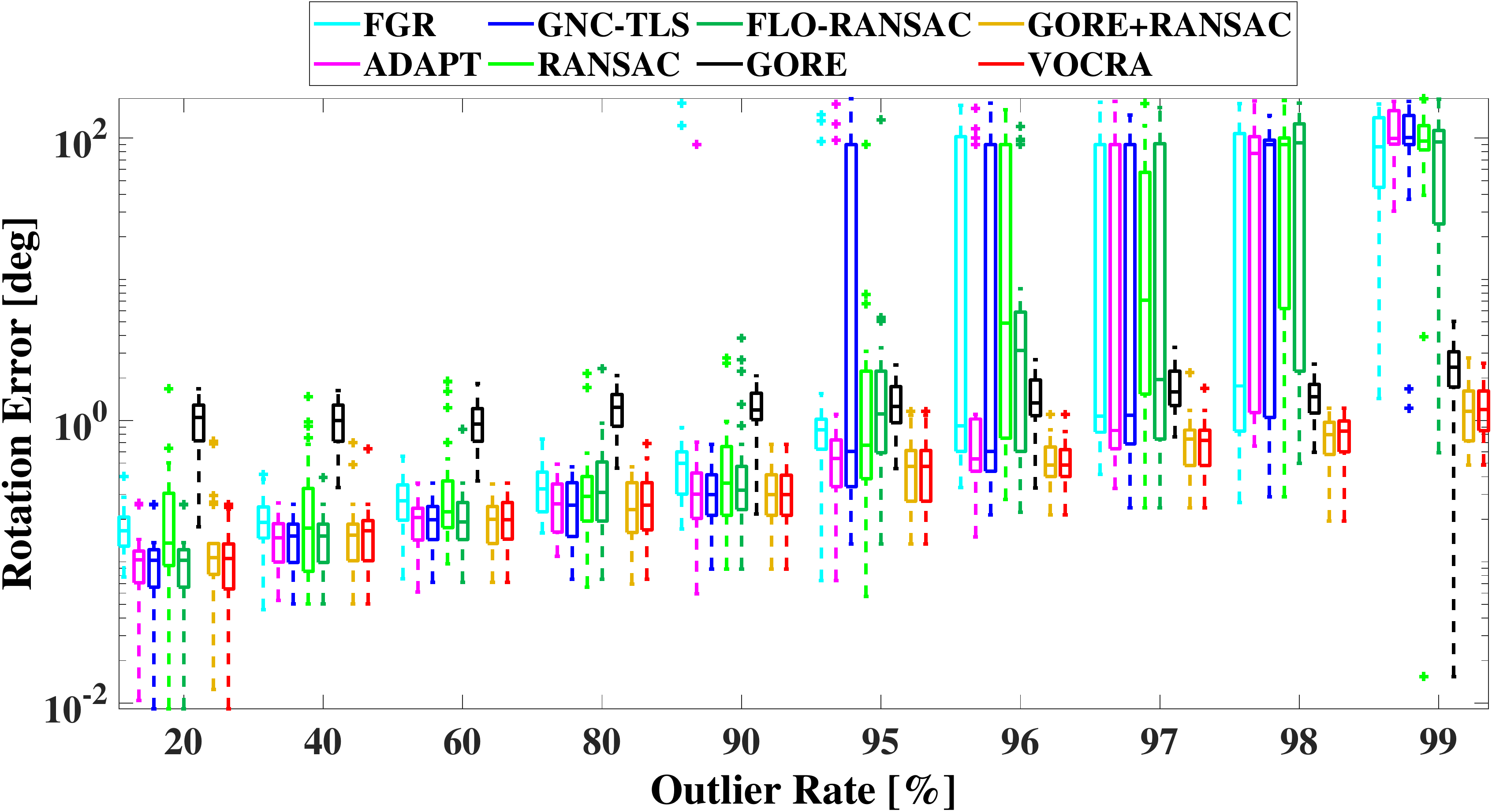}
\end{minipage}\\

\begin{minipage}[t]{0.42\linewidth}
\centering
\includegraphics[width=1\linewidth]{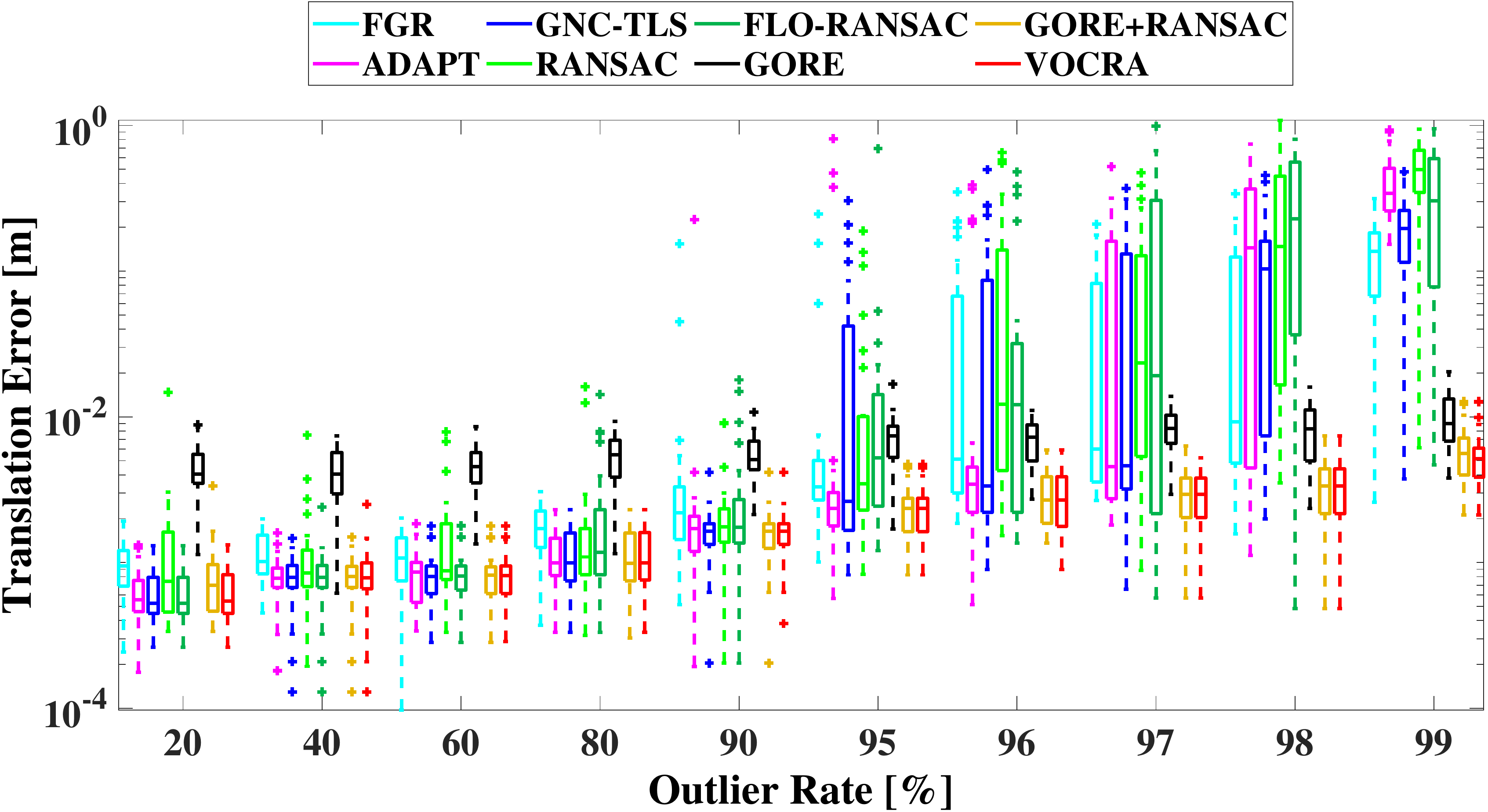}
\end{minipage}&

\begin{minipage}[t]{0.42\linewidth}
\centering
\includegraphics[width=1\linewidth]{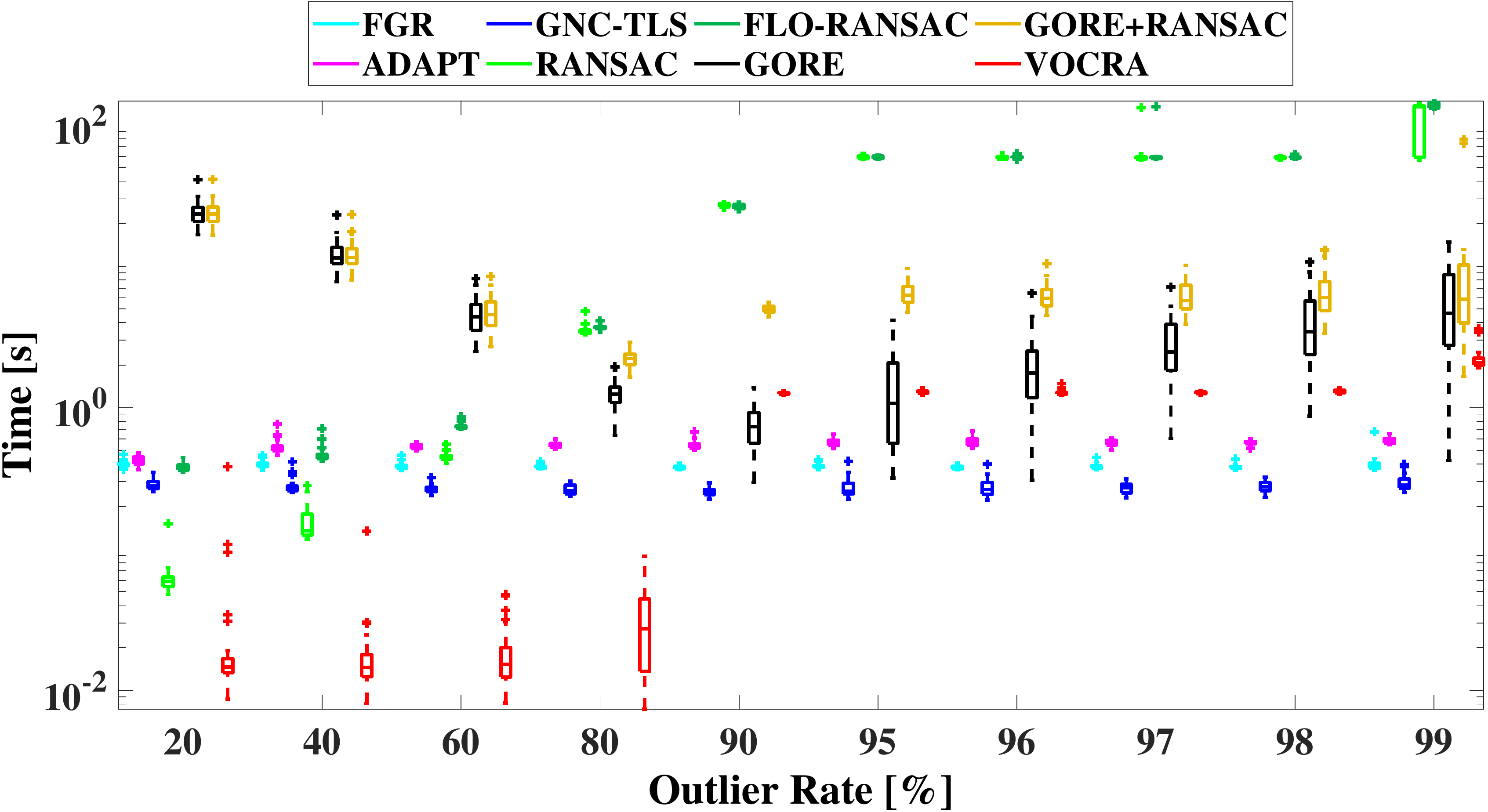}
\end{minipage}

\end{tabular}
\vspace{-2mm}
\caption{Standard benchmarking over the \textit{armadillo} point cloud~\cite{curless1996volumetric} using RANSAC~\cite{fischler1981random}, FLO-RANSAC~\cite{lebeda2012fixing}, FGR~\cite{zhou2016fast}, GNC-TLS~\cite{yang2020graduated}, ADAPT~\cite{tzoumas2019outlier}, GORE~\cite{bustos2017guaranteed}, GORE+RANSAC and VOCRA. The top-left image exemplifies correspondences with 95\% outliers in our setup where inliers are in green and outliers are in red.}
\label{bench-result2}
\vspace{-2mm}
\end{figure*}

\begin{figure*}[t]
\centering

\setlength\tabcolsep{1pt}
\addtolength{\tabcolsep}{0pt}
\begin{tabular}{cc}

%\multicolumn{2}{c}{\footnotesize{(c) dragon}}\\

\begin{minipage}[t]{0.42\linewidth}
\centering
\includegraphics[width=1\linewidth]{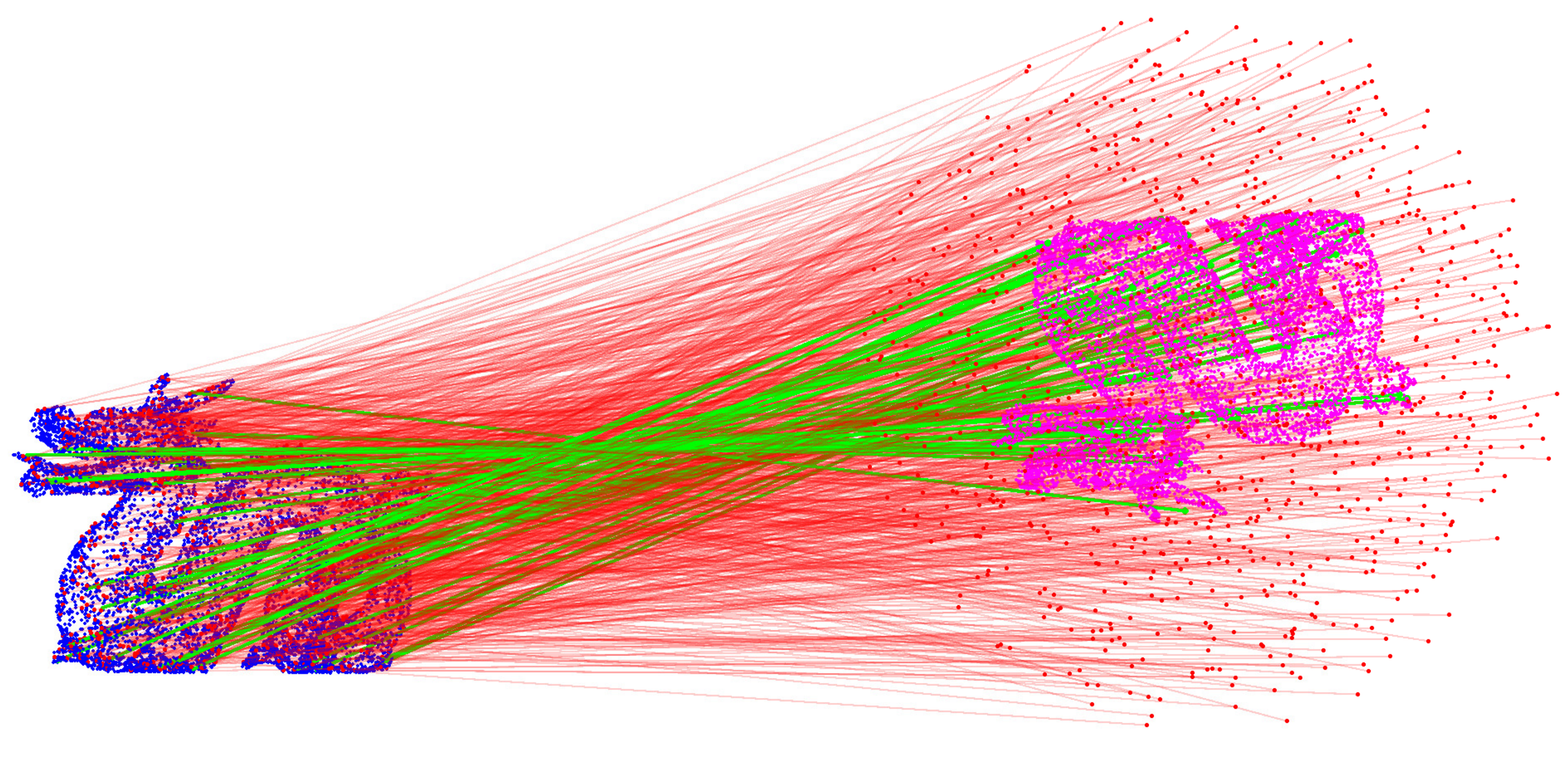}
\end{minipage}&

\begin{minipage}[t]{0.42\linewidth}
\centering
\includegraphics[width=1\linewidth]{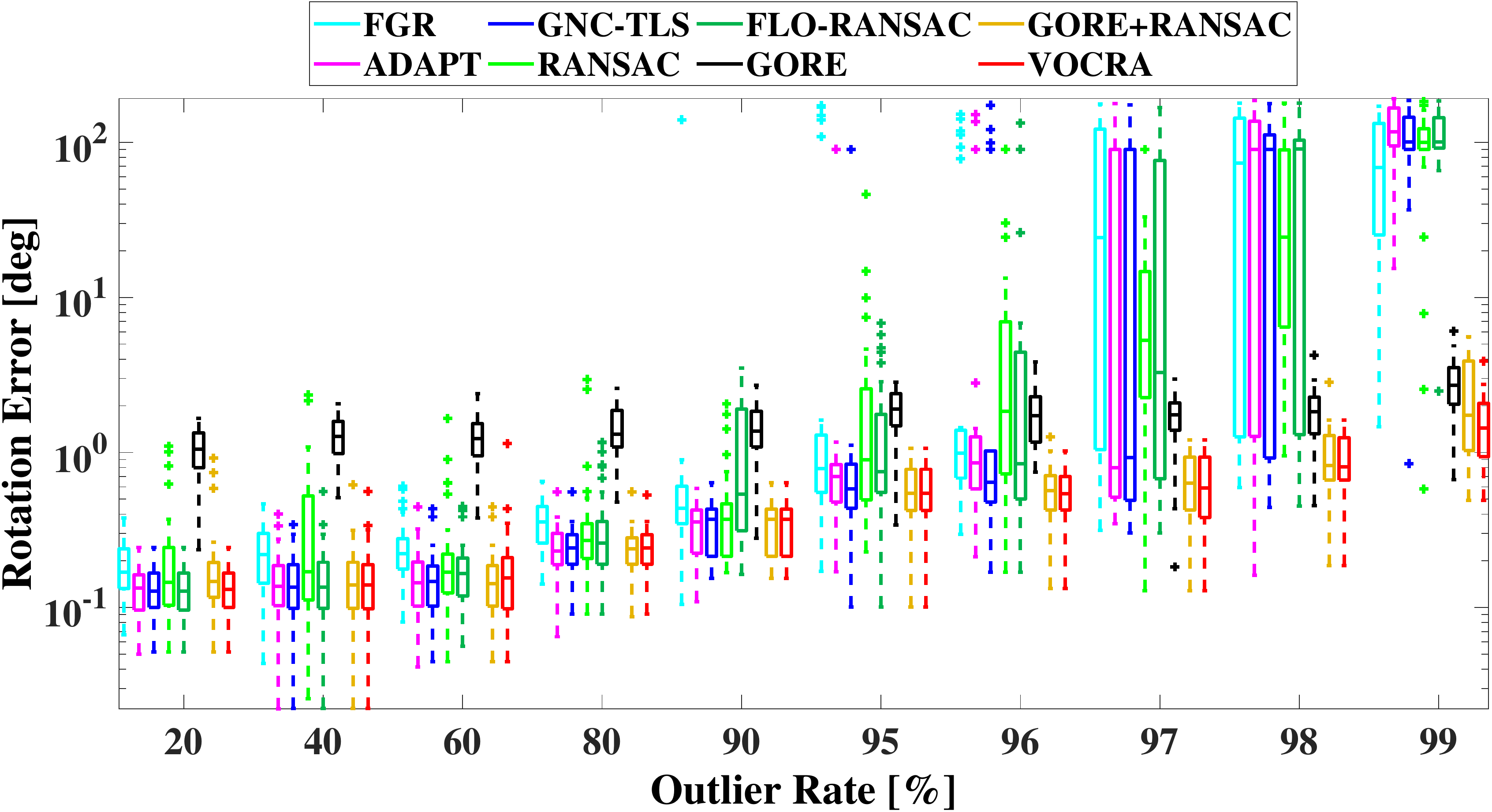}
\end{minipage}\\

\begin{minipage}[t]{0.42\linewidth}
\centering
\includegraphics[width=1\linewidth]{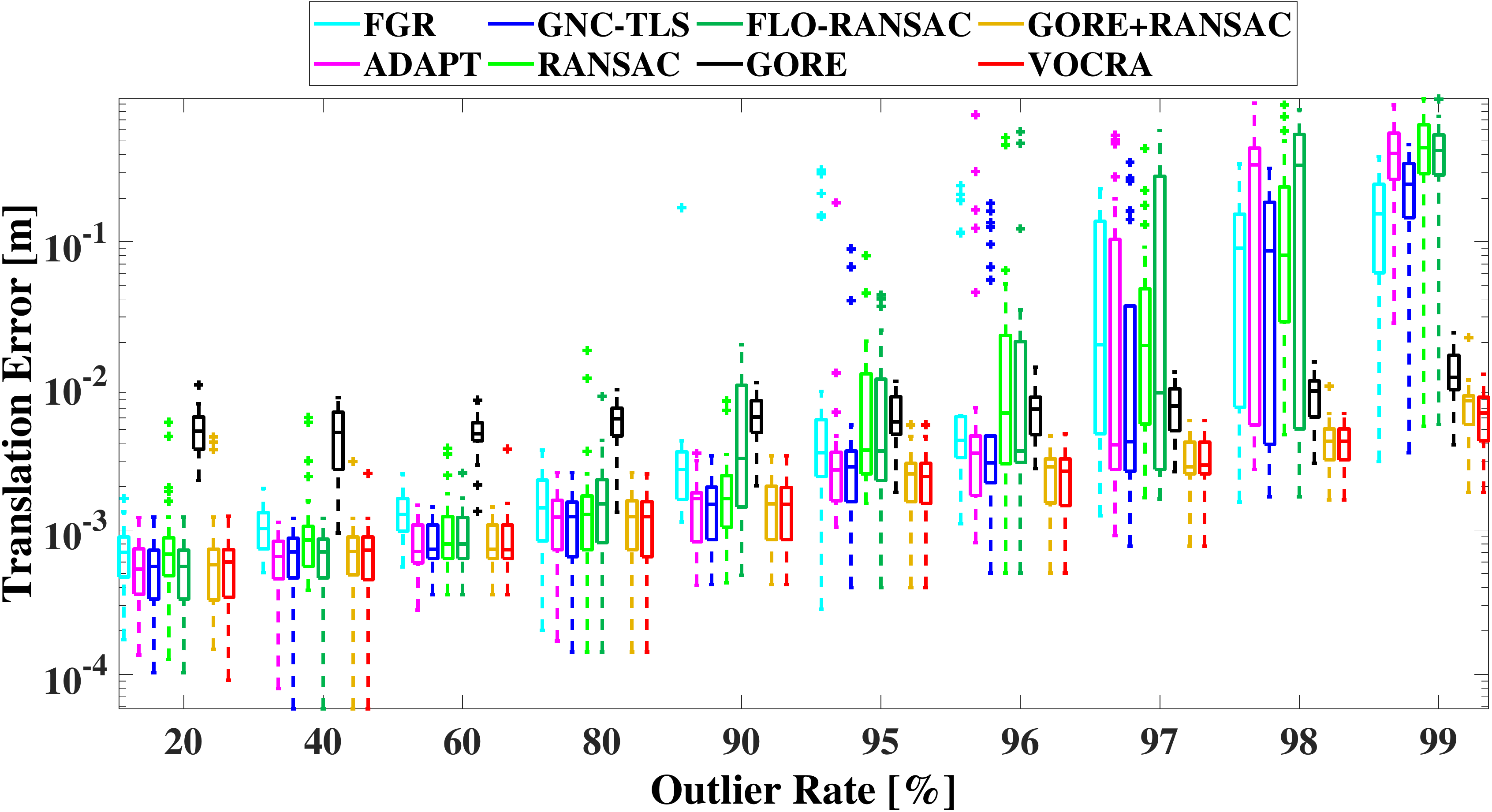}
\end{minipage}&

\begin{minipage}[t]{0.42\linewidth}
\centering
\includegraphics[width=1\linewidth]{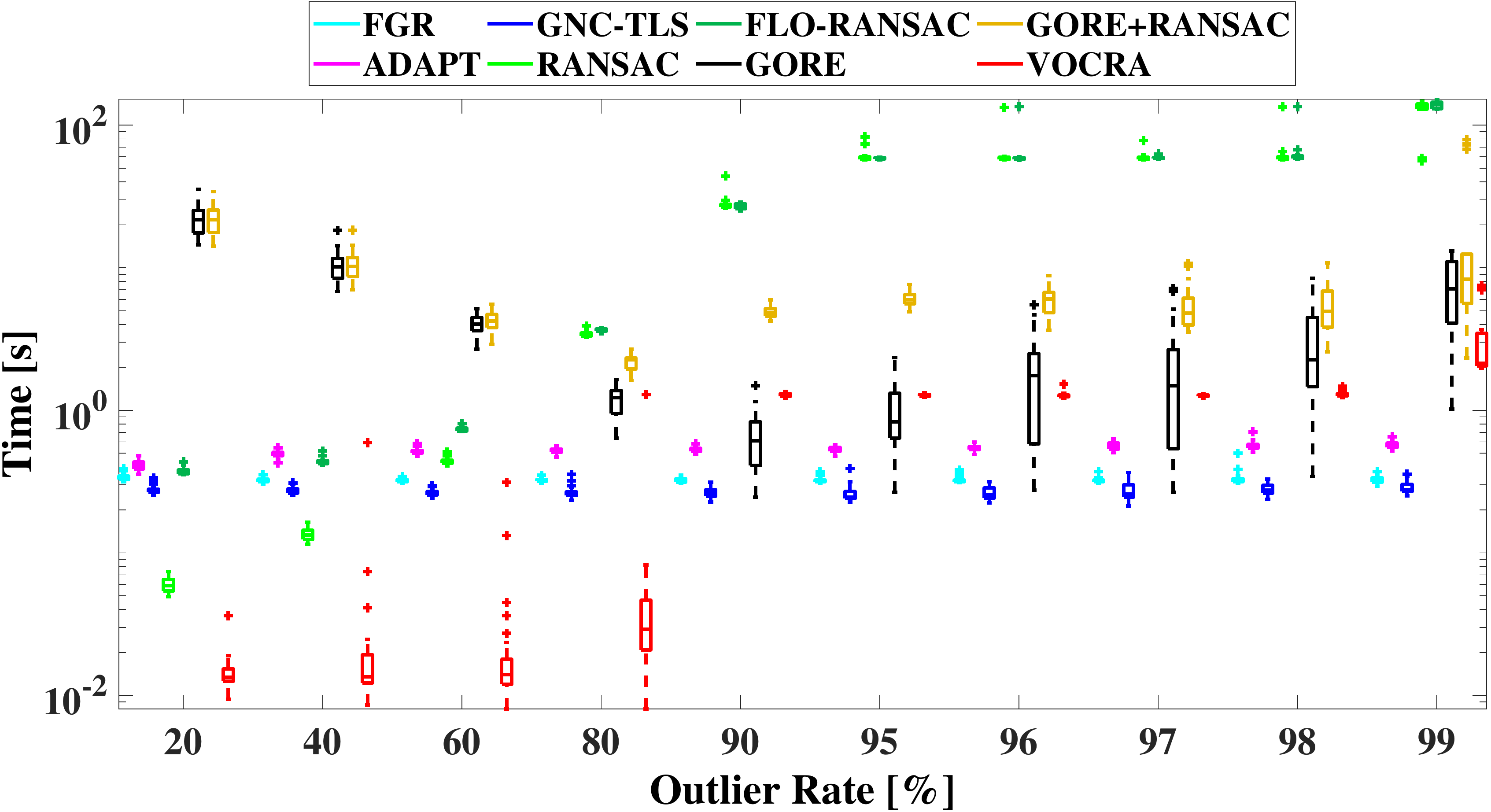}
\end{minipage}

\end{tabular}
\vspace{-2mm}
\caption{Standard benchmarking over the \textit{dragon} point cloud~\cite{curless1996volumetric} using RANSAC~\cite{fischler1981random}, FLO-RANSAC~\cite{lebeda2012fixing}, FGR~\cite{zhou2016fast}, GNC-TLS~\cite{yang2020graduated}, ADAPT~\cite{tzoumas2019outlier}, GORE~\cite{bustos2017guaranteed}, GORE+RANSAC and VOCRA. The top-left image exemplifies correspondences with 95\% outliers in our setup where inliers are in green and outliers are in red.}
\label{bench-result3}
\vspace{-2mm}
\end{figure*}

\begin{figure*}[t]
\centering

\setlength\tabcolsep{1pt}
\addtolength{\tabcolsep}{0pt}
\begin{tabular}{cc}

\begin{minipage}[t]{0.32\linewidth}
\centering
\includegraphics[width=1\linewidth]{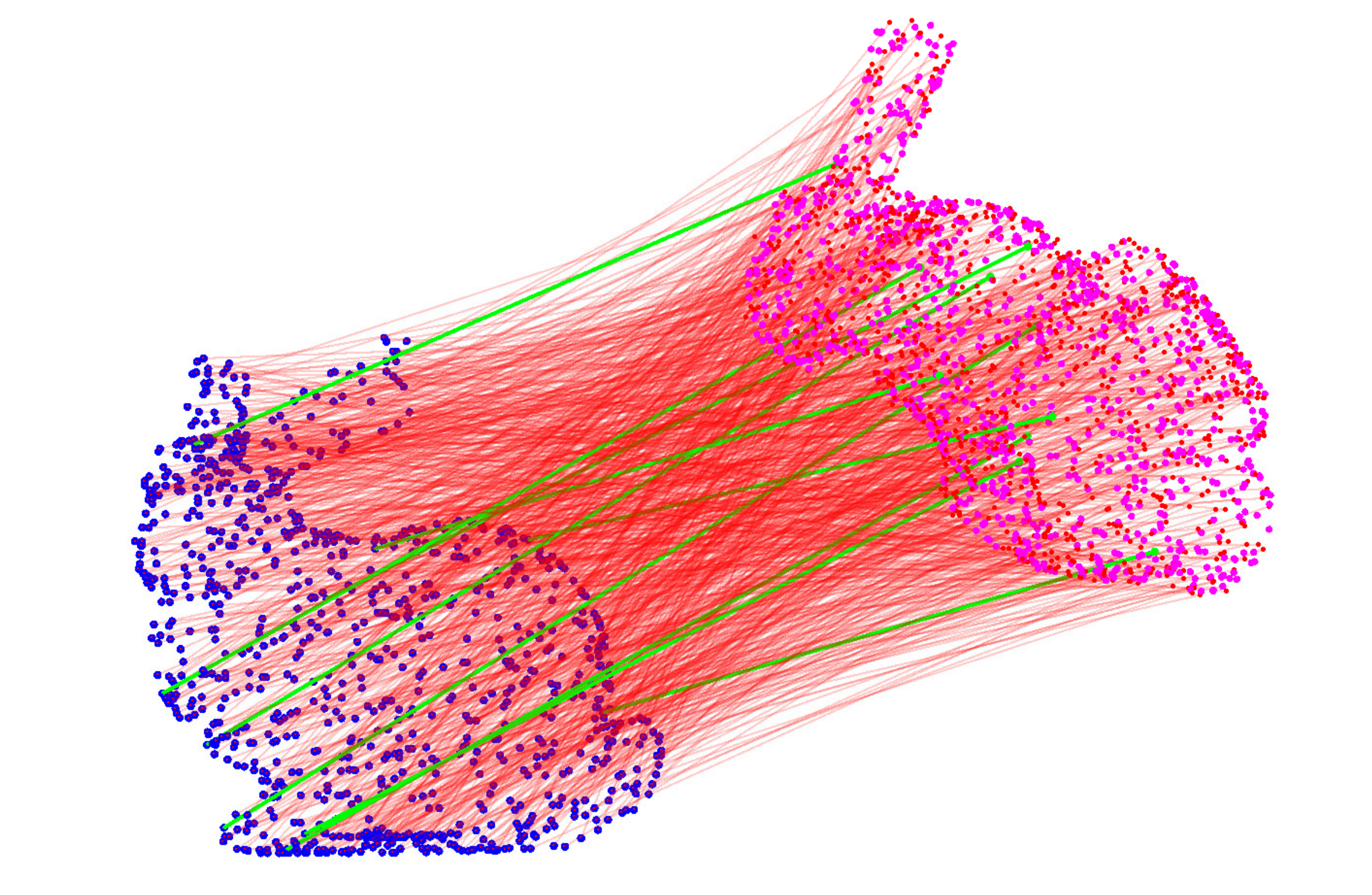}
\end{minipage}&

\begin{minipage}[t]{0.42\linewidth}
\centering
\includegraphics[width=1\linewidth]{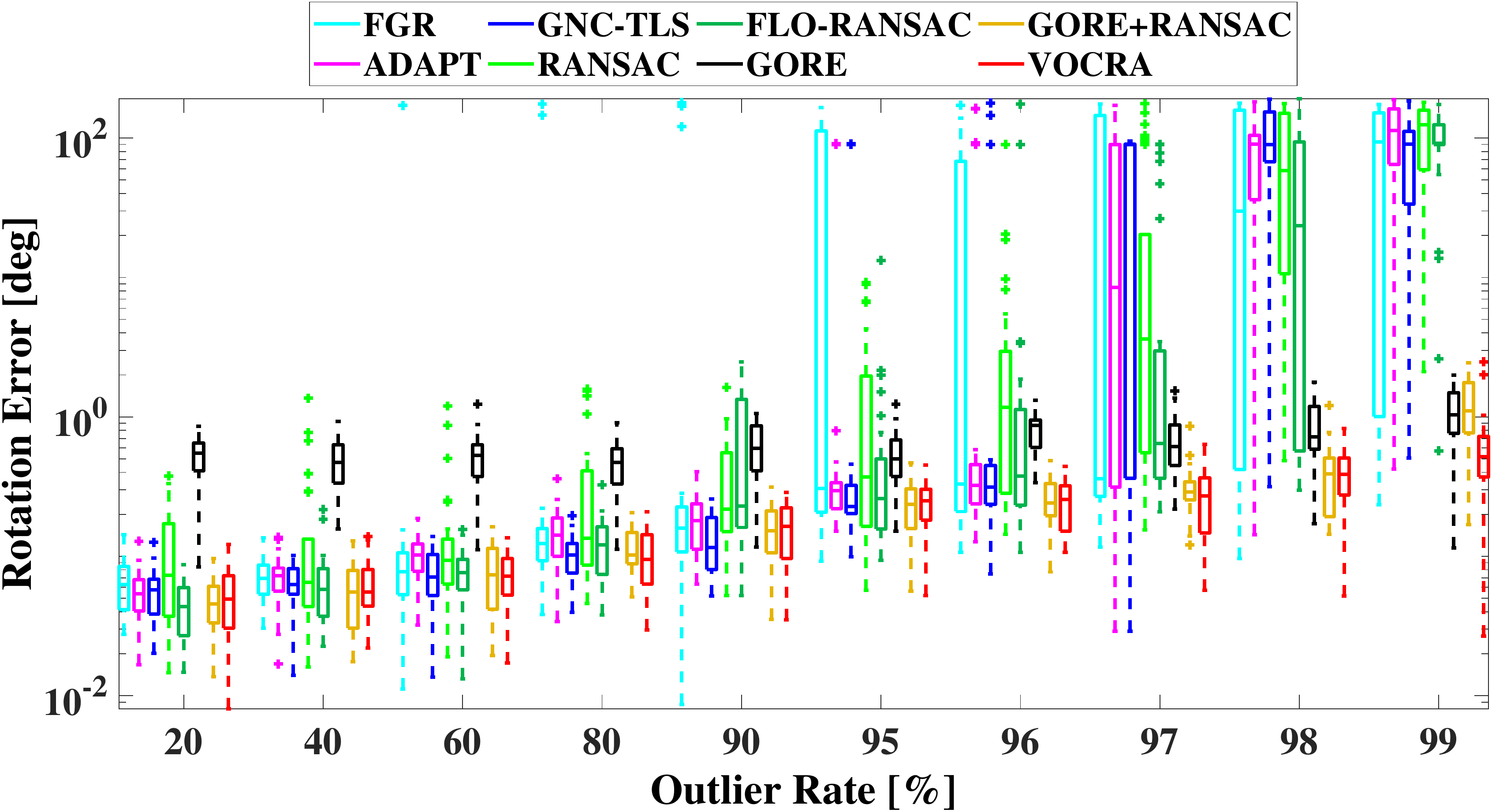}
\end{minipage}\\

\begin{minipage}[t]{0.42\linewidth}
\centering
\includegraphics[width=1\linewidth]{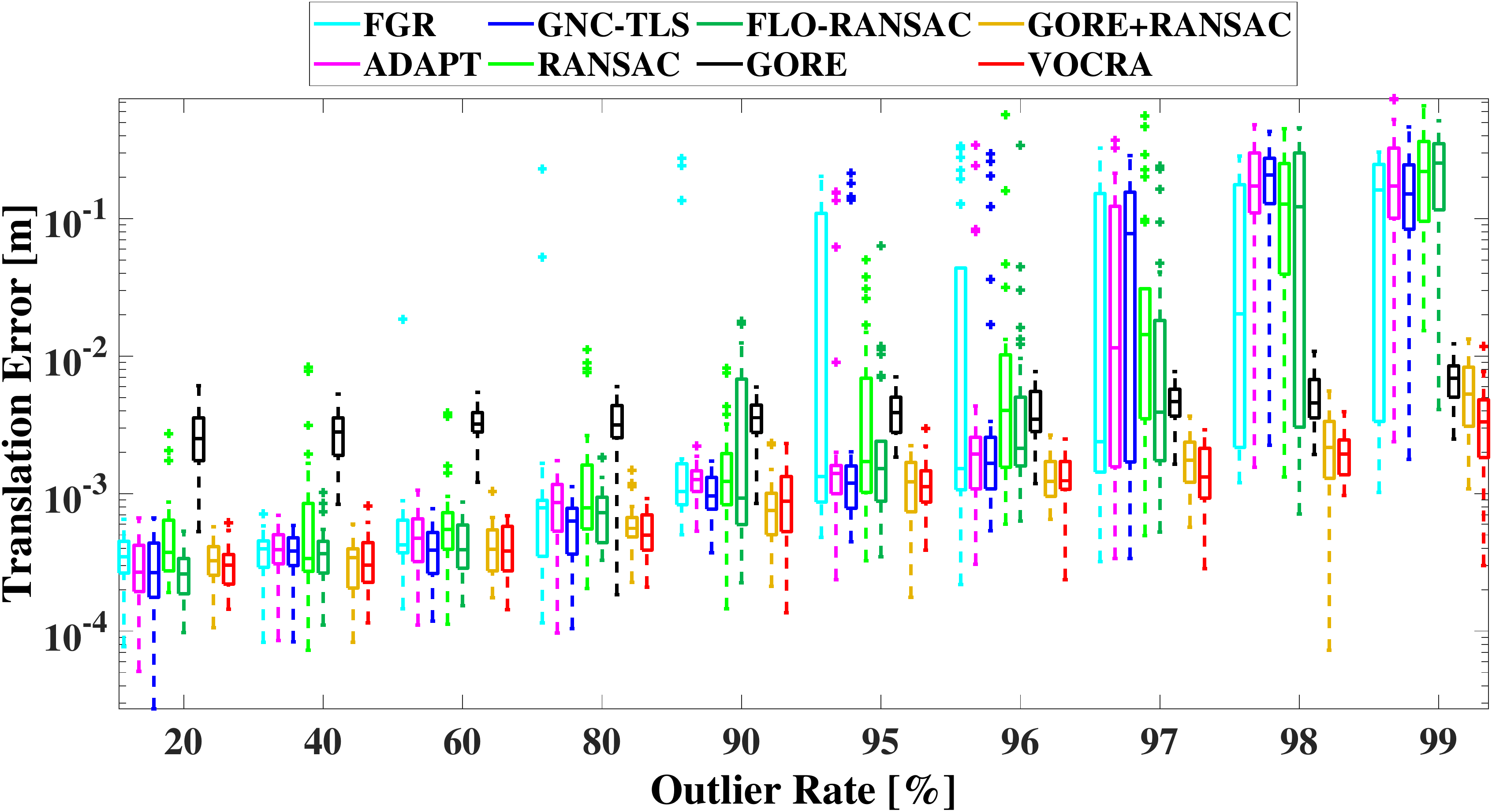}
\end{minipage}&

\begin{minipage}[t]{0.42\linewidth}
\centering
\includegraphics[width=1\linewidth]{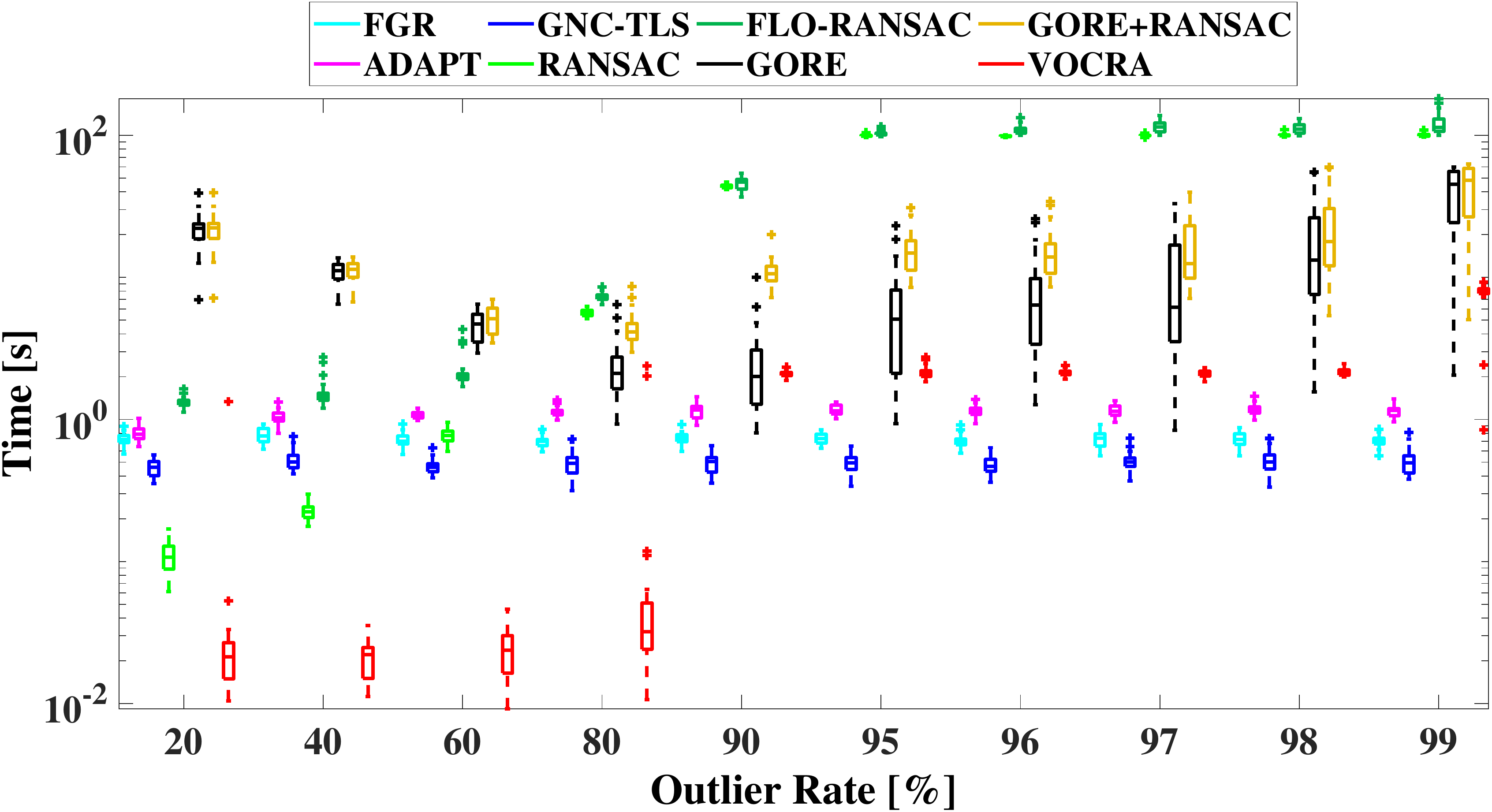}
\end{minipage} 

\end{tabular}
\vspace{-2mm}
\caption{On-surface benchmarking over the \textit{bunny} point cloud~\cite{curless1996volumetric} using RANSAC~\cite{fischler1981random}, FLO-RANSAC~\cite{lebeda2012fixing}, FGR~\cite{zhou2016fast}, GNC-TLS~\cite{yang2020graduated}, ADAPT~\cite{tzoumas2019outlier}, GORE~\cite{bustos2017guaranteed}, GORE+RANSAC and VOCRA. The top-left image exemplifies correspondences with 99\% outliers in our setup where inliers are in green and outliers are in red.}
\label{bench-surface}
\vspace{-2mm}
\end{figure*}

\subsection{To Further Prune Outliers: GNC-TB}\label{prune-outliers}

Despite the consensus reached over rotation within the inlier set candidate $\mathcal{I}^{\star}$, there might still exist outliers, but only in small numbers. Thus, we can apply our GNC-TB framework (Proposition~\ref{Prop1}) to find the true inliers from set $\mathcal{I}^{\star}$.

\begin{figure*}[t]
\centering
\setlength\tabcolsep{0pt}
\addtolength{\tabcolsep}{0pt}
\begin{tabular}{c|cccc|ccc}

\quad & \,\, & \footnotesize{FPFH Correspondences} & \footnotesize{Registration by VOCRA} & \quad & \,\, & \footnotesize{FPFH Correspondences} & \footnotesize{Registration by VOCRA} \\
\hline 
& & & & & & &
\\
\rotatebox{90}{\footnotesize{bunny}}\,
& &
\begin{minipage}[t]{0.22\linewidth}
\centering
\includegraphics[width=1\linewidth]{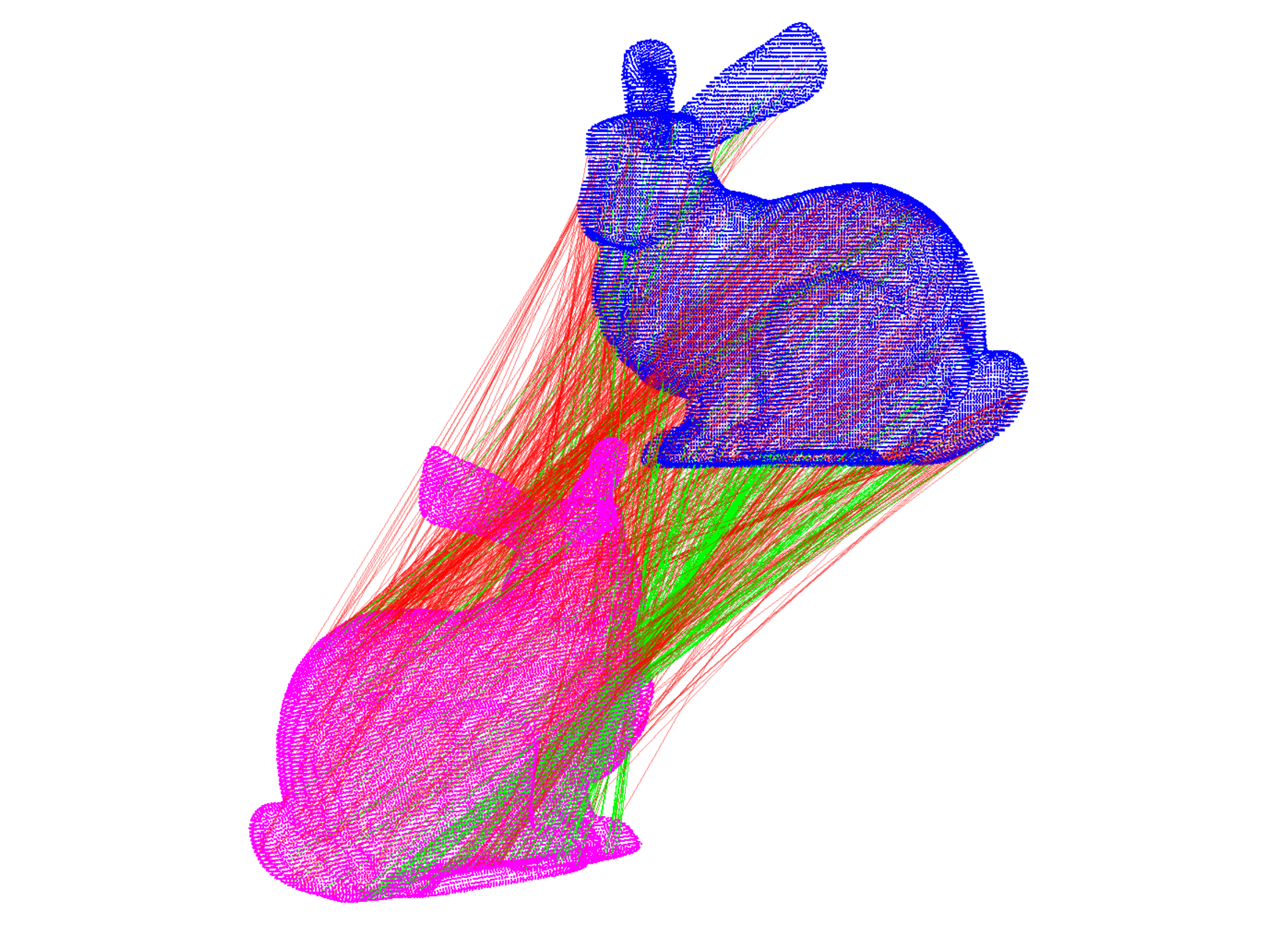}
\end{minipage}
& 
\begin{minipage}[t]{0.22\linewidth}
\centering
\includegraphics[width=1\linewidth]{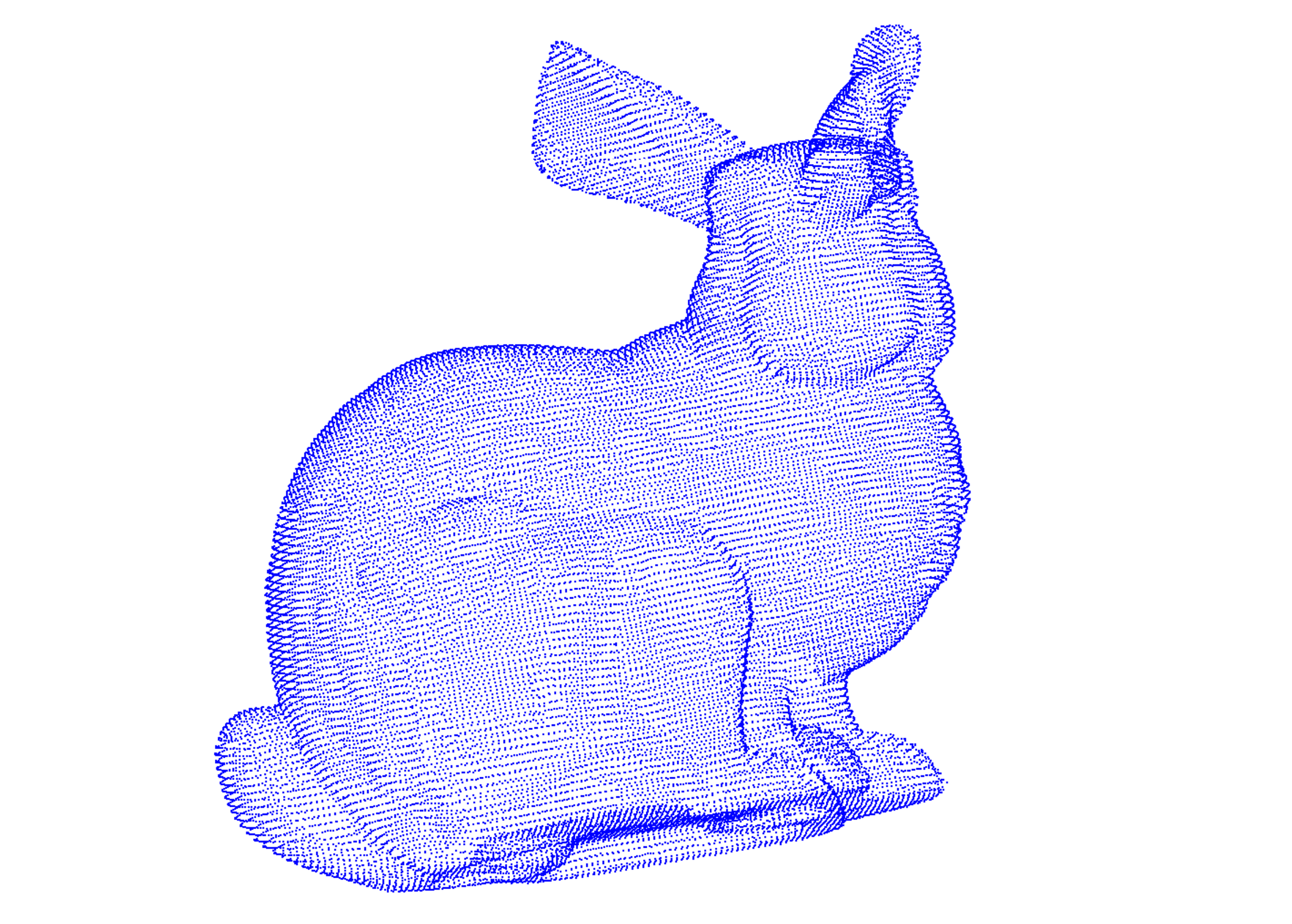}
\end{minipage}
&
\rotatebox{90}{\footnotesize{armadillo}}\,
& &
\begin{minipage}[t]{0.21\linewidth}
\centering
\includegraphics[width=1\linewidth]{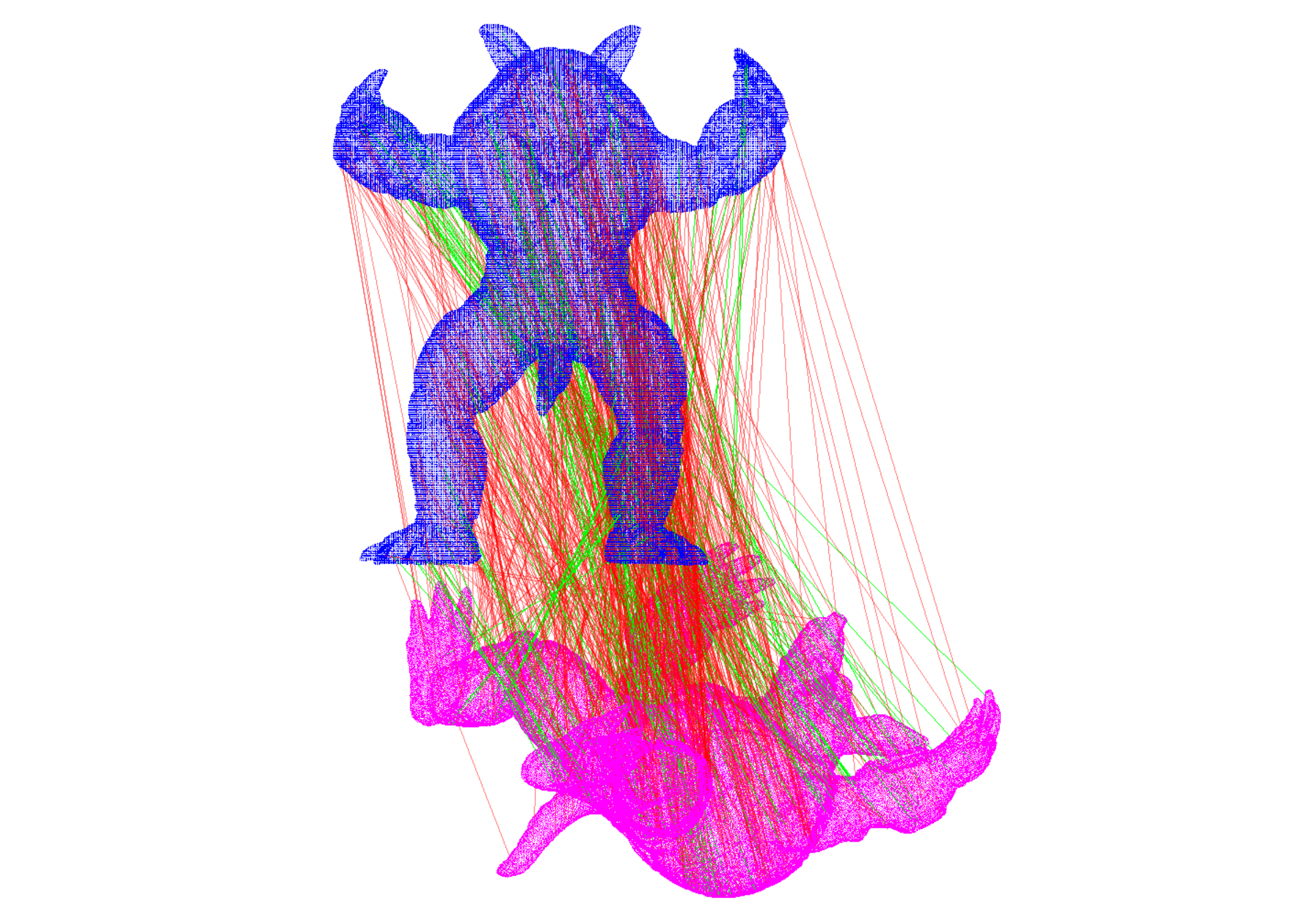}
\end{minipage}
&
\begin{minipage}[t]{0.20\linewidth}
\centering
\includegraphics[width=1\linewidth]{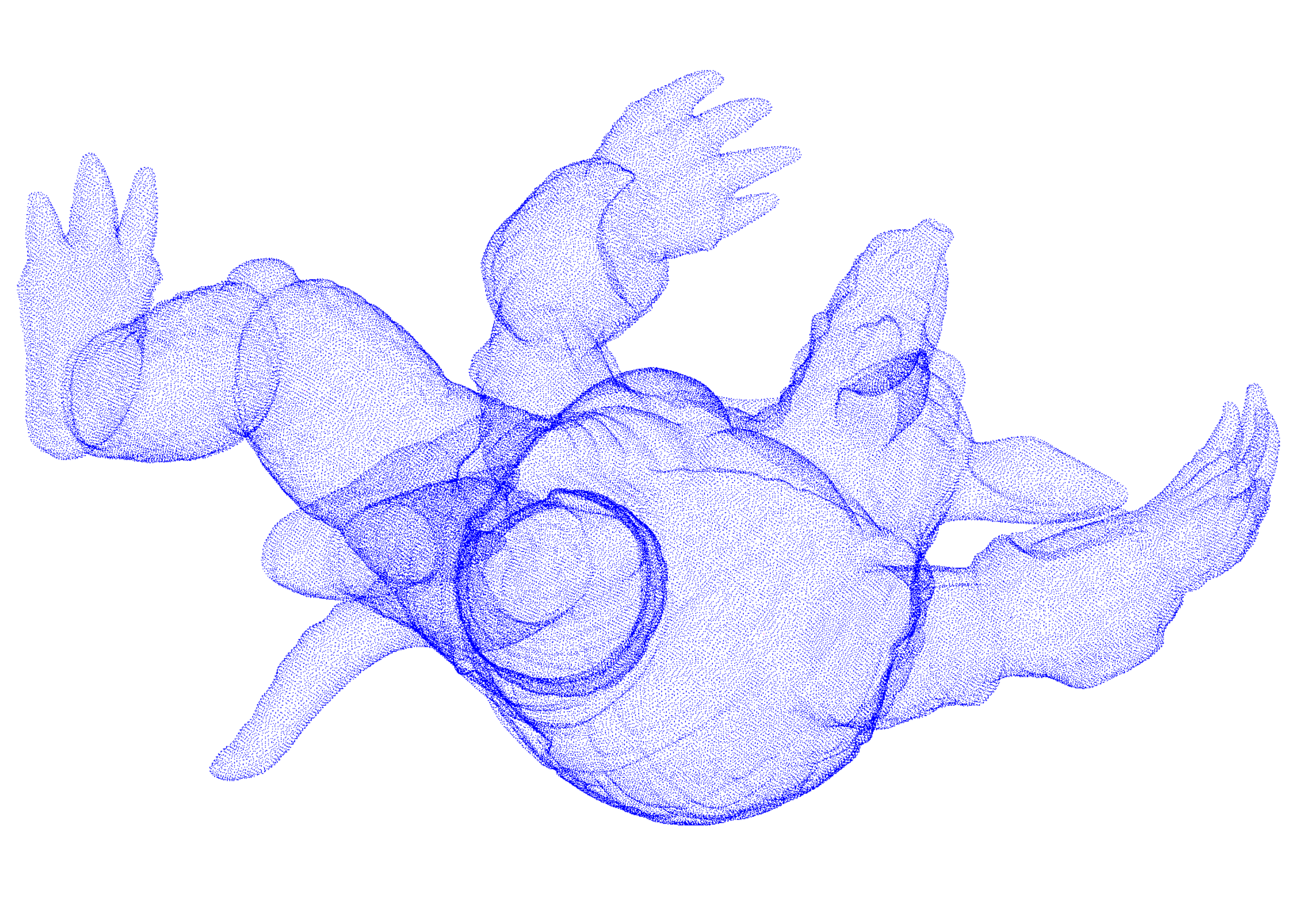}
\end{minipage}
\\
\rotatebox{90}{\footnotesize{dragon}}\,
& &
\begin{minipage}[t]{0.22\linewidth}
\centering
\includegraphics[width=1\linewidth]{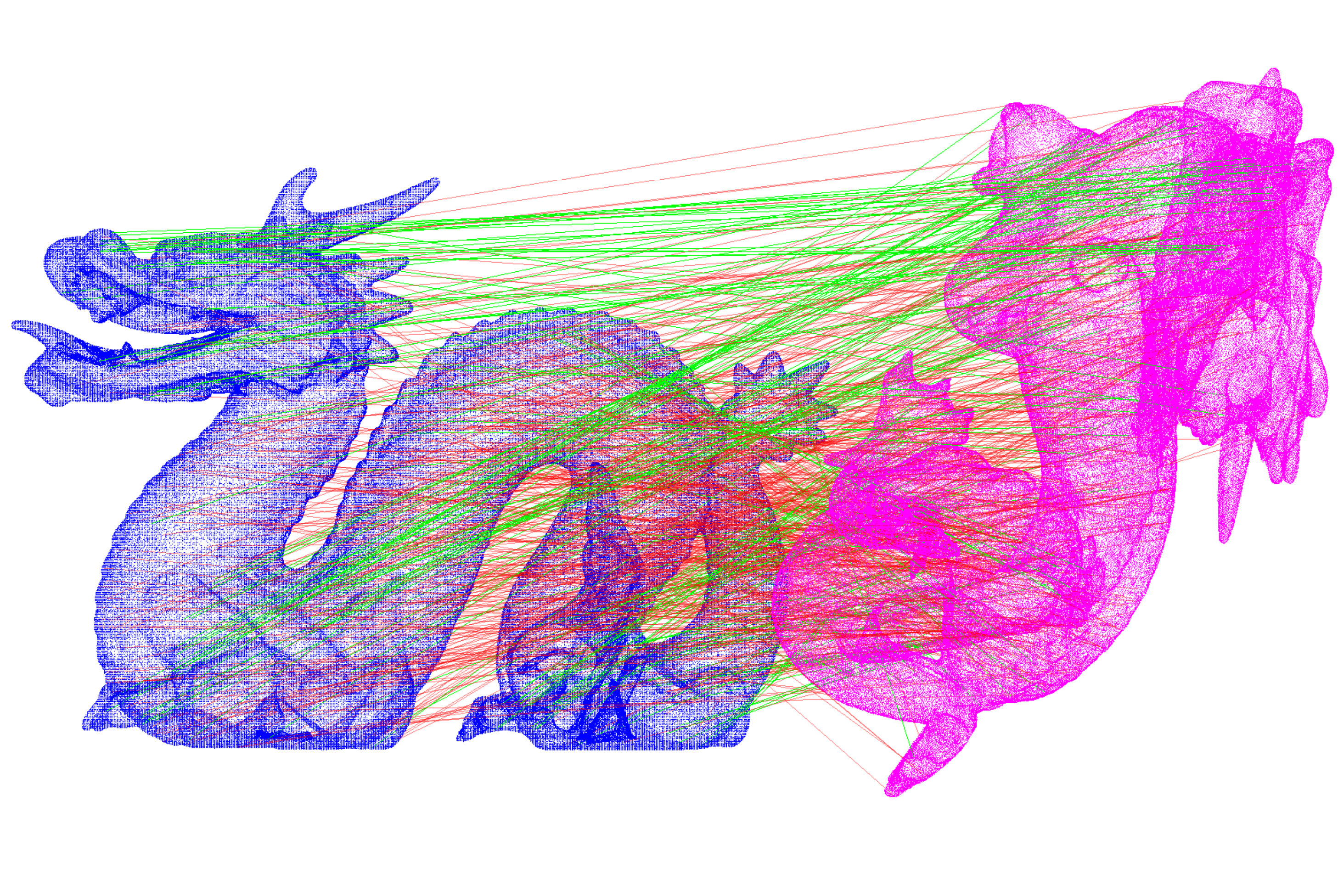}
\end{minipage}
&
\begin{minipage}[t]{0.23\linewidth}
\centering
\includegraphics[width=1\linewidth]{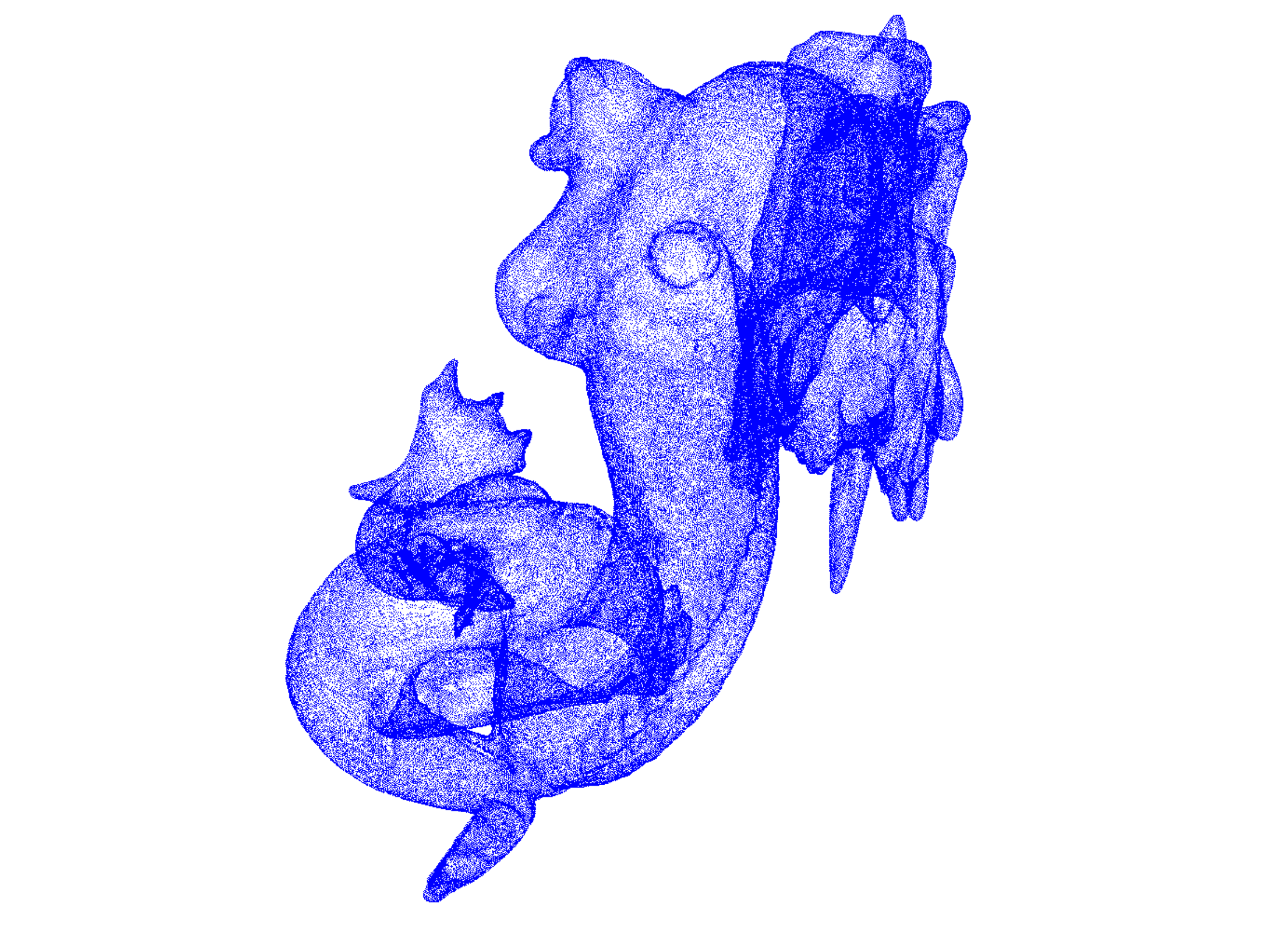}
\end{minipage}
&
\rotatebox{90}{\footnotesize{cheff}}\,
& &
\begin{minipage}[t]{0.22\linewidth}
\centering
\includegraphics[width=1\linewidth]{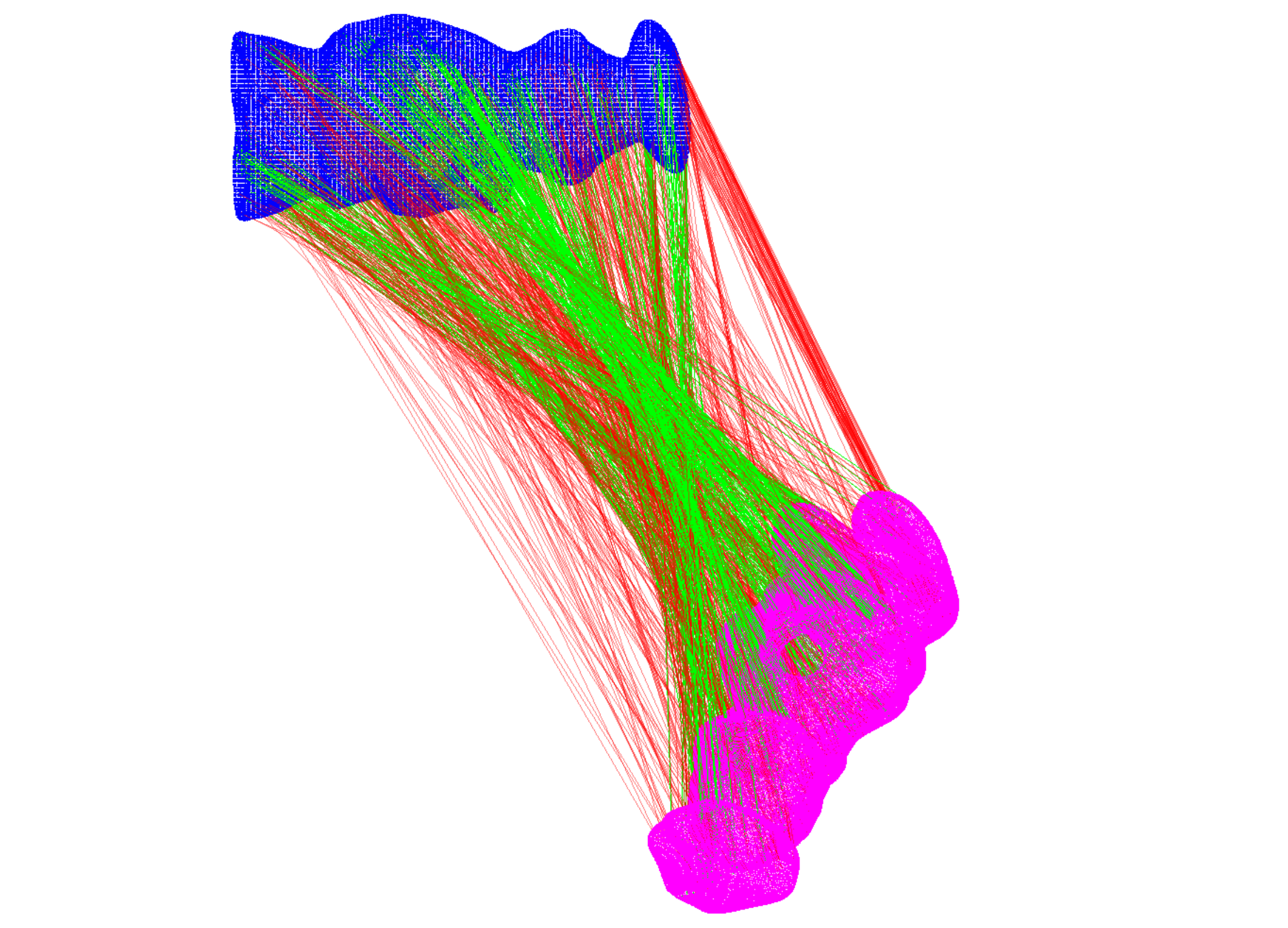}
\end{minipage}
&
\begin{minipage}[t]{0.225\linewidth}
\centering
\includegraphics[width=1\linewidth]{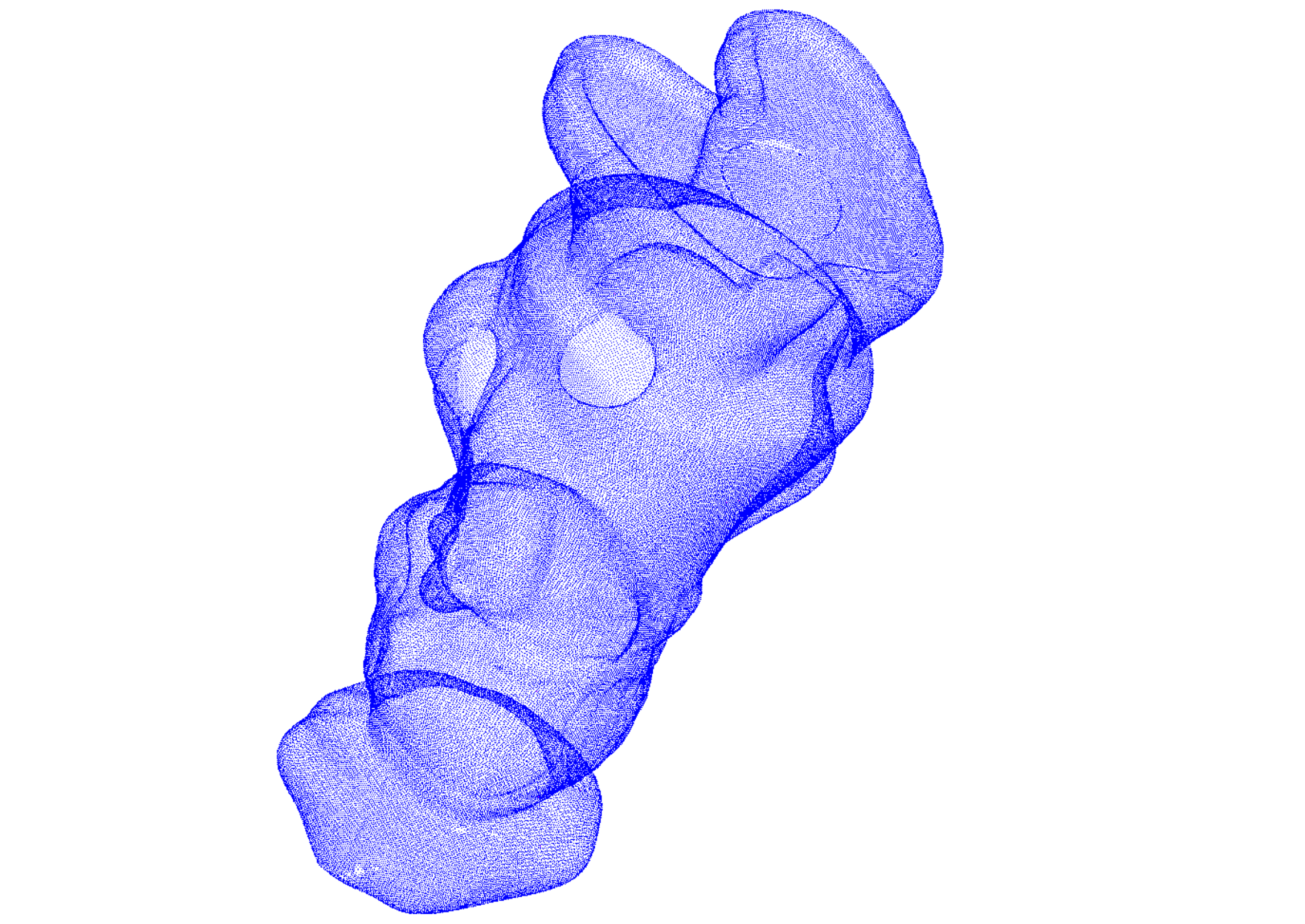}
\end{minipage}
\\
\rotatebox{90}{\footnotesize{chicken}}\,
& &
\begin{minipage}[t]{0.22\linewidth}
\centering
\includegraphics[width=1\linewidth]{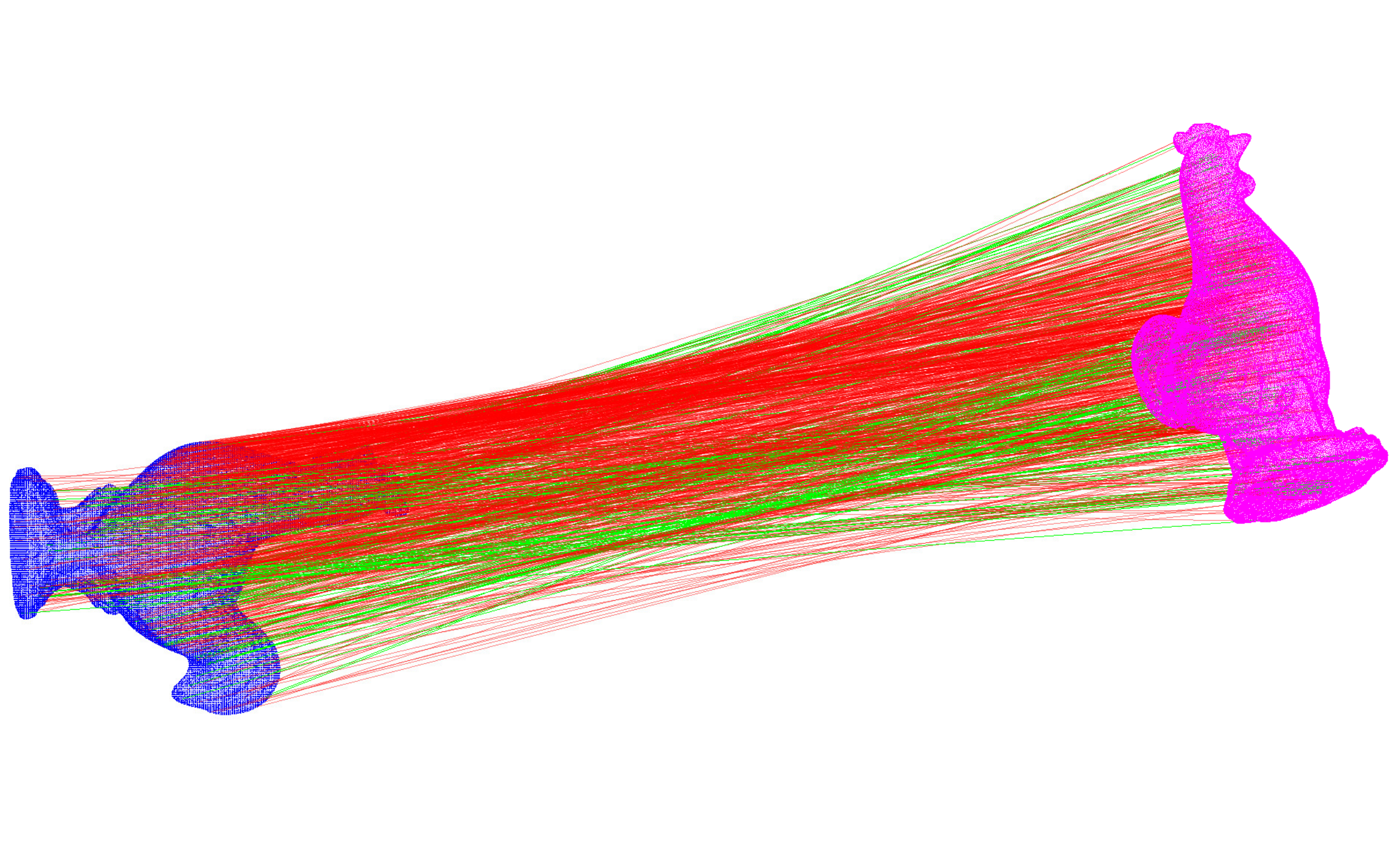}
\end{minipage}
&
\begin{minipage}[t]{0.225\linewidth}
\centering
\includegraphics[width=1\linewidth]{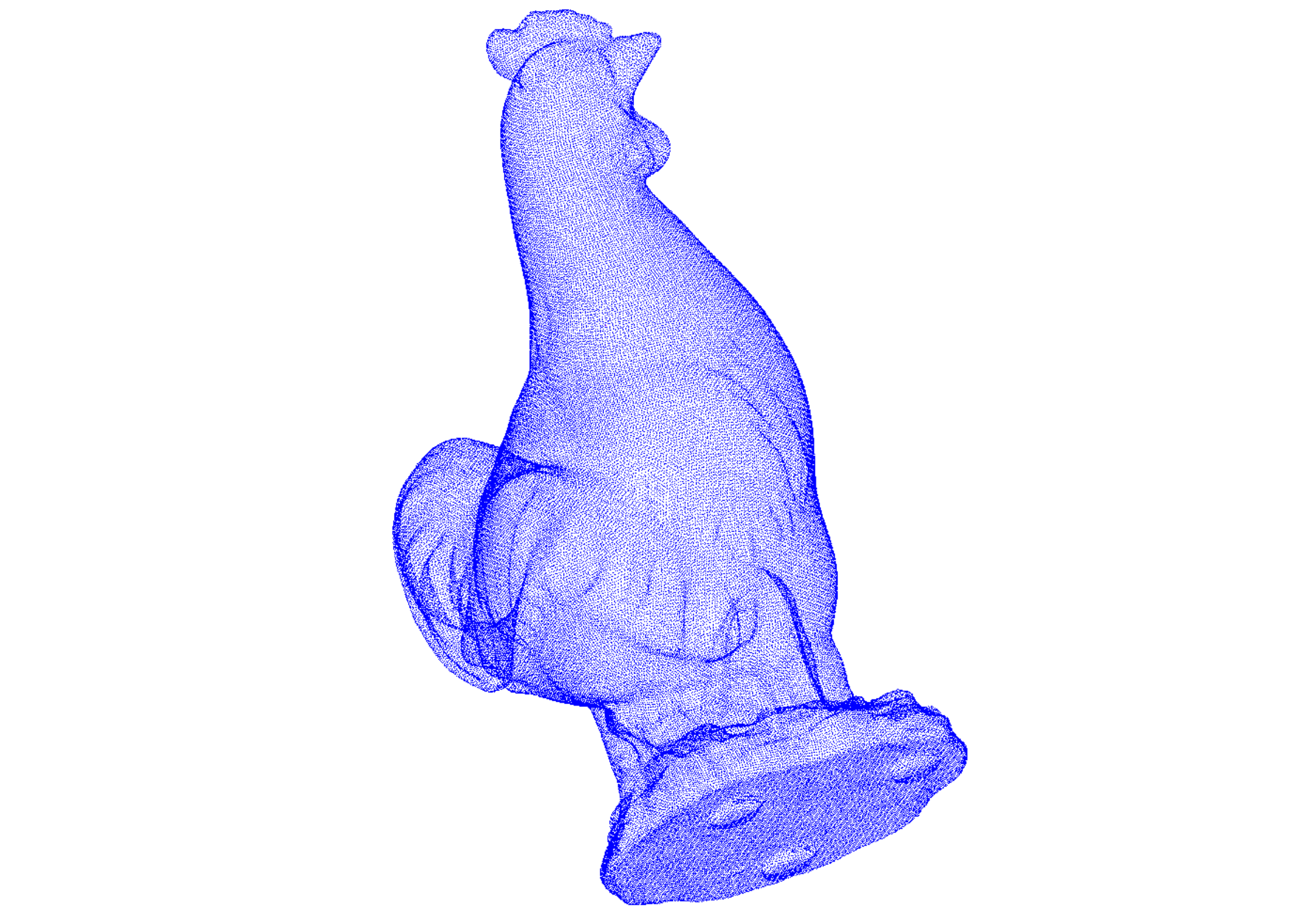}
\end{minipage}
&
\rotatebox{90}{\footnotesize{rhino}}\,
& &
\begin{minipage}[t]{0.22\linewidth}
\centering
\includegraphics[width=1\linewidth]{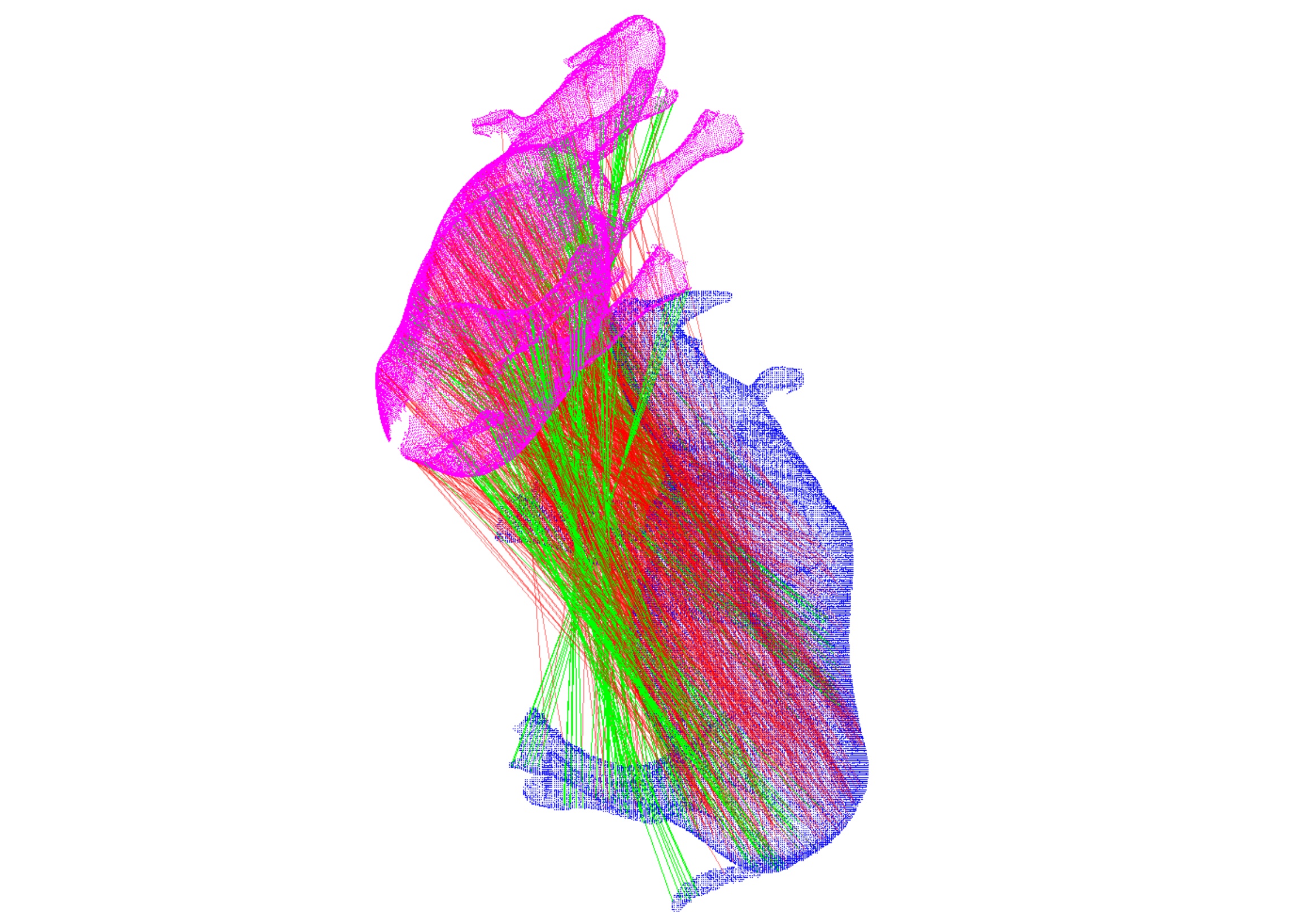}
\end{minipage}
&
\begin{minipage}[t]{0.23\linewidth}
\centering
\includegraphics[width=1\linewidth]{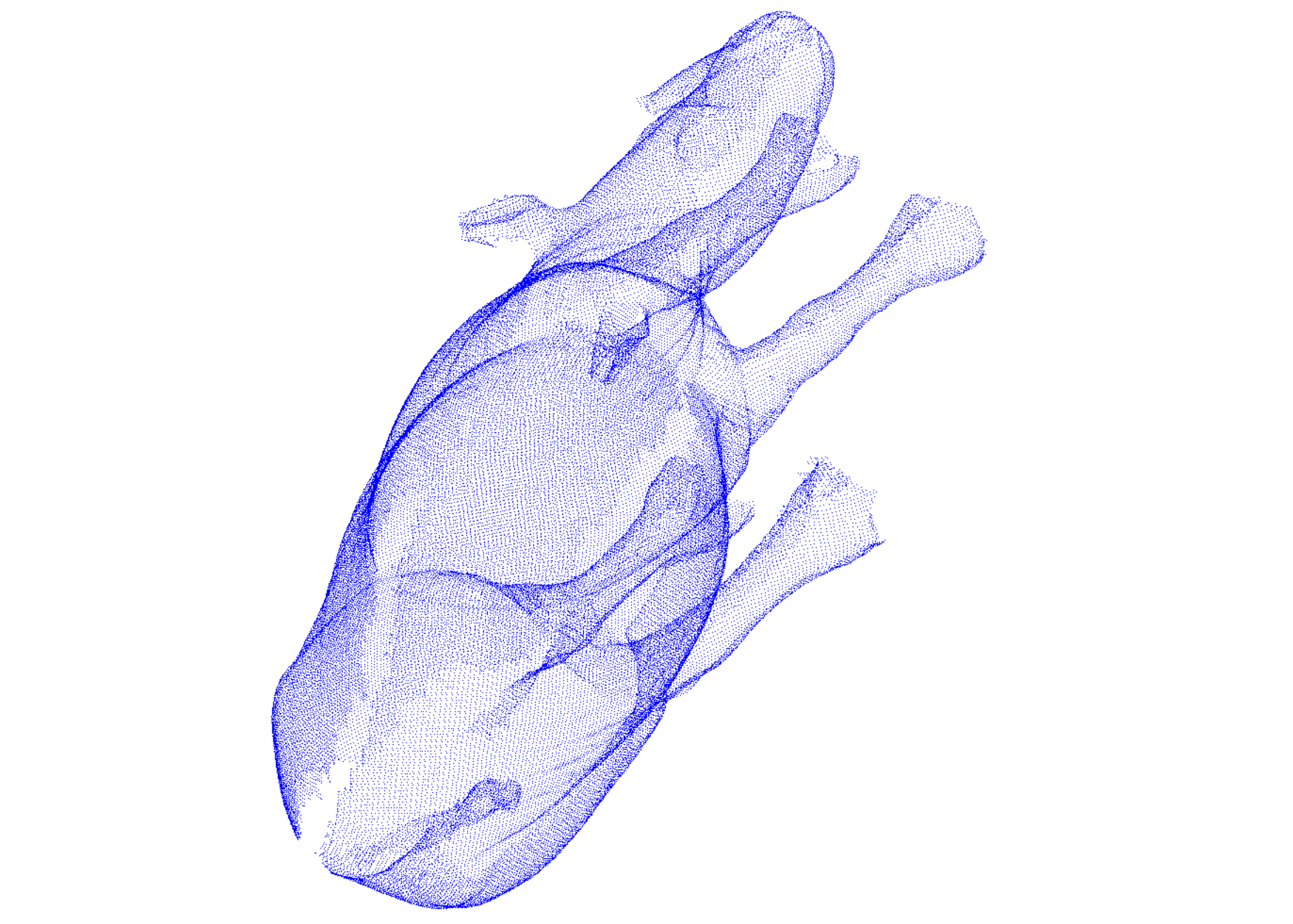}
\end{minipage}
\\
\rotatebox{90}{\footnotesize{parasauro}}\,
& &
\begin{minipage}[t]{0.22\linewidth}
\centering
\includegraphics[width=1\linewidth]{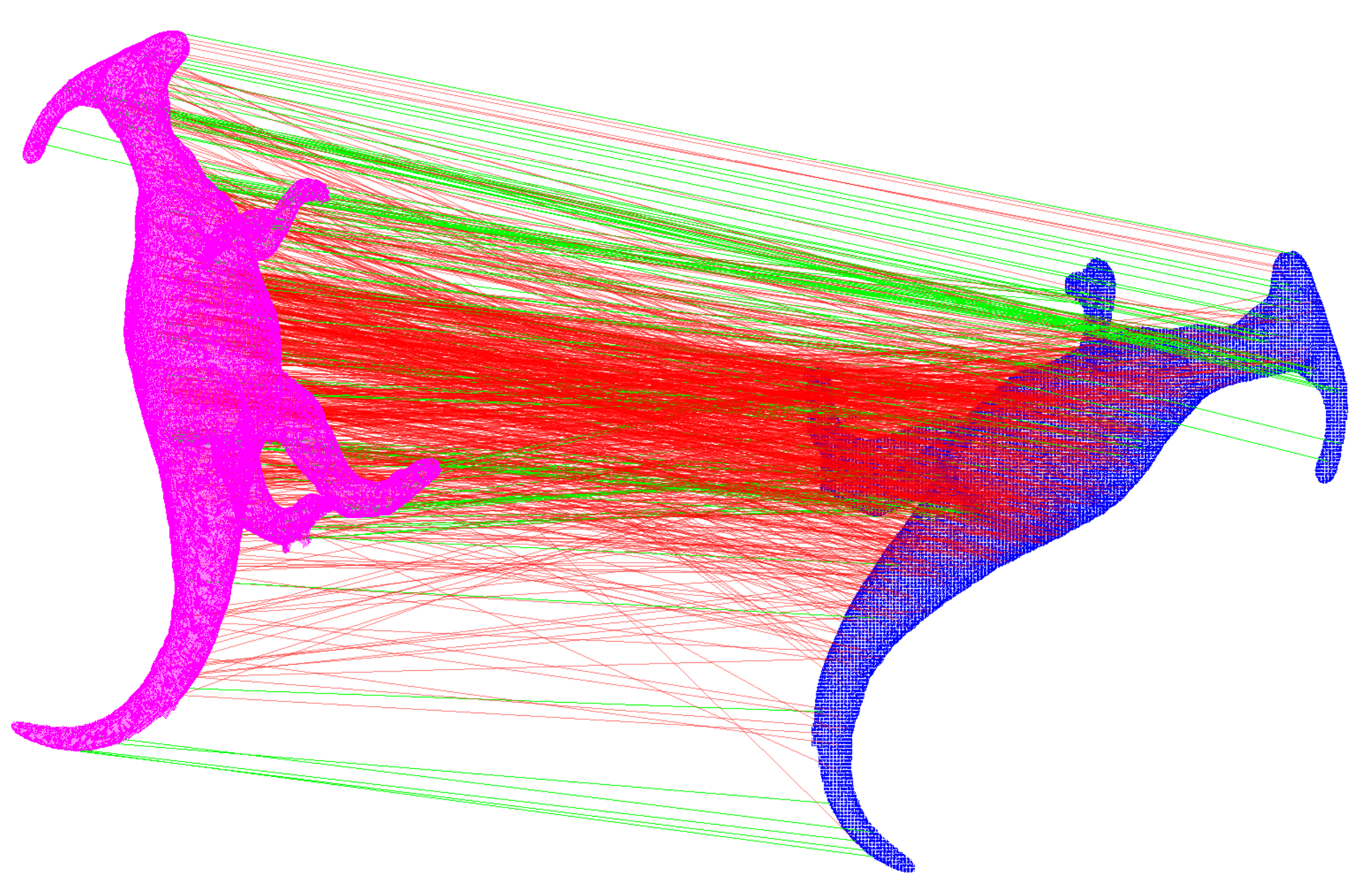}
\end{minipage}
&
\begin{minipage}[t]{0.23\linewidth}
\centering
\includegraphics[width=1\linewidth]{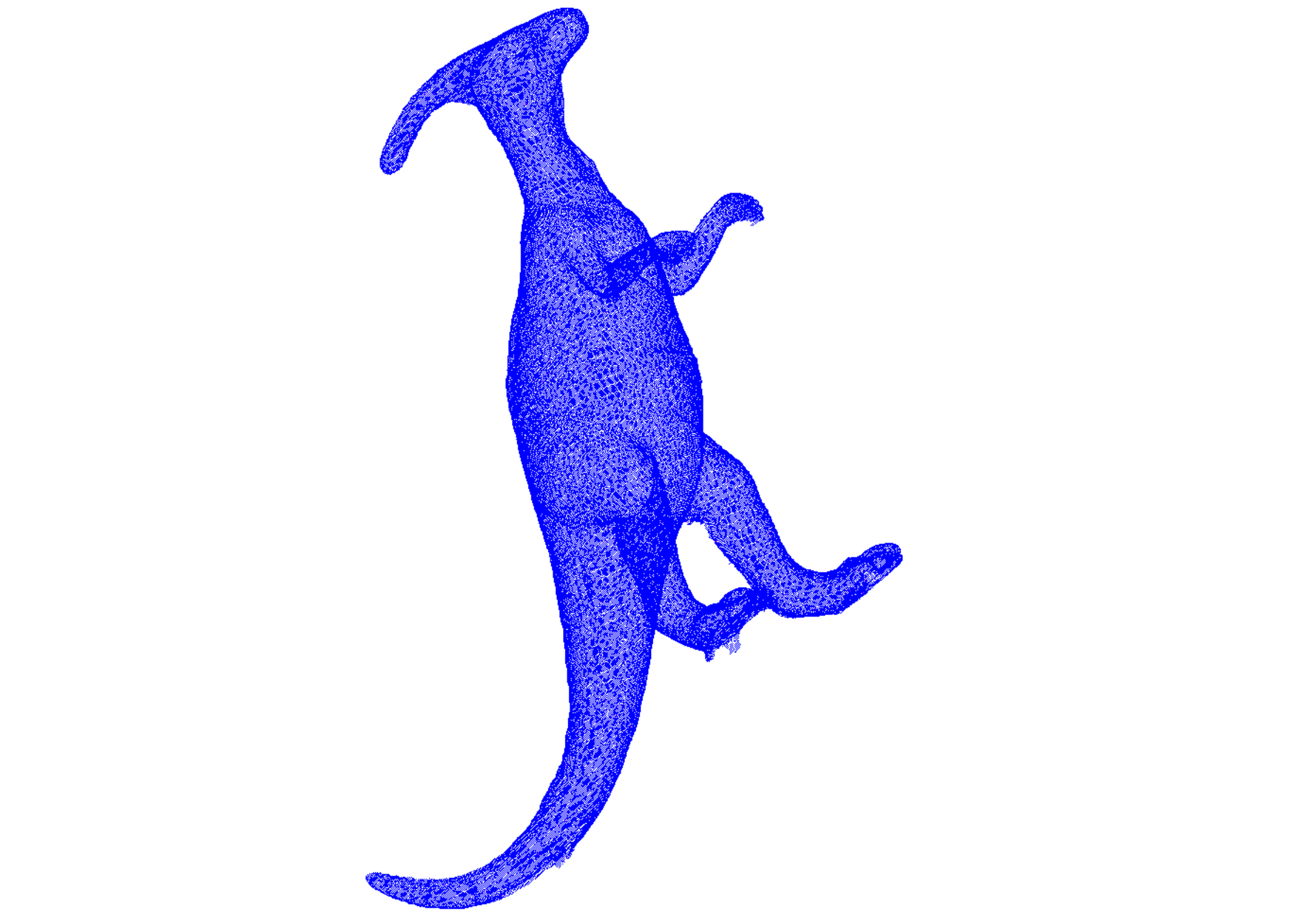}
\end{minipage}
&
\rotatebox{90}{\footnotesize{t-rex}}\,
& &
\begin{minipage}[t]{0.22\linewidth}
\centering
\includegraphics[width=1\linewidth]{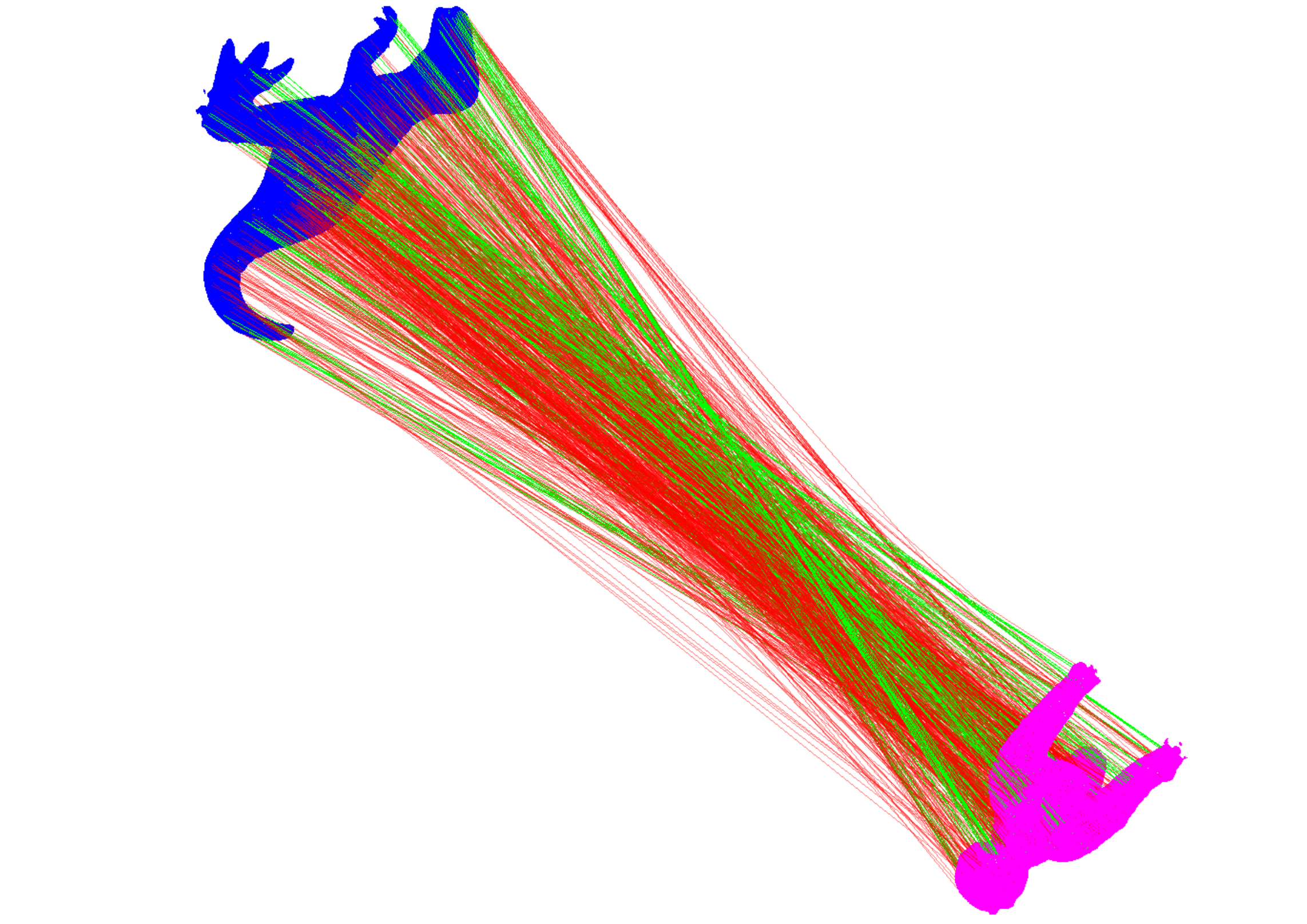}
\end{minipage}
&
\begin{minipage}[t]{0.23\linewidth}
\centering
\includegraphics[width=1\linewidth]{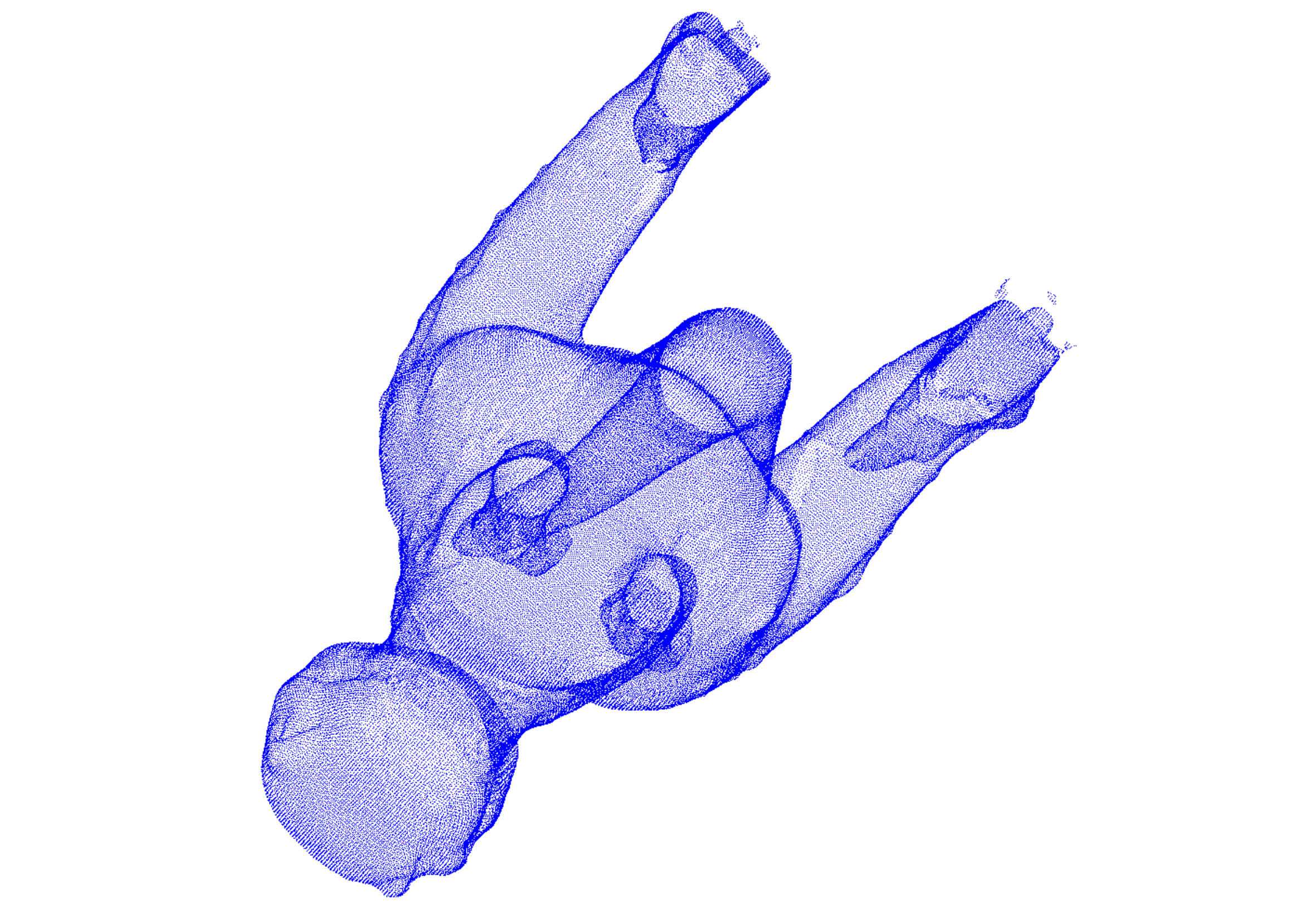}
\end{minipage}
\\
\rotatebox{90}{\footnotesize{city}}\,
& &
\begin{minipage}[t]{0.22\linewidth}
\centering
\includegraphics[width=1\linewidth]{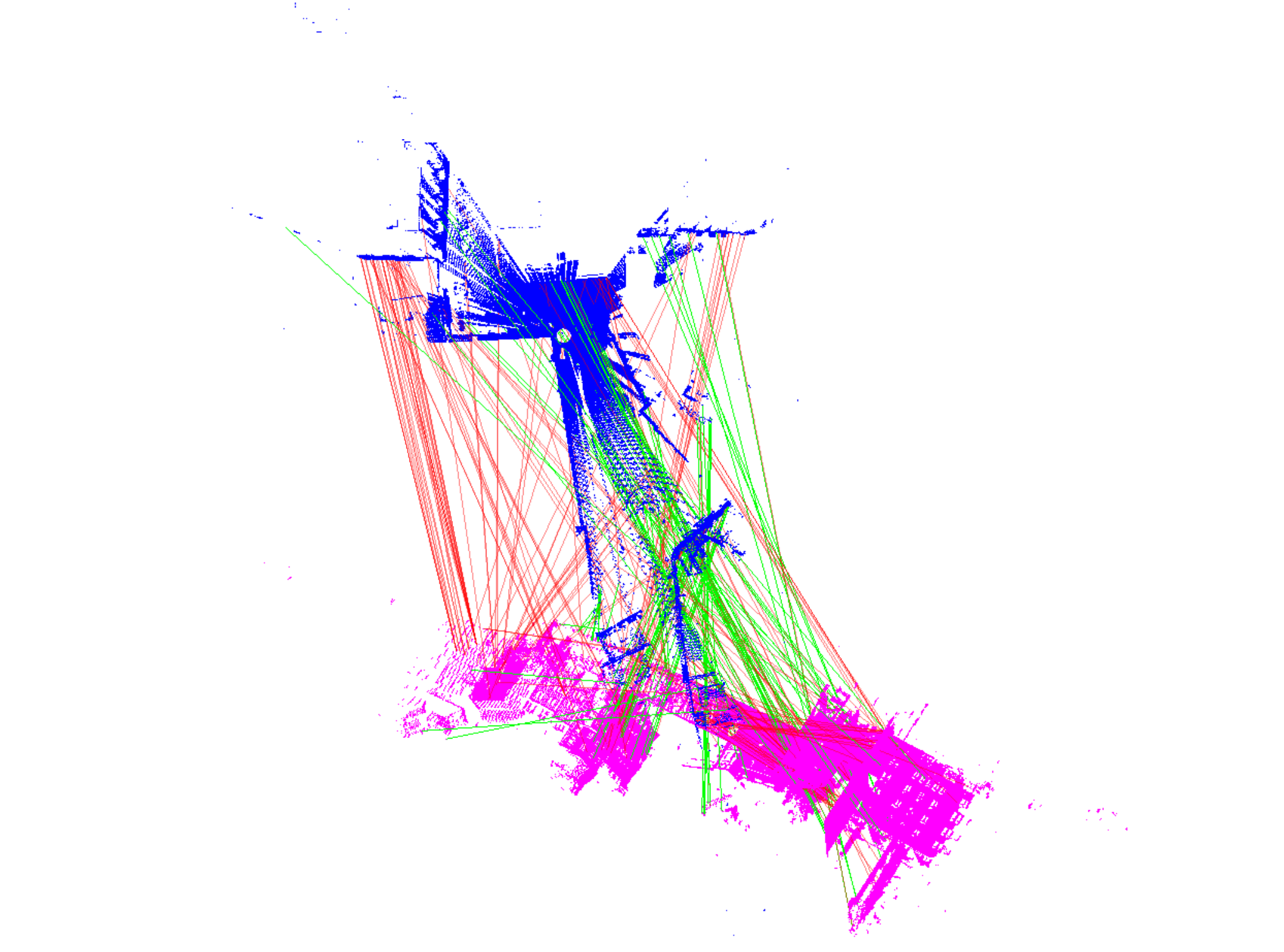}
\end{minipage}
&
\begin{minipage}[t]{0.23\linewidth}
\centering
\includegraphics[width=1\linewidth]{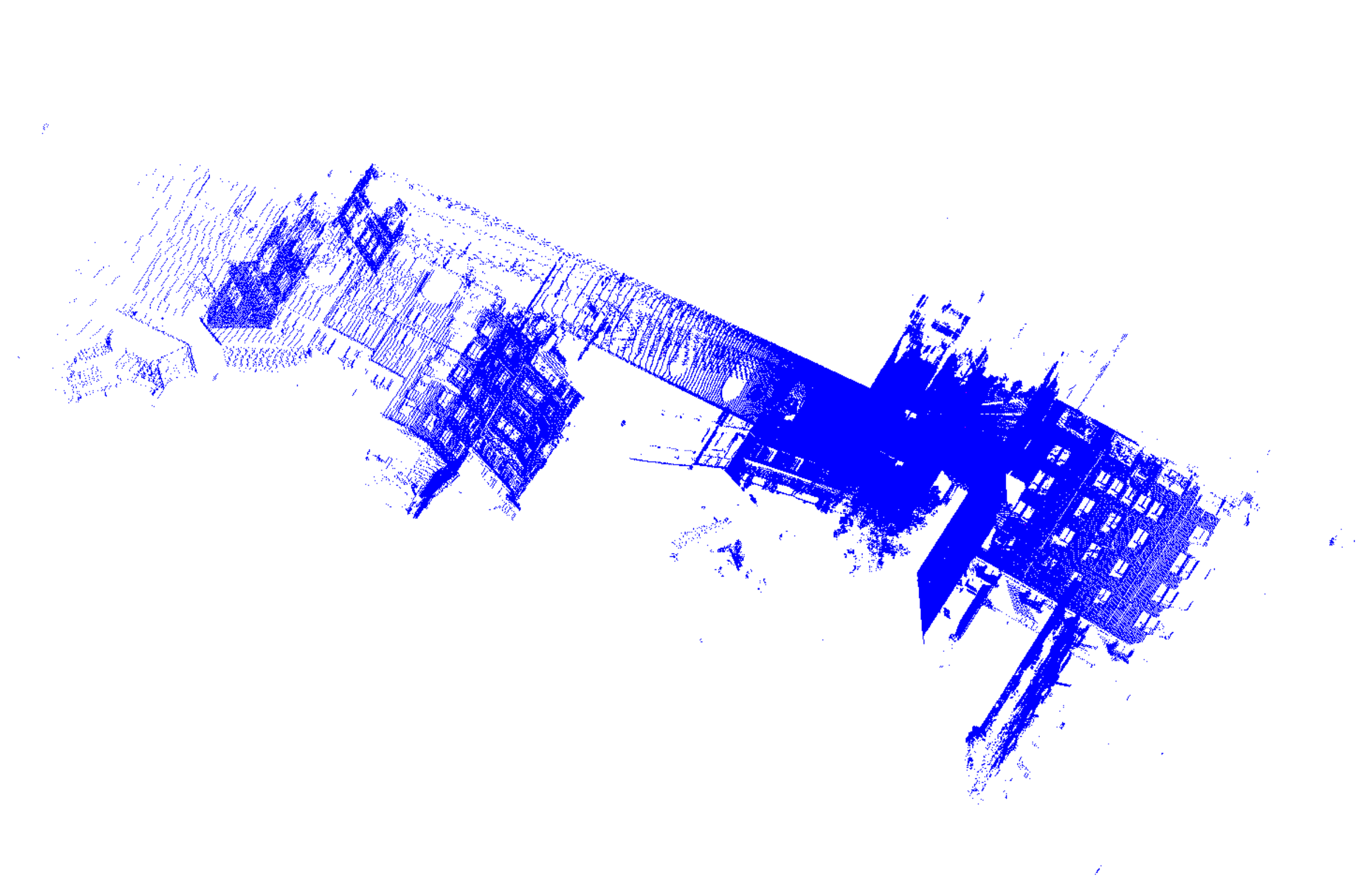}
\end{minipage}
&
\rotatebox{90}{\footnotesize{castle}}\,
& &
\begin{minipage}[t]{0.22\linewidth}
\centering
\includegraphics[width=1\linewidth]{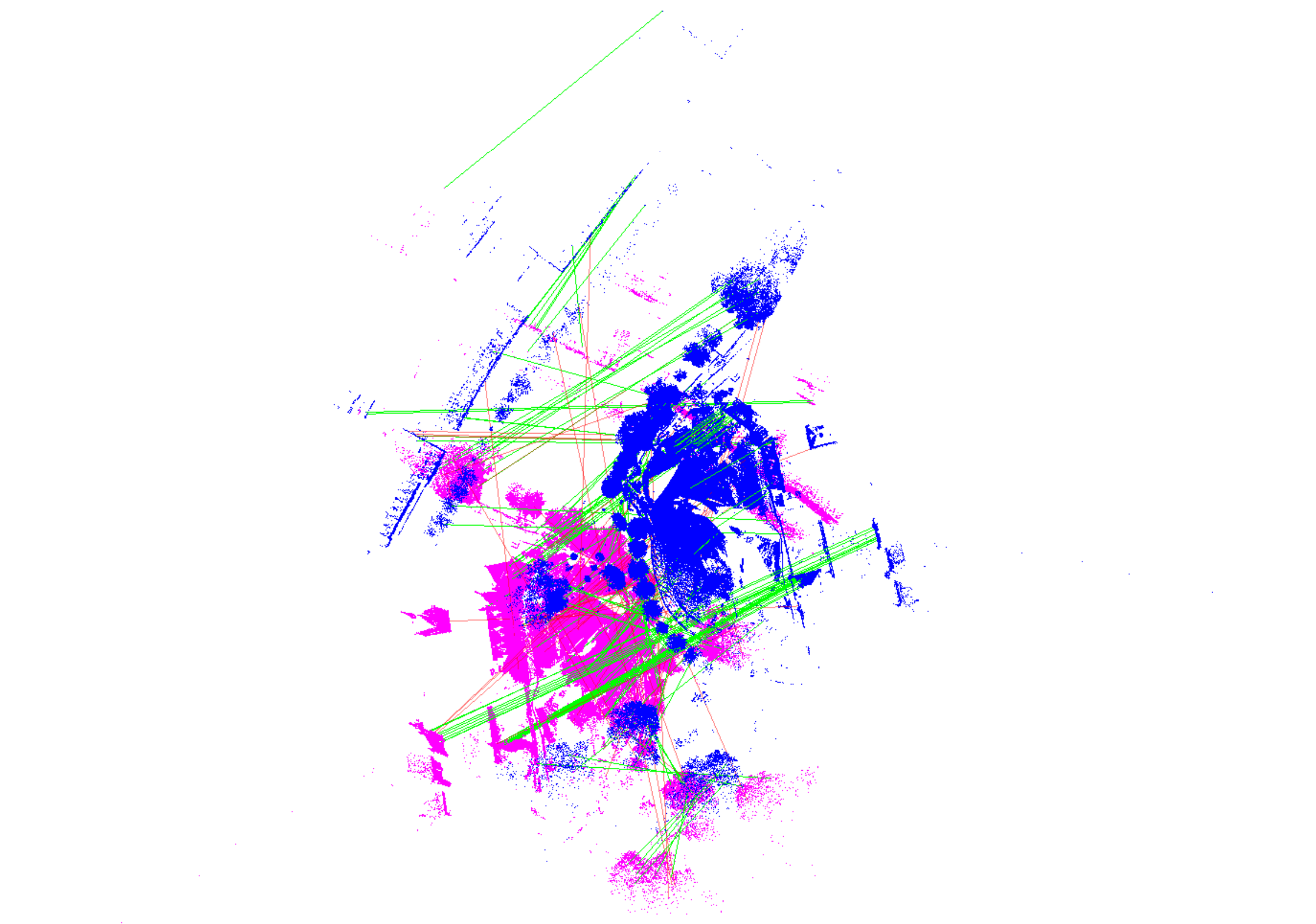}
\end{minipage}
&
\begin{minipage}[t]{0.23\linewidth}
\centering
\includegraphics[width=1\linewidth]{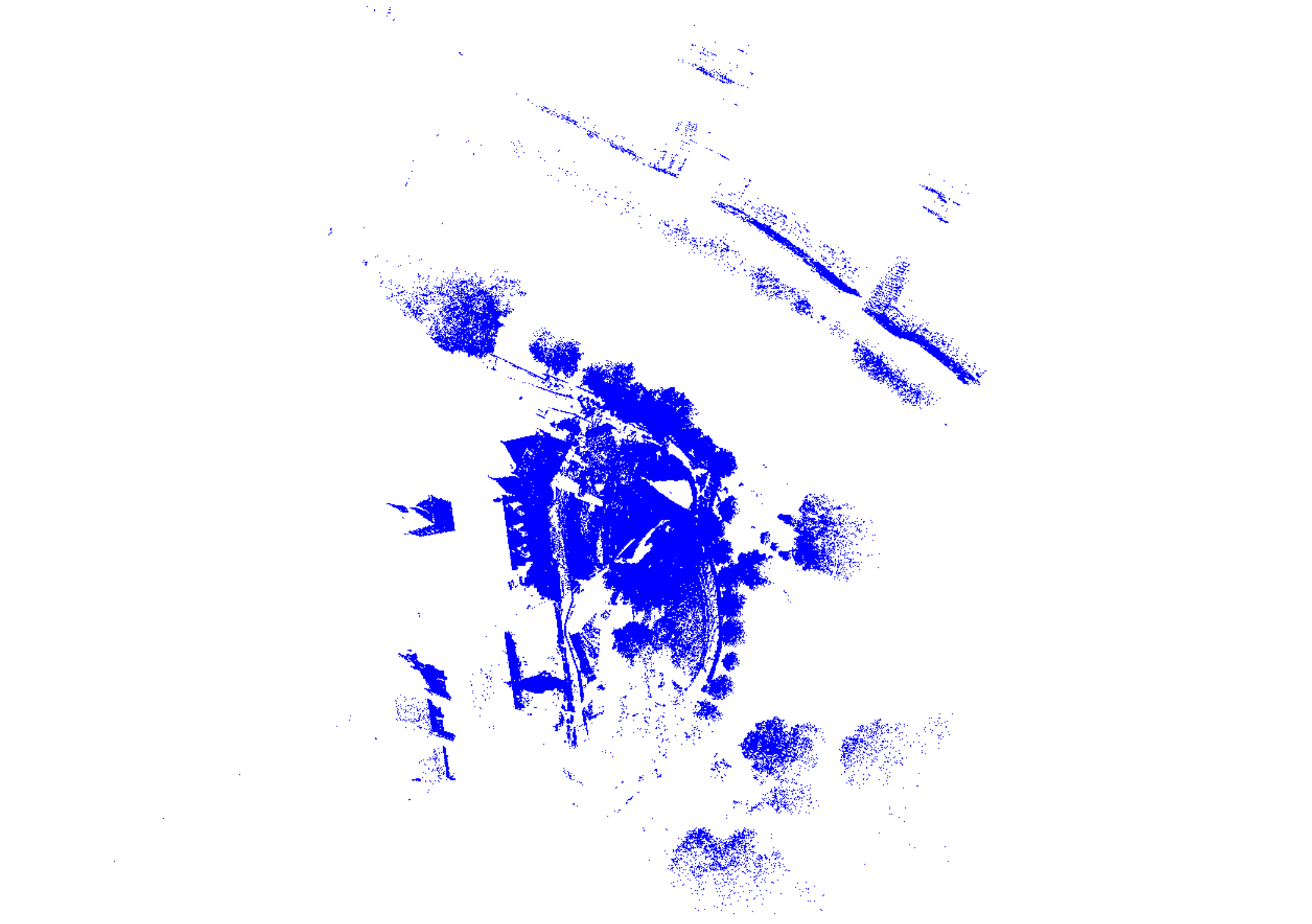}
\end{minipage}
\end{tabular}

\caption{Qualitative results of realistic normal registration over the 10 point clouds using our solver VOCRA. The first column shows the point correspondences matched by FPFH~\cite{rusu2009fast} where inliers are in green while outliers are in red. The second column displays the registration result using the pose (transformation) estimated by VOCRA.}
\label{qualit-normal}
\vspace{-2mm}
\end{figure*}

First, we compute the centroids w.r.t. the two point sets $\mathcal{P}=\{\boldsymbol{q}_i\}_{i=1}^{|\mathcal{I}^{\star}|}$ and $\mathcal{Q}=\{\boldsymbol{q}_i\}_{i=1}^{|\mathcal{I}^{\star}|}$, respectively, as:

\begin{equation}
\boldsymbol{\bar{p}}=\frac{1}{N} \sum^{|\mathcal{I}^{\star}|}_{i=1}\boldsymbol{p}_i, \,\,\,\boldsymbol{\bar{q}}=\frac{1}{N}\sum^{|\mathcal{I}^{\star}|}_{i=1}\boldsymbol{q}_i,
\end{equation}

\noindent where $i\in\mathcal{I}^{\star}$ so that we can derive the translation-free objective function to minimize such that

\begin{equation}\label{obj}
\underset{\boldsymbol{R}\in SO(3), \omega_i\in[0,1]}{\min}\,\sum_{i=1}^{|\mathcal{I}^{\star}|}\omega_i \left\|\boldsymbol{R}(\boldsymbol{p}_i-\boldsymbol{\bar{p}})-(\boldsymbol{q}_i-\boldsymbol{\bar{q}})\right\|,
\end{equation}

\noindent where we adopt the Singular Value Decomposition (SVD) non-minimal solver~\cite{arun1987least} to solve $\boldsymbol{R}$ efficiently in closed form. 

Subsequently, we can tailor the SVD solver to GNC-TB in order to robustly reject the outliers in the inlier set candidate $\mathcal{I}^{\star}$, which can be operated as in Algorithm~\ref{solveGNCTB}.

First, we set all the initial weights $\omega_i$ to be 1 and the controlling parameter $\mu$ to be large (e.g. $\mu=100$). Then, we start the GNC iterations. In each iteration, with current $\omega_i$, we first estimate the rotation $\boldsymbol{\hat{R}}$ based on objective~\eqref{obj} using SVD, and then update the weights $\omega_i$ for the next iteration according to the residual errors computed. After that, we decrease $\mu$ by $\mu=\mu/1.2$. In this manner, we stop iteration once the residual errors (or the objective) converge or $\mu$ is smaller than 1.

\begin{figure*}[t]
\centering
\setlength\tabcolsep{0pt}
\addtolength{\tabcolsep}{0pt}
\begin{tabular}{c|cccc|ccc}

\quad & \,\, & \footnotesize{FPFH Correspondences} & \footnotesize{Registration by VOCRA} & \quad & \,\, & \footnotesize{FPFH Correspondences} & \footnotesize{Registration by VOCRA} \\
\hline 
& & & & & & &
\\
\rotatebox{90}{\footnotesize{bunny}}\,
& &
\begin{minipage}[t]{0.22\linewidth}
\centering
\includegraphics[width=1\linewidth]{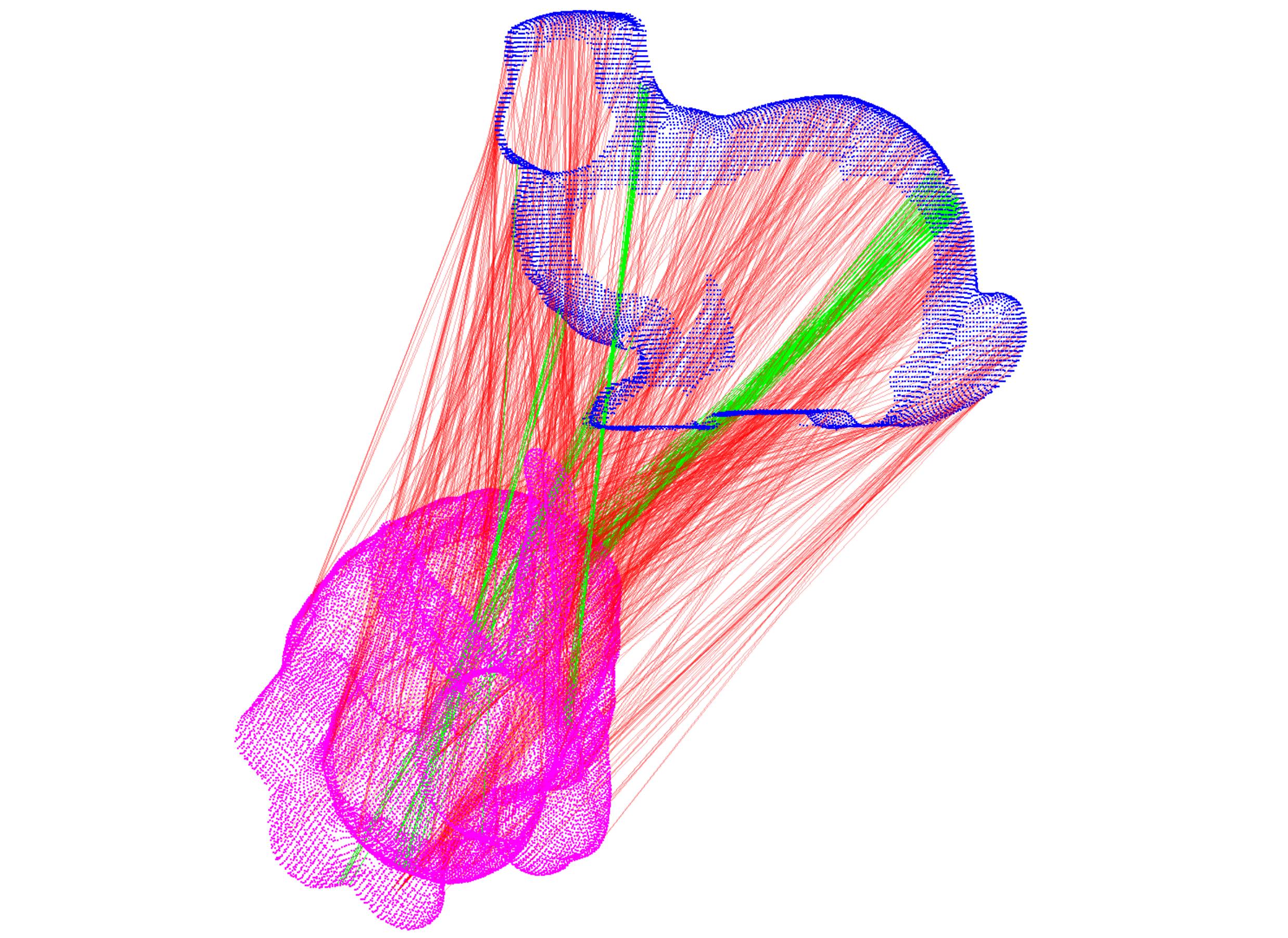}
\end{minipage}
& 
\begin{minipage}[t]{0.22\linewidth}
\centering
\includegraphics[width=1\linewidth]{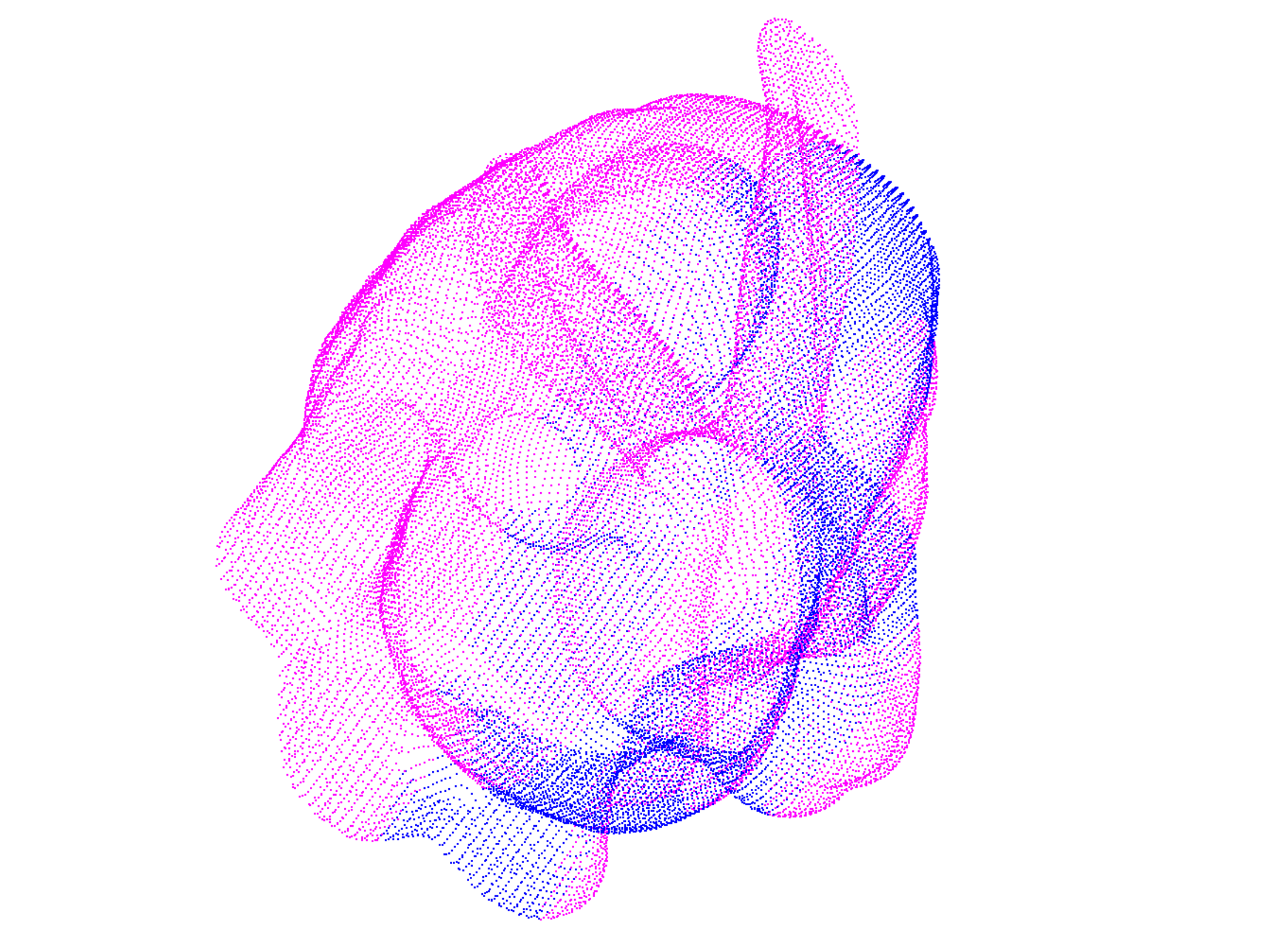}
\end{minipage}
&
\rotatebox{90}{\footnotesize{armadillo}}\,
& &
\begin{minipage}[t]{0.21\linewidth}
\centering
\includegraphics[width=1\linewidth]{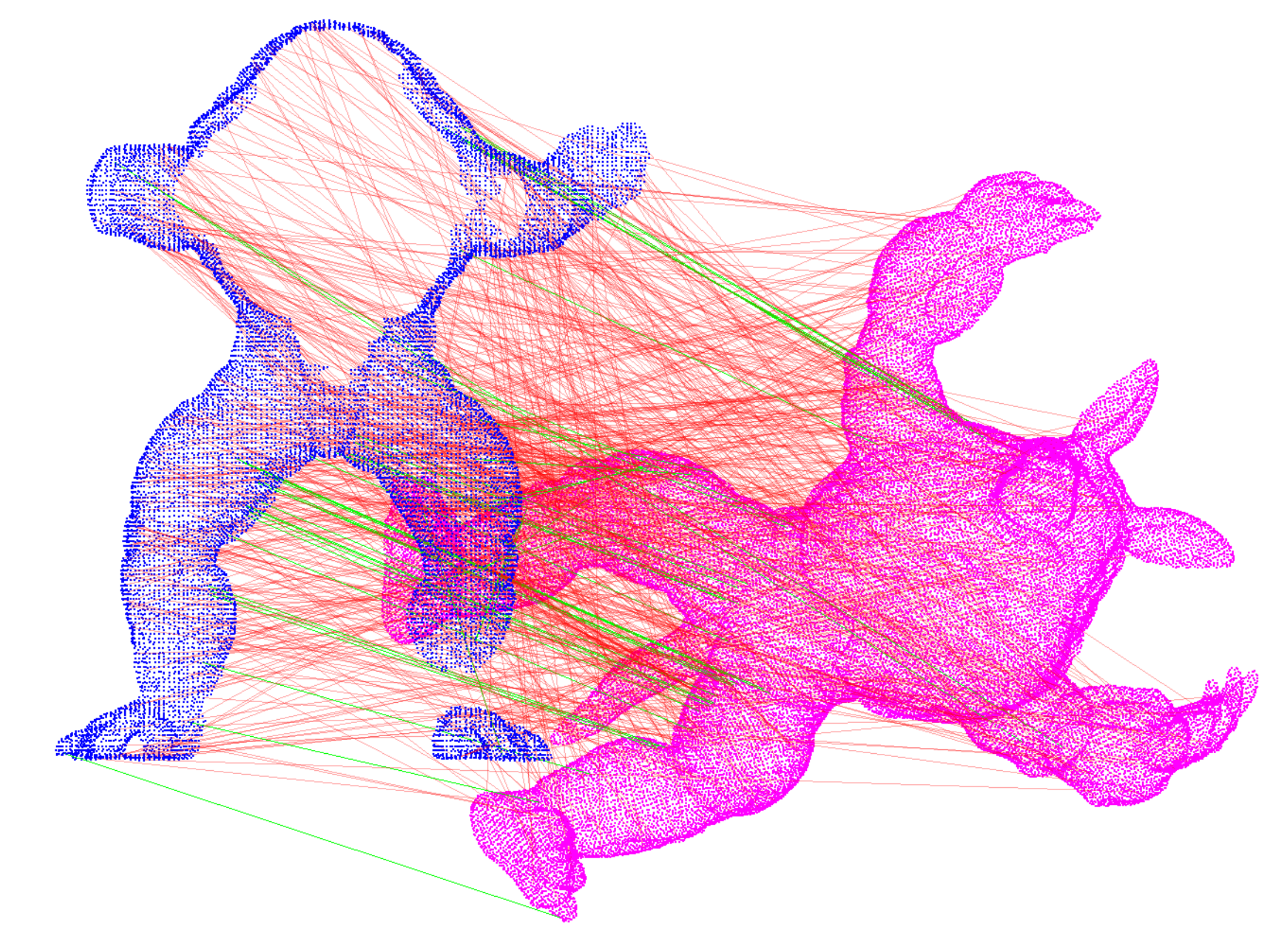}
\end{minipage}
&
\begin{minipage}[t]{0.20\linewidth}
\centering
\includegraphics[width=1\linewidth]{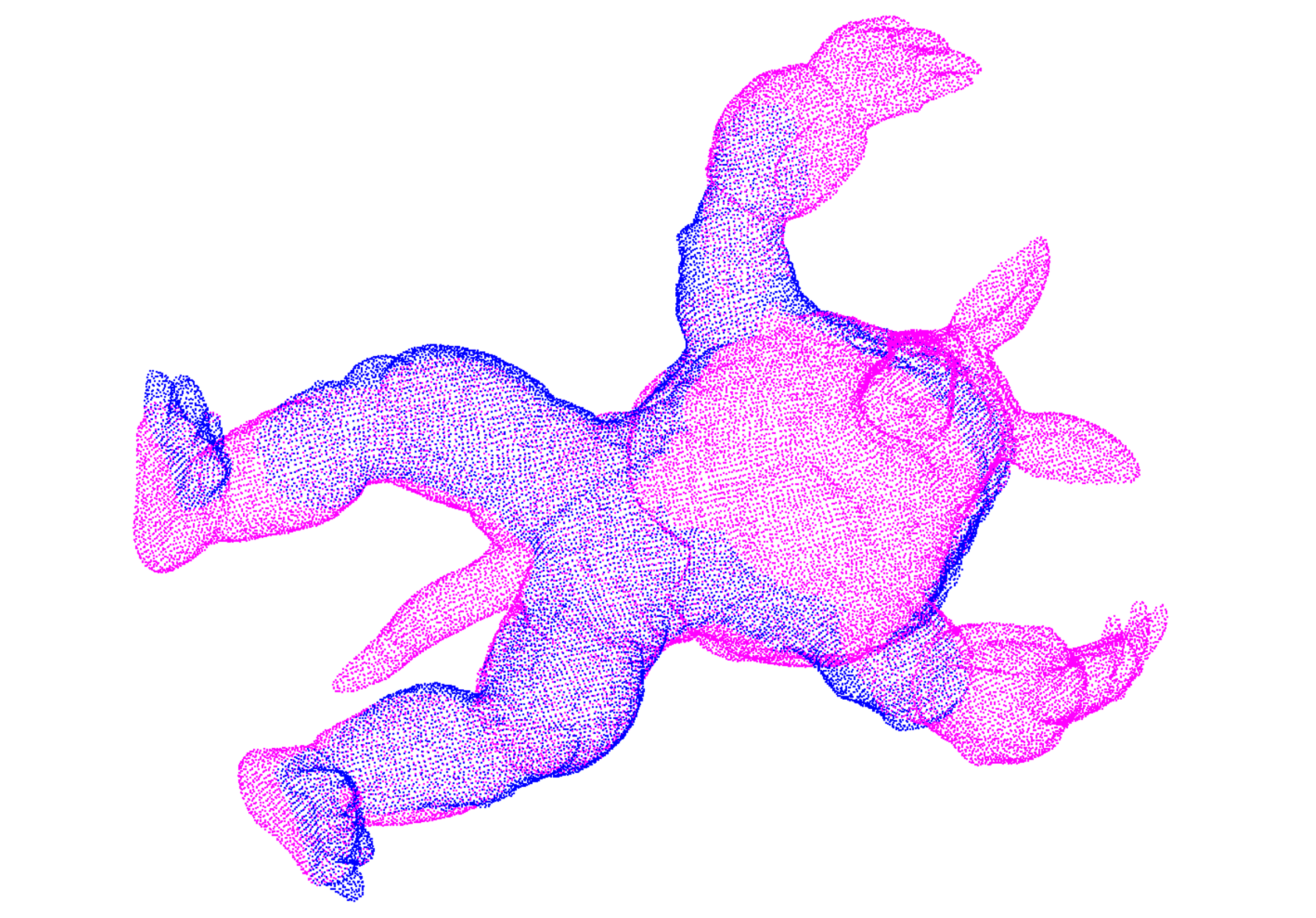}
\end{minipage}
\\
\rotatebox{90}{\footnotesize{dragon}}\,
& &
\begin{minipage}[t]{0.22\linewidth}
\centering
\includegraphics[width=1\linewidth]{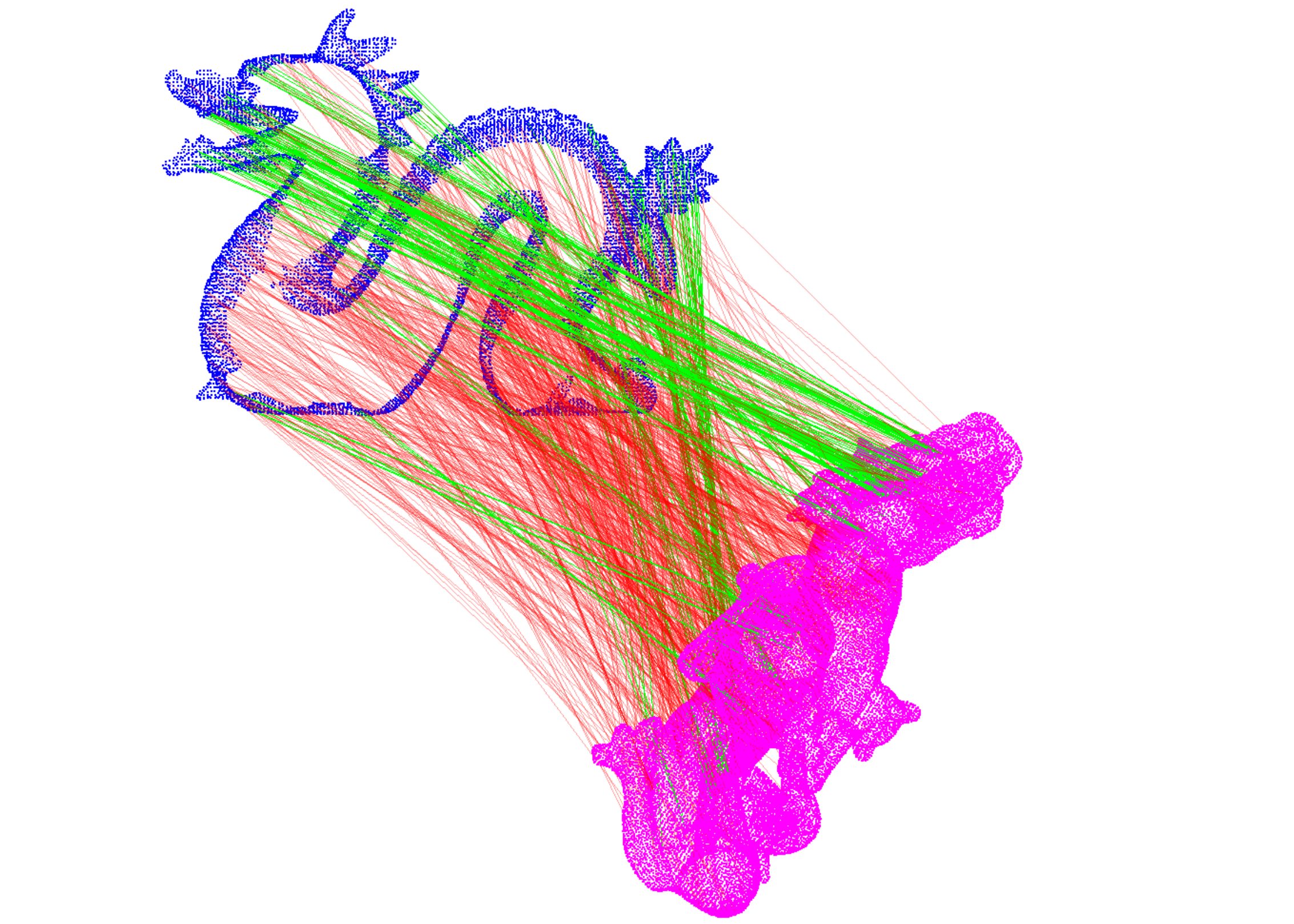}
\end{minipage}
&
\begin{minipage}[t]{0.23\linewidth}
\centering
\includegraphics[width=1\linewidth]{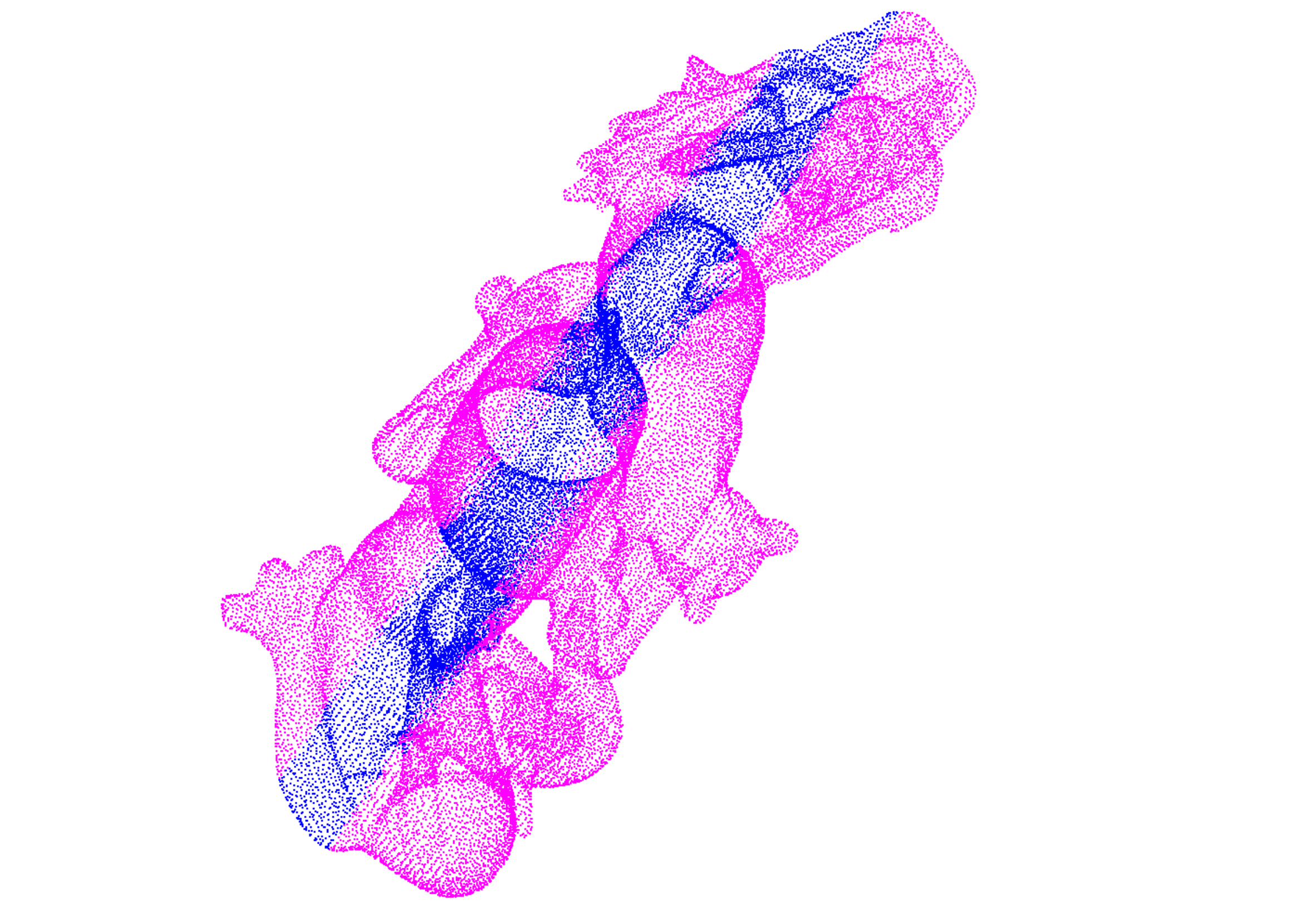}
\end{minipage}
&
\rotatebox{90}{\footnotesize{cheff}}\,
& &
\begin{minipage}[t]{0.22\linewidth}
\centering
\includegraphics[width=1\linewidth]{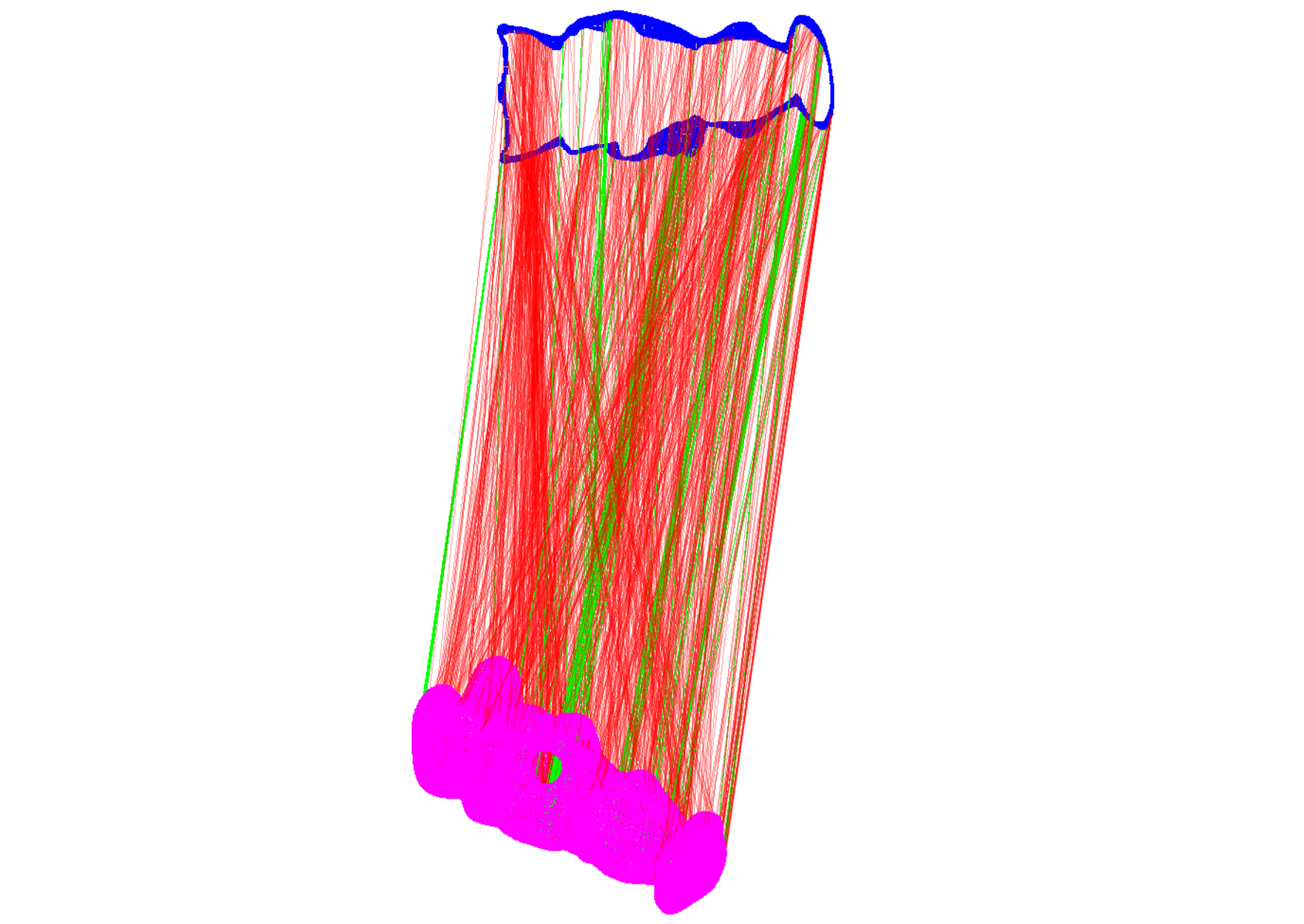}
\end{minipage}
&
\begin{minipage}[t]{0.225\linewidth}
\centering
\includegraphics[width=1\linewidth]{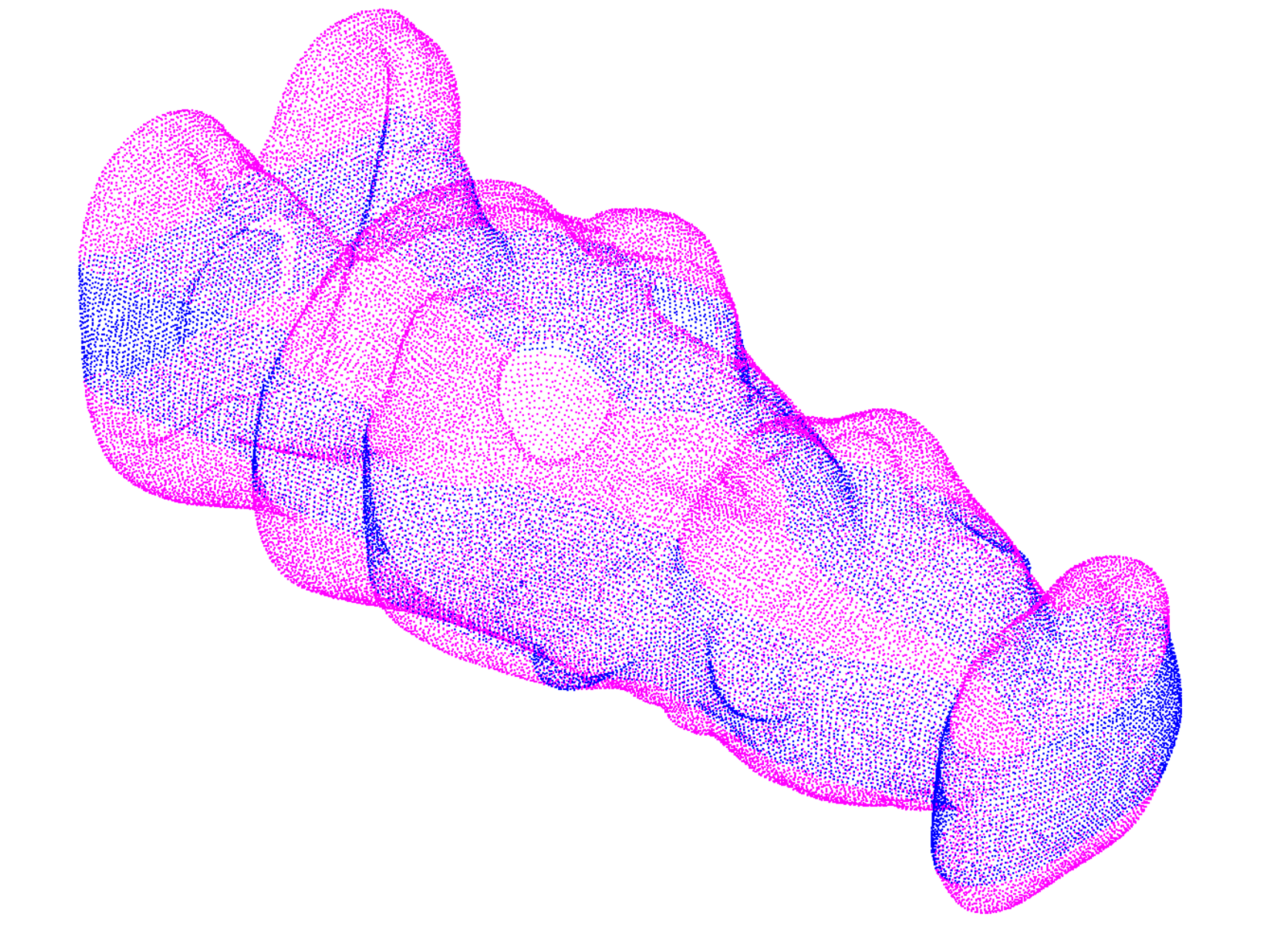}
\end{minipage}
\\
\rotatebox{90}{\footnotesize{chicken}}\,
& &
\begin{minipage}[t]{0.22\linewidth}
\centering
\includegraphics[width=1\linewidth]{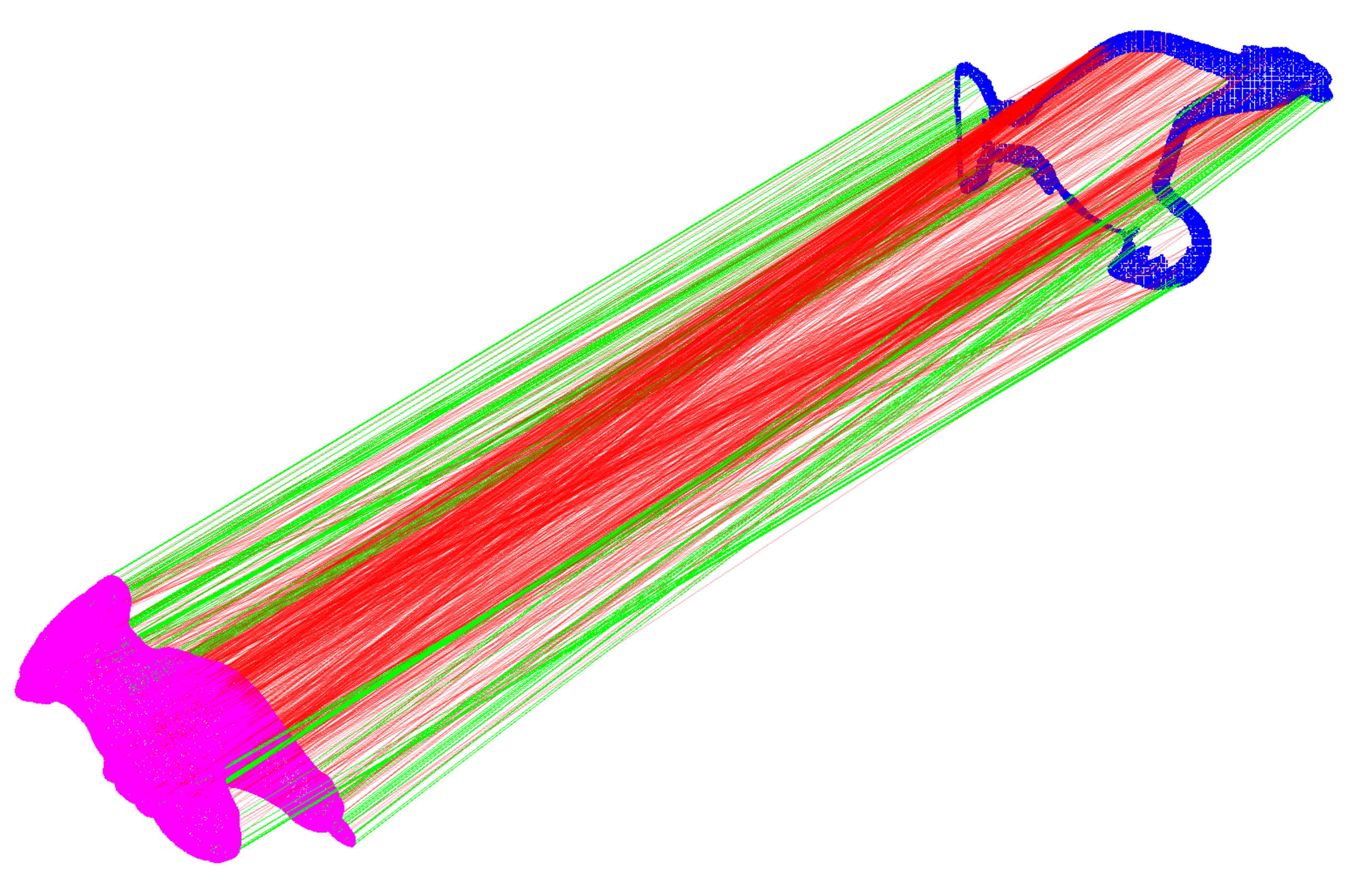}
\end{minipage}
&
\begin{minipage}[t]{0.225\linewidth}
\centering
\includegraphics[width=1\linewidth]{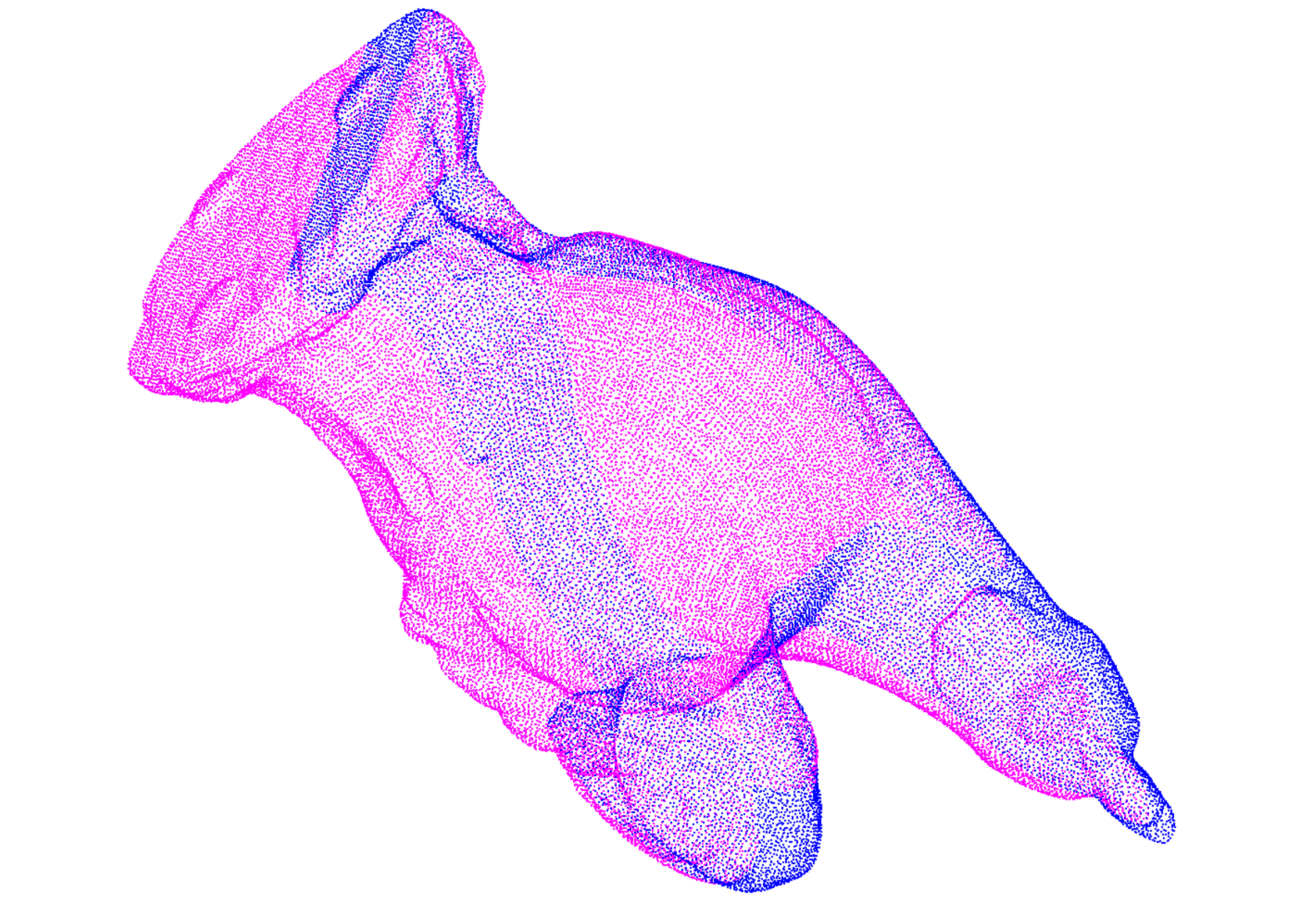}
\end{minipage}
&
\rotatebox{90}{\footnotesize{rhino}}\,
& &
\begin{minipage}[t]{0.22\linewidth}
\centering
\includegraphics[width=1\linewidth]{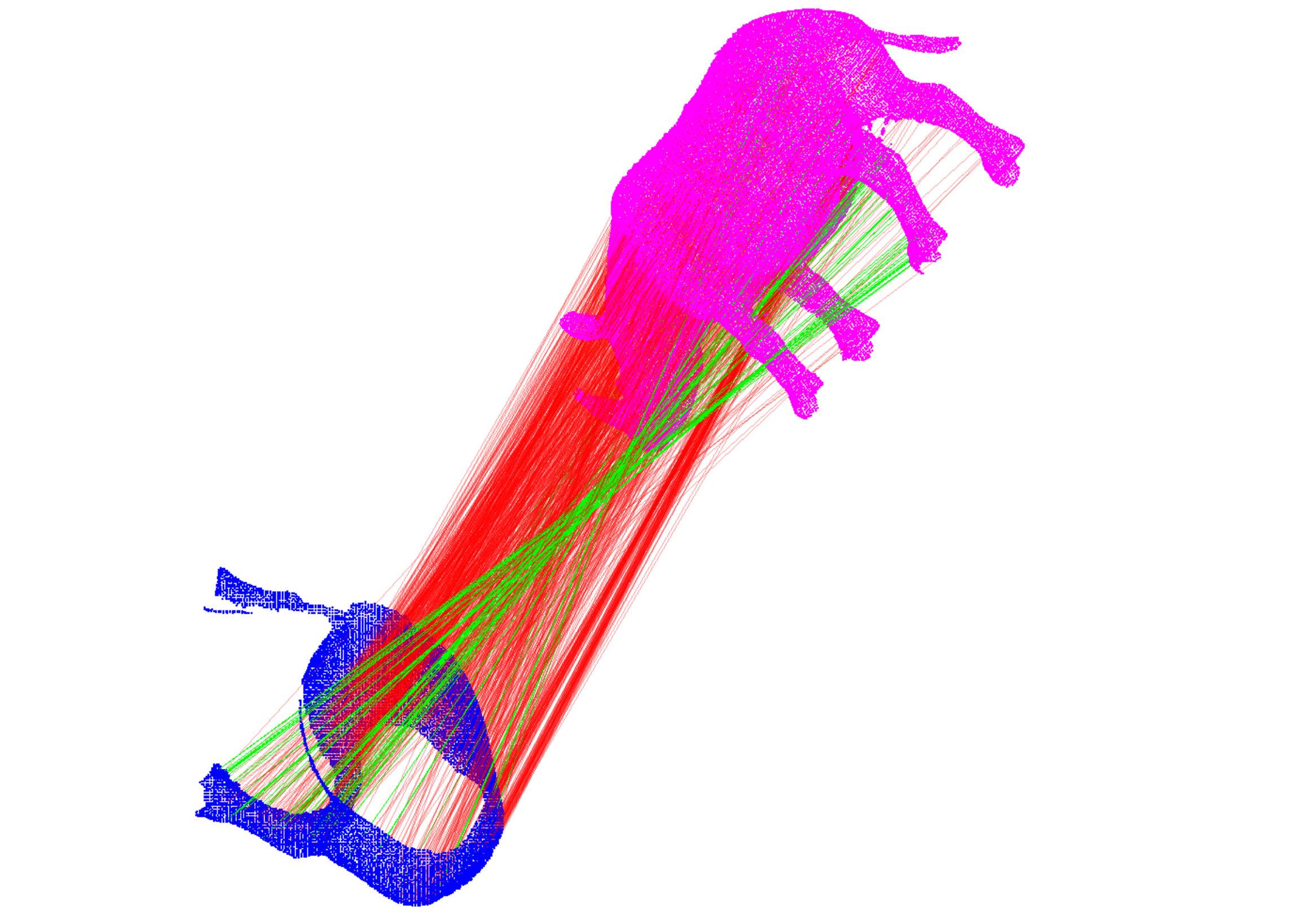}
\end{minipage}
&
\begin{minipage}[t]{0.23\linewidth}
\centering
\includegraphics[width=1\linewidth]{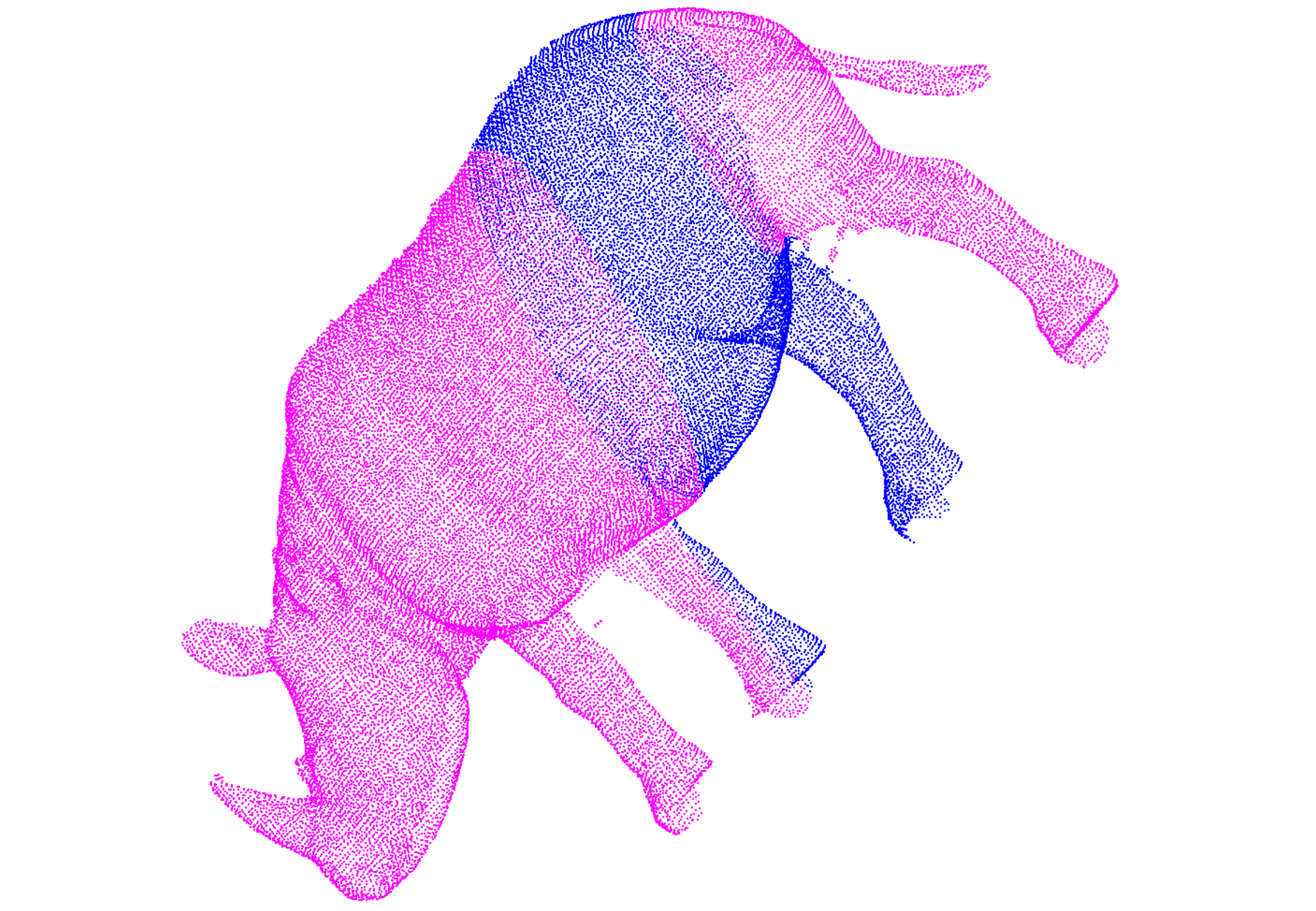}
\end{minipage}
\\
\rotatebox{90}{\footnotesize{parasauro}}\,
& &
\begin{minipage}[t]{0.22\linewidth}
\centering
\includegraphics[width=1\linewidth]{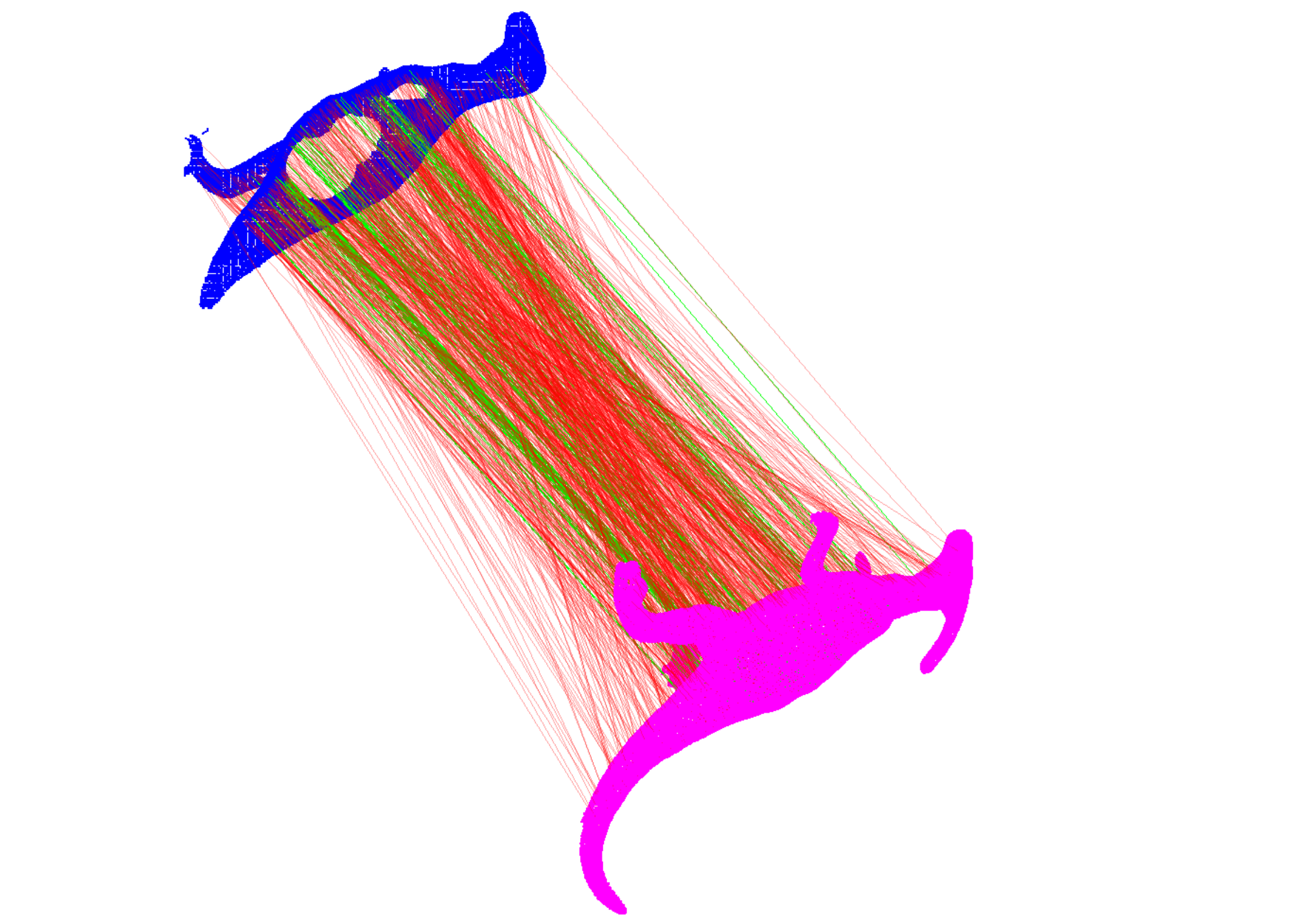}
\end{minipage}
&
\begin{minipage}[t]{0.23\linewidth}
\centering
\includegraphics[width=1\linewidth]{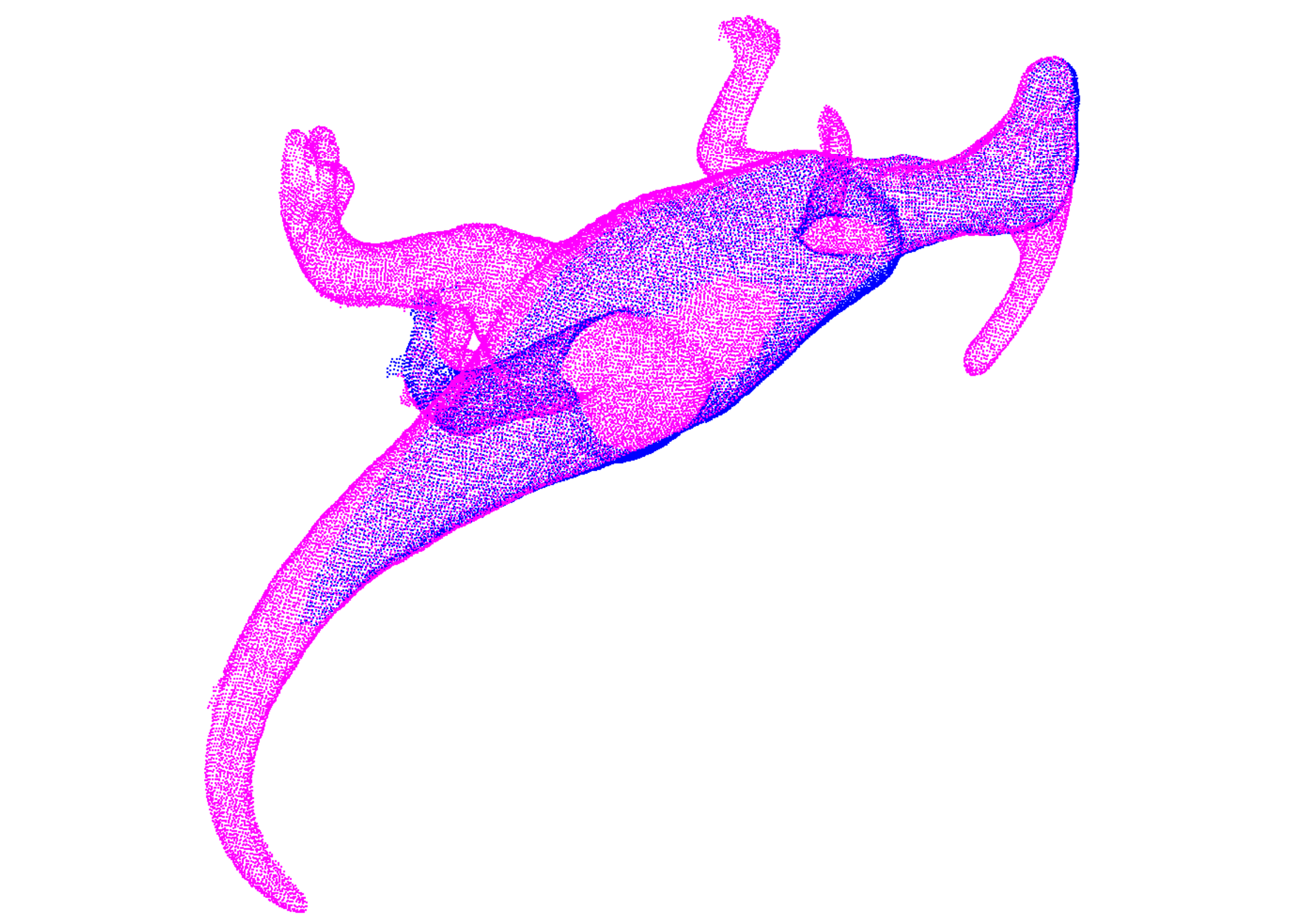}
\end{minipage}
&
\rotatebox{90}{\footnotesize{t-rex}}\,
& &
\begin{minipage}[t]{0.22\linewidth}
\centering
\includegraphics[width=1\linewidth]{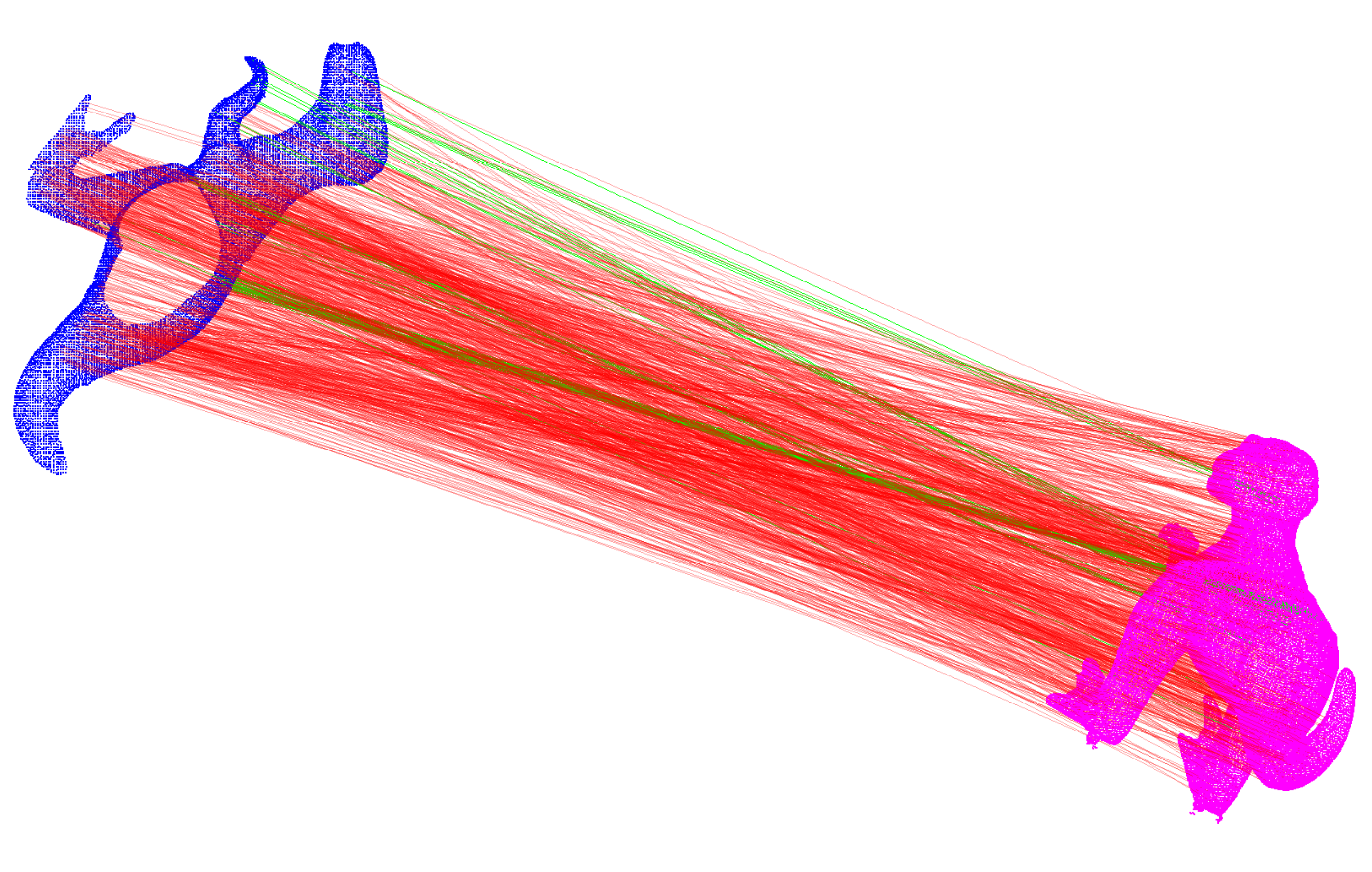}
\end{minipage}
&
\begin{minipage}[t]{0.23\linewidth}
\centering
\includegraphics[width=1\linewidth]{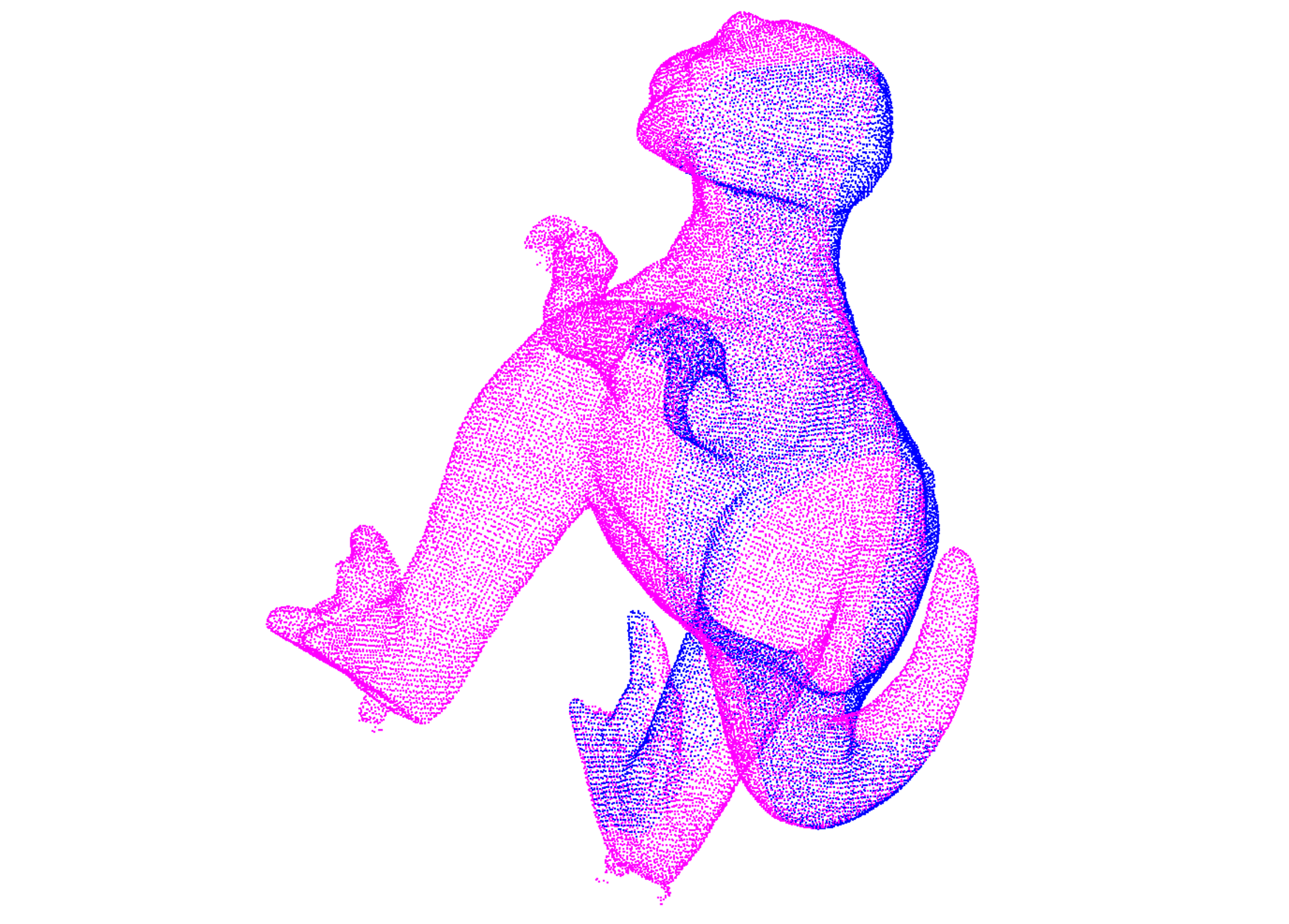}
\end{minipage}
\\
\rotatebox{90}{\footnotesize{city}}\,
& &
\begin{minipage}[t]{0.22\linewidth}
\centering
\includegraphics[width=1\linewidth]{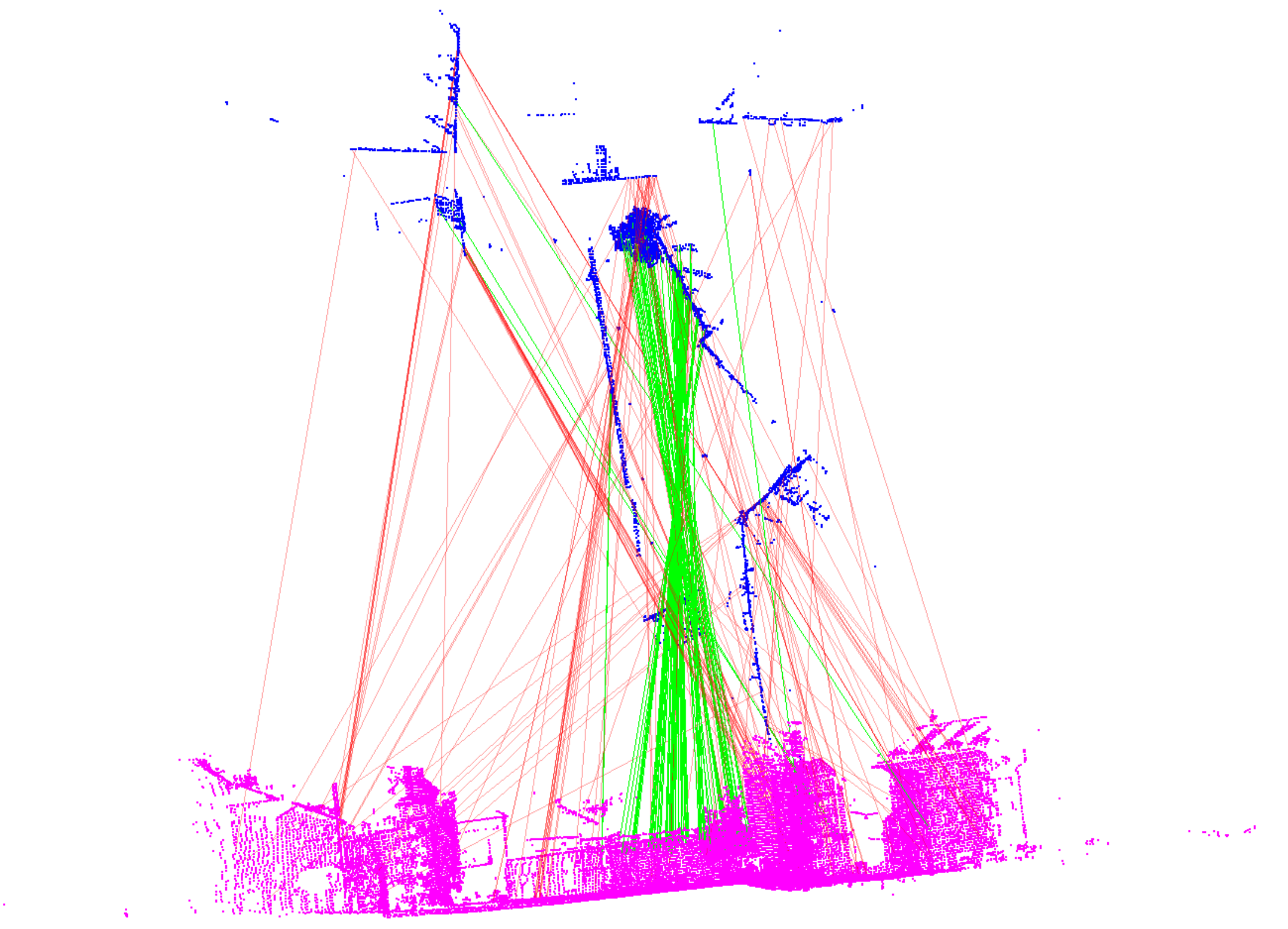}
\end{minipage}
&
\begin{minipage}[t]{0.23\linewidth}
\centering
\includegraphics[width=1\linewidth]{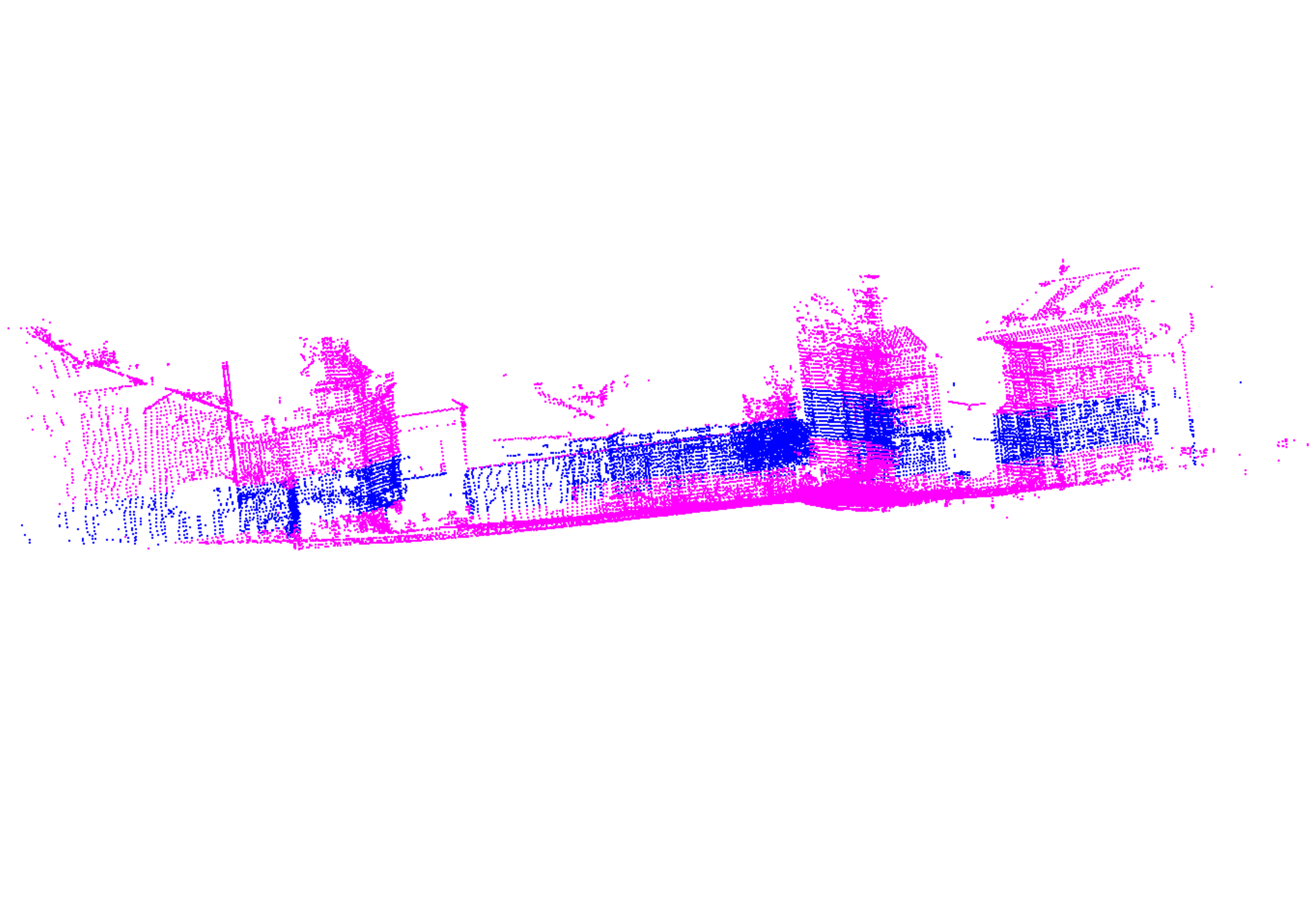}
\end{minipage}
&
\rotatebox{90}{\footnotesize{castle}}\,
& &
\begin{minipage}[t]{0.22\linewidth}
\centering
\includegraphics[width=1\linewidth]{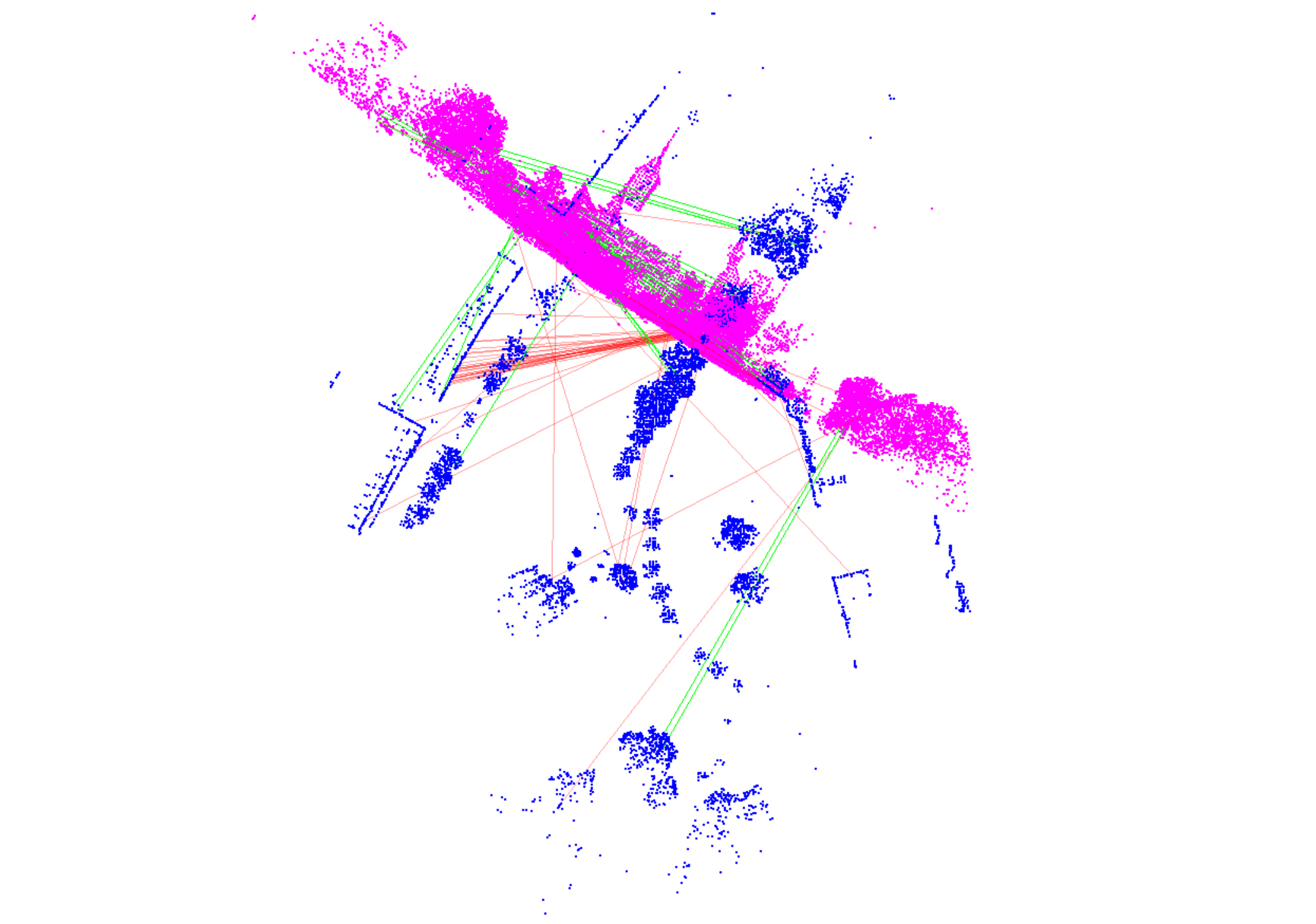}
\end{minipage}
&
\begin{minipage}[t]{0.23\linewidth}
\centering
\includegraphics[width=1\linewidth]{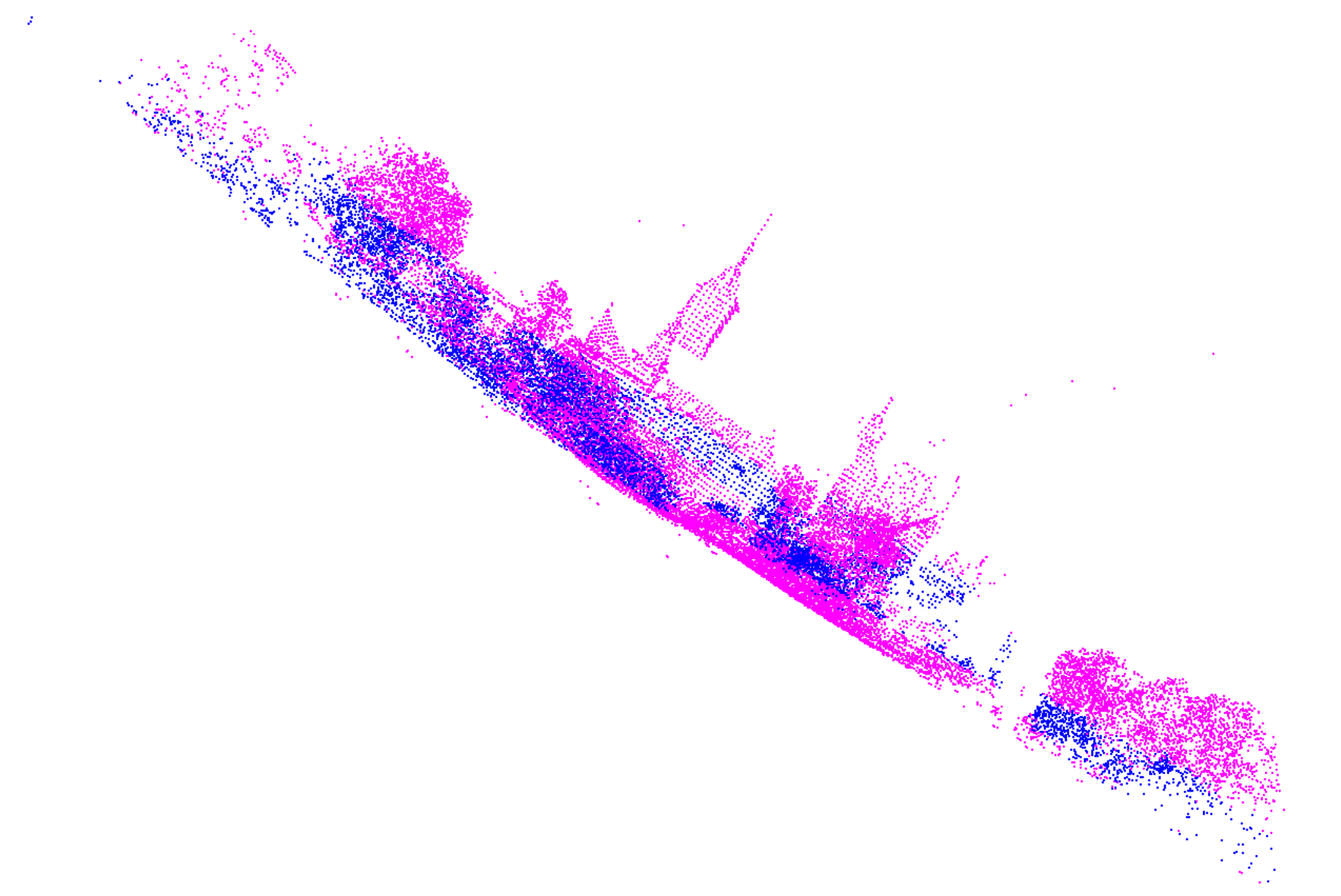}
\end{minipage}

\end{tabular}

\caption{Qualitative results for registration with partiality (low overlapping ratio) over the 10 point clouds using our VOCRA. The first column shows the point correspondences matched by FPFH~\cite{rusu2009fast} where inliers are in green while outliers are in red. The second column displays the registration result using the pose (transformation) estimated by VOCRA.}
\label{qualit-partial}
\vspace{-2mm}
\end{figure*}

\subsection{Main Algorithm: VOCRA}

Up to now, we can render the pseudocode for the main algorithm of VOCRA in Algorithm~\ref{VOCRA}. To summarize, VOCRA is implemented as follows: (a) using \textit{votingTB} (Algorithm~\ref{voting}) to sort correspondences, (b) using \textit{maxRotConsensus} (Algorithm~\ref{MaximizeRotConsensus}) to maximize the consensus set, (c) using \textit{solveGNCTB} to obtain a correct estimate (Algorithm~\ref{solveGNCTB}), and (d) find the complete inlier set (line 4).

\section{Experiments and Applications}\label{experiments}

In this section, we test and evaluate our solver VOCRA in both standard benchmarking and realistic application experiments. All the experiments are implemented in Matlab on a laptop with a i7-7700HQ CPU and 16GB of RAM.

We compare VOCRA against multiple existing state-of-the-art robust registration solvers, including RANSAC~\cite{fischler1981random}, FLO-RANSAC~\cite{lebeda2012fixing} (a variant of RANSAC with speeded-up local optimization, also known as LO$^+$-RANSAC elsewhere), FGR~\cite{zhou2016fast}, GNC-TLS~\cite{yang2020graduated}, ADAPT~\cite{tzoumas2019outlier}, GORE~\cite{bustos2017guaranteed} and GORE+RANSAC (using RANSAC to further refine the remaining correspondences after the outlier removal of GORE). BnB~\cite{parra2014fast} is not adopted for evaluation since it runs in hours with 1000 correspondence, and TEASER~\cite{yang2019polynomial,yang2020graduated} is also not used because there is no standard maximum clique solver in Matlab. Note that all the solvers are implemented with single thread and no parallelism programming is applied. 

In order to quantify the accuracy of estimation, we adopt the following errors for rotation and translation such that

\begin{equation}
\begin{gathered}
\boldsymbol{R}.\text{Err}=\angle_{geo}(\boldsymbol{R}_{gt}, \boldsymbol{R}^{\star})\cdot {\frac{180}{\pi}}^{\circ},\\
\boldsymbol{t}.\text{Err}=\left\|\boldsymbol{t}_{gt}-\boldsymbol{t}^{\star}\right\|,
\end{gathered}
\end{equation}

\noindent where $_{gt}$ denotes the ground-truth values and $^{\star}$ denotes the optimal values solved.

\subsection{Standard Benchmarking over Real Point Clouds}
\label{exp-benchmark}

We first comprehensively benchmark all the robust solvers in a fair experimental environment to comparatively evaluate their robustness against outliers, accuracy of estimation and efficiency.

\begin{figure*}[t]
\centering

\setlength\tabcolsep{1.5pt}
\addtolength{\tabcolsep}{0pt}
\begin{tabular}{ccc}

\multicolumn{3}{c}{\footnotesize{(a) Quantitative results of normal registration}} \\

\begin{minipage}[t]{0.32\linewidth}
\centering
\includegraphics[width=1\linewidth]{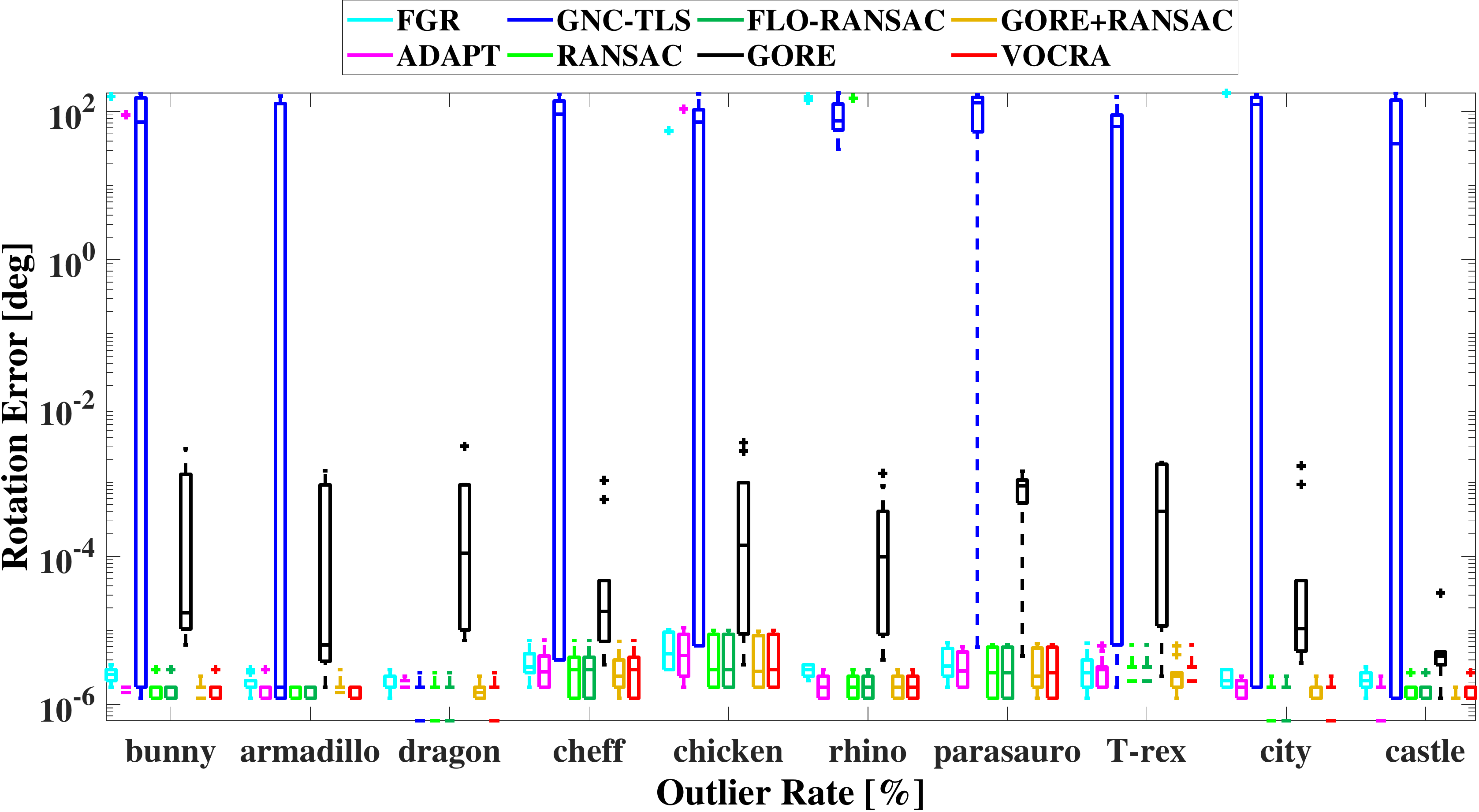}
\end{minipage}&
\begin{minipage}[t]{0.32\linewidth}
\centering
\includegraphics[width=1\linewidth]{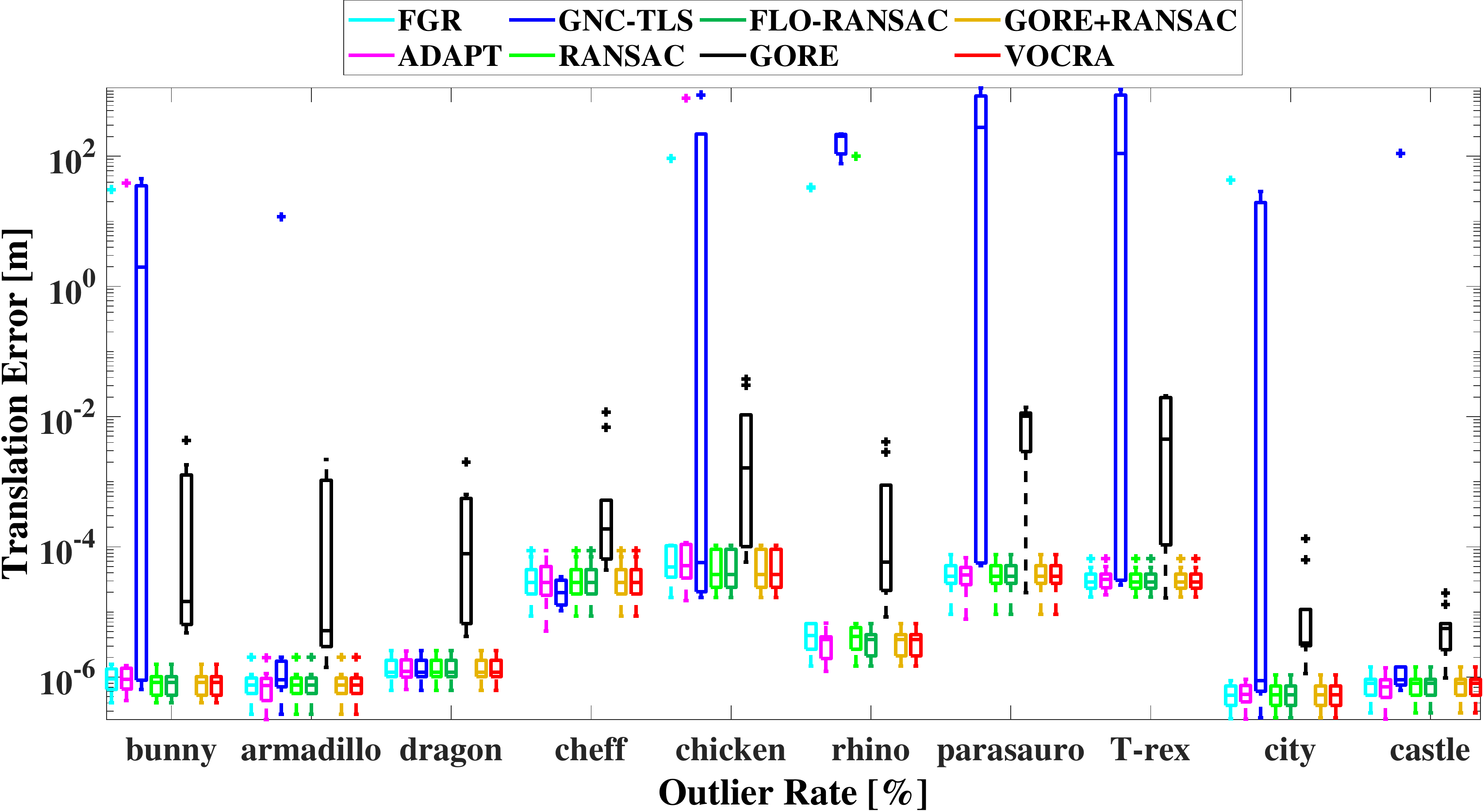}
\end{minipage}&
\begin{minipage}[t]{0.32\linewidth}
\centering
\includegraphics[width=1\linewidth]{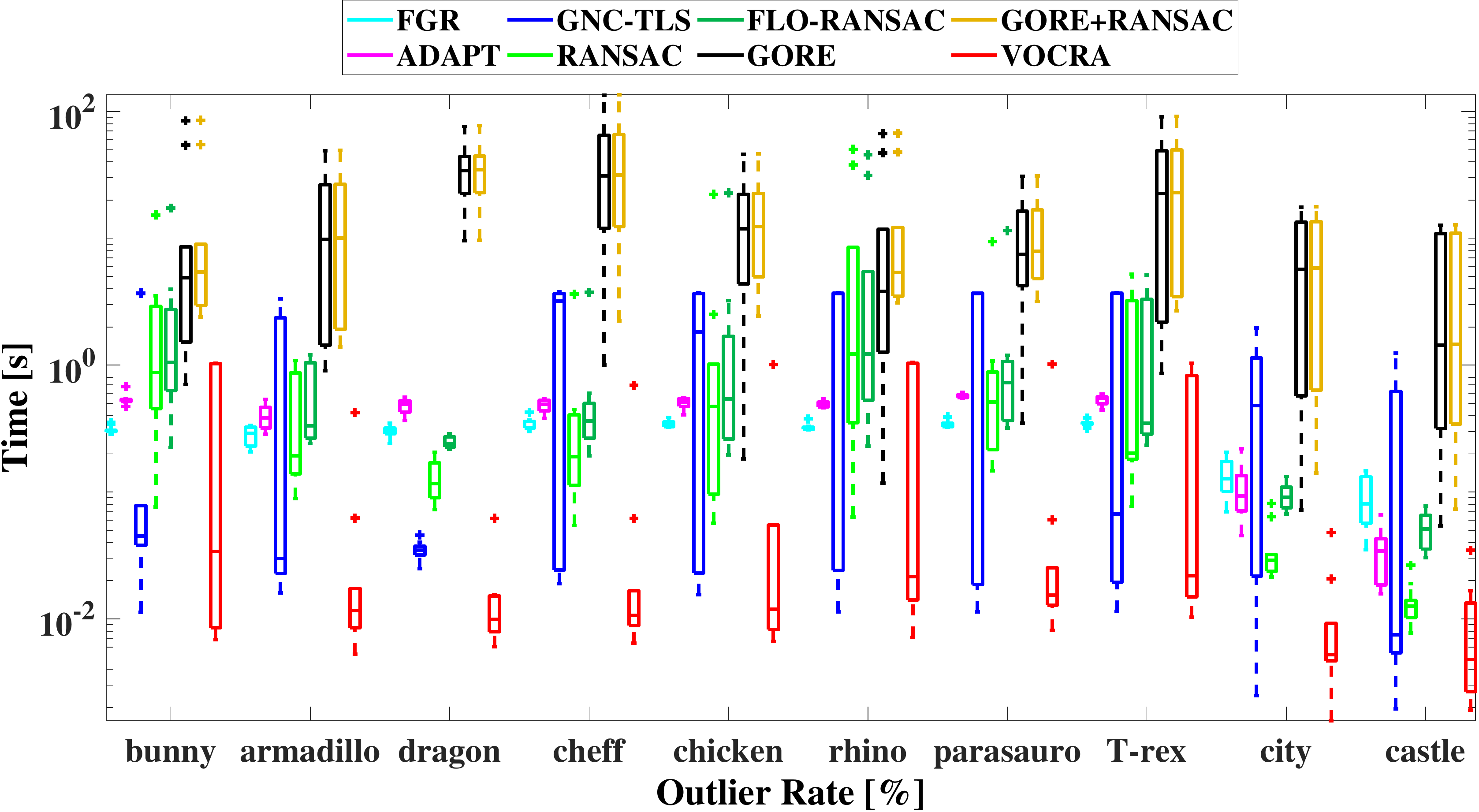}
\end{minipage} \\

\multicolumn{3}{c}{\footnotesize{(b) Quantitative results of partial registration}} \\

\begin{minipage}[t]{0.32\linewidth}
\centering
\includegraphics[width=1\linewidth]{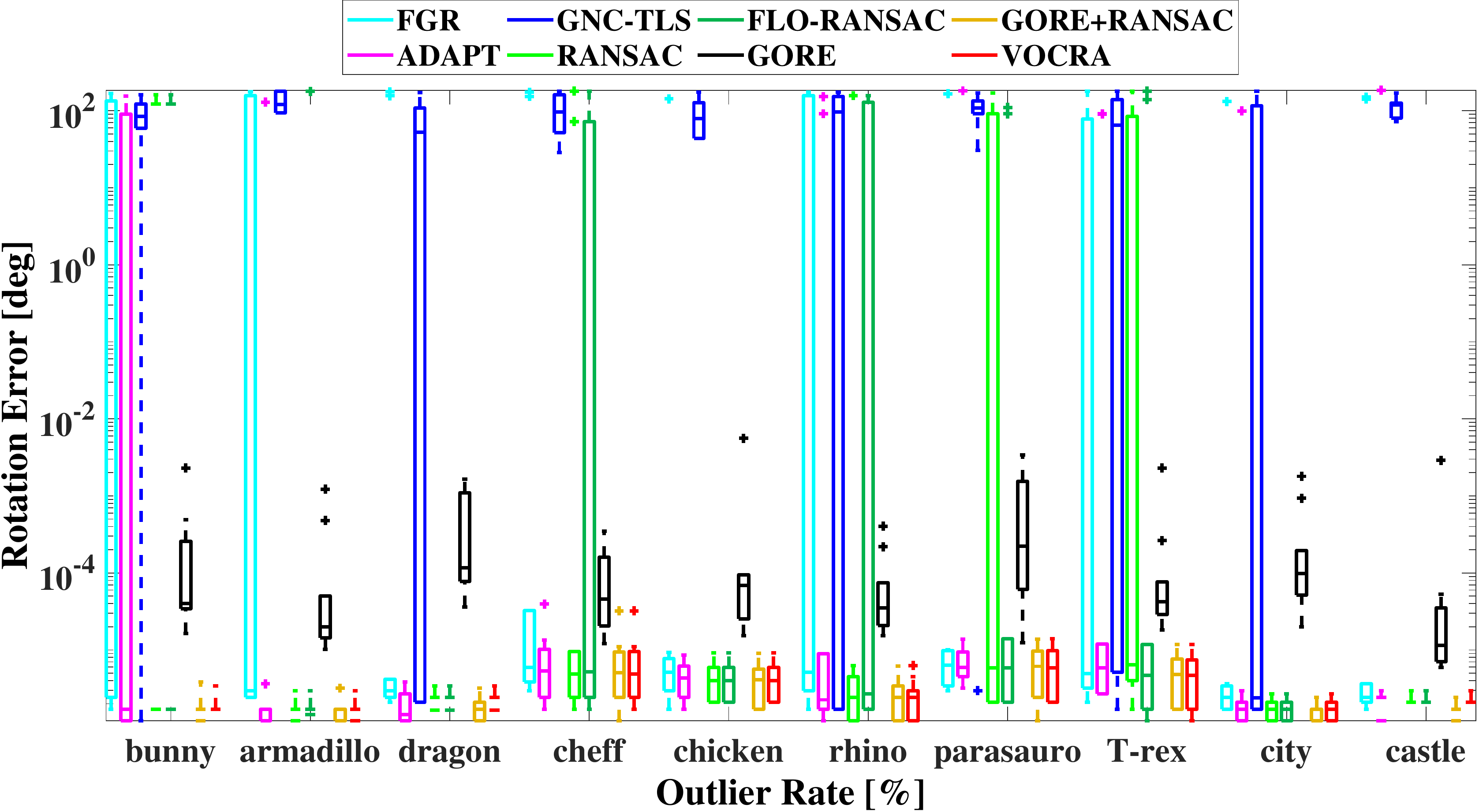}
\end{minipage}&

\begin{minipage}[t]{0.32\linewidth}
\centering
\includegraphics[width=1\linewidth]{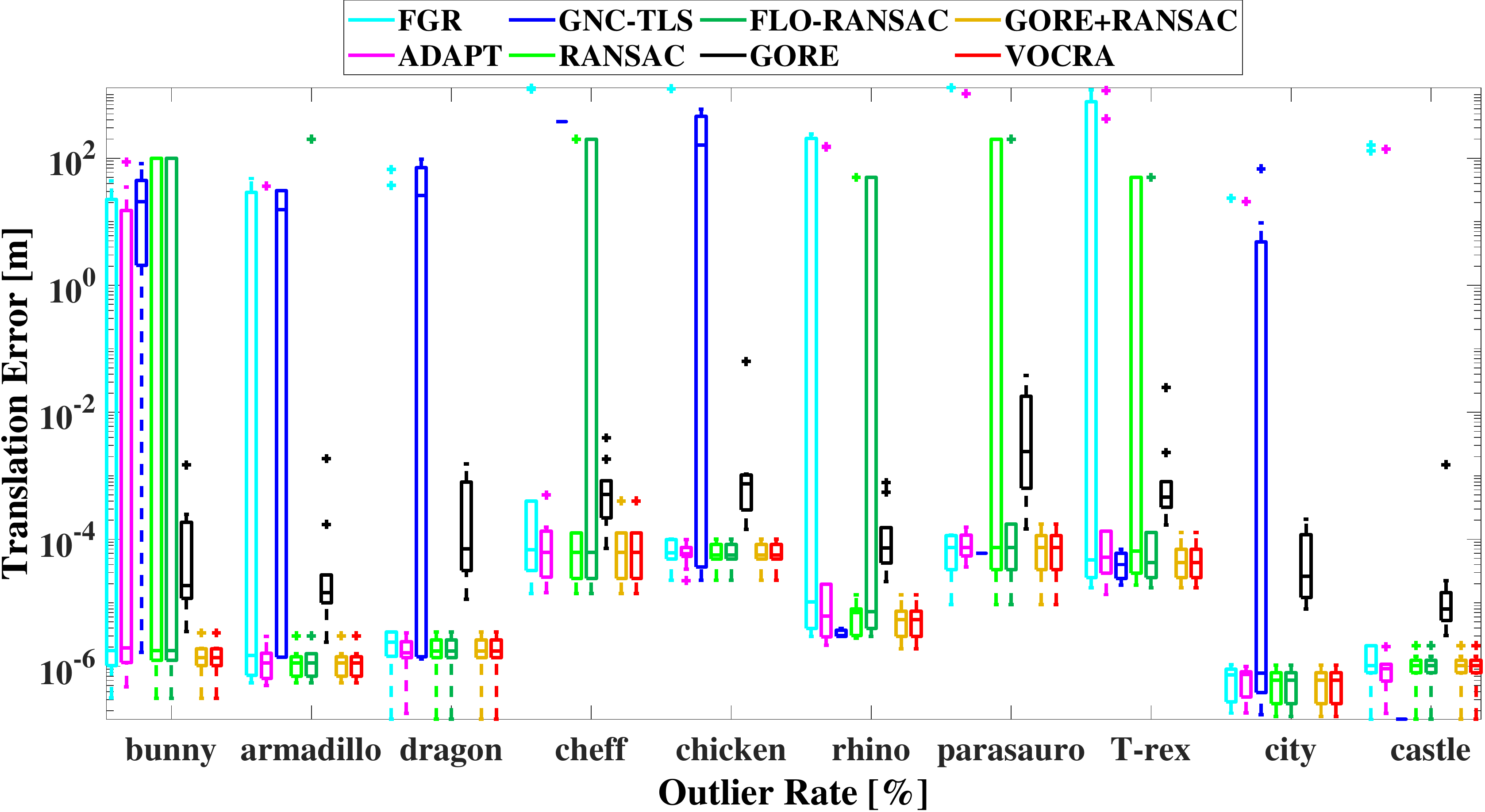}
\end{minipage}&

\begin{minipage}[t]{0.32\linewidth}
\centering
\includegraphics[width=1\linewidth]{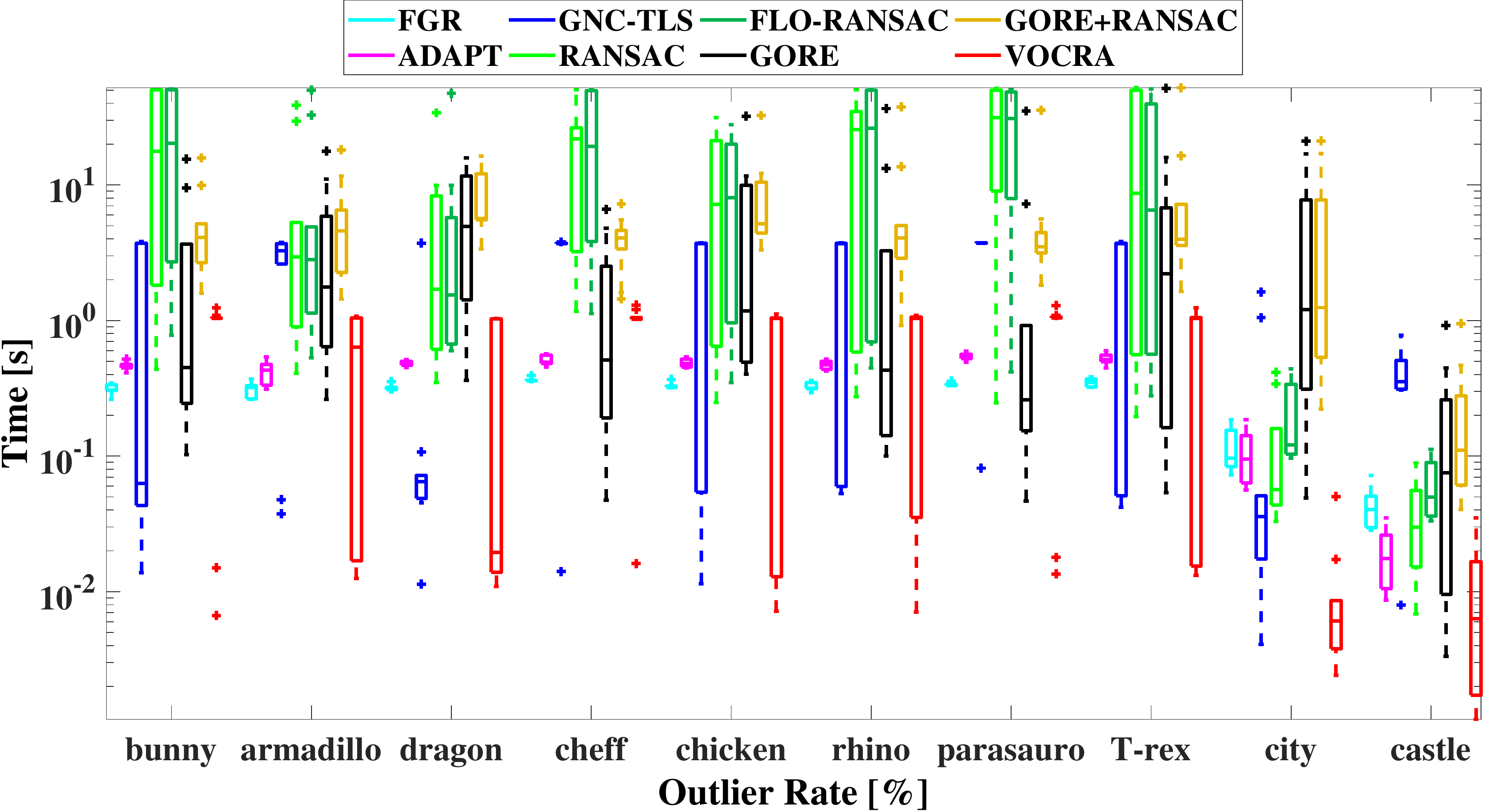}
\end{minipage}

\end{tabular}
\vspace{-2mm}
\caption{Quantitative results of both normal and partial registration over the 10 point clouds using RANSAC~\cite{fischler1981random}, FLO-RANSAC~\cite{lebeda2012fixing}, FGR~\cite{zhou2016fast}, GNC-TLS~\cite{yang2020graduated}, ADAPT~\cite{tzoumas2019outlier}, GORE~\cite{bustos2017guaranteed}, GORE+RANSAC and VOCRA, corresponding to Figure~\ref{qualit-normal} and~\ref{qualit-partial}, respectively.}
\label{quant-both}
\vspace{-2mm}
\end{figure*}

\begin{figure*}[h]
\centering
\setlength\tabcolsep{0.0pt}
\addtolength{\tabcolsep}{0pt}
\begin{tabular}{c|cc|c|c|c|c}
\quad &\,\footnotesize{FPFH}\, &\,&  \footnotesize{FLO-RANSAC~\cite{lebeda2012fixing}} & \footnotesize{GNC-TLS~\cite{yang2020graduated}} & \footnotesize{GORE+RANSAC~\cite{bustos2017guaranteed}} & \footnotesize{VOCRA} \\
\hline
&&&&&&
\\
\rotatebox{90}{\,\,\footnotesize{\textit{Scene 1-2}}\,}\,
&
\,\,
\begin{minipage}[t]{0.1\linewidth}
\centering
\includegraphics[width=1\linewidth]{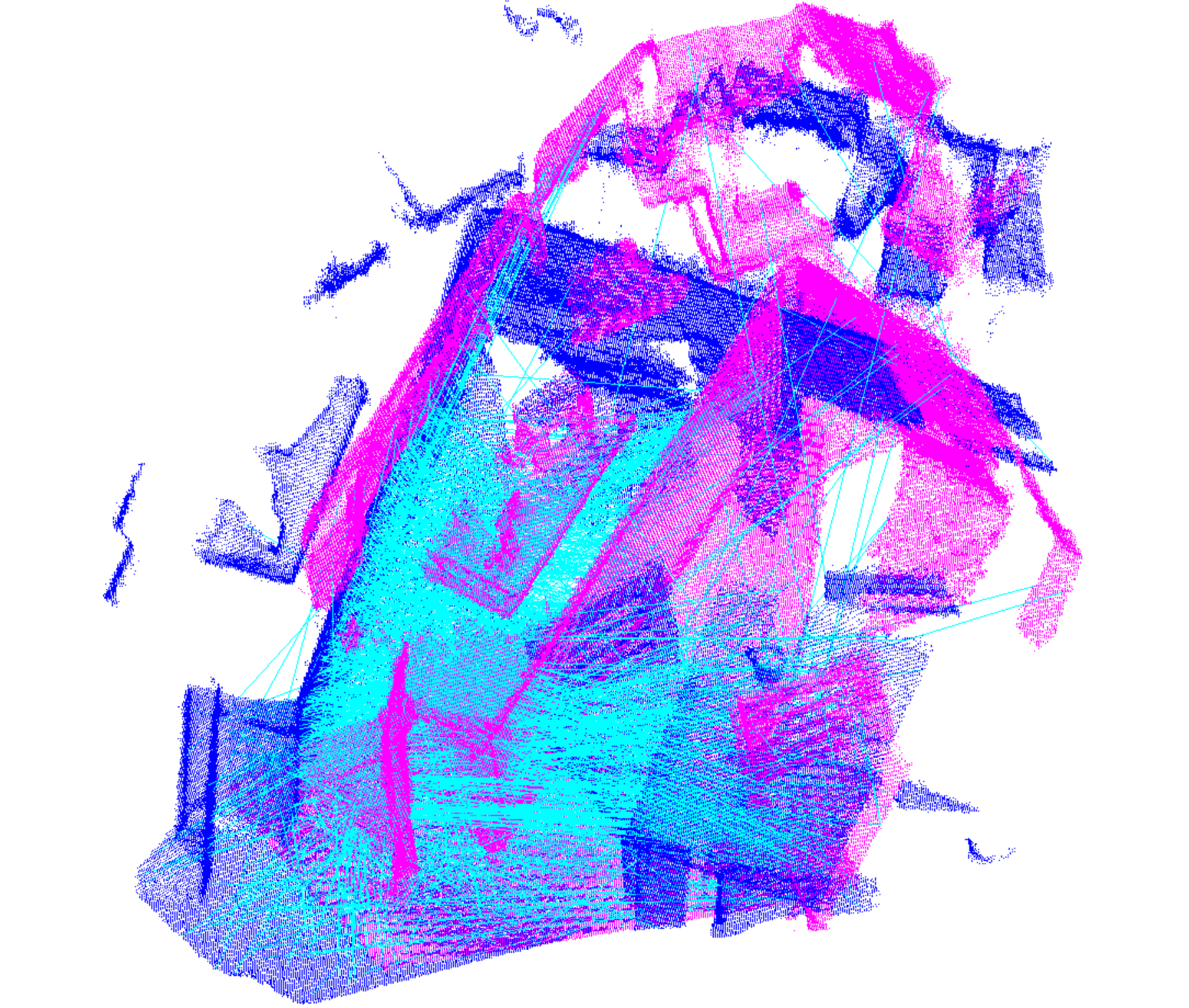}
\end{minipage}\,\,
& &
\,\,
\begin{minipage}[t]{0.19\linewidth}
\centering
\includegraphics[width=.48\linewidth]{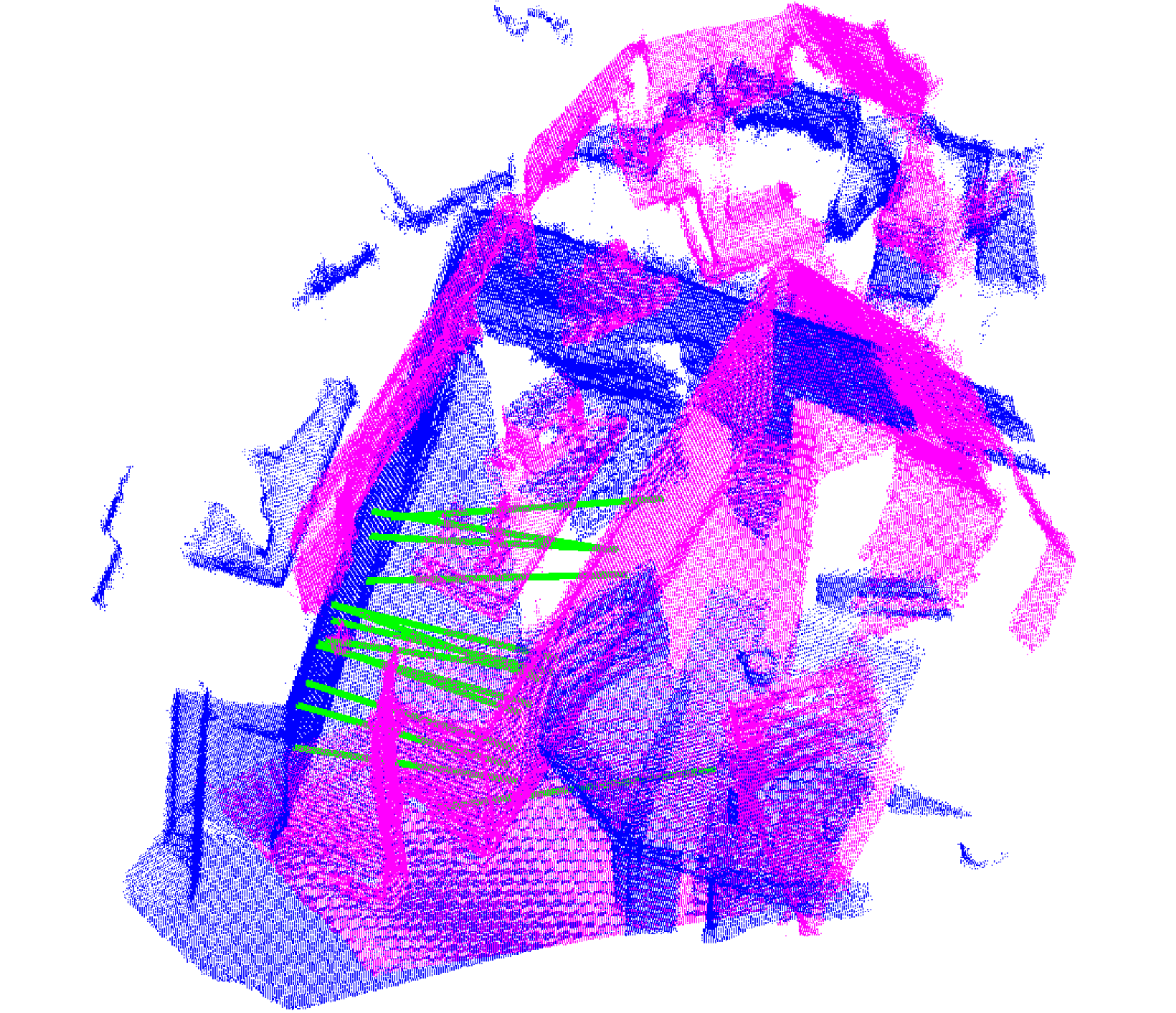}
\includegraphics[width=.48\linewidth]{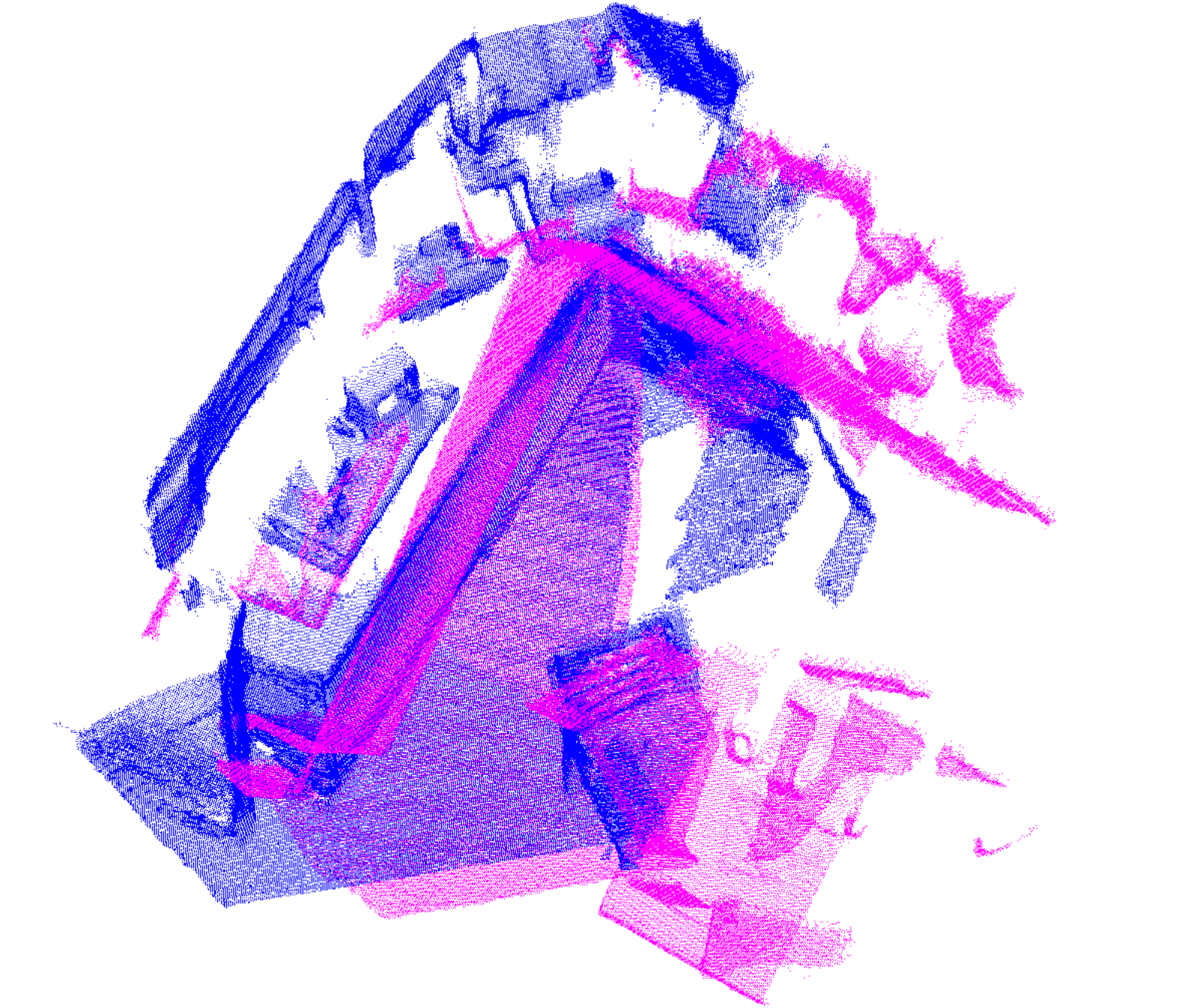}
\end{minipage}\,\,
&
\,\,
\begin{minipage}[t]{0.19\linewidth}
\centering
\includegraphics[width=.48\linewidth]{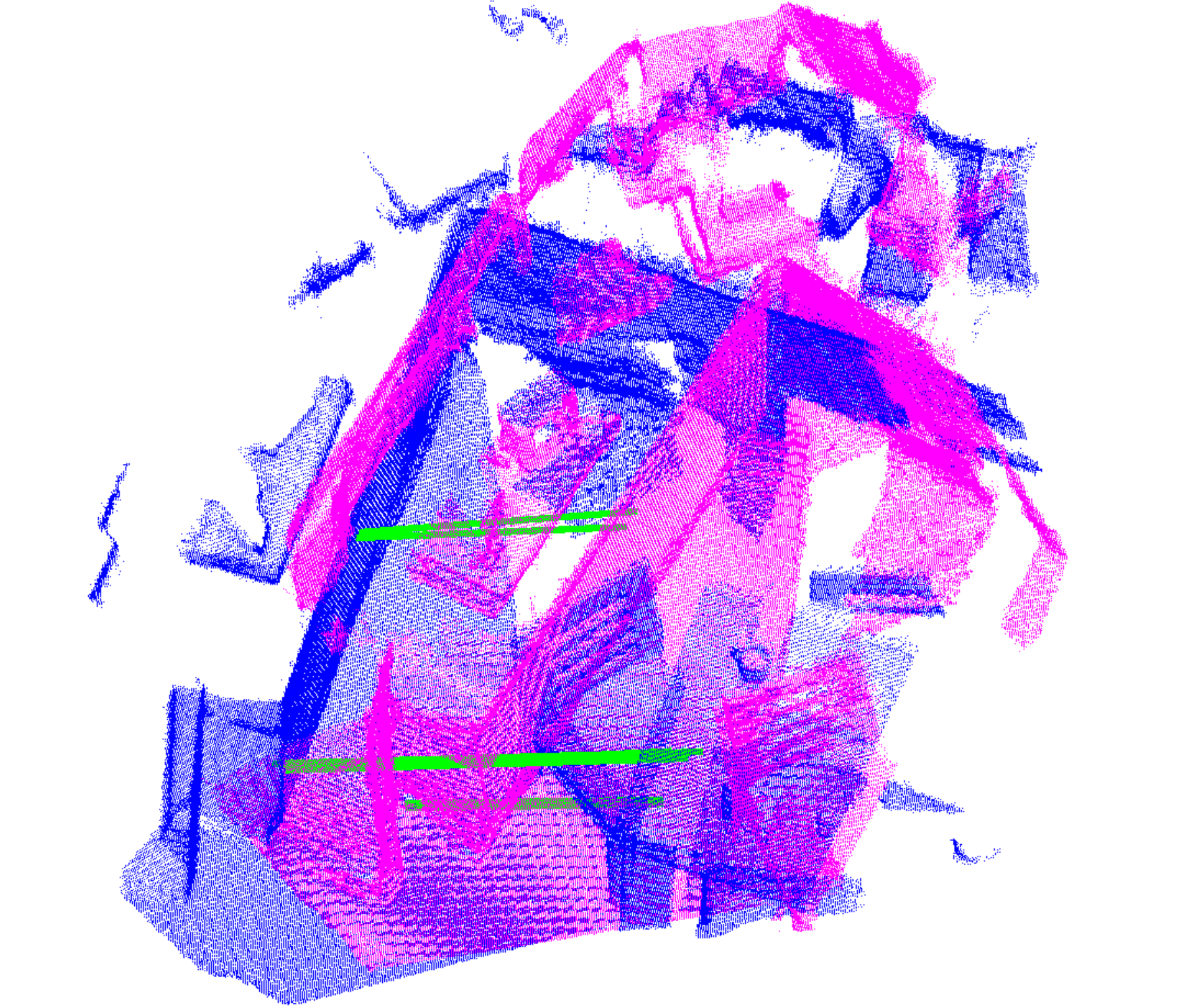}
\includegraphics[width=.48\linewidth]{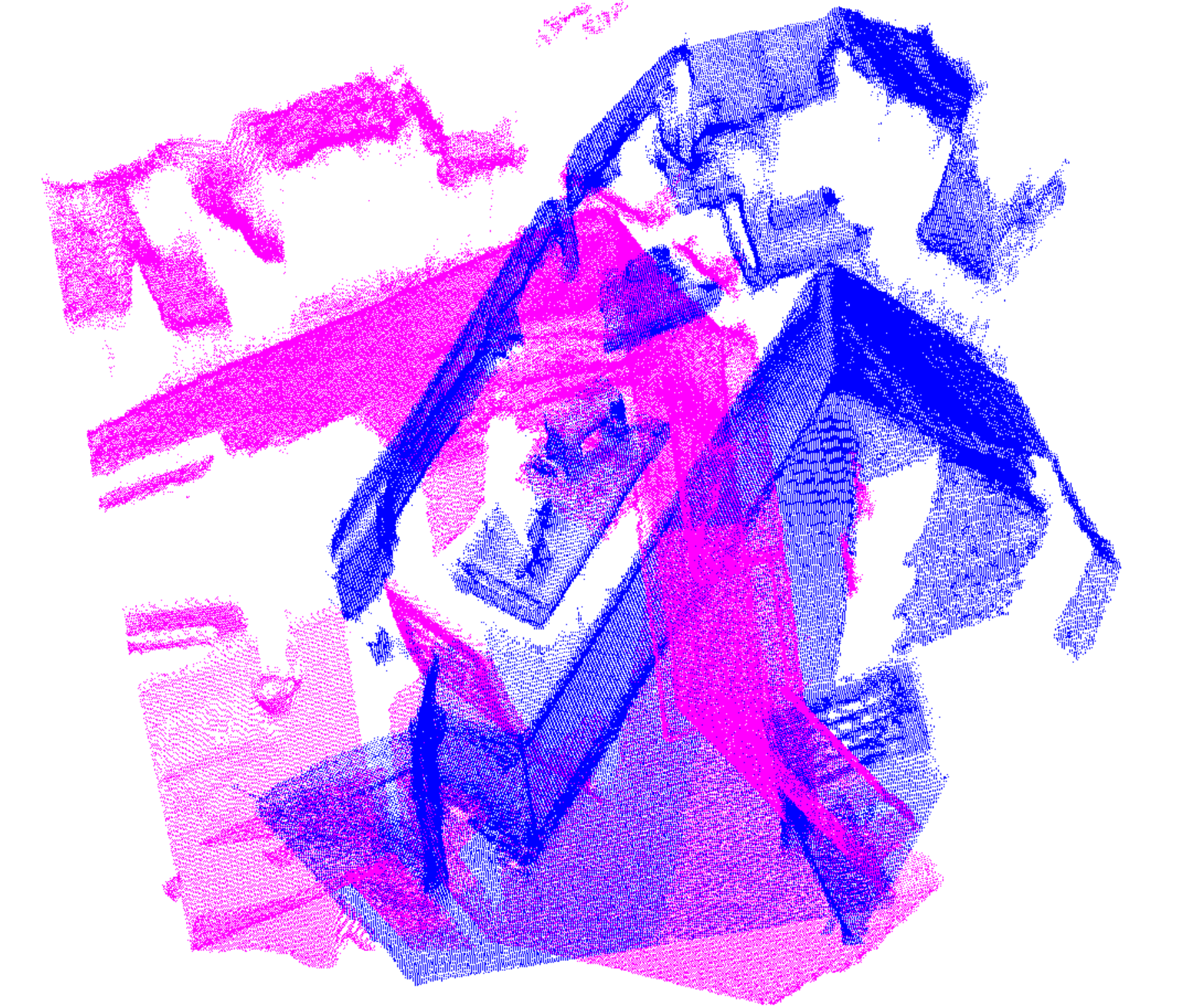}
\end{minipage}\,\,
&
\,\,
\begin{minipage}[t]{0.19\linewidth}
\centering
\includegraphics[width=.48\linewidth]{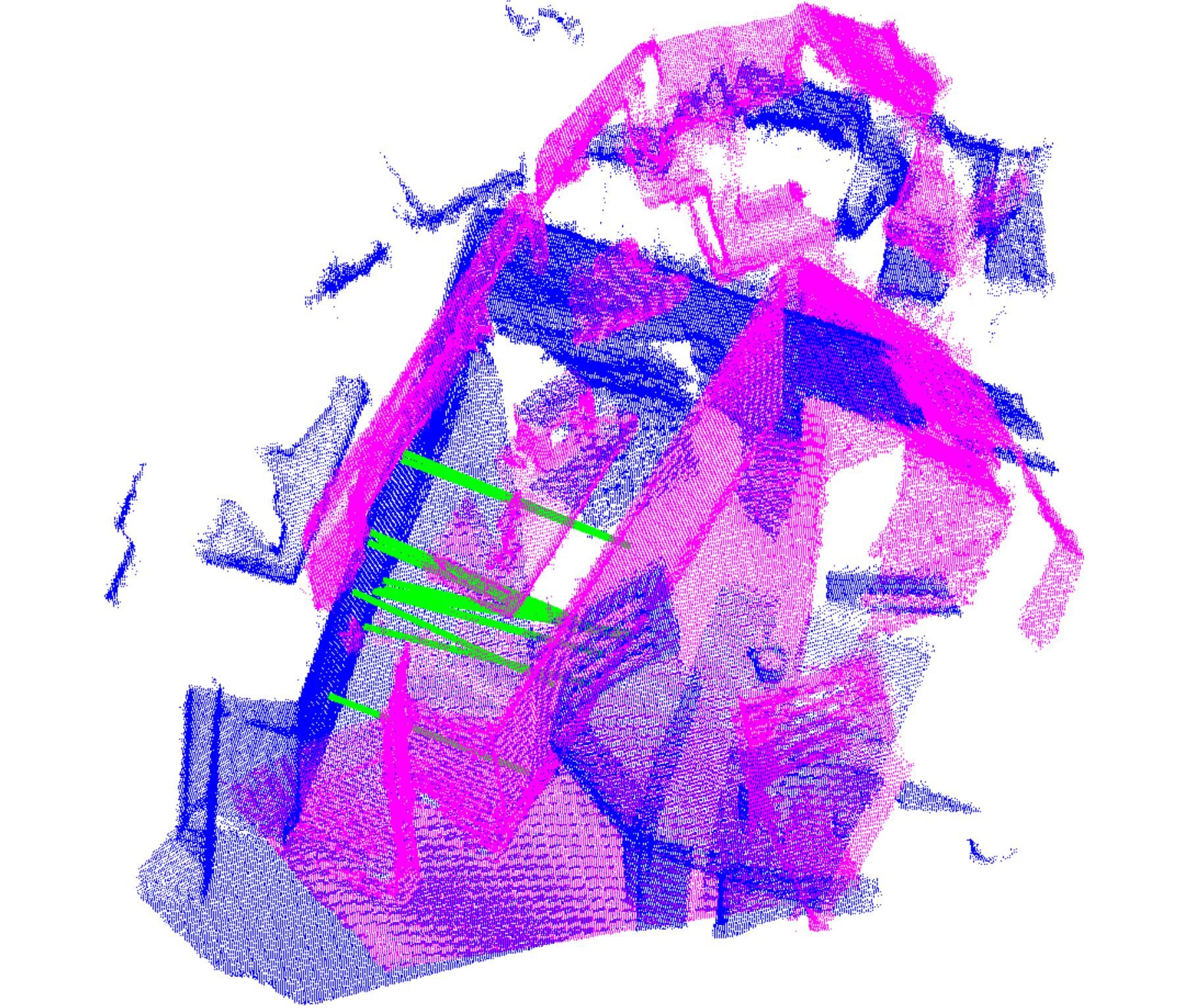}
\includegraphics[width=.48\linewidth]{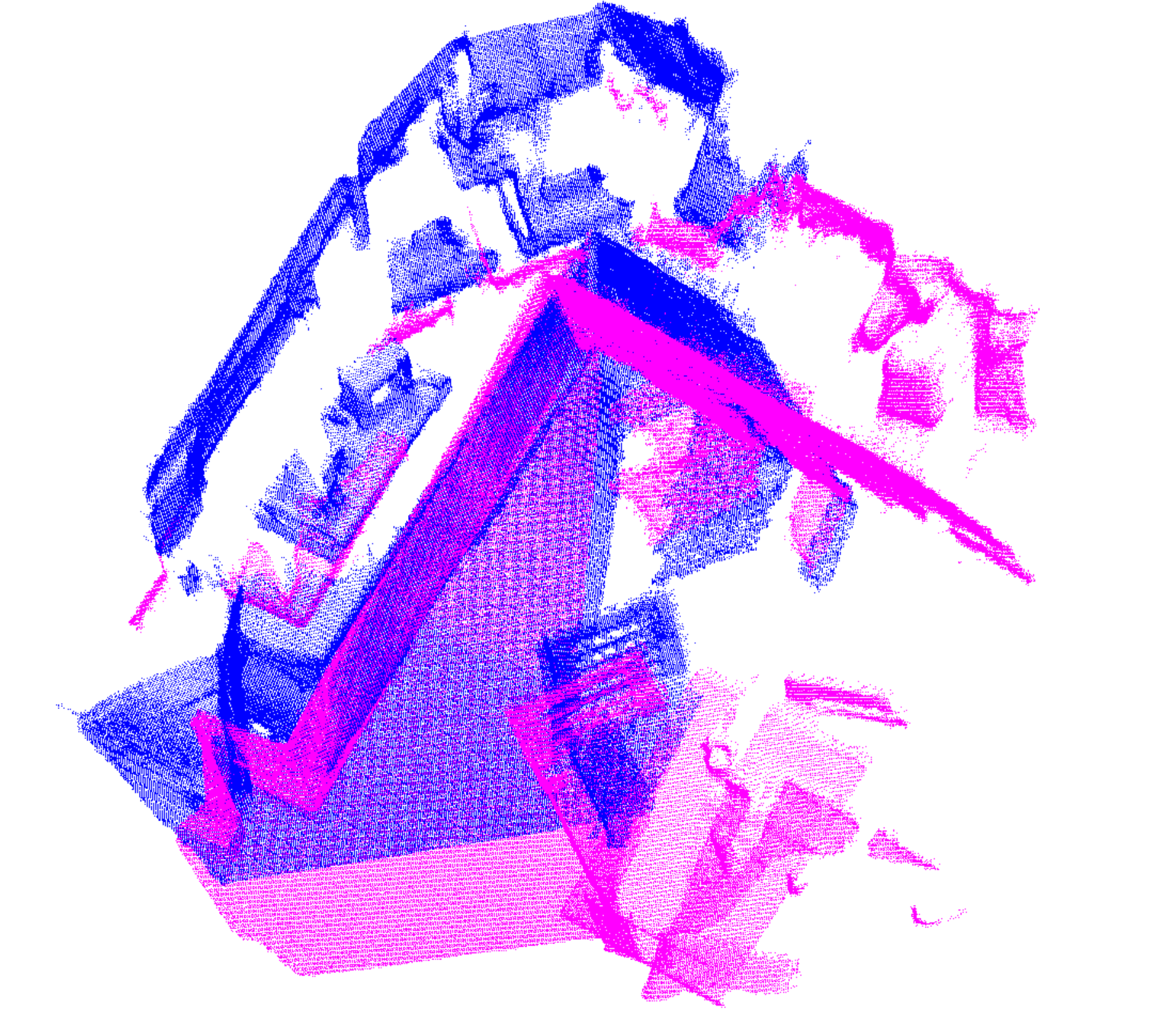}
\end{minipage}\,\,
&
\,\,
\begin{minipage}[t]{0.19\linewidth}
\centering
\includegraphics[width=.48\linewidth]{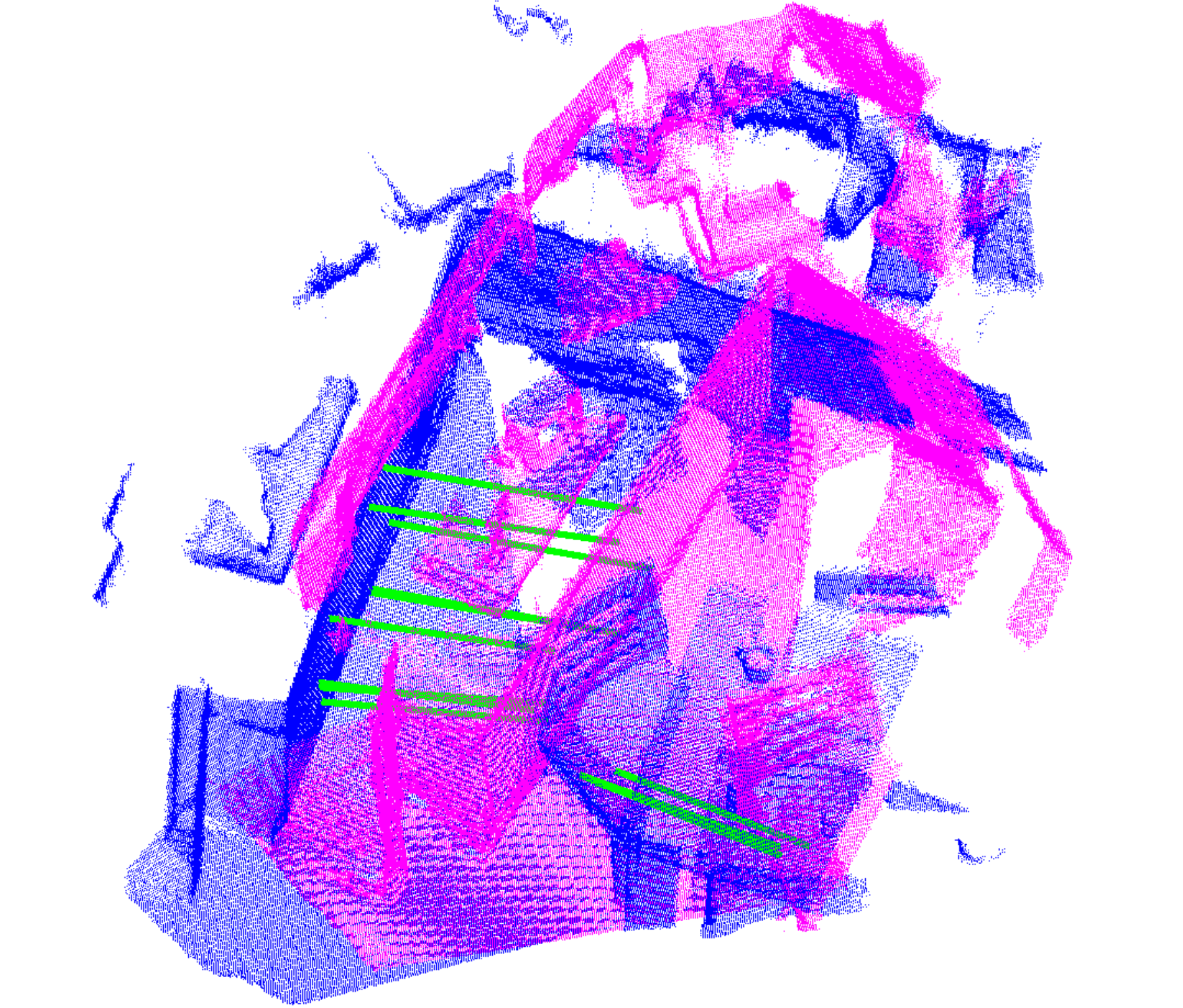}
\includegraphics[width=.48\linewidth]{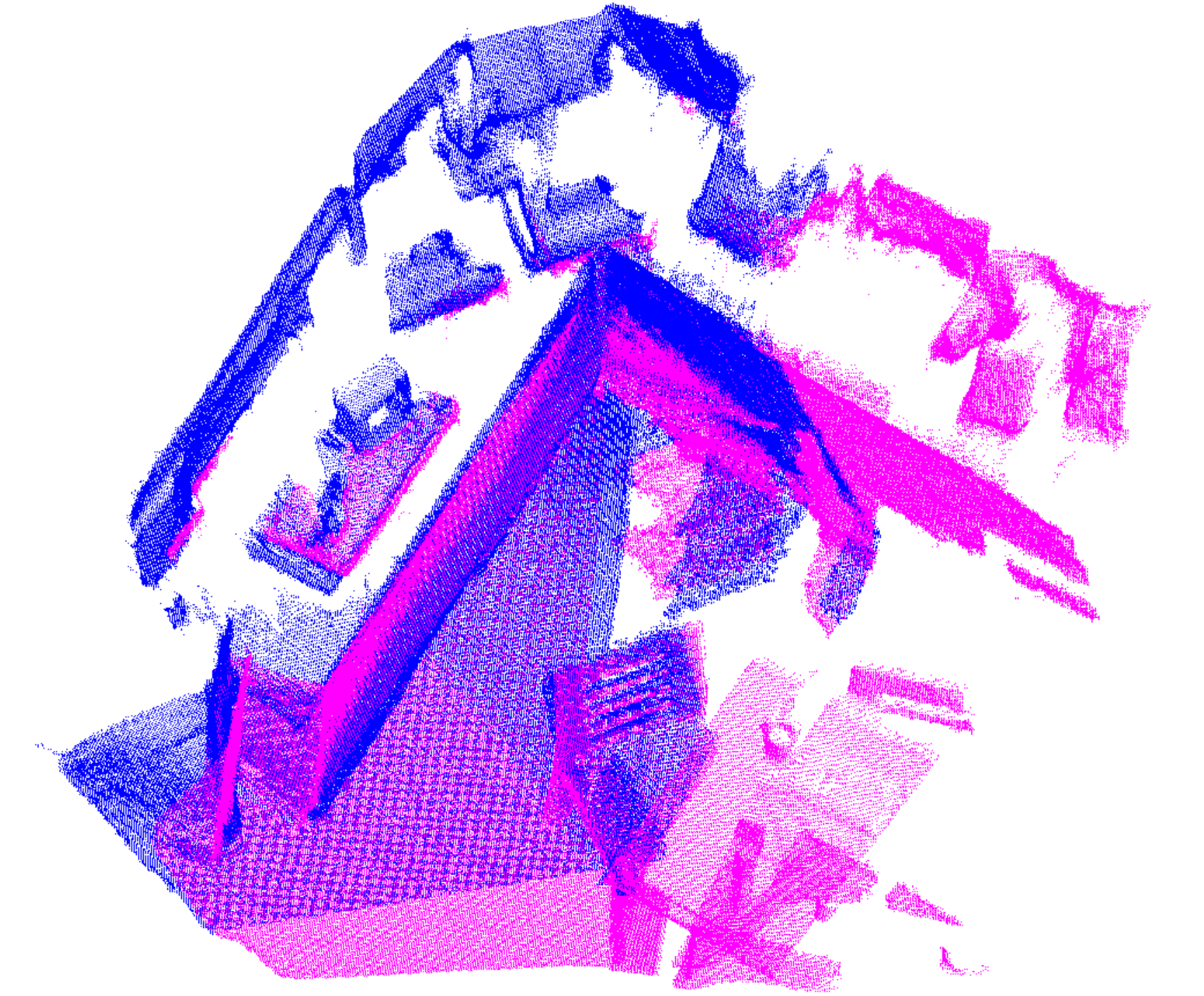}
\end{minipage}\,\,
\\
\rotatebox{90}{\,\,\footnotesize{\textit{Scene 4-5}}\,}\,
&
\,\,
\begin{minipage}[t]{0.1\linewidth}
\centering
\includegraphics[width=1\linewidth]{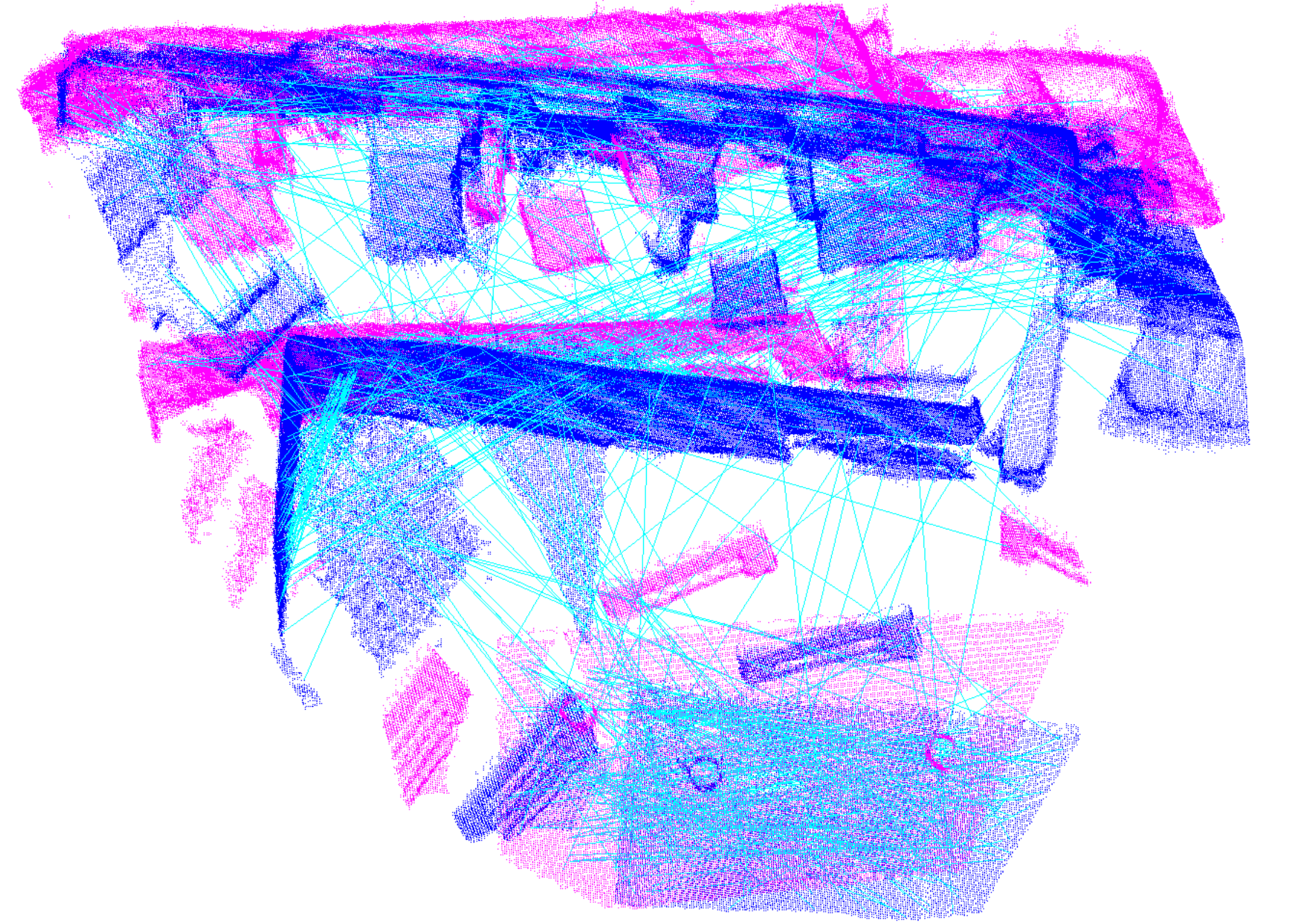}
\end{minipage}\,\,
& &
\,\,
\begin{minipage}[t]{0.19\linewidth}
\centering
\includegraphics[width=.48\linewidth]{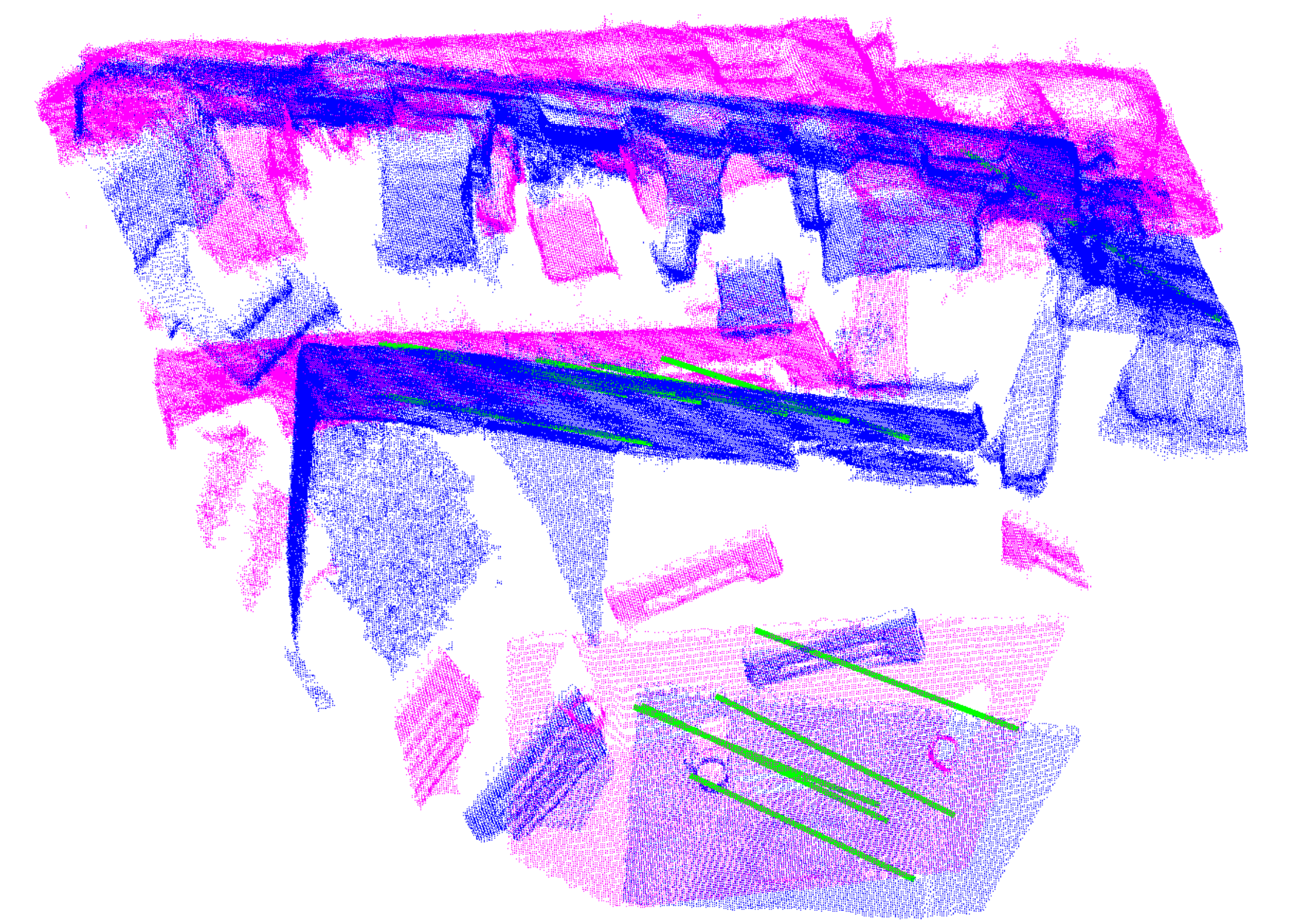}
\includegraphics[width=.48\linewidth]{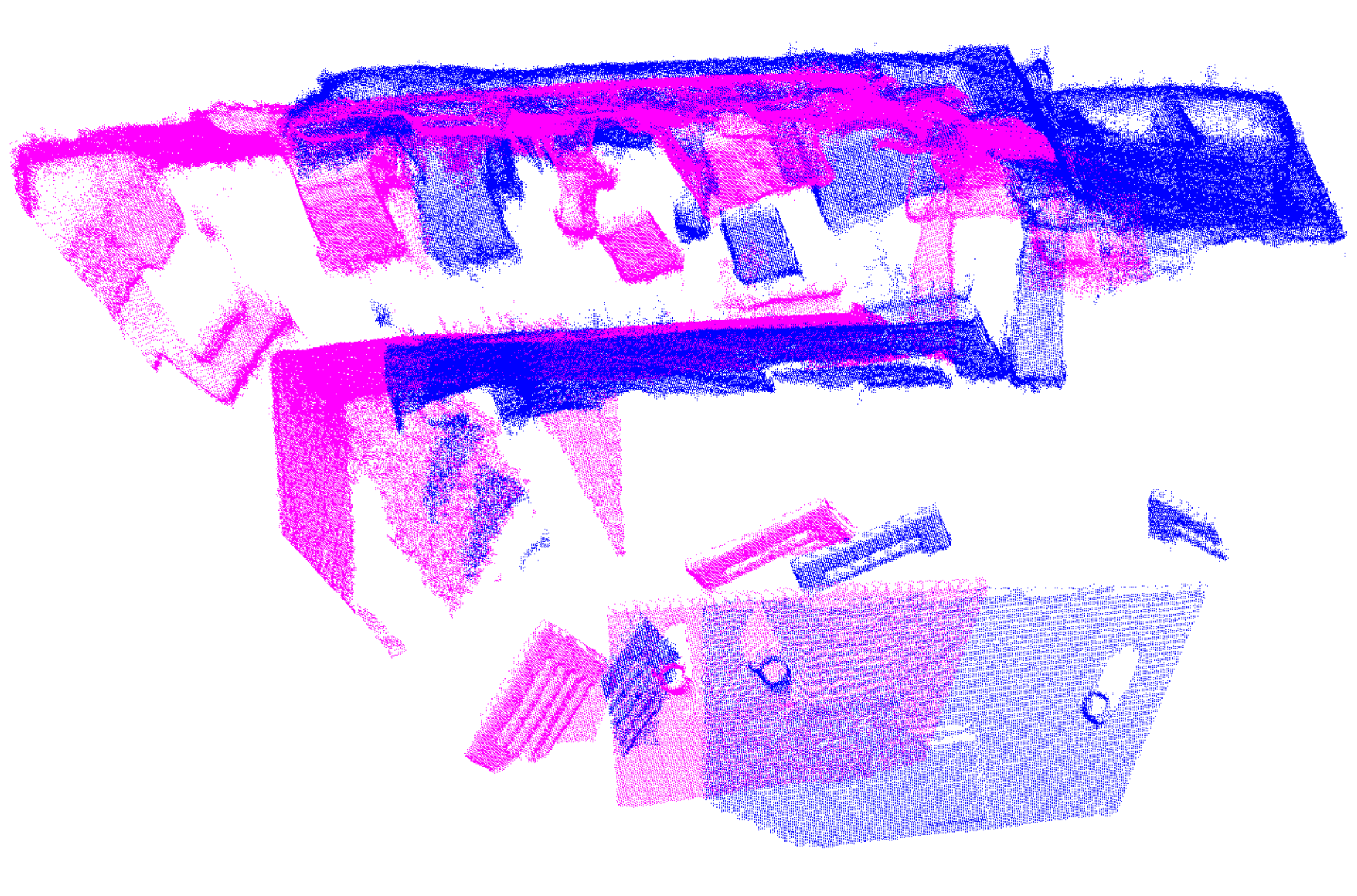}
\end{minipage}\,\,
&
\,\,
\begin{minipage}[t]{0.19\linewidth}
\centering
\includegraphics[width=.48\linewidth]{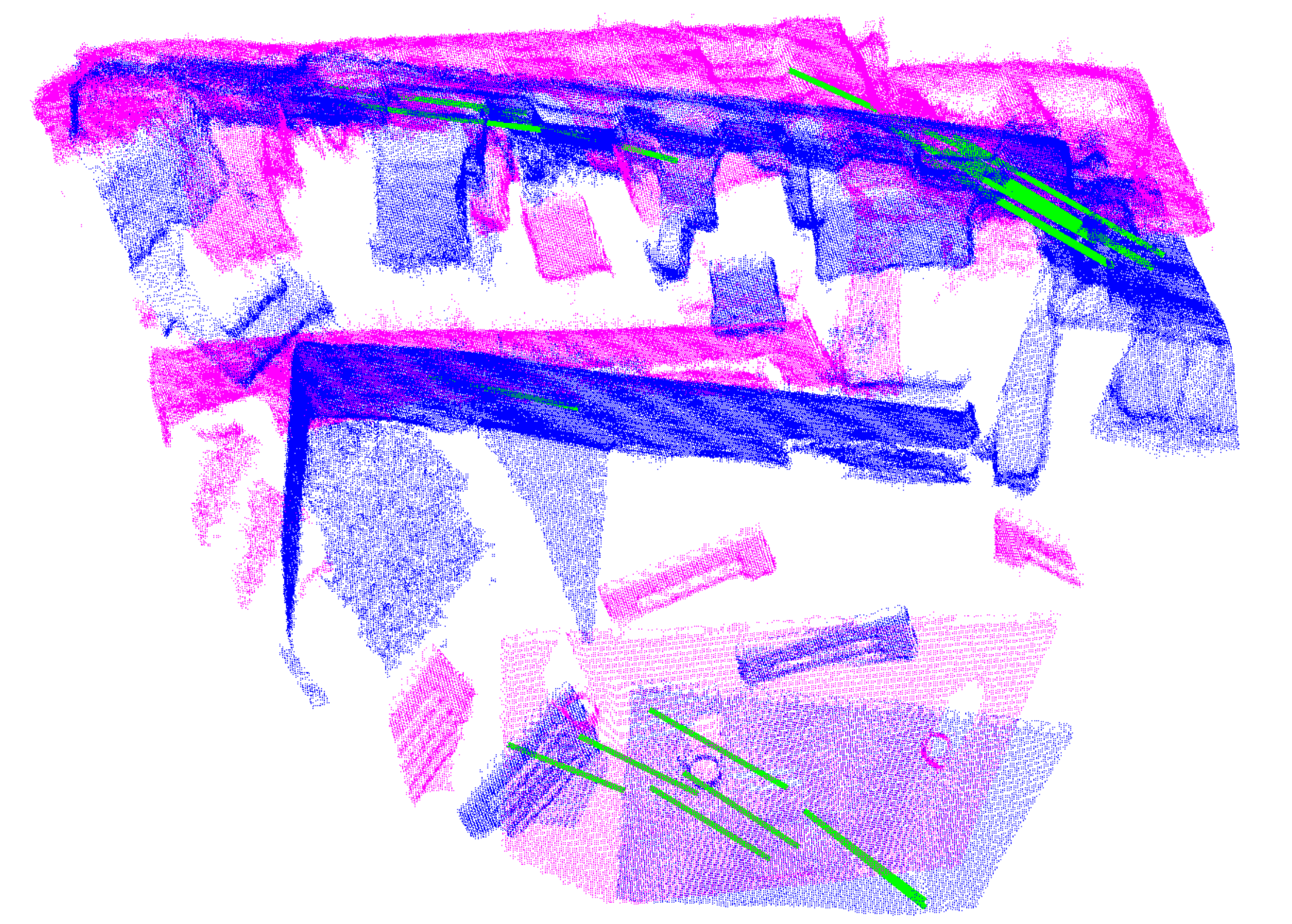}
\includegraphics[width=.48\linewidth]{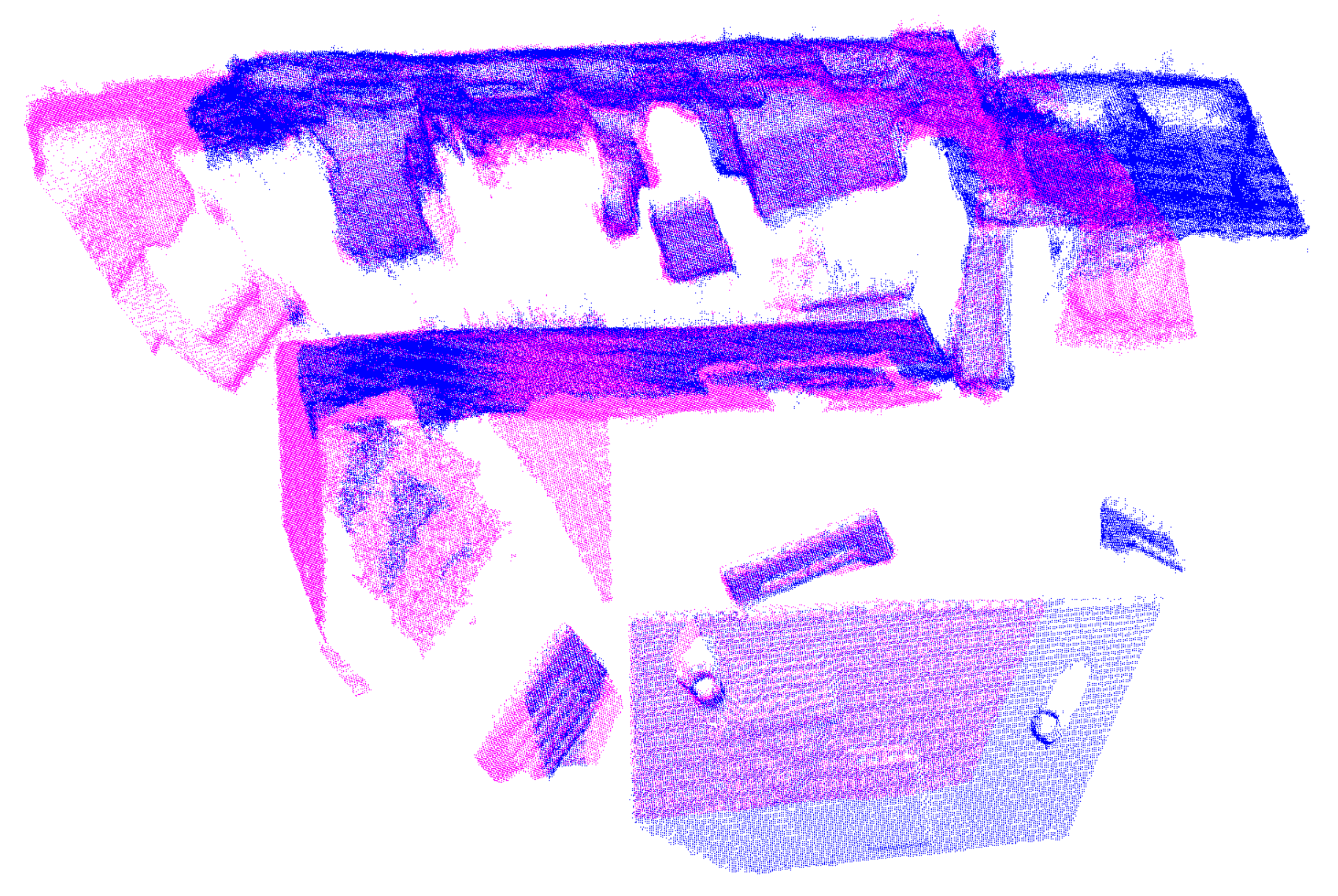}
\end{minipage}\,\,
&
\,\,
\begin{minipage}[t]{0.19\linewidth}
\centering
\includegraphics[width=.48\linewidth]{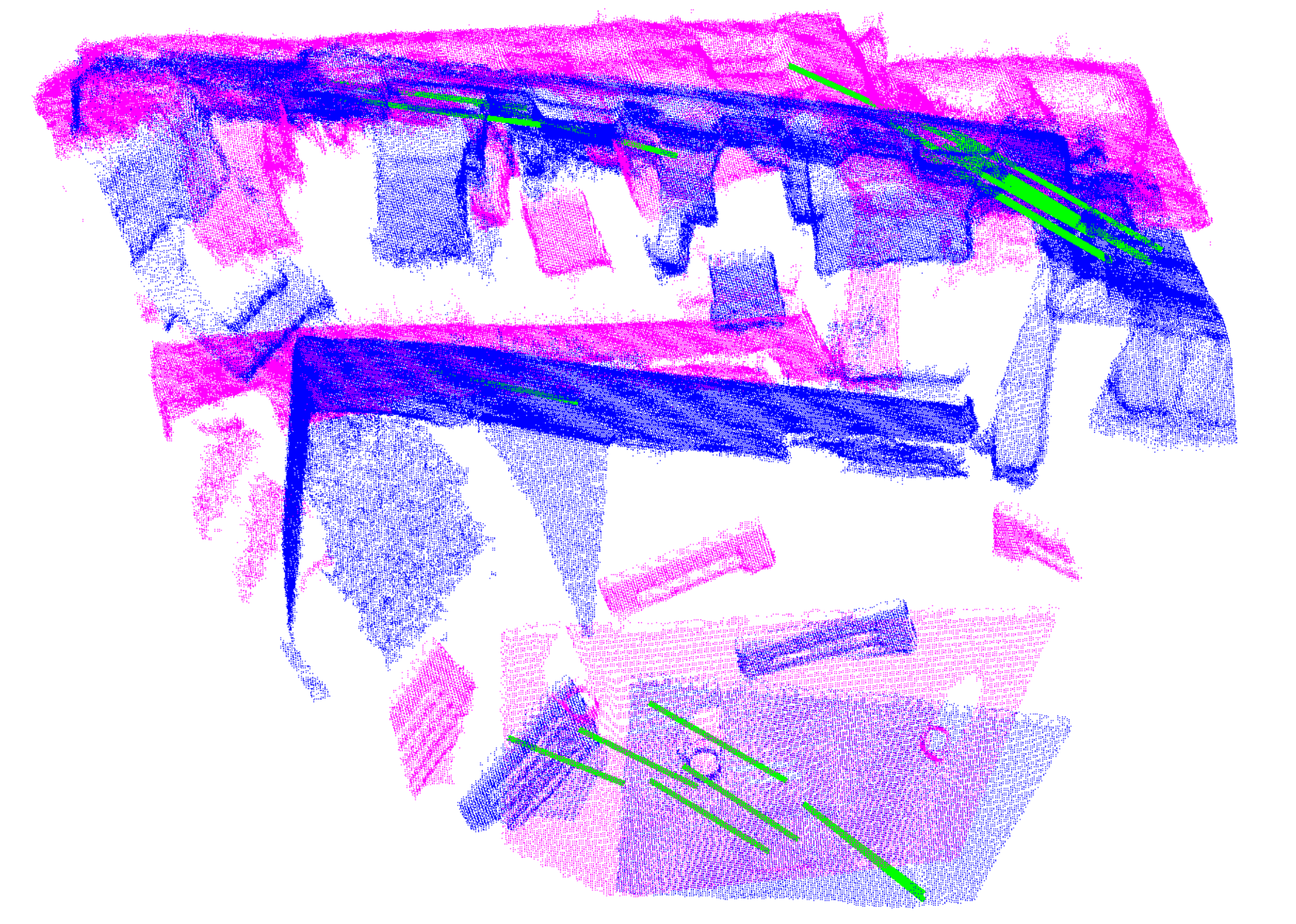}
\includegraphics[width=.48\linewidth]{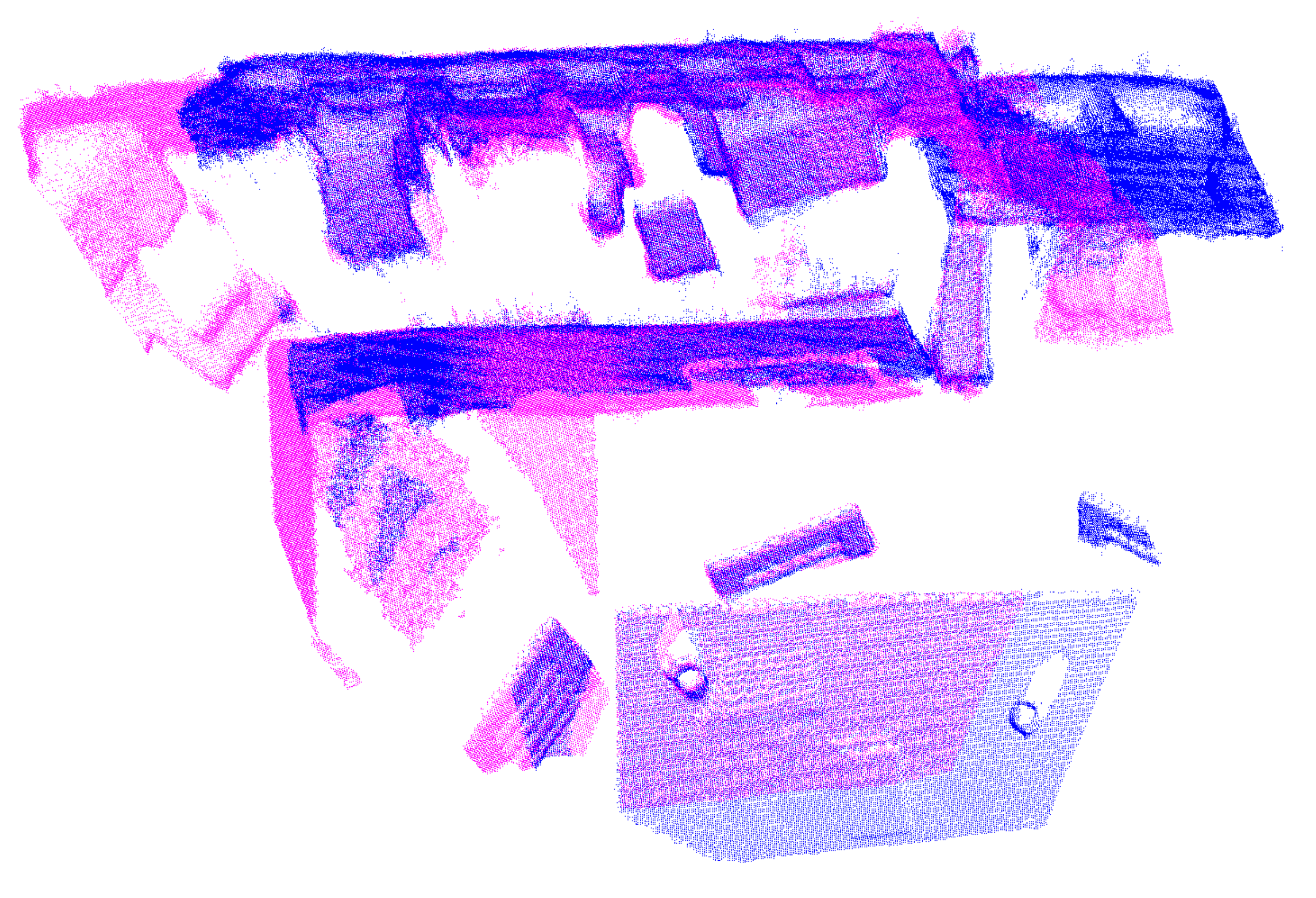}
\end{minipage}\,\,
&
\,\,
\begin{minipage}[t]{0.19\linewidth}
\centering
\includegraphics[width=.48\linewidth]{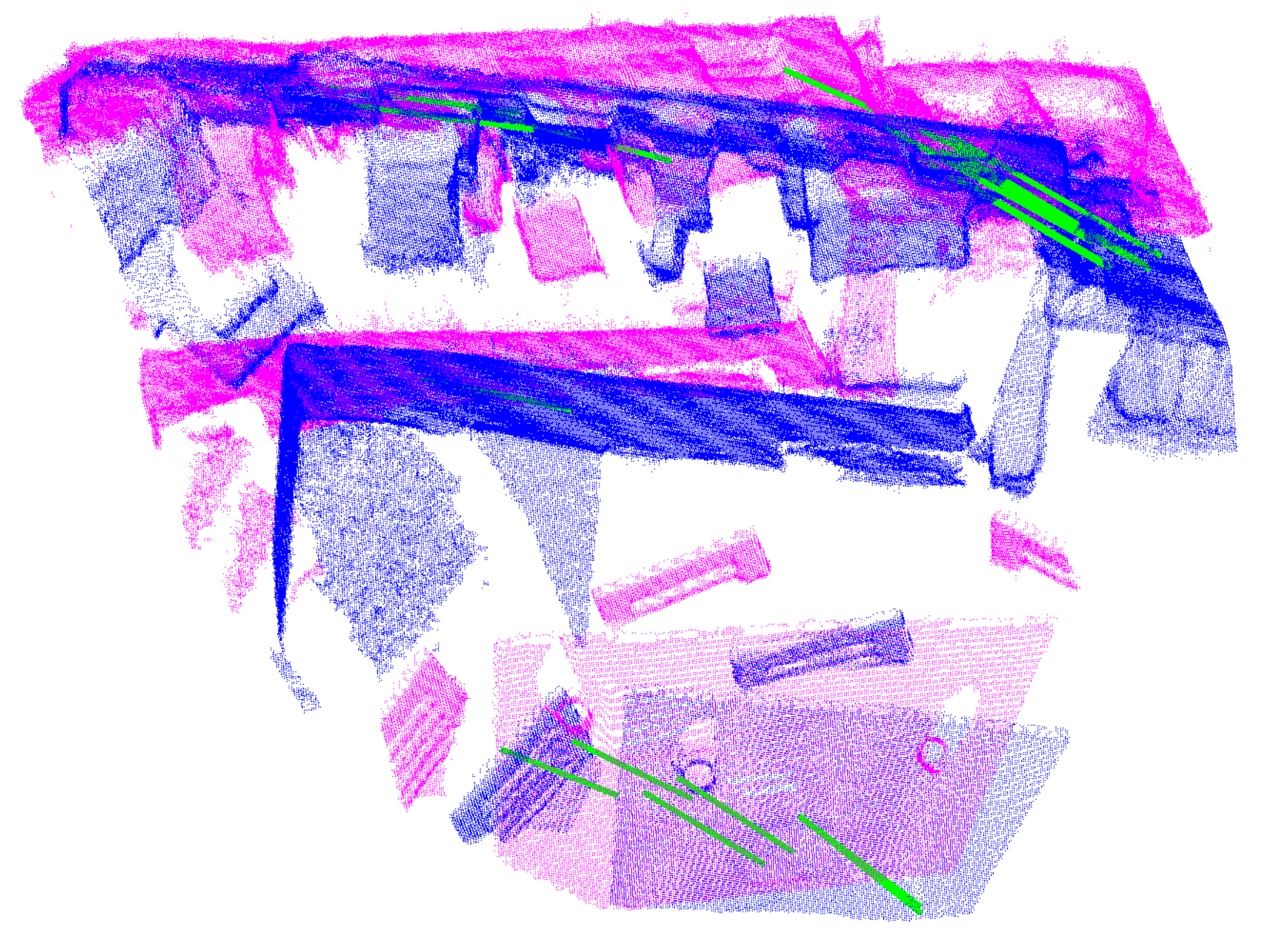}
\includegraphics[width=.48\linewidth]{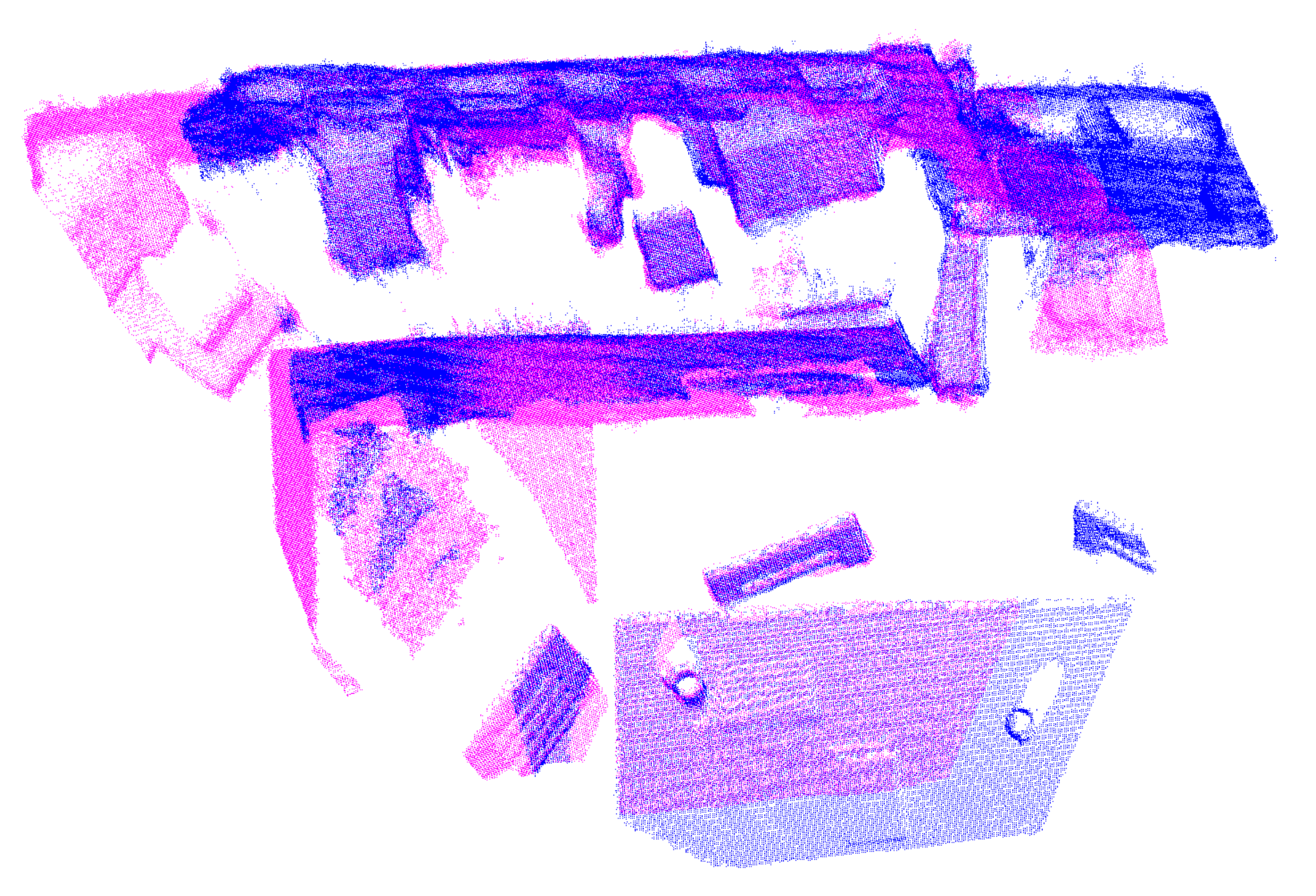}
\end{minipage}\,\,
\\
\rotatebox{90}{\,\,\footnotesize{\textit{Scene 9-10}}\,}\,
&
\,\,
\begin{minipage}[t]{0.1\linewidth}
\centering
\includegraphics[width=1\linewidth]{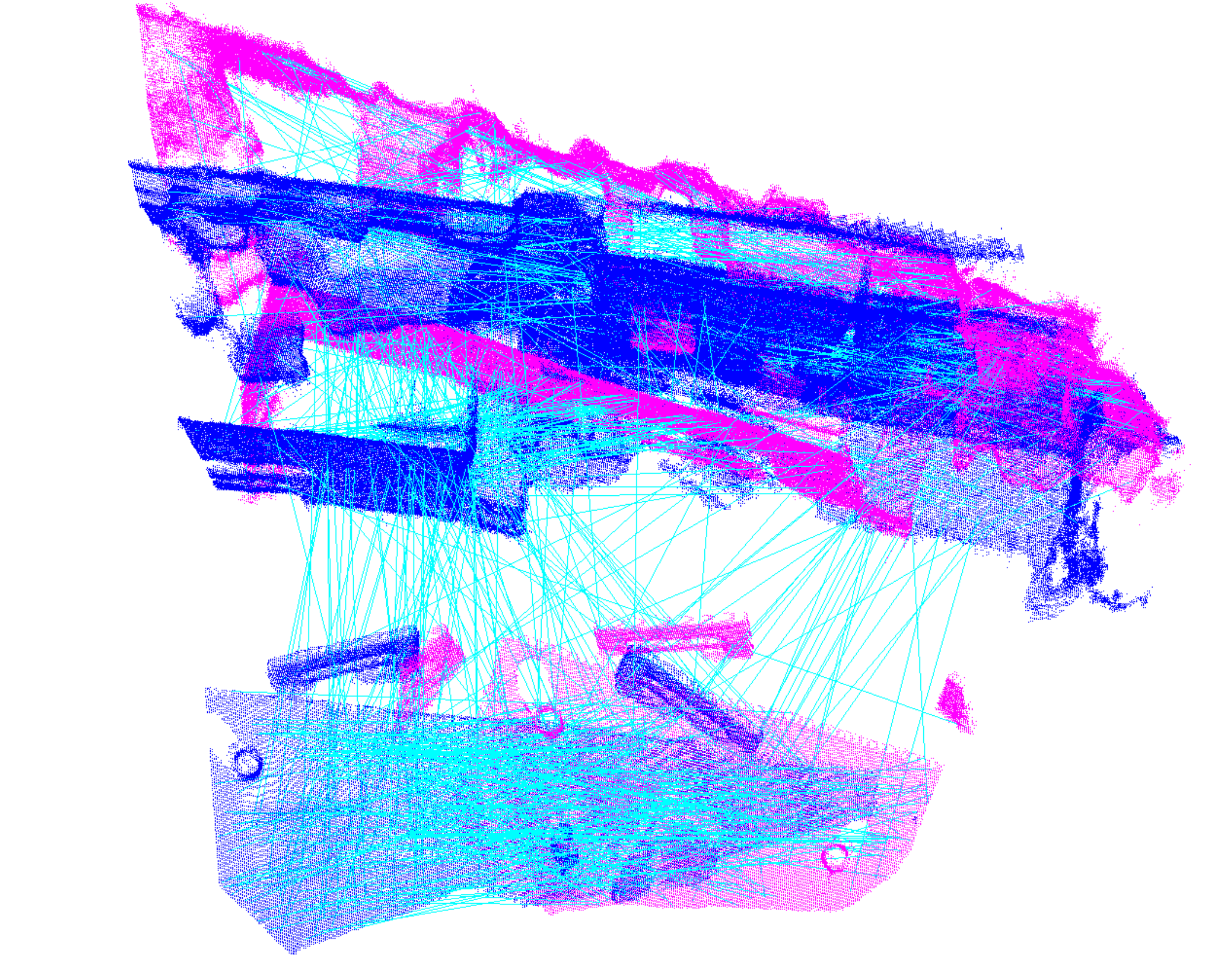}
\end{minipage}\,\,
& &
\,\,
\begin{minipage}[t]{0.19\linewidth}
\centering
\includegraphics[width=.48\linewidth]{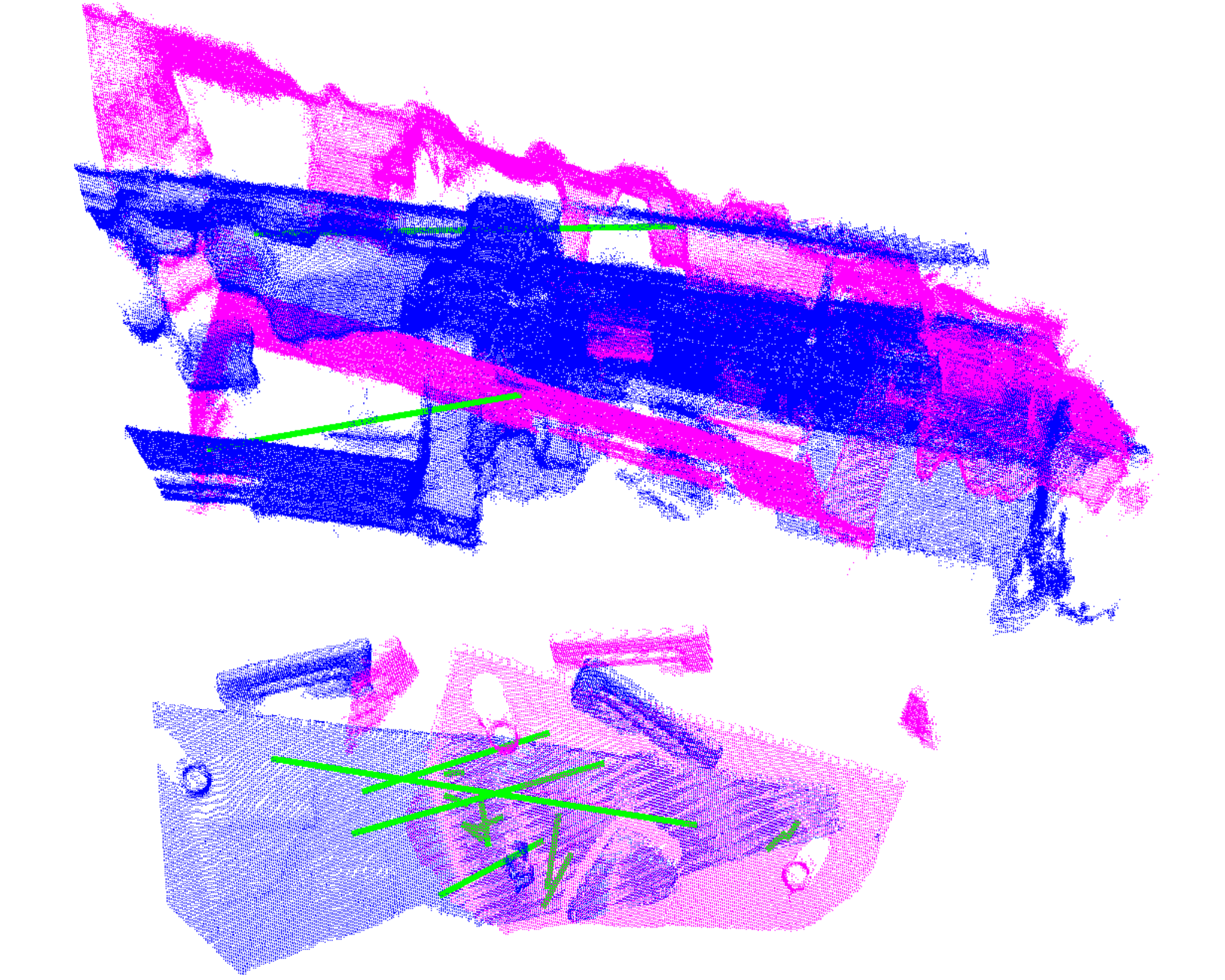}
\includegraphics[width=.48\linewidth]{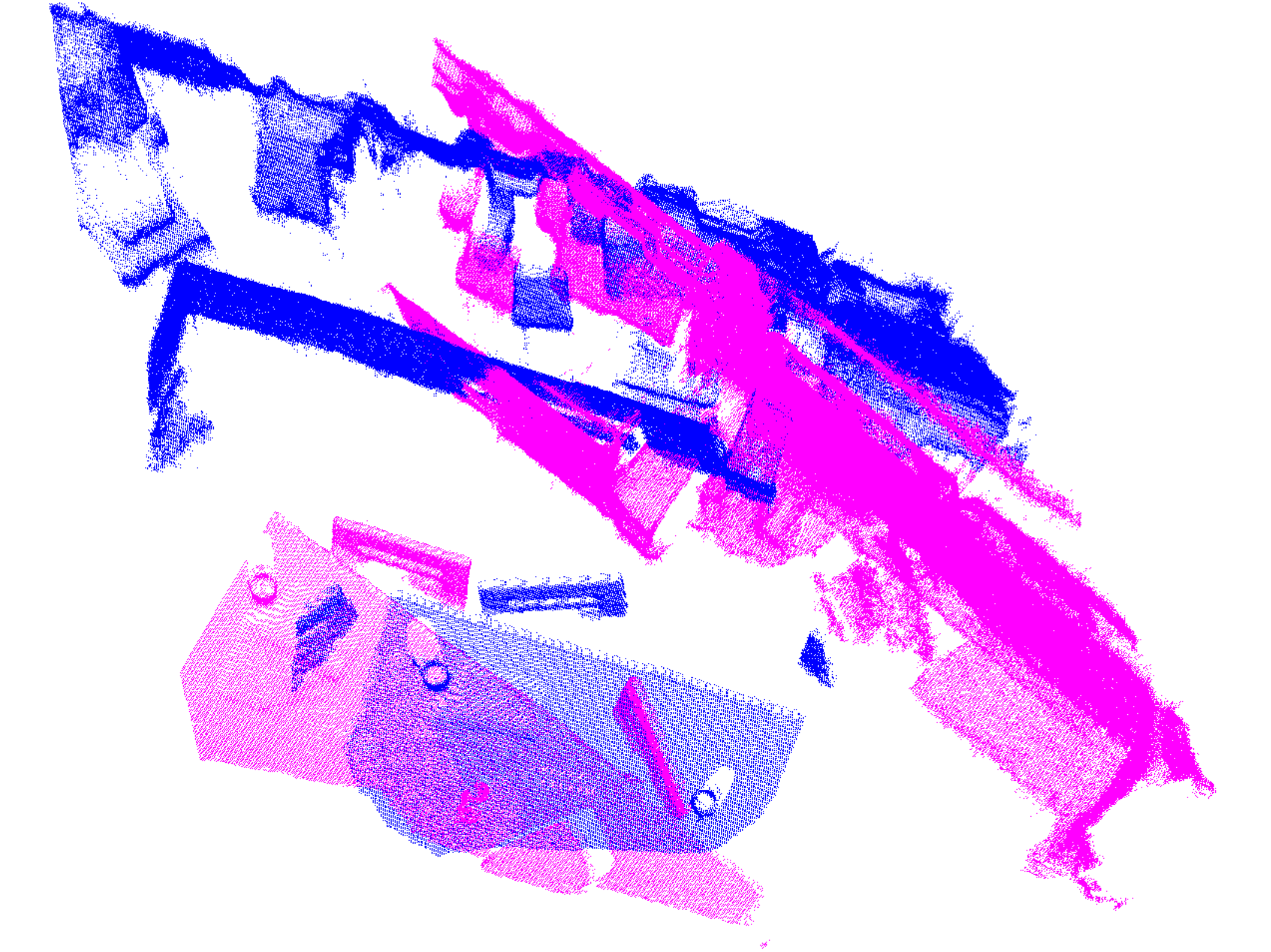}
\end{minipage}\,\,
&
\,\,
\begin{minipage}[t]{0.19\linewidth}
\centering
\includegraphics[width=.48\linewidth]{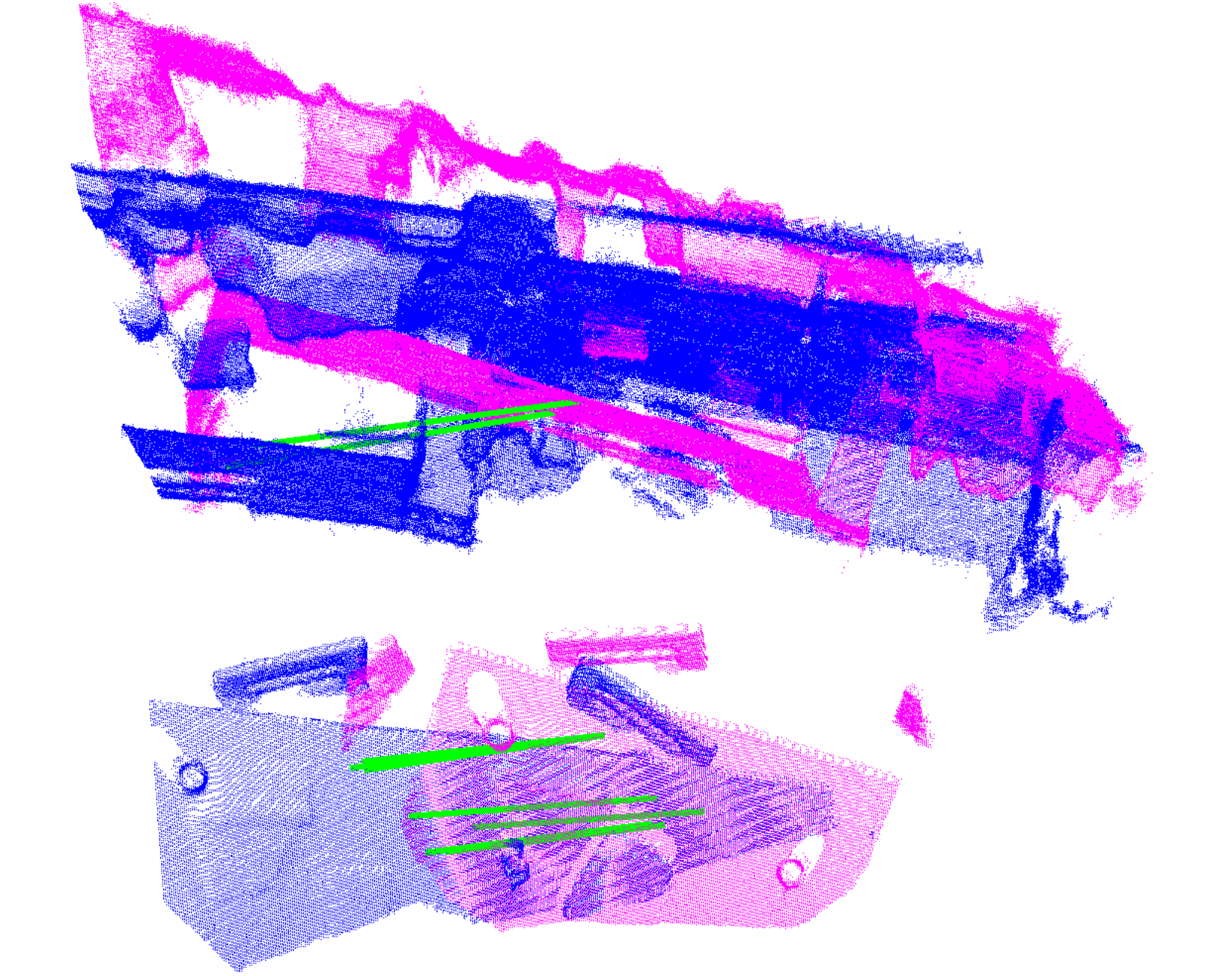}
\includegraphics[width=.48\linewidth]{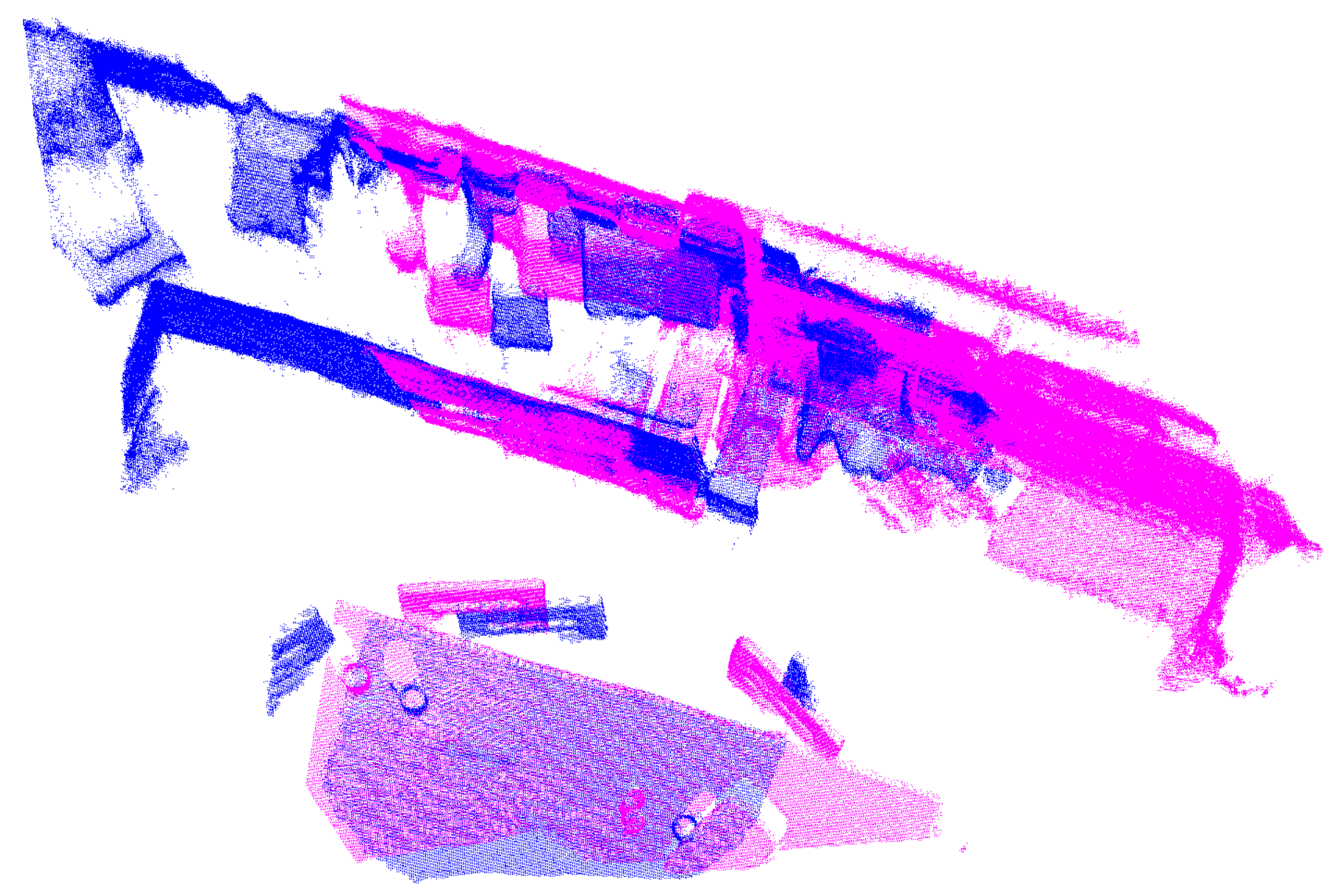}
\end{minipage}\,\,
&
\,\,
\begin{minipage}[t]{0.19\linewidth}
\centering
\includegraphics[width=.48\linewidth]{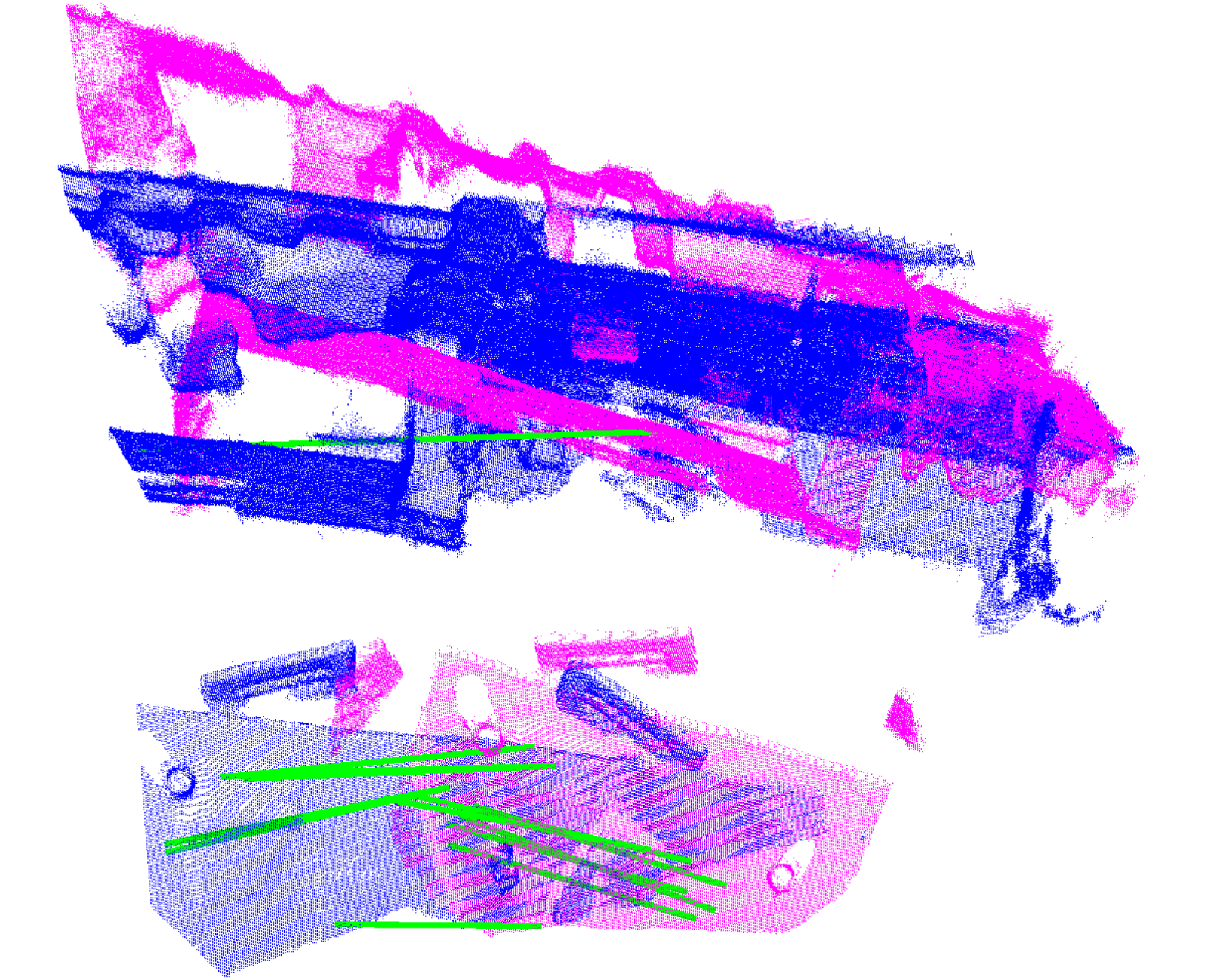}
\includegraphics[width=.48\linewidth]{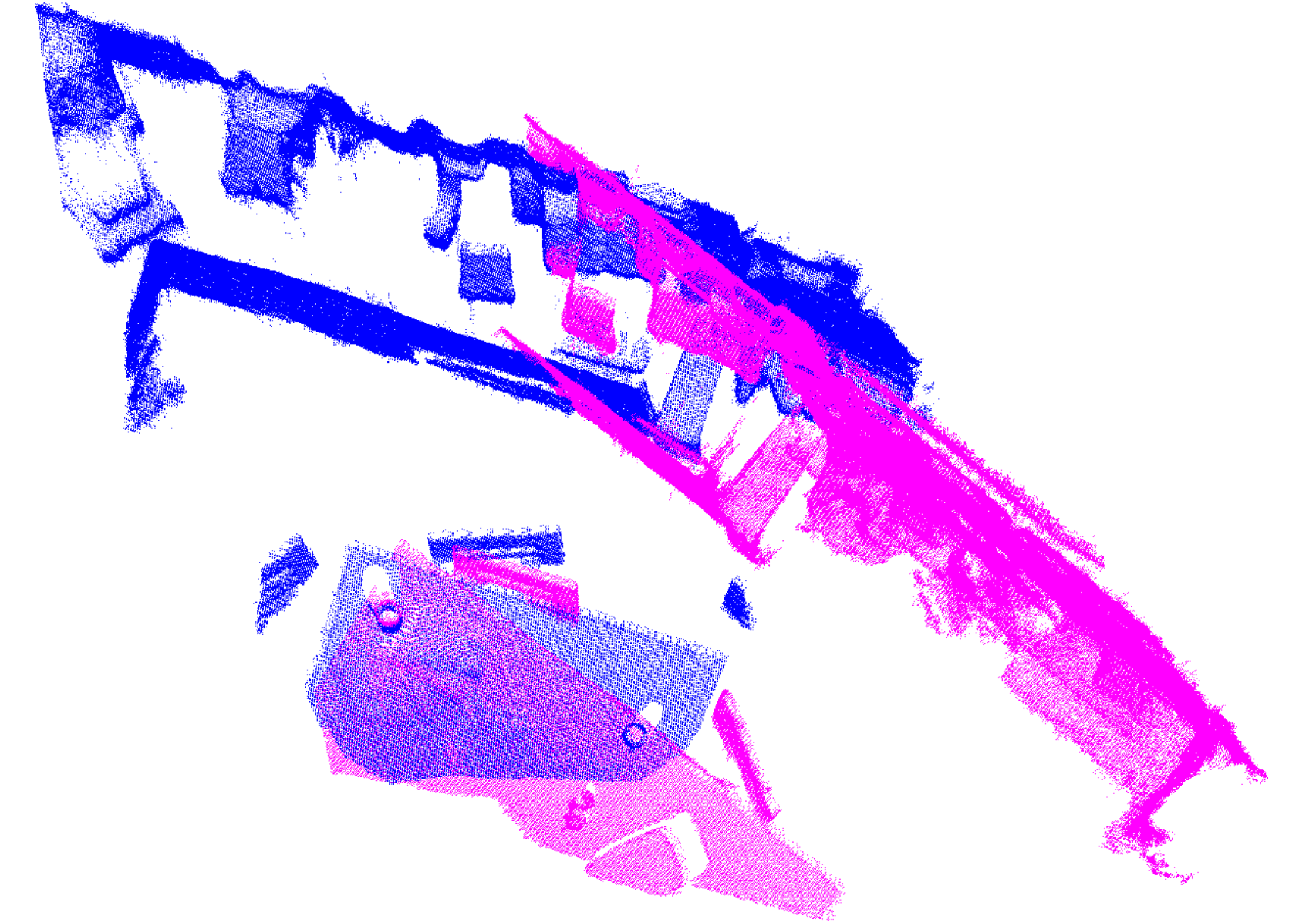}
\end{minipage}\,\,
&
\,\,
\begin{minipage}[t]{0.19\linewidth}
\centering
\includegraphics[width=.48\linewidth]{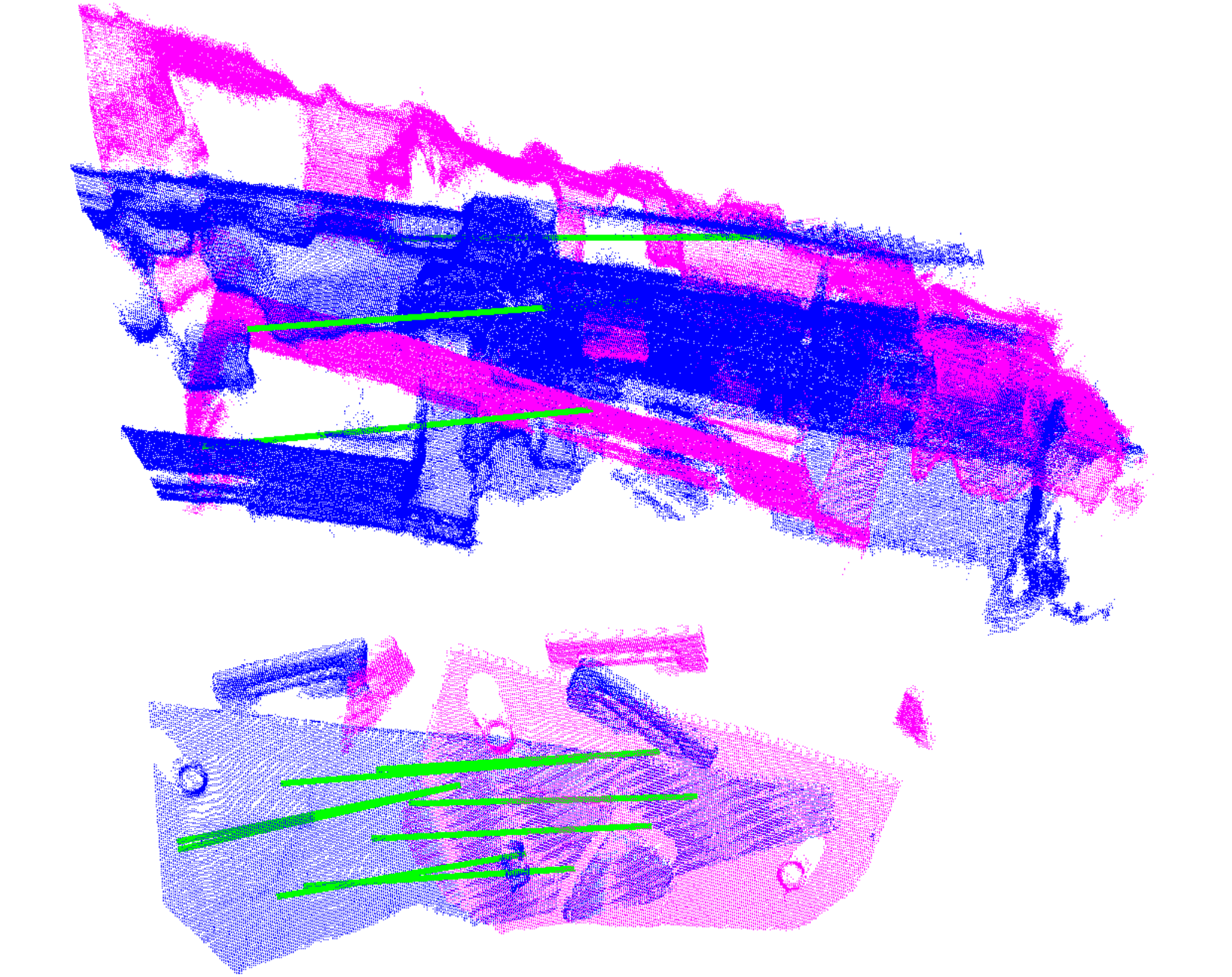}
\includegraphics[width=.48\linewidth]{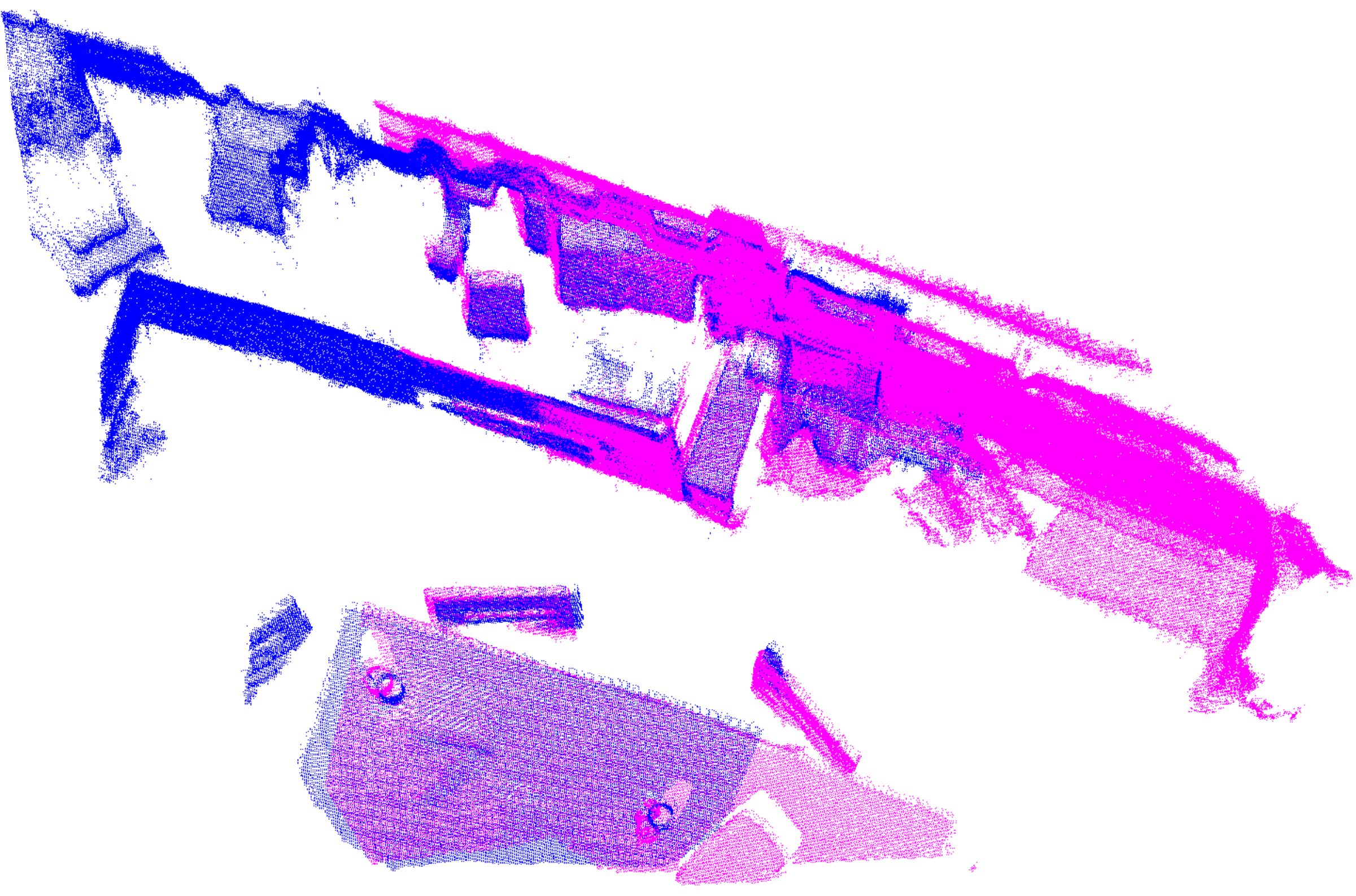}
\end{minipage}\,\,
\\
\rotatebox{90}{\,\,\footnotesize{\textit{Scene 11-12}}\,}\,
&
\,\,
\begin{minipage}[t]{0.1\linewidth}
\centering
\includegraphics[width=1\linewidth]{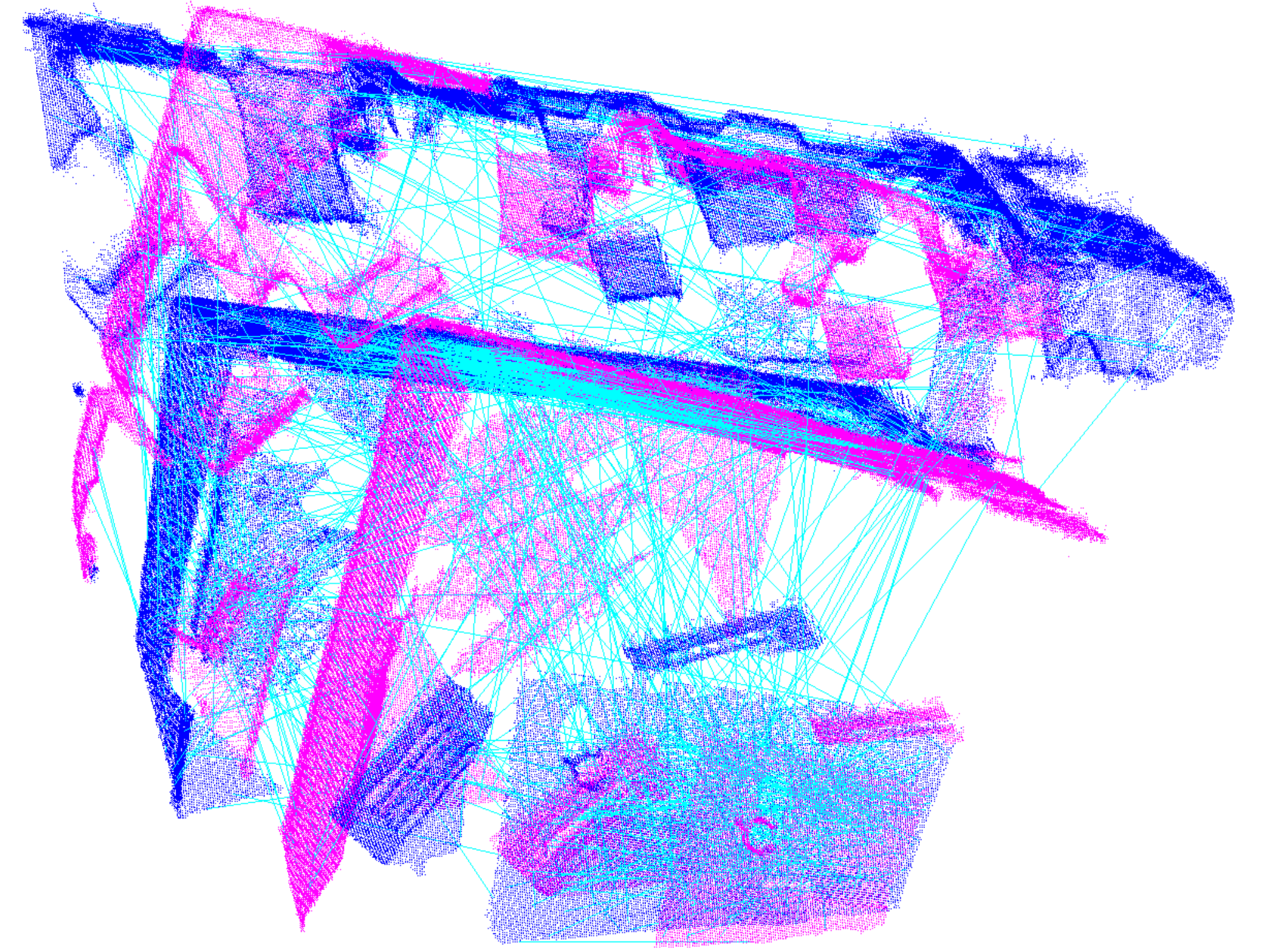}
\end{minipage}\,\,
& &
\,\,
\begin{minipage}[t]{0.19\linewidth}
\centering
\includegraphics[width=.48\linewidth]{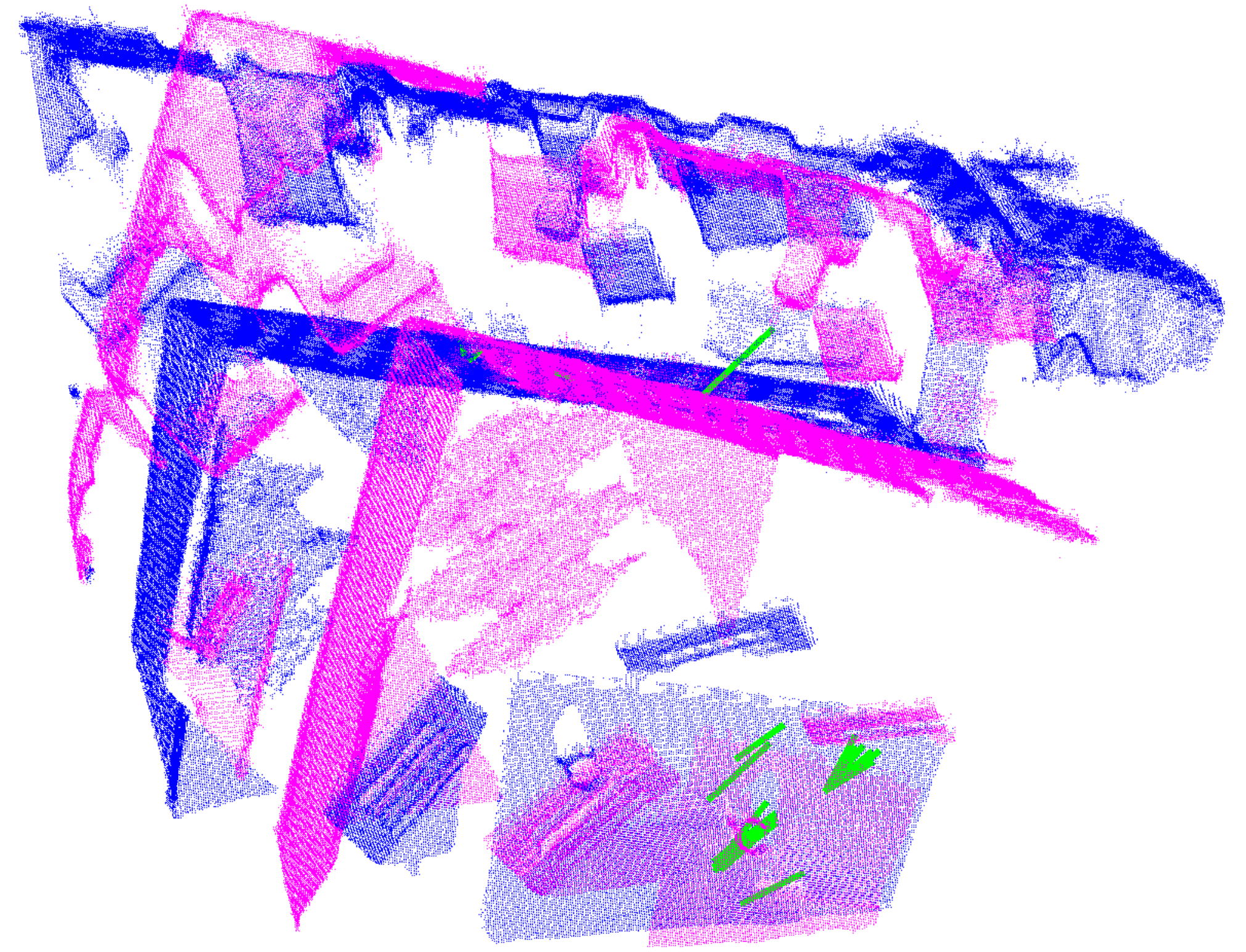}
\includegraphics[width=.48\linewidth]{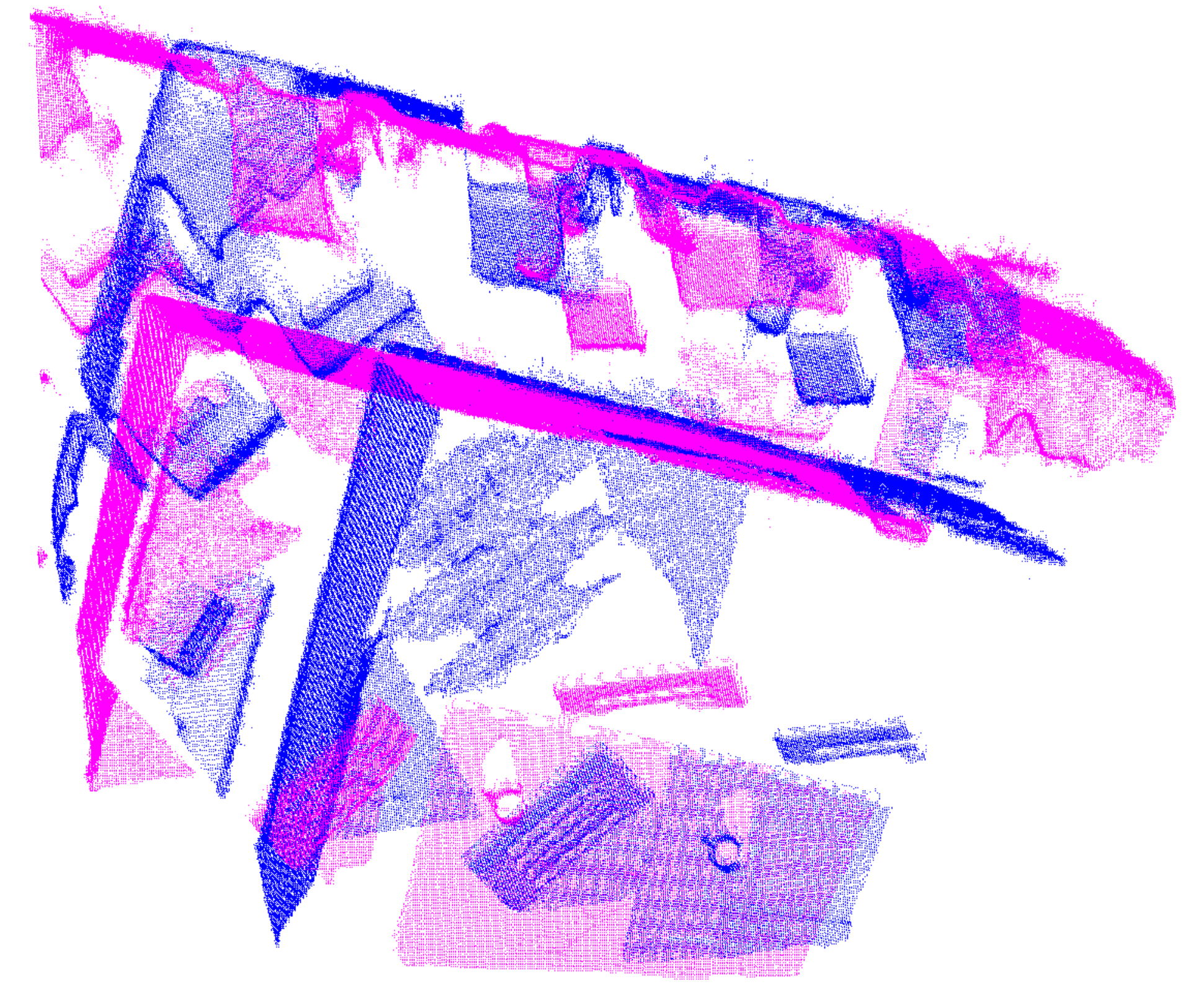}
\end{minipage}\,\,
&
\,\,
\begin{minipage}[t]{0.19\linewidth}
\centering
\includegraphics[width=.48\linewidth]{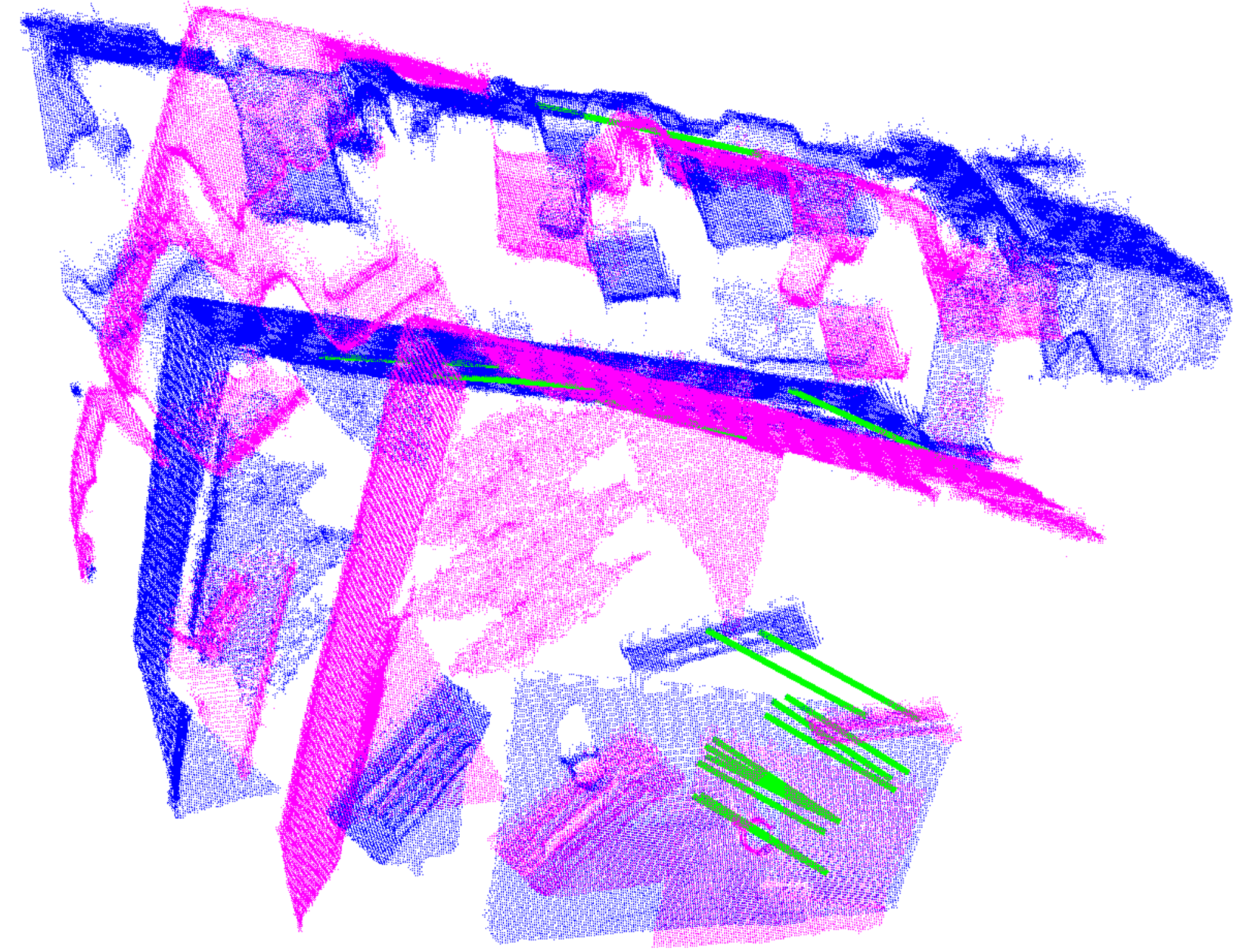}
\includegraphics[width=.48\linewidth]{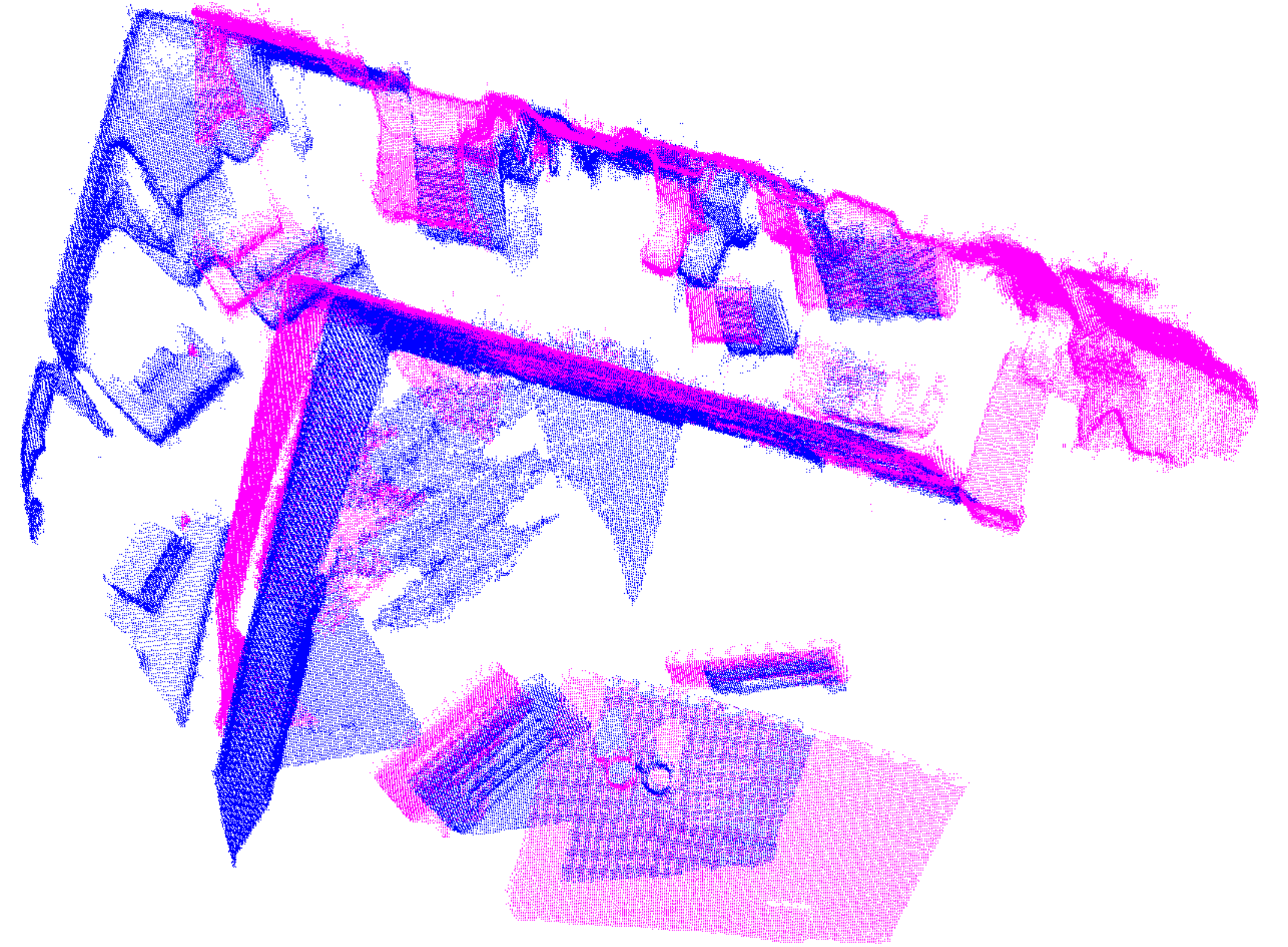}
\end{minipage}\,\,
&
\,\,
\begin{minipage}[t]{0.19\linewidth}
\centering
\includegraphics[width=.48\linewidth]{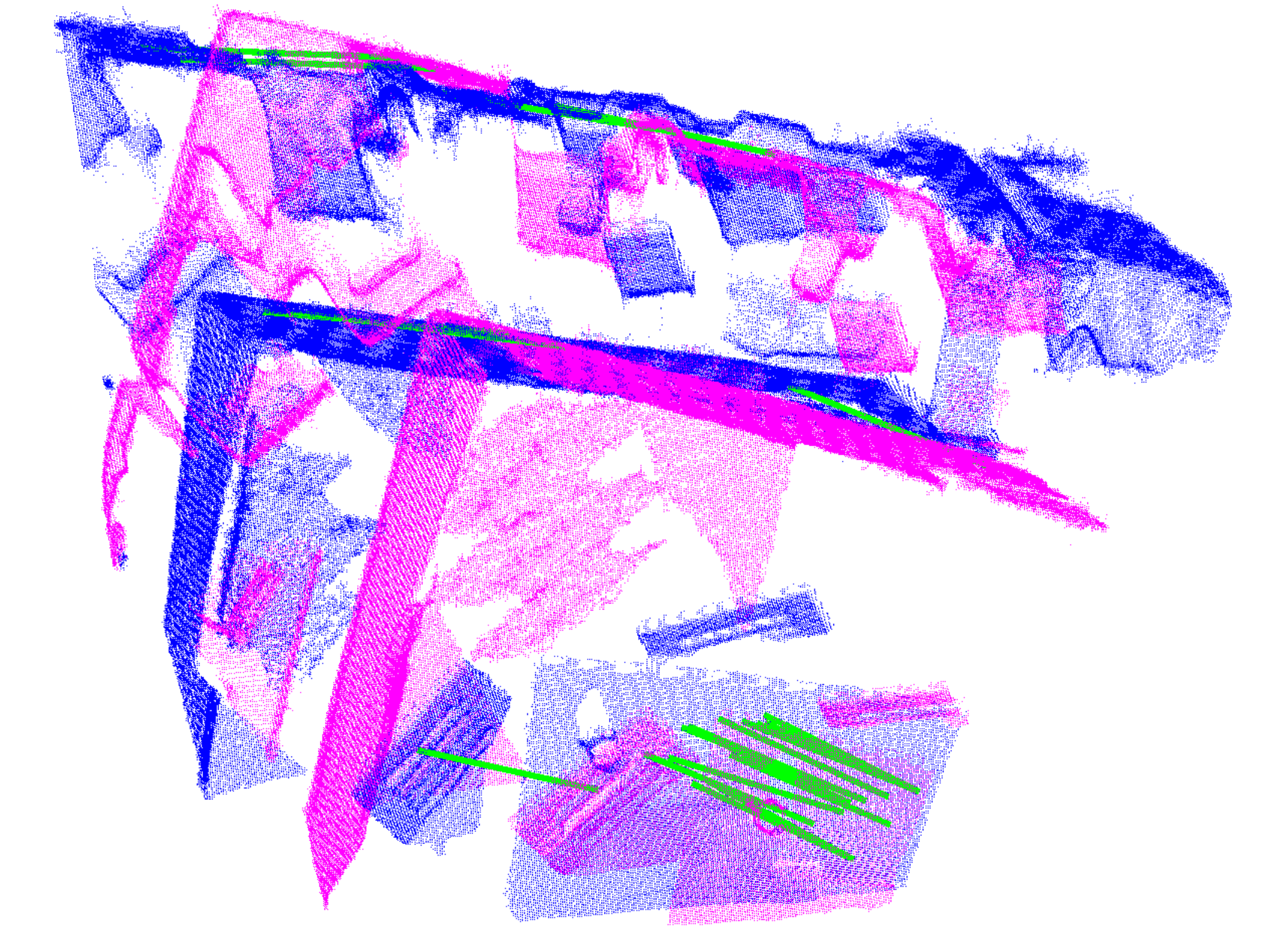}
\includegraphics[width=.48\linewidth]{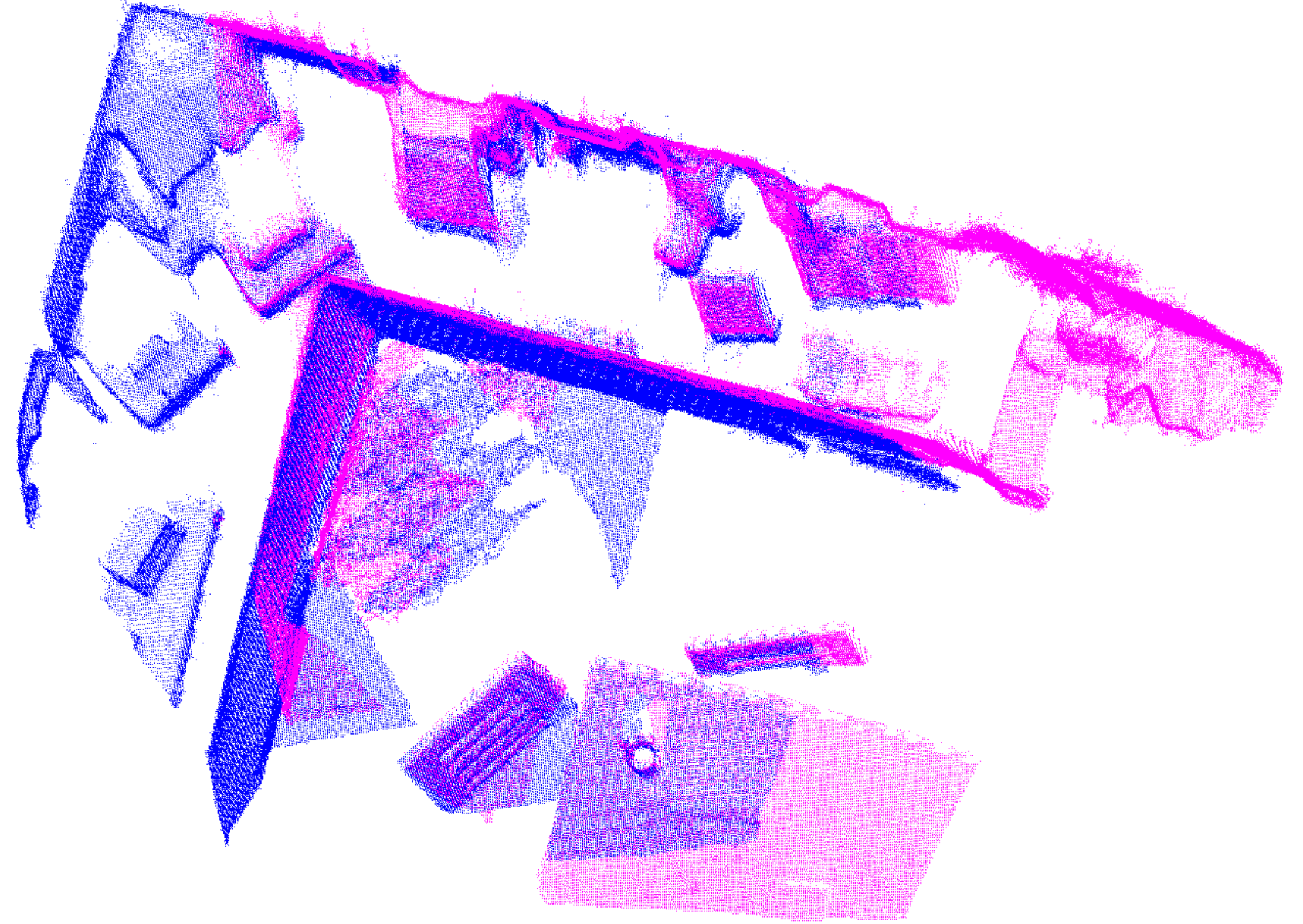}
\end{minipage}\,\,
&
\,\,
\begin{minipage}[t]{0.19\linewidth}
\centering
\includegraphics[width=.48\linewidth]{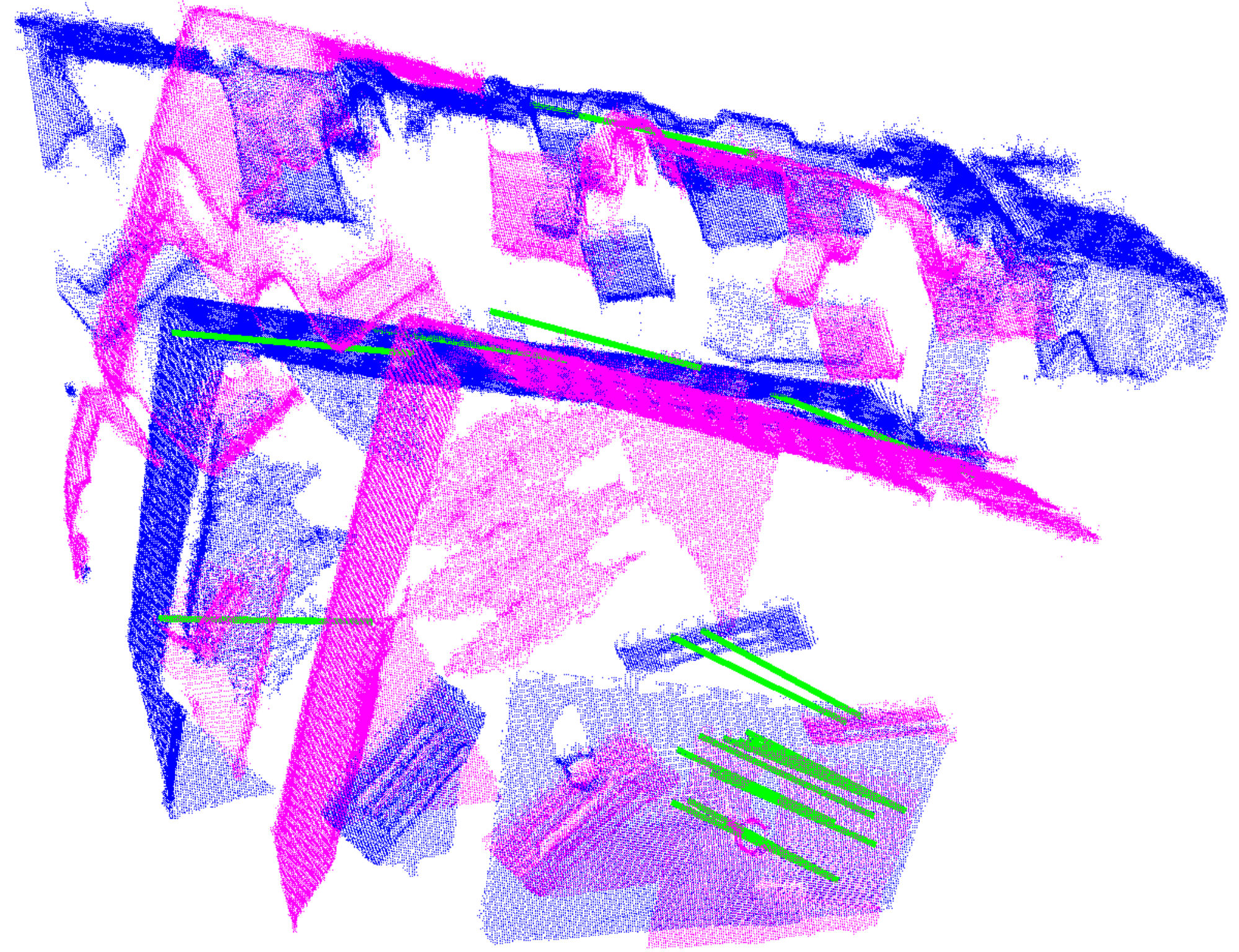}
\includegraphics[width=.48\linewidth]{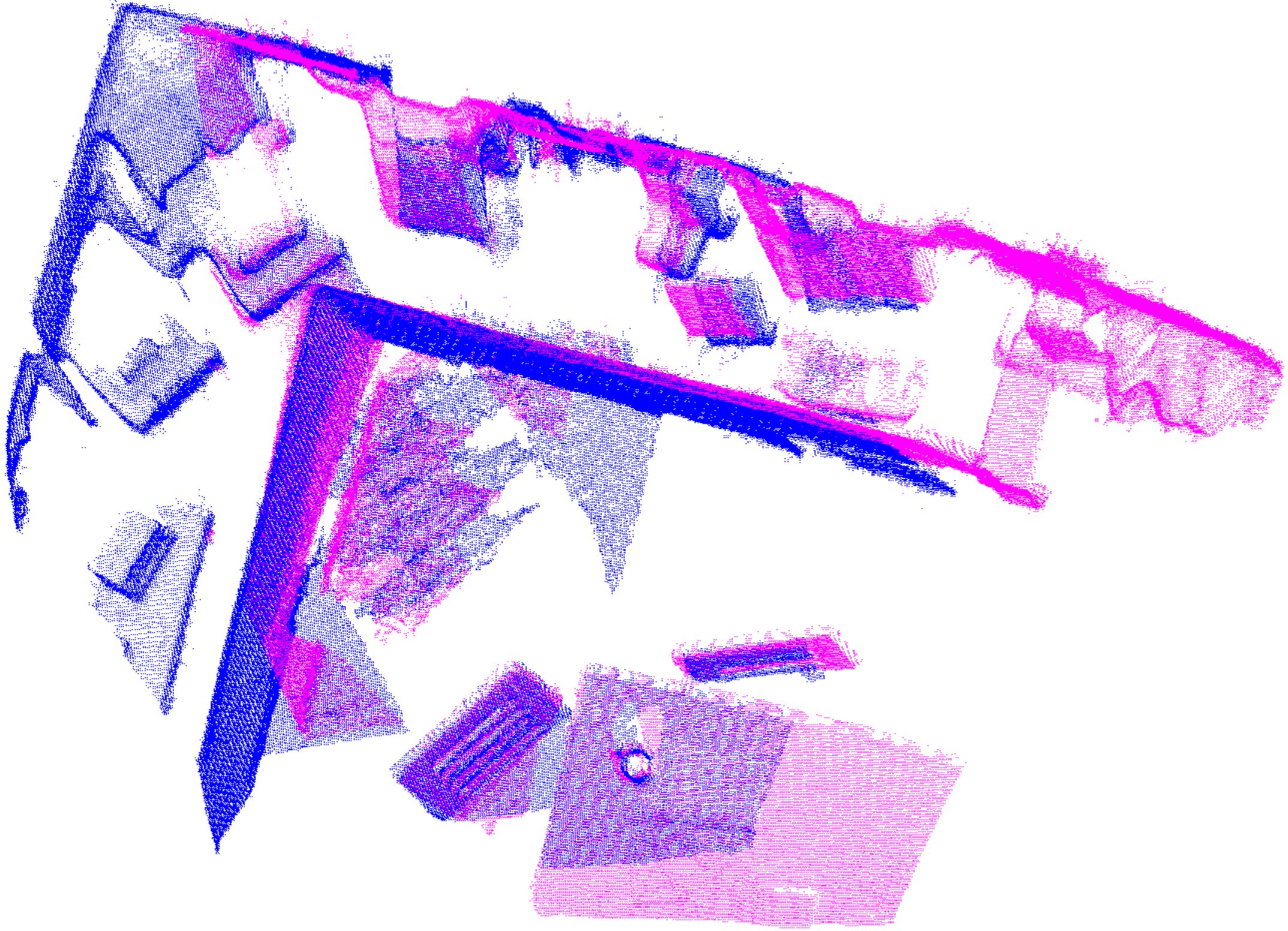}
\end{minipage}\,\,
\\
\rotatebox{90}{\,\,\footnotesize{\textit{Scene 37-38}}\,}\,
&
\,\,
\begin{minipage}[t]{0.1\linewidth}
\centering
\includegraphics[width=1\linewidth]{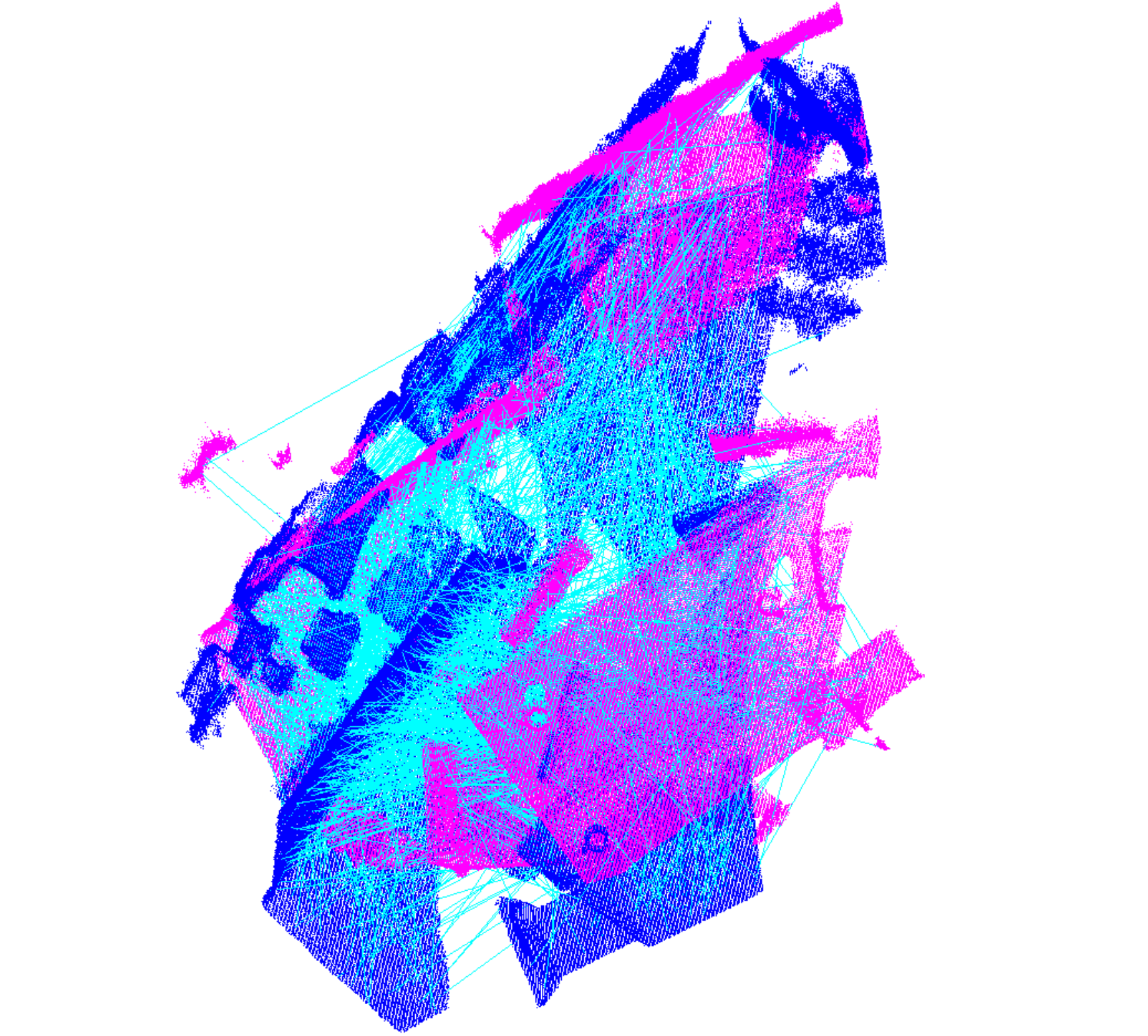}
\end{minipage}\,\,
& &
\,\,
\begin{minipage}[t]{0.19\linewidth}
\centering
\includegraphics[width=.48\linewidth]{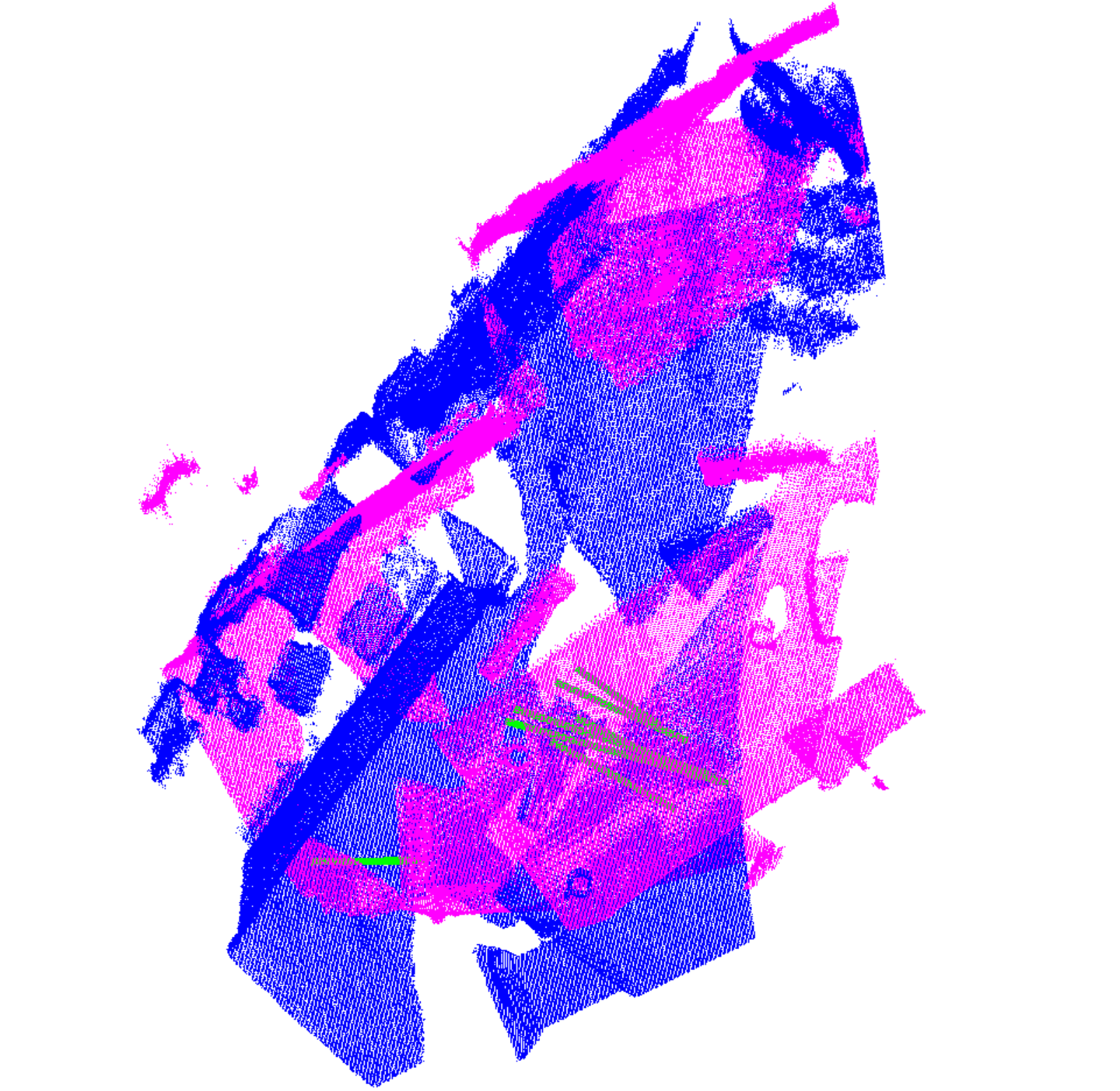}
\includegraphics[width=.48\linewidth]{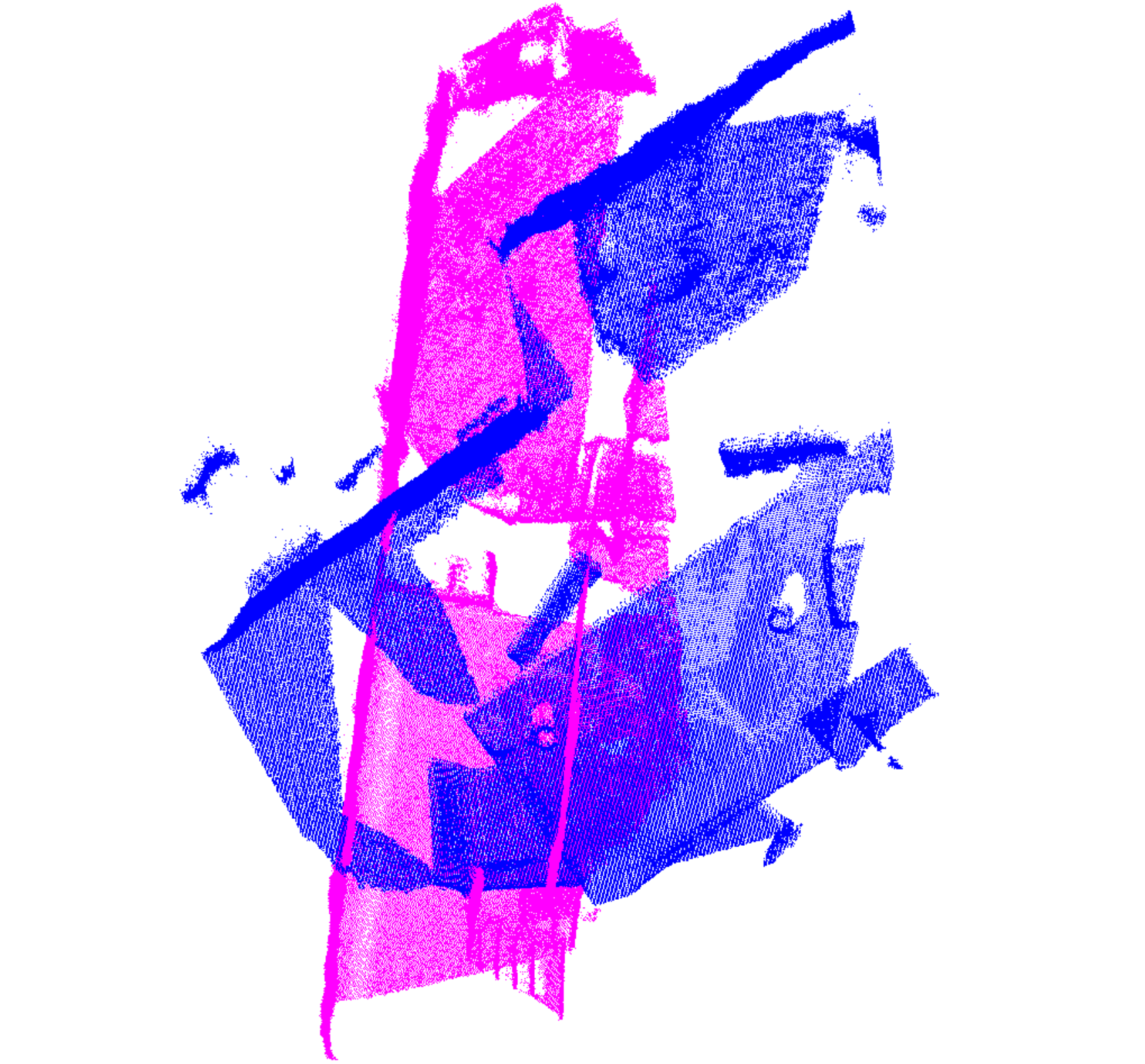}
\end{minipage}\,\,
&
\,\,
\begin{minipage}[t]{0.19\linewidth}
\centering
\includegraphics[width=.48\linewidth]{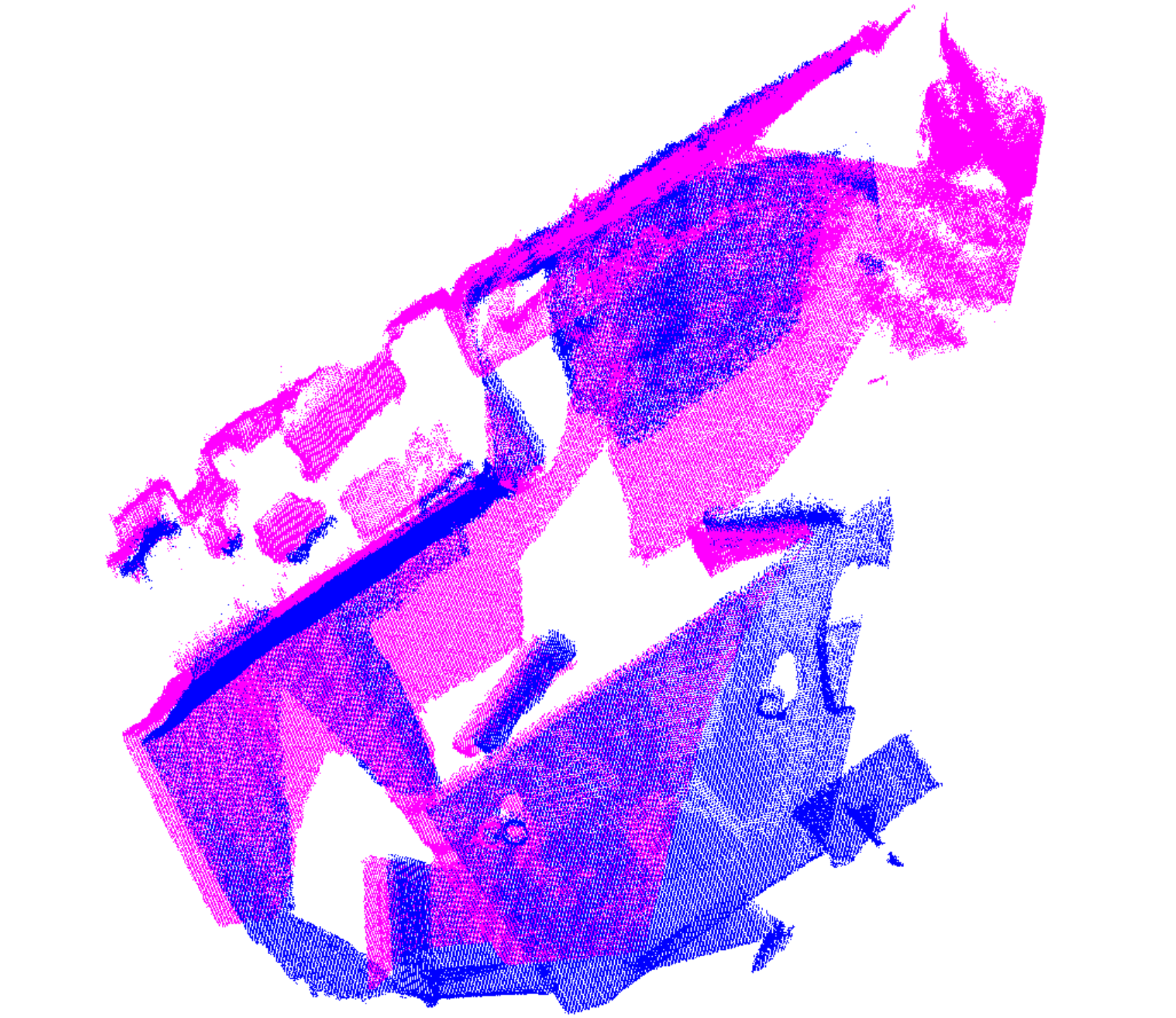}
\includegraphics[width=.48\linewidth]{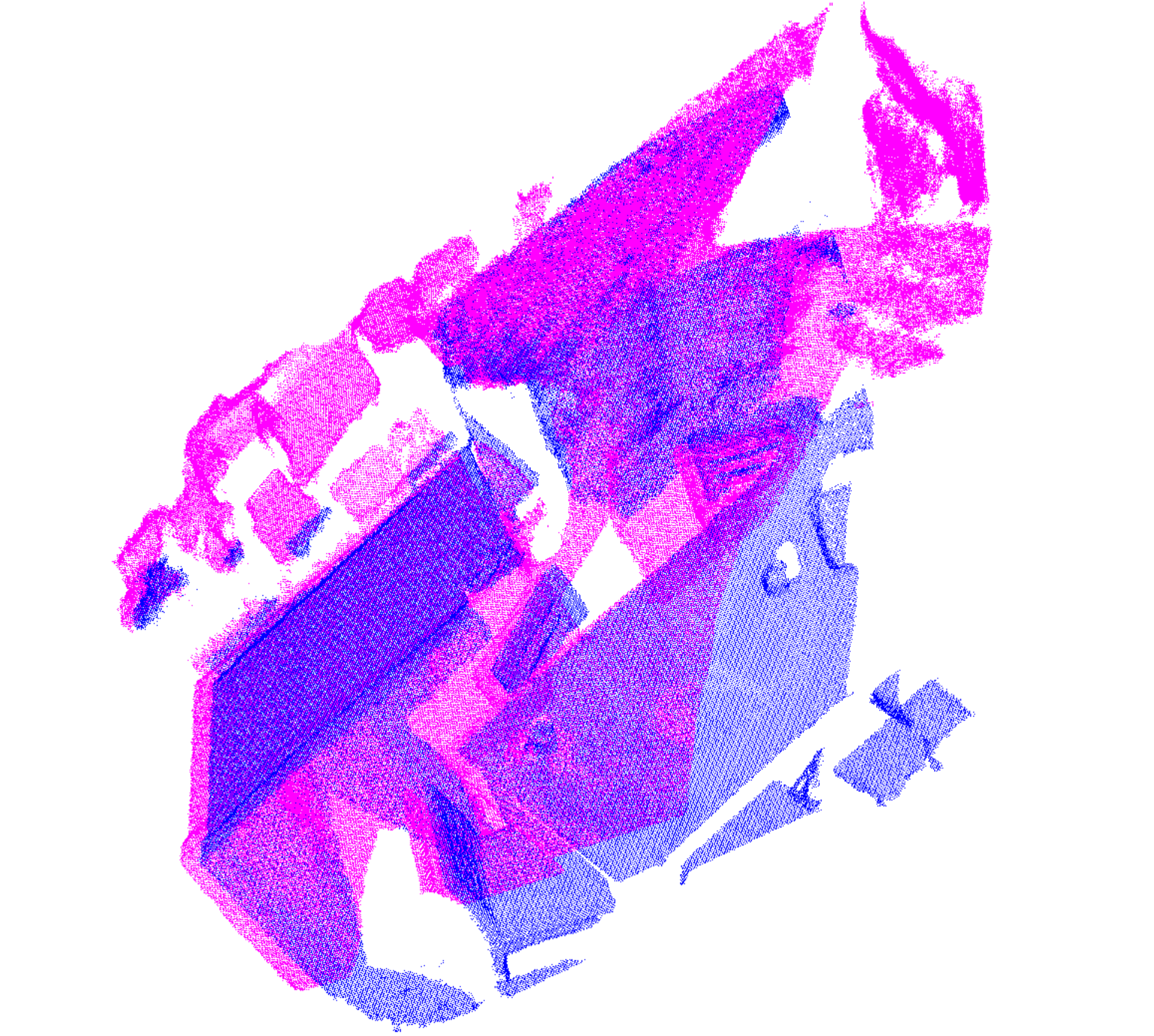}
\end{minipage}\,\,
&
\,\,
\begin{minipage}[t]{0.19\linewidth}
\centering
\includegraphics[width=.48\linewidth]{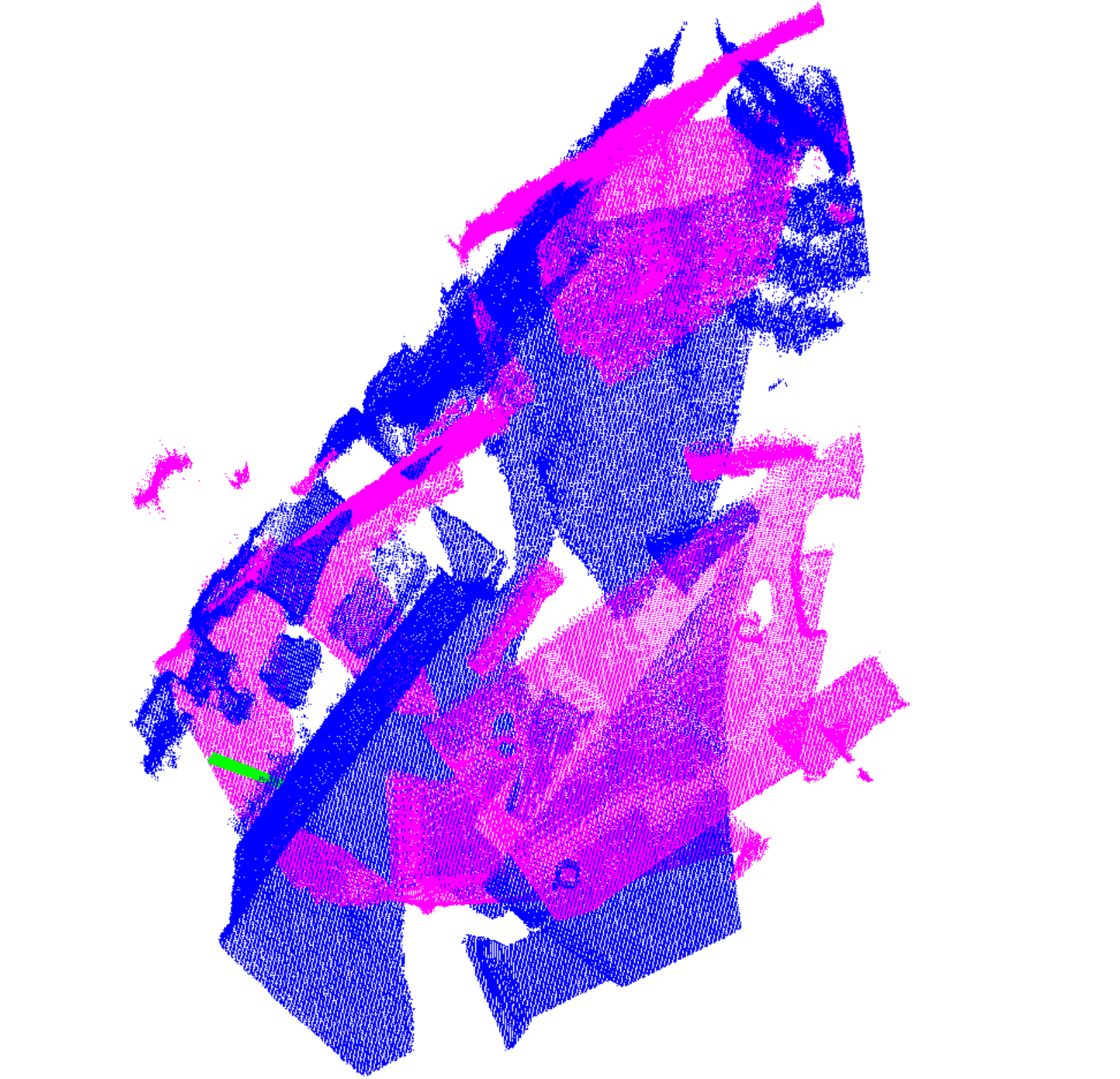}
\includegraphics[width=.48\linewidth]{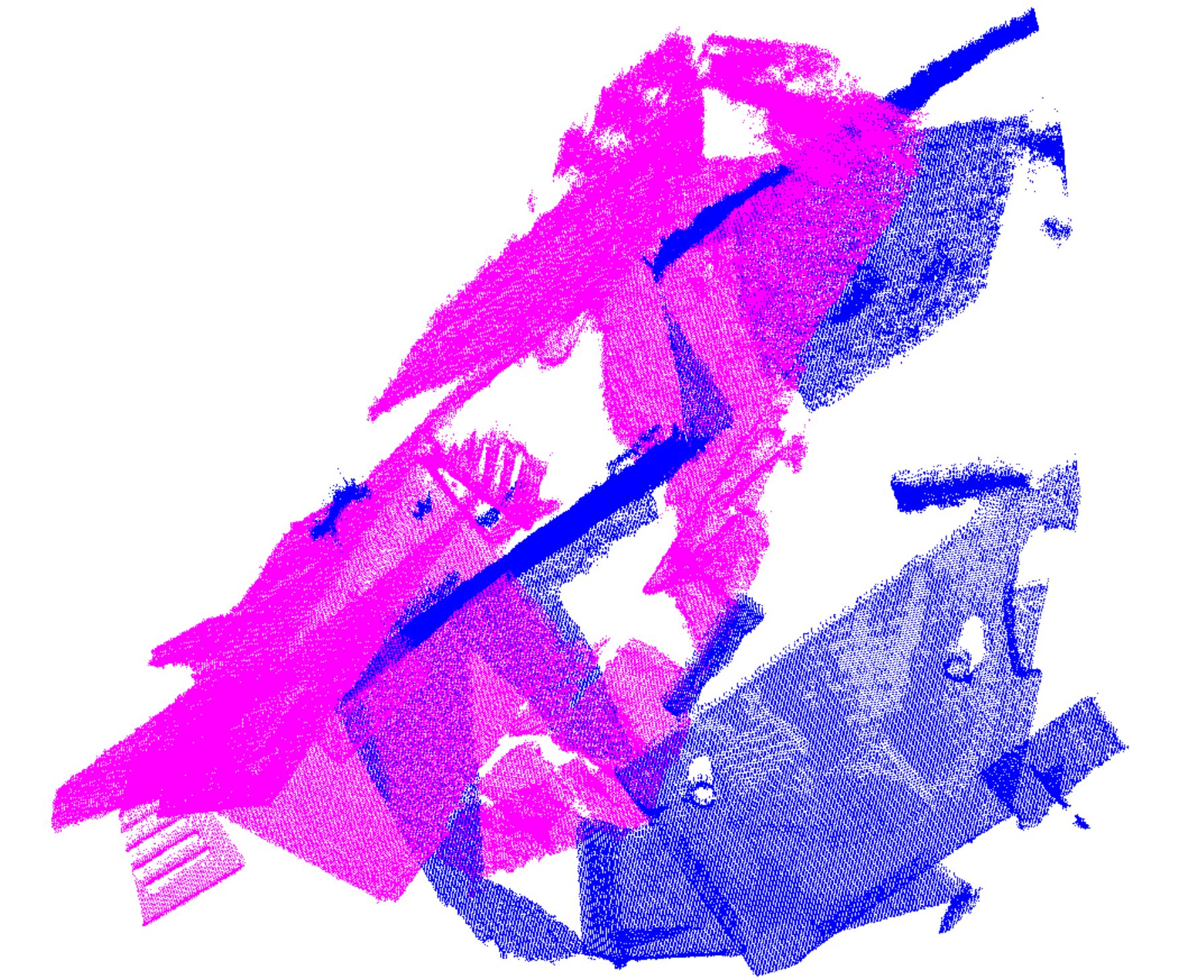}
\end{minipage}\,\,
&
\,\,
\begin{minipage}[t]{0.19\linewidth}
\centering
\includegraphics[width=.48\linewidth]{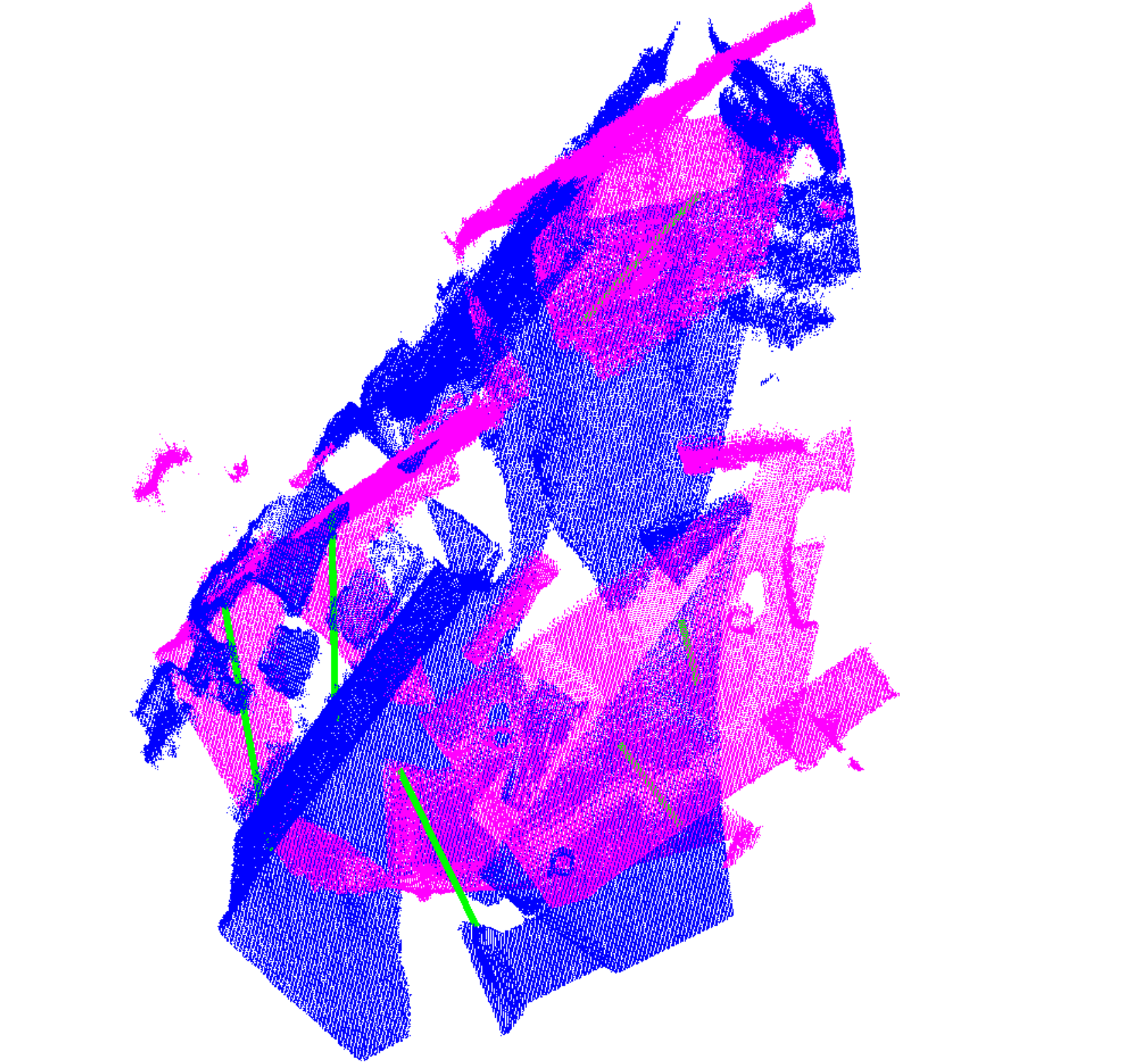}
\includegraphics[width=.48\linewidth]{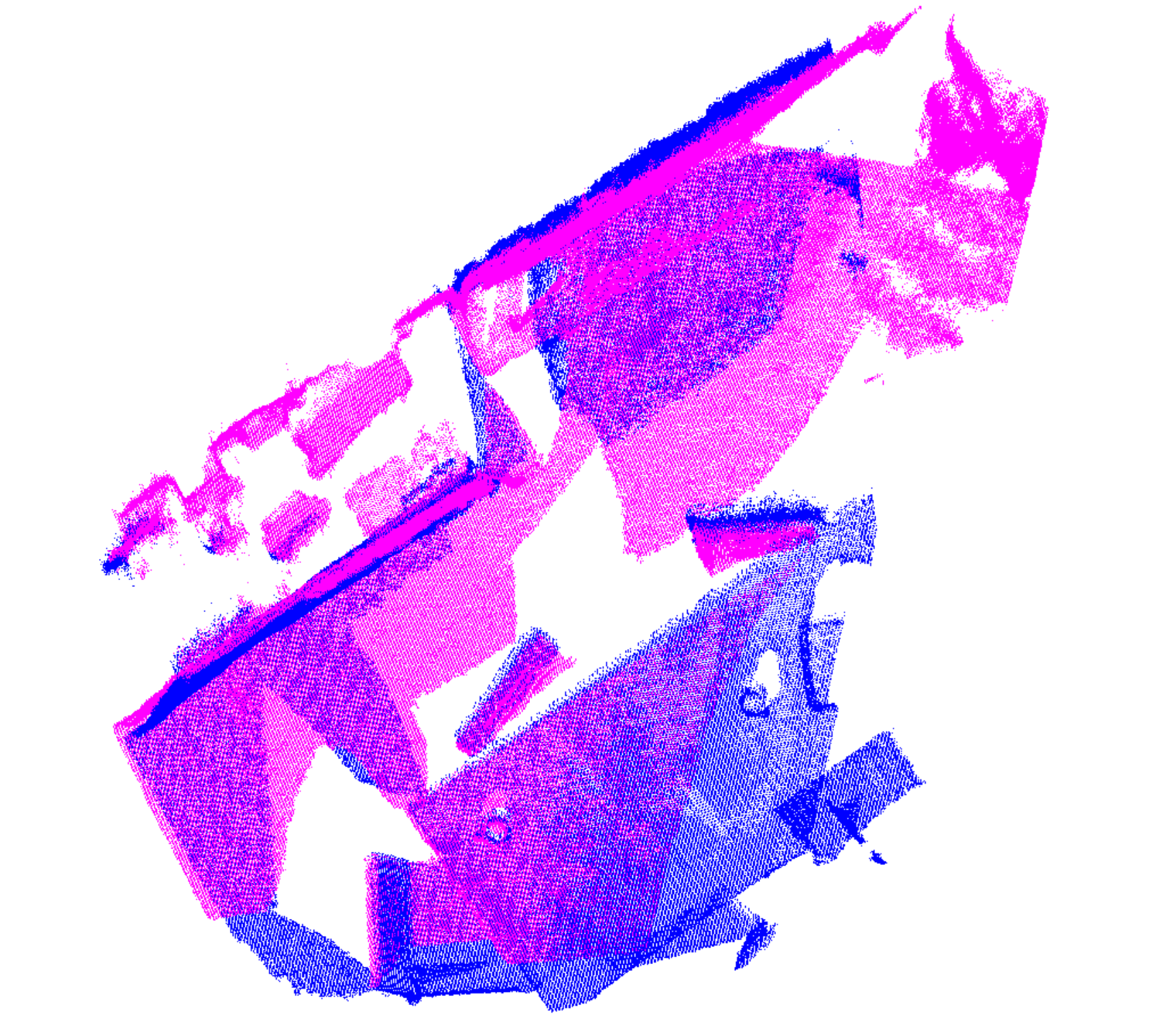}
\end{minipage}\,\,
\\
\rotatebox{90}{\,\,\footnotesize{\textit{Scene 58-59}}\,}\,
&
\,\,
\begin{minipage}[t]{0.1\linewidth}
\centering
\includegraphics[width=1\linewidth]{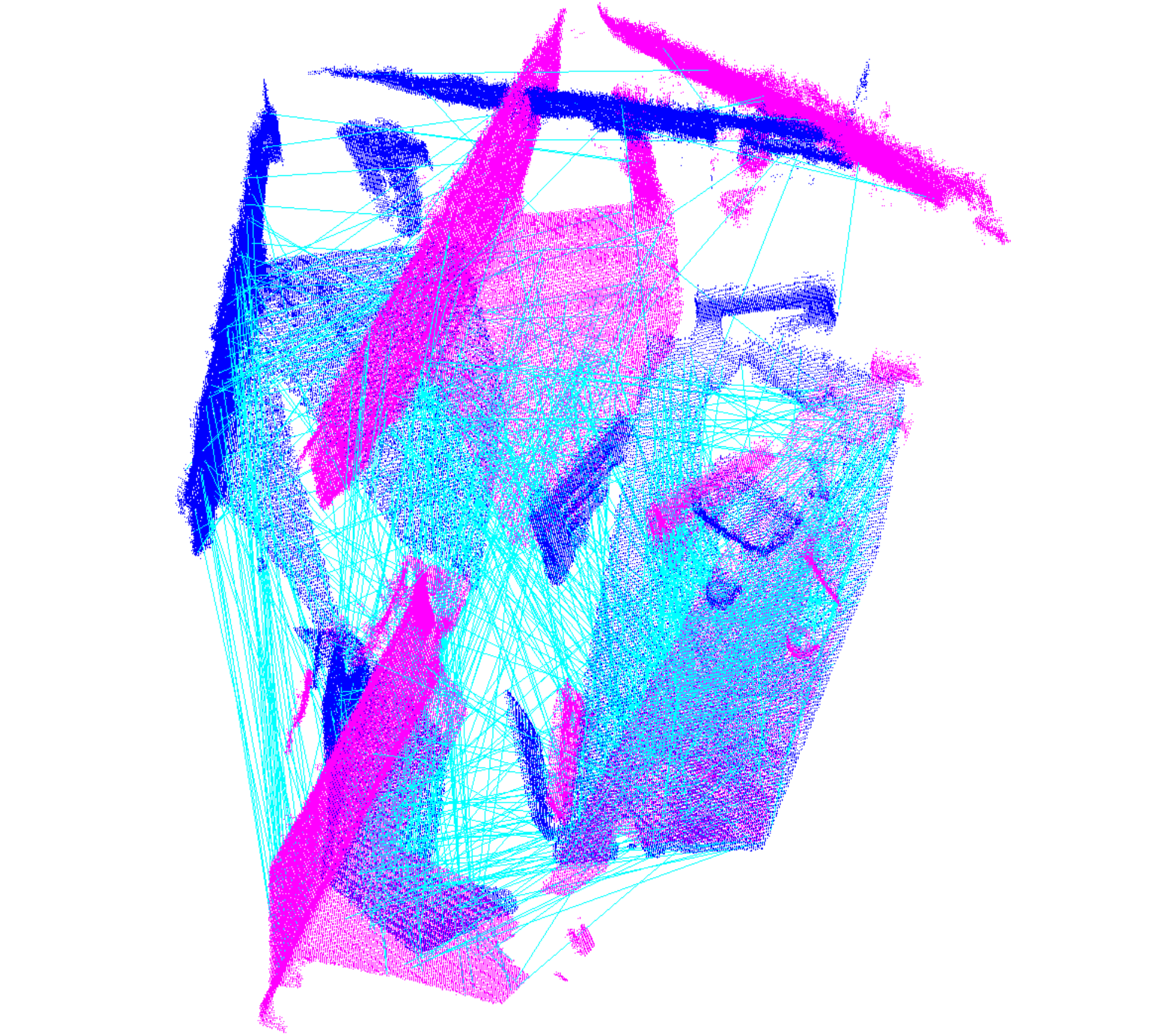}
\end{minipage}\,\,
& &
\,\,
\begin{minipage}[t]{0.19\linewidth}
\centering
\includegraphics[width=.48\linewidth]{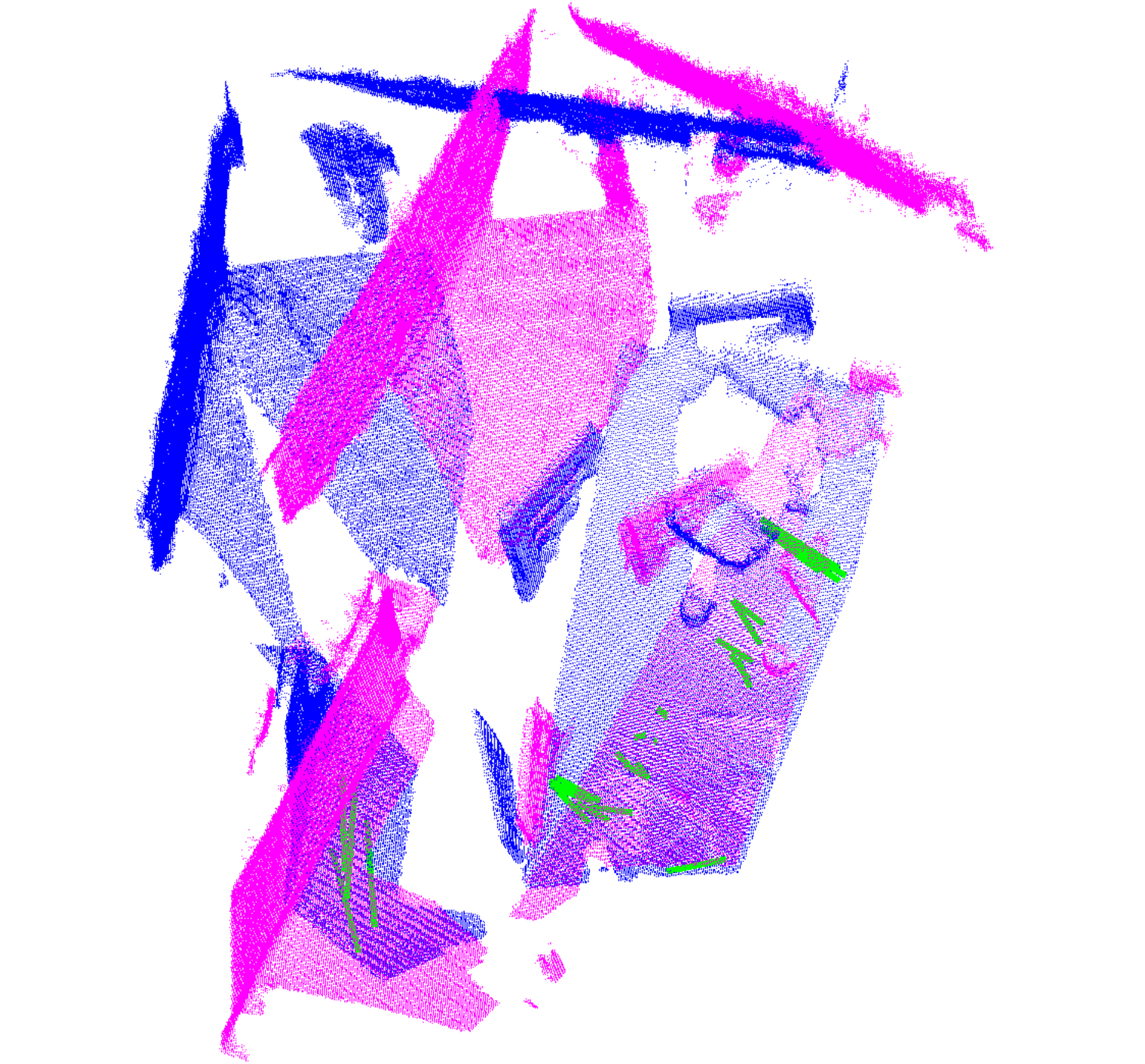}
\includegraphics[width=.48\linewidth]{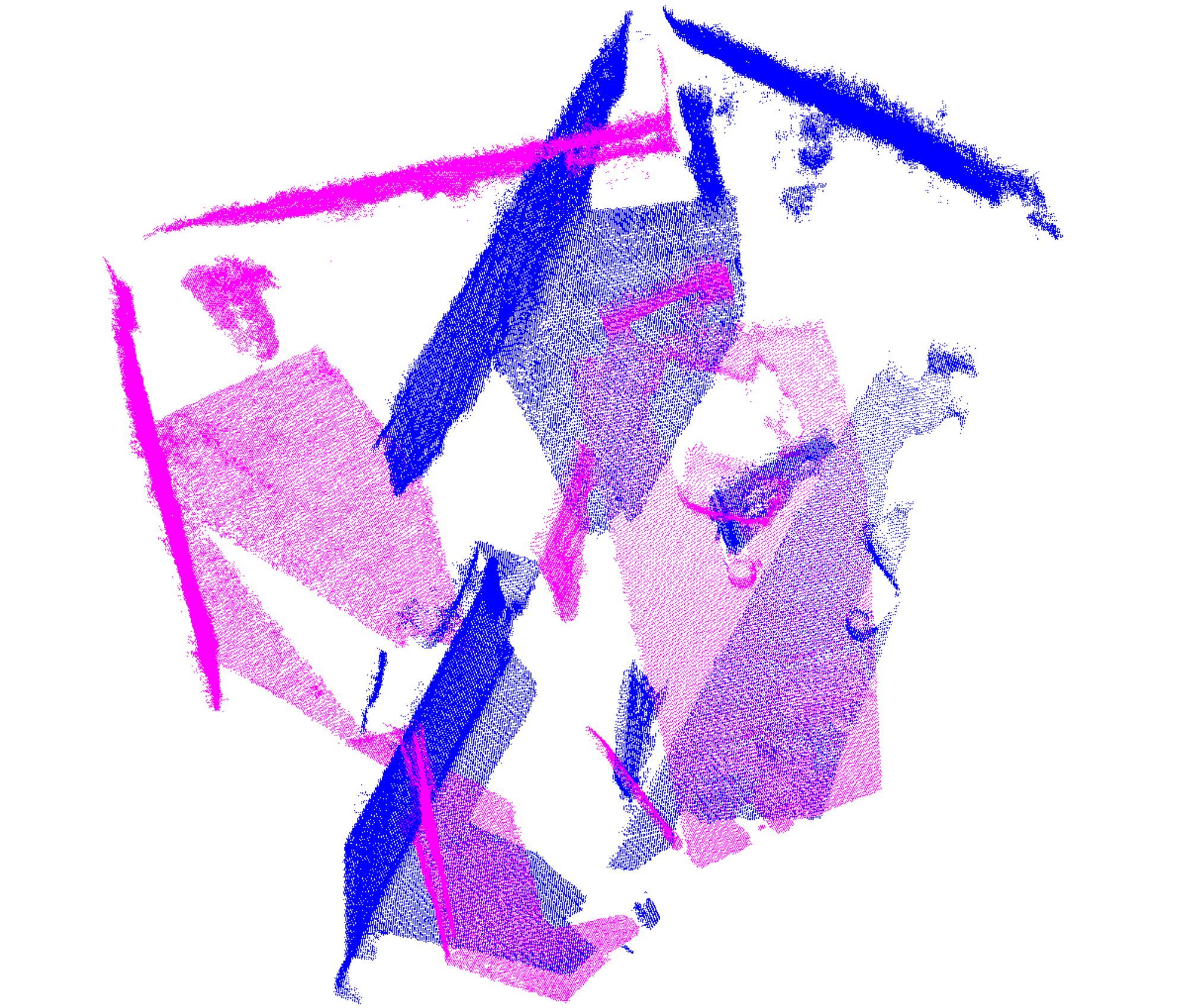}
\end{minipage}\,\,
&
\,\,
\begin{minipage}[t]{0.19\linewidth}
\centering
\includegraphics[width=.48\linewidth]{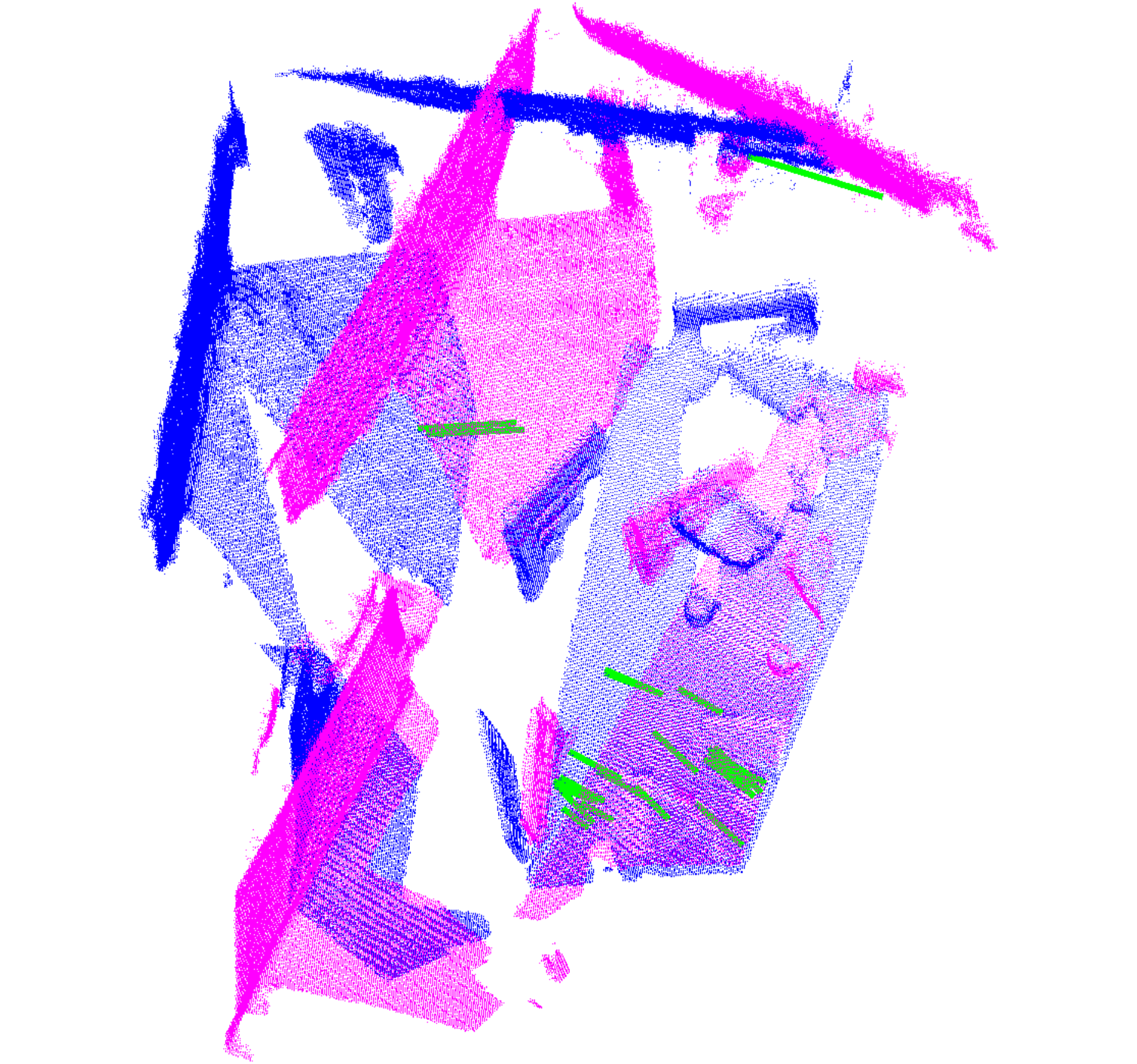}
\includegraphics[width=.48\linewidth]{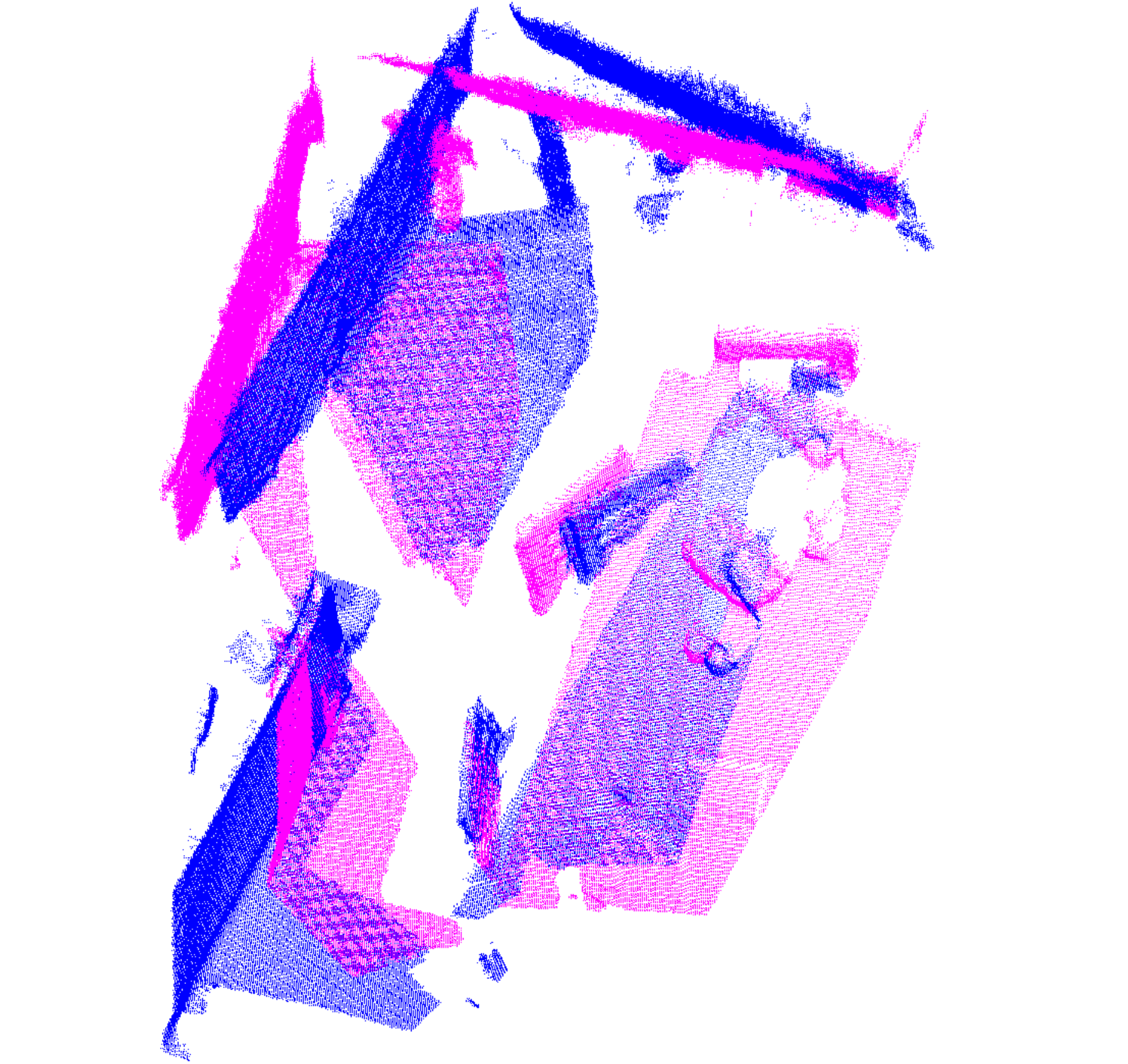}
\end{minipage}\,\,
&
\,\,
\begin{minipage}[t]{0.19\linewidth}
\centering
\includegraphics[width=.48\linewidth]{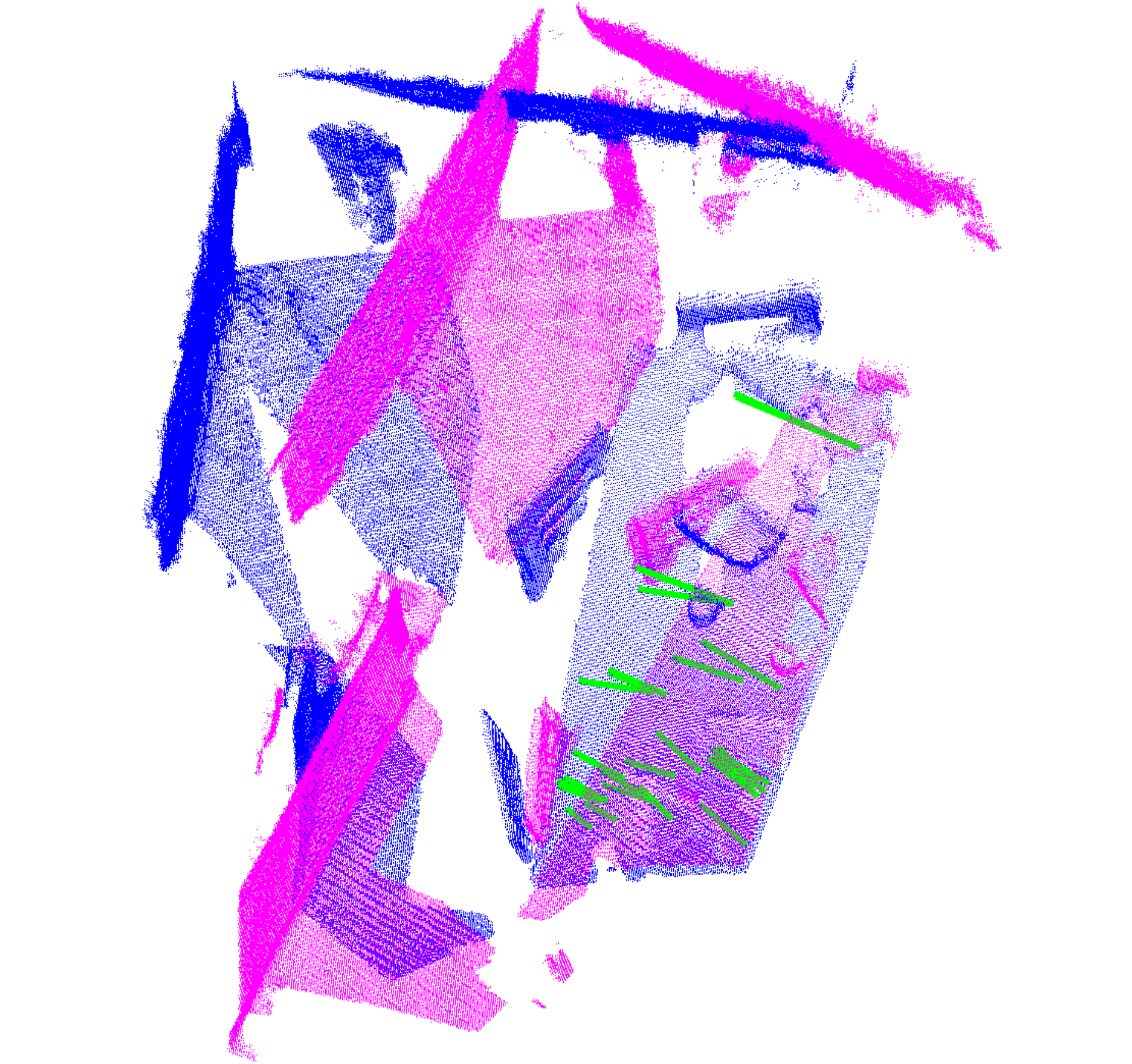}
\includegraphics[width=.48\linewidth]{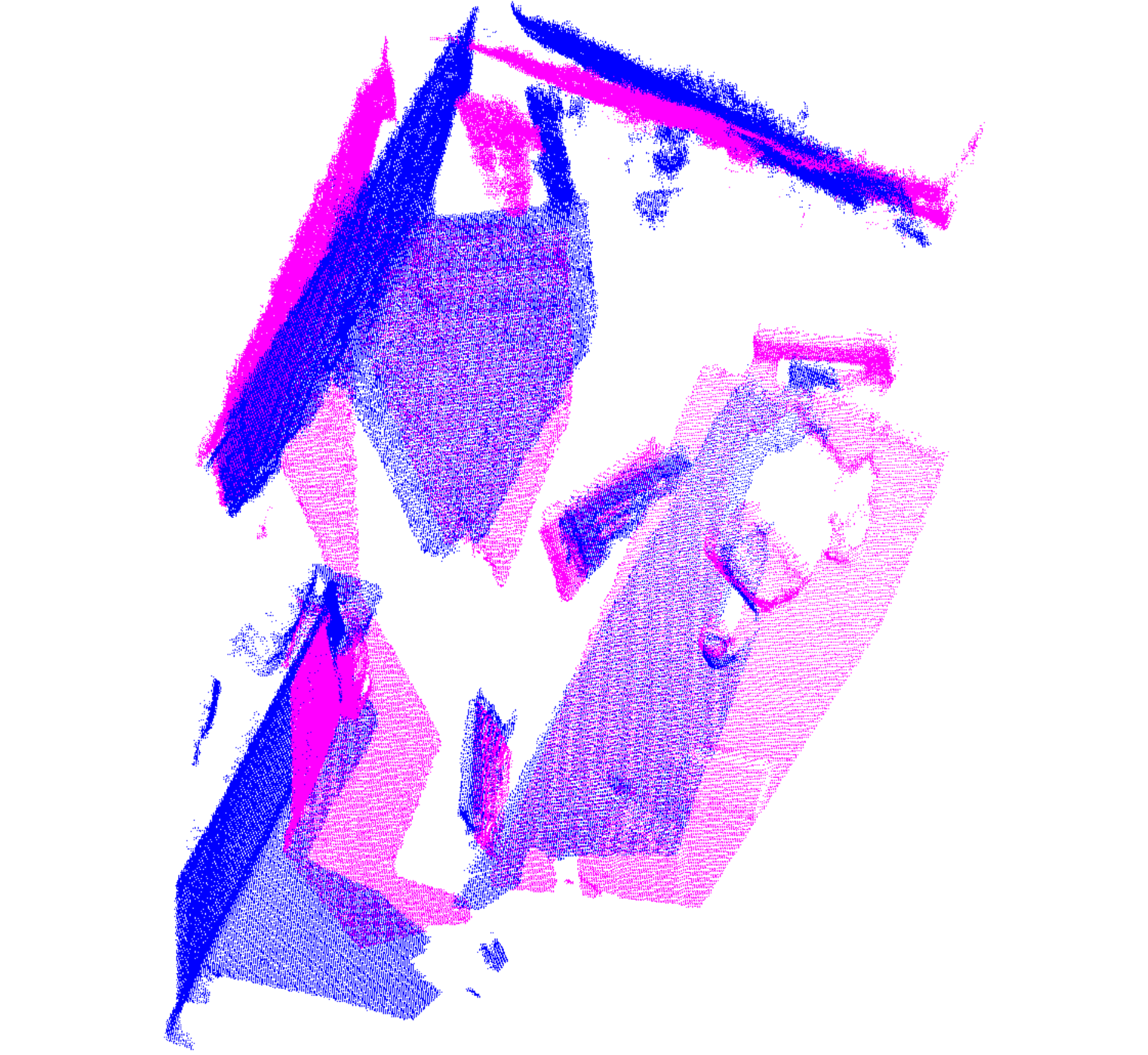}
\end{minipage}\,\,
&
\,\,
\begin{minipage}[t]{0.19\linewidth}
\centering
\includegraphics[width=.48\linewidth]{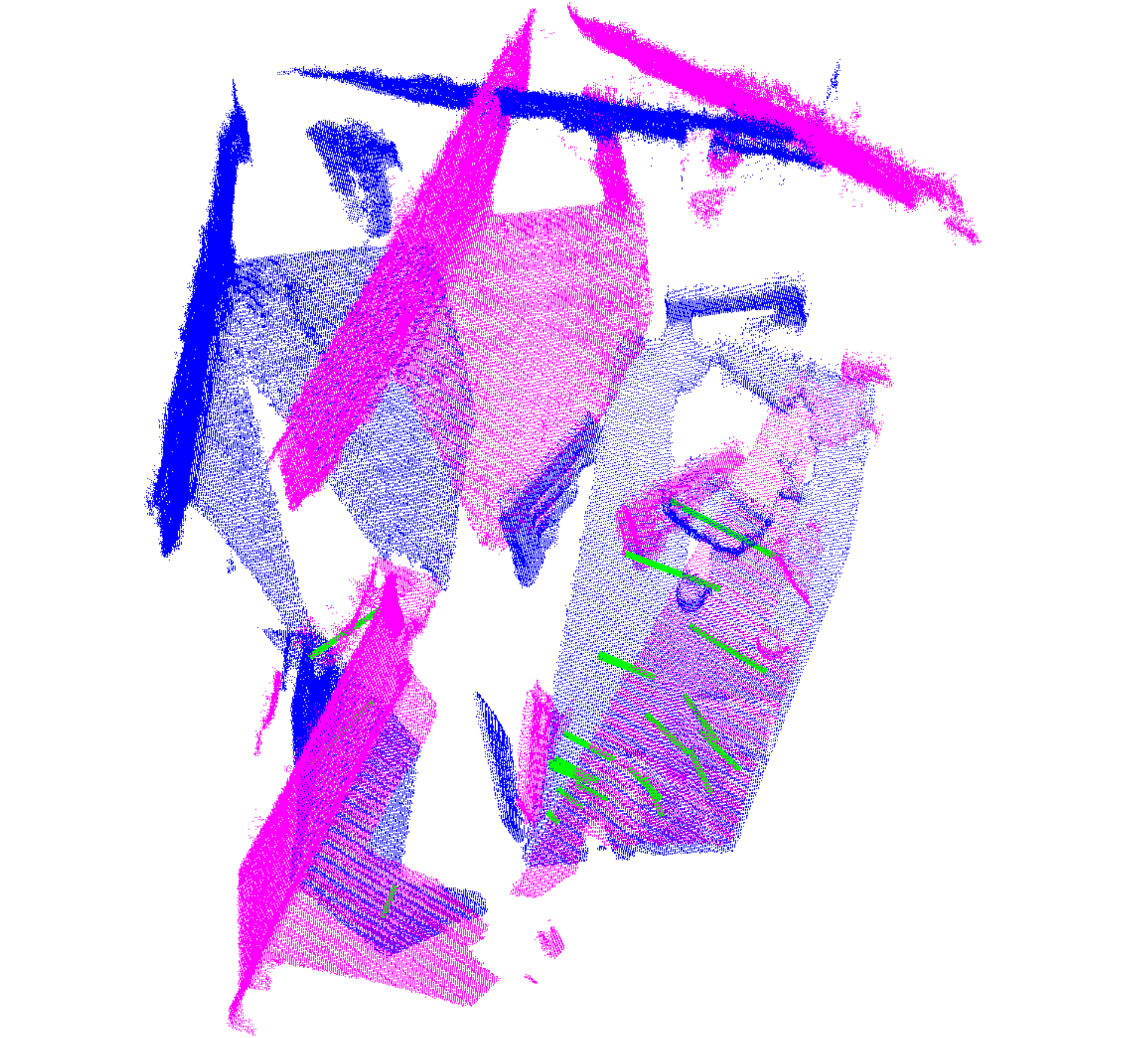}
\includegraphics[width=.48\linewidth]{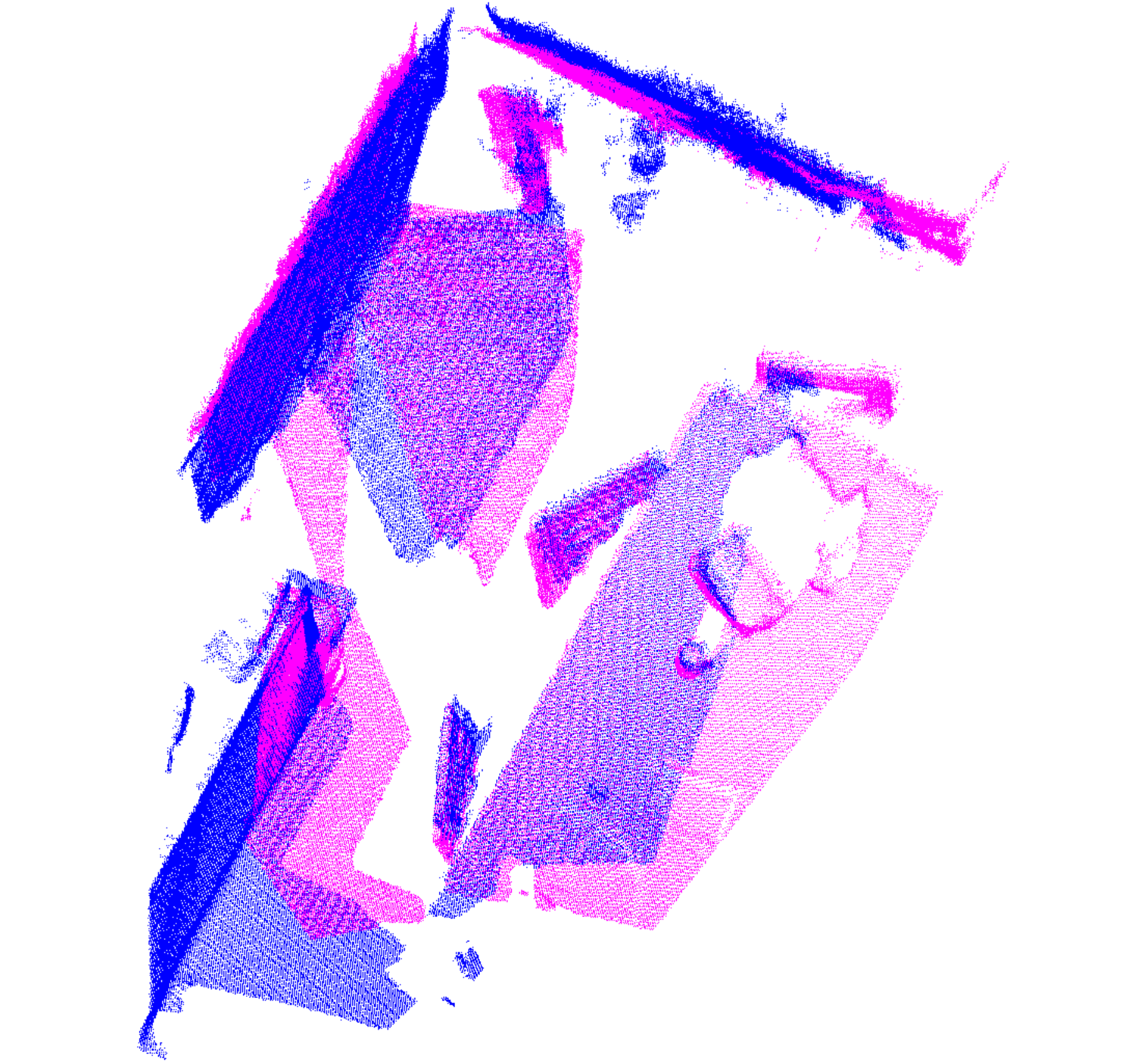}
\end{minipage}\,\,
\end{tabular}

\caption{Qualitative scan matching results over the \textit{kitchen} scene from the Microsoft 7-scenes dataset~\cite{shotton2013scene}. The left-most column shows the putative correspondences (in cyan color) matched by FPFH, and the rest columns show: (a) the inliers found and (b) the qualitative stitching results, using FLO-RANSAC, GNC-TLS, GORE+RANSAC and VOCRA, respectively.}
\label{qualit-scan-matching}
\vspace{-2mm}
\end{figure*}

\subsubsection{Environmental Setup}

We adopt \textit{bunny}, \textit{armadillo} and \textit{dragon} from the Stanford 3D Scanning Repository~\cite{curless1996volumetric} as our point clouds. For each point cloud, we downsample it to $N=1000$ and resize it to be put inside a $[-0.5,0.5]^3$ cube as the first point set $\mathcal{P}=\{\boldsymbol{p}_i\}_{i=1}^{N}$. Then we transform $\mathcal{P}$ with random rigid transformation: $\boldsymbol{R}\in SO(3)$ and $\boldsymbol{t}\in\mathbb{R}^3$ ($||\boldsymbol{t}||\leq 3$), and add random noise with $\sigma=0.01$ to the transformed point set $\mathcal{Q}=\{\boldsymbol{q}_i\}_{i=1}^{N}$ and set $\theta=0.15$. To create outliers in clutter, from 20\% up to 99\% of the points in $\mathcal{Q}$ are replaced by random points generated inside a 3D sphere of radius 1. Three examples of the correspondences are shown in the left-top images of Figure~\ref{bench-result1},~\ref{bench-result2} and~\ref{bench-result3}. We obtain our results (including rotation and translation errors and runtime) in boxplot based on 30 Monte Carlo runs over each point cloud (hence 90 runs in total), which are shown in Figure~\ref{bench-result1},~\ref{bench-result2} and~\ref{bench-result3}.

\subsubsection{Benchmarking Results}

From the results in Figure~\ref{bench-result1},~\ref{bench-result2} and~\ref{bench-result3}, we find that the non-minimal robust solvers (FGR, GNC-TLS and ADAPT) as well as the RANSAC solvers (RANSAC and FLO-RANSAC) all fail at 95\% outliers, while GORE (GORE+RANSAC also) and our VOCRA are highly robust against as many as 99\% outliers. More importantly, VOCRA has the highest accuracy (in line with GORE+RANSAC) with all outlier rates, and is 2 or 3 orders of magnitude faster than GORE when the outlier rate is not high ($\leq$80\%) and several times faster than GORE when the outlier rate is extreme ($>$95\%), showing the most state-of-the-art performance. Specifically, with 99\% outliers, VOCRA can return the accurate results in 3 seconds.

\subsubsection{On-Surface Benchmarking}

In addition, since in real-world applications the outliers are highly likely to lie on the surface of the registered 3D objects, we supplement another benchmarking experiment where all the outliers are generated on the surface rather than randomly in the 3D sphere, as demonstrated in Figure~\ref{bench-surface}. We can see that the results are similar to that in the standard benchmaring and our VOCRA is still the most outstanding solver overall.

\subsection{Realistic Point Cloud Registration}

\subsubsection{Normal Registration}

In addition to benchmarking the solvers with artificial outliers and noise, we conduct realistic registration experiments by using the 3D feature descriptor FPFH~\cite{rusu2009fast} (function \textit{extractFPFHFeatures} in Matlab) to generate the correspondences. 

Apart from the three point clouds in Section~\ref{exp-benchmark}, we also include the \textit{cheff}, \textit{chicken}, \textit{rhino}, \textit{parasauro}, and \textit{T-rex} from Mian's dataset~\cite{mian2006three,mian2010repeatability}, and the \textit{city} and \textit{castle} scans from the ETHZ LiDAR dataset~\cite{zeisl2013automatic}. For each point cloud, we conduct 10 independent runs where in each run we transform it with a new random rigid transformation and use FPFH to match the correspondences. Then, the raw putative correspondences are directly fed to the different solvers with $\sigma=0.003$ and $\theta=0.15$ constantly.

Figure~\ref{qualit-partial} shows the examples (one example for each point cloud) of qualitative registration results by projecting the initial point cloud to the transformed point cloud with the transformation estimated by the respective solvers. VOCRA is able to estimate the exact transformation(s), making the projected initial point cloud overlap with the transformed one so well that no deviation can be observed. Besides, quantitative statistics over the 10 runs are supplemented in Figure~\ref{quant-both}(a), where VOCRA can always render the lowest estimation errors and the shortest runtime just as in the benchmarking experiments.

\begin{table*}[h]

\centering

\caption{Quantitative scan matching results  corresponding to Figure~\ref{qualit-scan-matching}.}
\label{quant-scan-matching}

\setlength\tabcolsep{2.3pt}
%\addtolength{\tabcolsep}{-0pt}

\begin{tabular}{|cc|ccc|ccc|ccc|ccc|ccc|}
\hline\rule{0pt}{6pt}
\quad & \quad & \multicolumn{3}{c|}{\scriptsize{FLO-RANSAC}~\cite{lebeda2012fixing}} & \multicolumn{3}{c|}{\scriptsize{GNC-TLS}~\cite{yang2020graduated}} & \multicolumn{3}{c|}{\scriptsize{GORE+RANSAC}~\cite{bustos2017guaranteed}} & \multicolumn{3}{c|}{\scriptsize{VOCRA}} \\ \hline

\scriptsize{Scene Pair} & \scriptsize{$N$} & \scriptsize{Recall} & \scriptsize{Registration Status}  & \scriptsize{Time [\textit{s}]} & \scriptsize{Recall} & \scriptsize{Registration Status}  & \scriptsize{Time [\textit{s}]} & \scriptsize{Recall} & \scriptsize{Registration Status}  & \scriptsize{Time [\textit{s}]} & \scriptsize{Recall} & \scriptsize{Registration Status}  & \scriptsize{Time [\textit{s}]} \\ \hline

\scriptsize{1-2} & \scriptsize{1500}  & \scriptsize{14} & \scriptsize{\textcolor[rgb]{1,0,1}{Large Error}} & \scriptsize{37.992} & \scriptsize{7} & \scriptsize{\textcolor[rgb]{1,0,0}{Fail}}  & \scriptsize{196.234} & \scriptsize{13}  & \scriptsize{\textcolor[rgb]{0,0.8,0}{Succeed}} & \scriptsize{6.736} & \scriptsize{13} & \scriptsize{\textcolor[rgb]{0,0.8,0}{Succeed}}  & \scriptsize{0.407} \\ \hline

\scriptsize{4-5} & \scriptsize{750}  & \scriptsize{13} & \scriptsize{\textcolor[rgb]{1,0,0}{Fail}} & \scriptsize{19.319} & \scriptsize{19} & \scriptsize{\textcolor[rgb]{0,0.8,0}{Succeed}} & \scriptsize{0.123}  & \scriptsize{18} & \scriptsize{\textcolor[rgb]{0,0.8,0}{Succeed}} & \scriptsize{3.197} & \scriptsize{18} & \scriptsize{\textcolor[rgb]{0,0.8,0}{Succeed}} & \scriptsize{0.650} \\ \hline

\scriptsize{9-10} & \scriptsize{796}  & \scriptsize{17} & \scriptsize{\textcolor[rgb]{1,0,0}{Fail}} & \scriptsize{20.165} & \scriptsize{9} & \scriptsize{\textcolor[rgb]{1,0,1}{Large Error}} & \scriptsize{0.136}  & \scriptsize{13} & \scriptsize{\textcolor[rgb]{1,0,0}{Fail}} & \scriptsize{31.858} & \scriptsize{11} & \scriptsize{\textcolor[rgb]{0,0.8,0}{Succeed}} & \scriptsize{2.023} \\ \hline

\scriptsize{11-12} & \scriptsize{846}  & \scriptsize{20} & \scriptsize{\textcolor[rgb]{1,0,0}{Fail}} & \scriptsize{21.486} & \scriptsize{16} & \scriptsize{\textcolor[rgb]{1,0,1}{Large Error}} & \scriptsize{0.141}  & \scriptsize{22} & \scriptsize{\textcolor[rgb]{0,0.8,0}{Succeed}} & \scriptsize{49.418} & \scriptsize{17} & \scriptsize{\textcolor[rgb]{0,0.8,0}{Succeed}} & \scriptsize{0.863} \\ \hline

\scriptsize{37-38} & \scriptsize{1475}  & \scriptsize{7} & \scriptsize{\textcolor[rgb]{1,0,0}{Fail}} & \scriptsize{37.465} & \scriptsize{7} & \scriptsize{\textcolor[rgb]{1,0,1}{Large Error}} & \scriptsize{0.219}  & \scriptsize{4} & \scriptsize{\textcolor[rgb]{1,0,0}{Fail}} & \scriptsize{68.232} & \scriptsize{6} & \scriptsize{\textcolor[rgb]{0,0.8,0}{Succeed}} & \scriptsize{3.743} \\ \hline

\scriptsize{58-59} & \scriptsize{1092}  & \scriptsize{24} & \scriptsize{\textcolor[rgb]{1,0,0}{Fail}} & \scriptsize{2.317} & \scriptsize{30} & \scriptsize{\textcolor[rgb]{1,0,1}{Large Error}} & \scriptsize{27.845}  & \scriptsize{23} & \scriptsize{\textcolor[rgb]{1,0,1}{Large Error}} & \scriptsize{0.208} & \scriptsize{24} & \scriptsize{\textcolor[rgb]{0,0.8,0}{Succeed}} & \scriptsize{2.371} \\ \hline

\end{tabular}

\vspace{-2mm}
\end{table*}

\subsubsection{Registration with Low Overlapping}

In real-world applications, partiality or low overlapping ratio between point clouds is a common but challenging issue. Hence, we supplement partiality registration experiments  to further evaluate the solvers. This time we only preserve a small portion of the initial point cloud and fix the overlapping ratio as $\approx$30\%, and then use FPFH to establish correspondences, which is highly likely to triggers high outlier rates.

For each point cloud, 10 different overlapping regions (all with 30\% overlapping ratio) are tested with 10 different random transformations. The mean outlier rates for \textit{bunny}, \textit{armadillo}, \textit{dragon}, \textit{cheff}, \textit{chicken}, \textit{rhino}, \textit{parasauro}, \textit{T-rex}, \textit{city} and \textit{castle} are: 90.2\%, 82.2\%, 79.3\%, 90.69\%, 87.4\%, 92.6\%, 94.9\%, 85.9\%, 56.0\%, 56.9\%, respectively. We also display both qualitative and quantitative results, as shown in Figure~\ref{qualit-partial} and~\ref{quant-both}(b).

Even if the outlier rates are high (even more than 98\% in some cases) due to the low overlapping ratio, we can see that VOCRA is still the most (or at least one of the most) robust and efficient solver throughout the partiality experiments.

\subsection{Real-Data Applications}

To validate the practicality of VOCRA in real scenes, we test it on two application problems: scan matching and 3D object localization over real-world datasets.

\subsubsection{Scan Matching}

Scan matching (also called scene stitching) is an important problem in 3D reconstruction and loop closure detection (SLAM). We evaluate our VOCRA over the \textit{kitchen} scene from the Microsoft 7-scenes dataset~\cite{shotton2013scene} in comparison with a RANSAC-based solver FLO-RANSAC, a non-minimal solver GNC-TLS, and a combination solver GORE+RANSAC (the most robust solver excluding VOCRA in the benchmarking).

Each time we select one pair of scenes (with overlapping regions), downsample them with box grid filter of size 0.02, and then use FPFH to establish the correspondences. Since GORE generally gets slower with increasing correspondences, we set the maximum correspondence number to $N=1500$. After that, we feed these correspondences (images in the first column of Figure~\ref{qualit-scan-matching}) to the four solvers with noise set to $\sigma=0.01$ constantly.

We report the qualitative scan matching results over 6 scene pairs in Figure~\ref{qualit-scan-matching} where we show the FPFH correspondences and the inliers found as well as the registration results by the different solvers. We also provide the corresponding quantitative data in Table~\ref{quant-scan-matching}, where we specify the $N$ for each scene pair, and the number of inliers recalled, registration status and runtime by each solver. For the registration status, \textcolor[rgb]{1,0,1}{Large Error} means that the registration is almost acceptable but with relatively high estimation errors, \textcolor[rgb]{1,0,0}{Fail} means that the registration is completely wrong, and \textcolor[rgb]{0,0.8,0}{Succeed} means that the registration is successful and with good accuracy, and this status is judged manually since the ground-truth relative poses between the scene pairs are not given.

According to Figure~\ref{qualit-scan-matching} and Table~\ref{quant-scan-matching}, VOCRA succeeds in stitching all the scene pairs, while all the other competitors fail at least once. In \textit{Scene 1-2}, we can see that there are only 13 inliers out of the 1500 correspondences (outlier rate is over 99.1\%) recalled successfully by VOCRA, manifesting that VOCRA remains highly robust in practical application.

\subsubsection{3D Object Localization}

In addition, we test VOCRA in the practical problem of localizing a 3D object with a RGB-D scene by adopting the RGB-D Scenes dataset~\cite{lai2011large}. We make use of the ground-truth labels provided to pick out and build the point cloud of the target object from the RGB-D scene, where we build three differently-shaped objects: \textit{cereal box, cap} and \textit{table}. Then we impose a random transformation on the object to generate an independent object in the 3D space. We employ FPFH to build correspondences between the scene and transformed target object. Afterwards, FLO-RANSAC, GNC-TLS, GORE+RANSAC and VOCRA are used to estimate the pose (transformation) between the object and the scene with noise all set to $\sigma=0.001$.

\begin{figure*}[t]
\centering
\setlength\tabcolsep{0.1pt}
\addtolength{\tabcolsep}{0pt}
\begin{tabular}{c|cc|ccccc}

\quad &\,&\,\footnotesize{FPFH}\, &\,&  \footnotesize{FLO-RANSAC~\cite{lebeda2012fixing}} & \footnotesize{GNC-TLS~\cite{yang2020graduated}} & \footnotesize{GORE+RANSAC~\cite{bustos2017guaranteed}} & \footnotesize{VOCRA} \\

\hline
&&&&&&&
\\

\rotatebox{90}{\,\,\footnotesize{\textit{Scene 01}}\,}\,

& &

\begin{minipage}[t]{0.18\linewidth}
\centering
\includegraphics[width=1\linewidth]{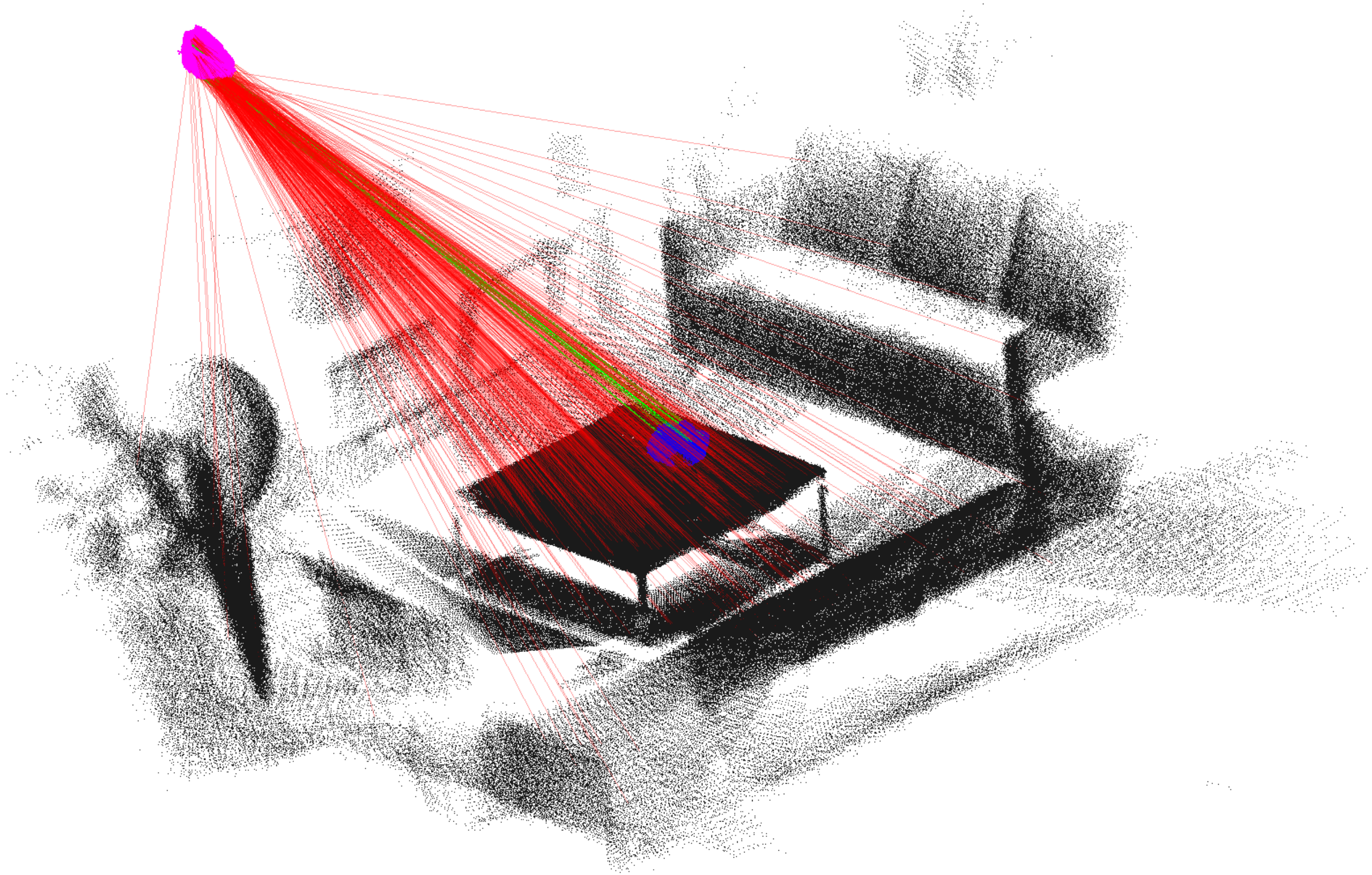}
\end{minipage}

& &

\begin{minipage}[t]{0.18\linewidth}
\centering
\includegraphics[width=1\linewidth]{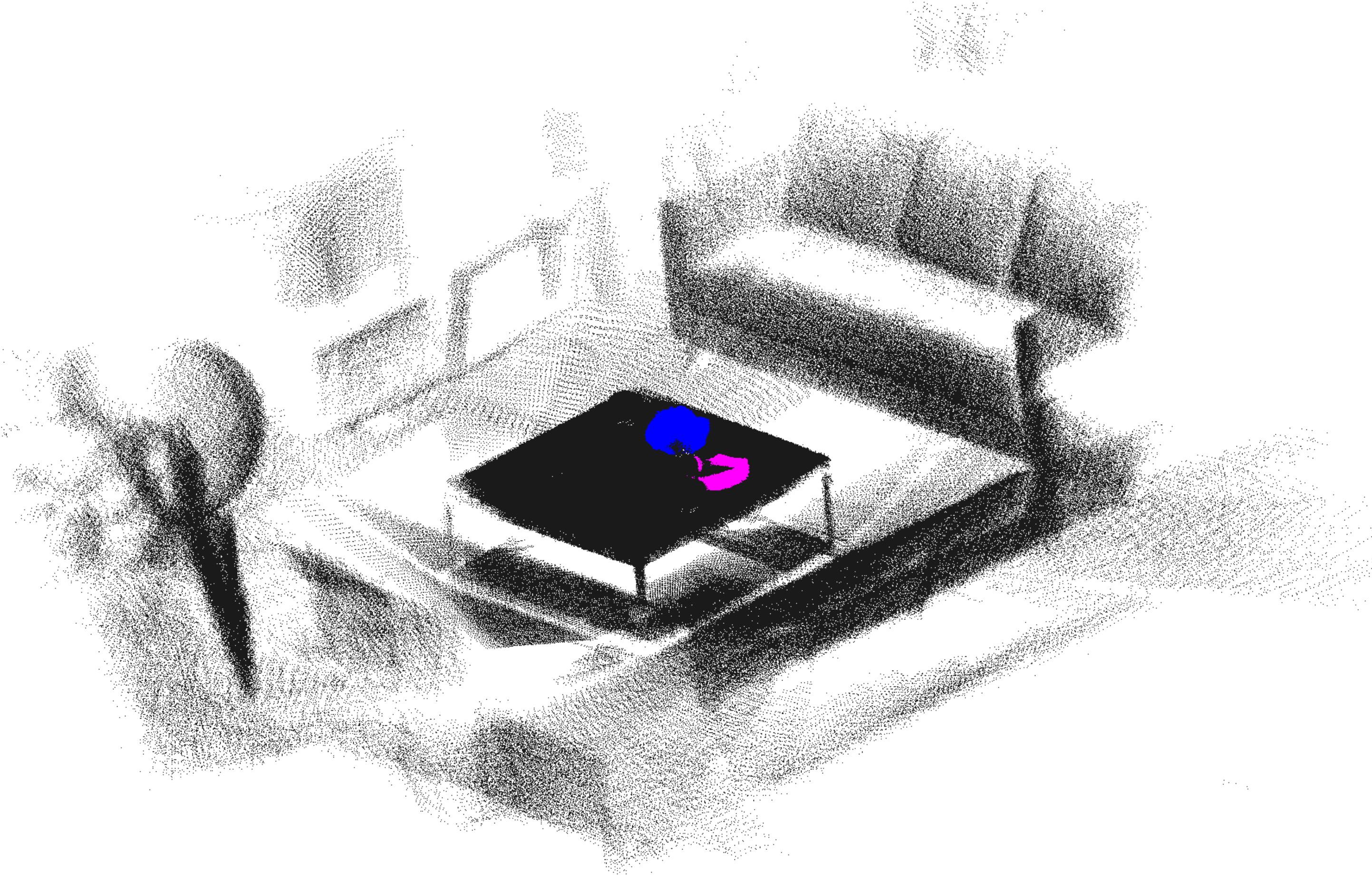}
\end{minipage}

&

\begin{minipage}[t]{0.18\linewidth}
\centering
\includegraphics[width=1\linewidth]{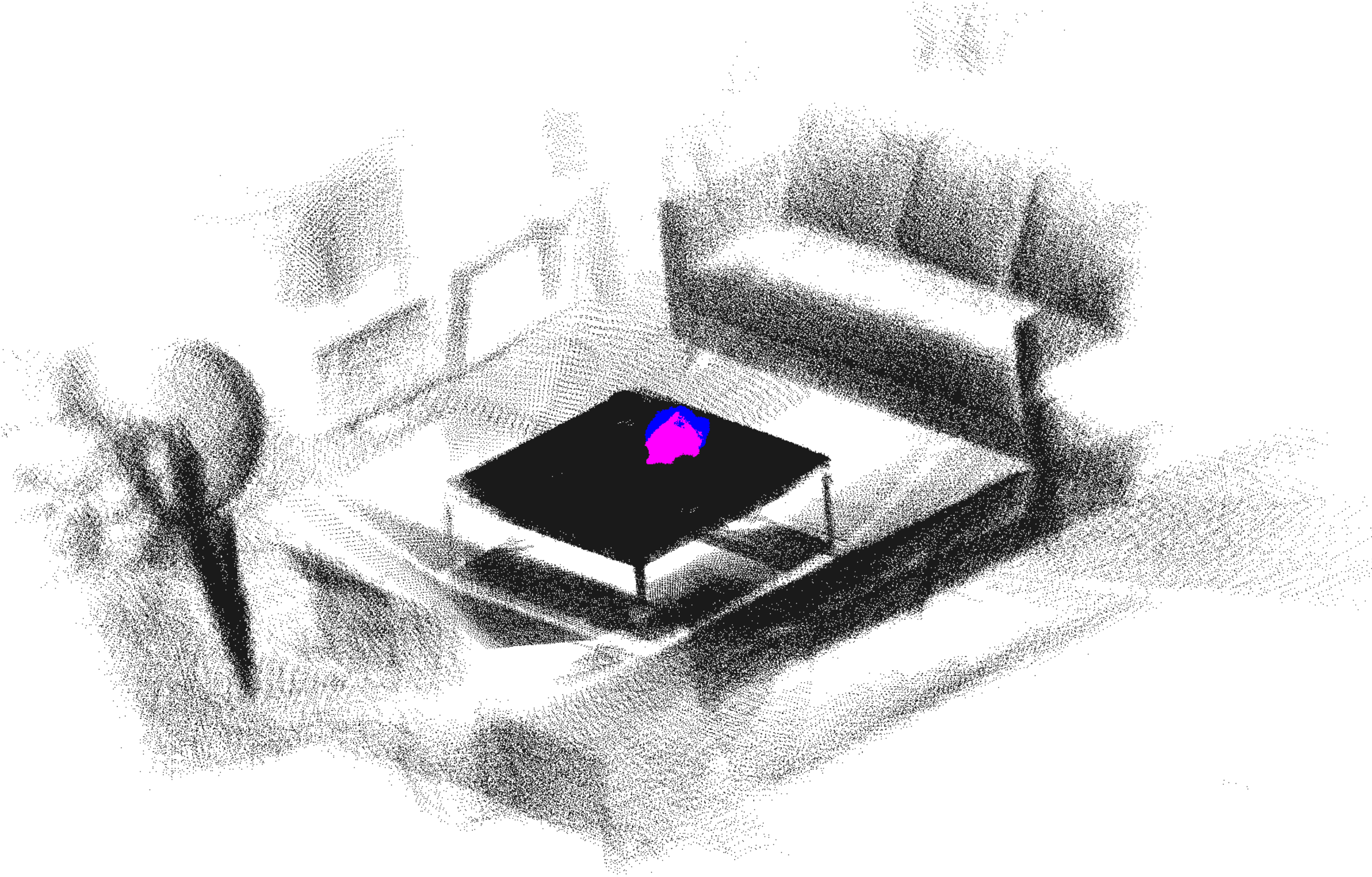}
\end{minipage}

&

\begin{minipage}[t]{0.18\linewidth}
\centering
\includegraphics[width=1\linewidth]{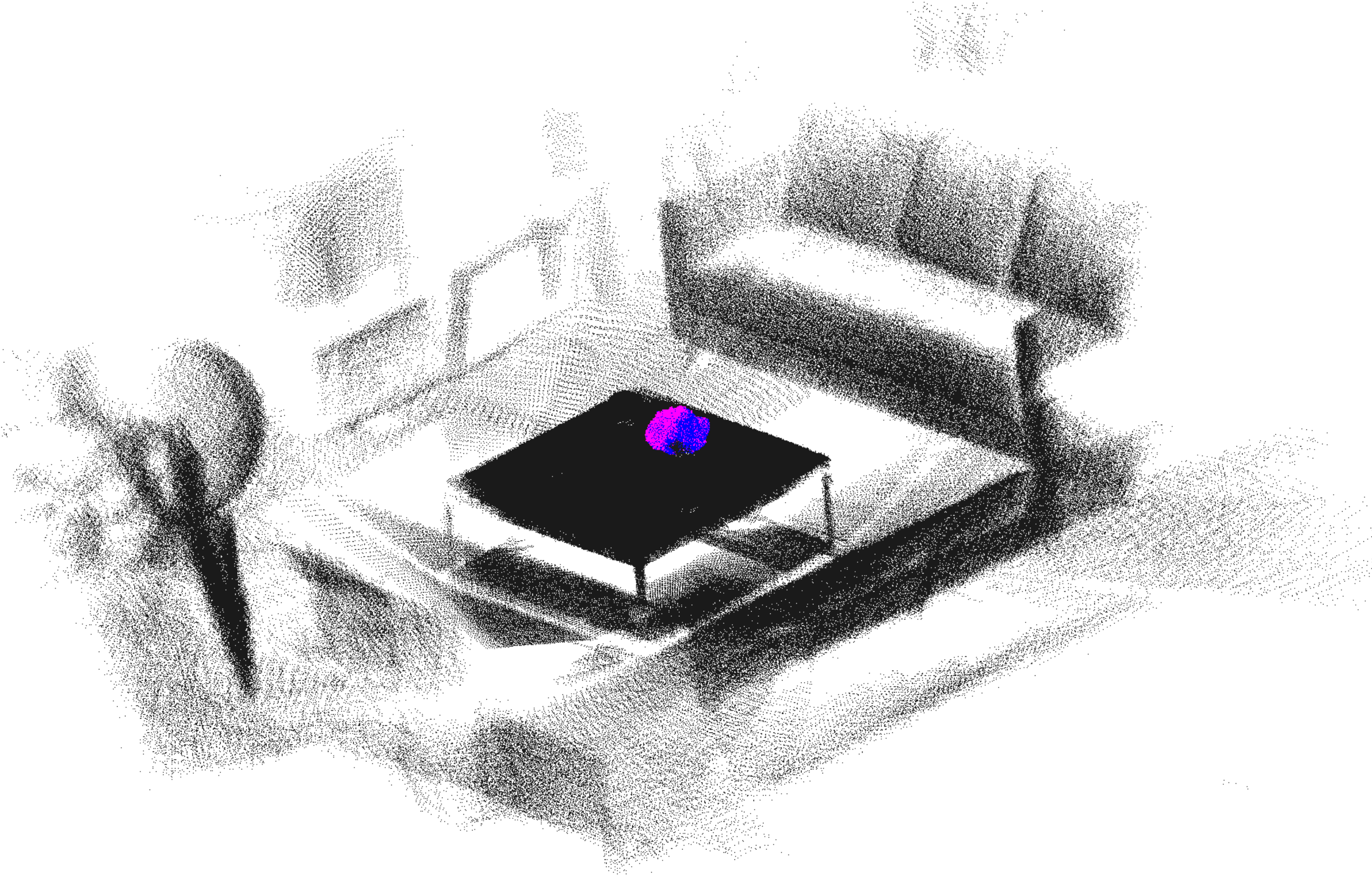}
\end{minipage}

&

\begin{minipage}[t]{0.18\linewidth}
\centering
\includegraphics[width=1\linewidth]{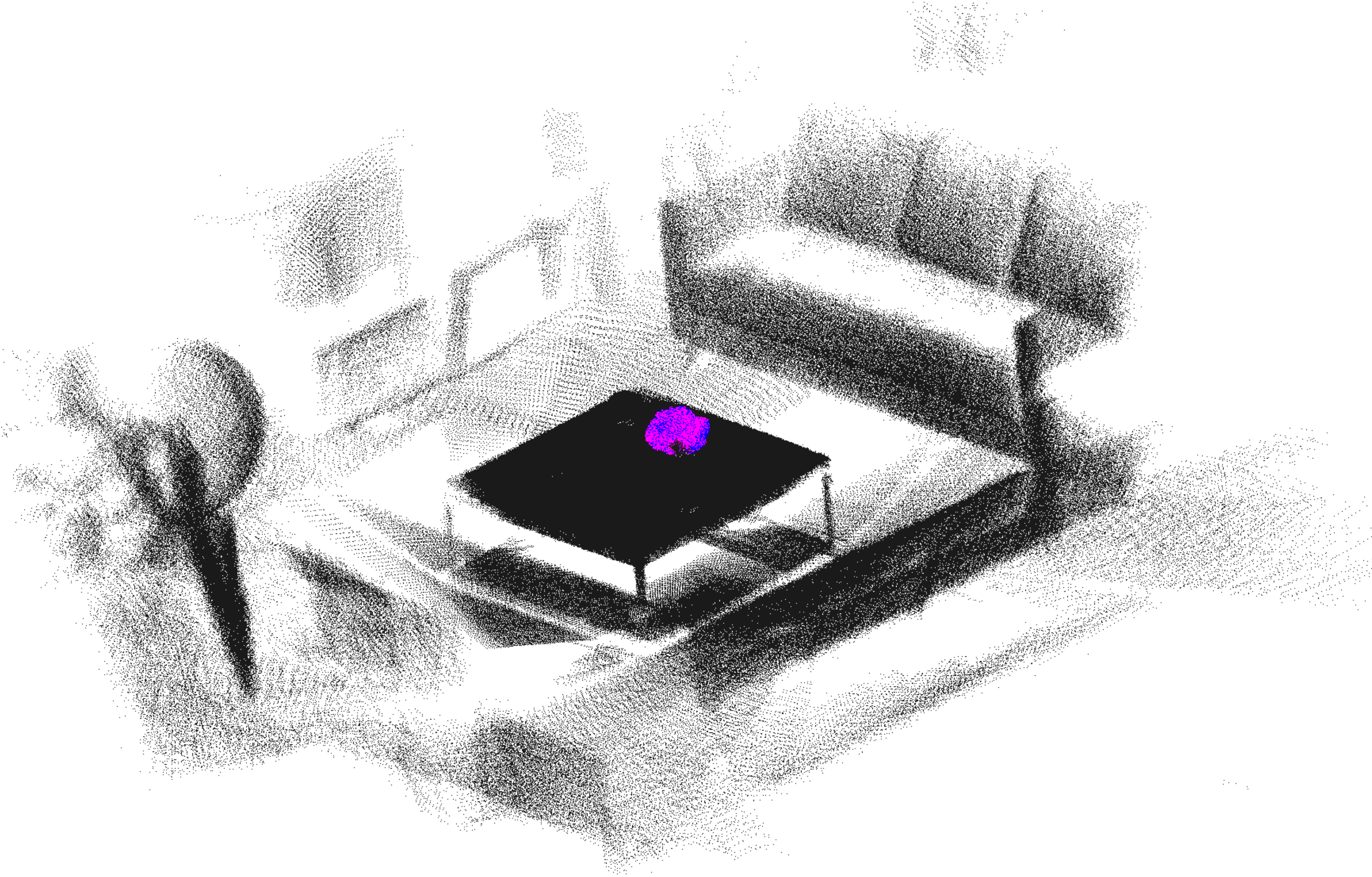}
\end{minipage}

\\

\rotatebox{90}{\,\,\footnotesize{\textit{Scene 02}}\,}\,

& &

\begin{minipage}[t]{0.18\linewidth}
\centering
\includegraphics[width=1\linewidth]{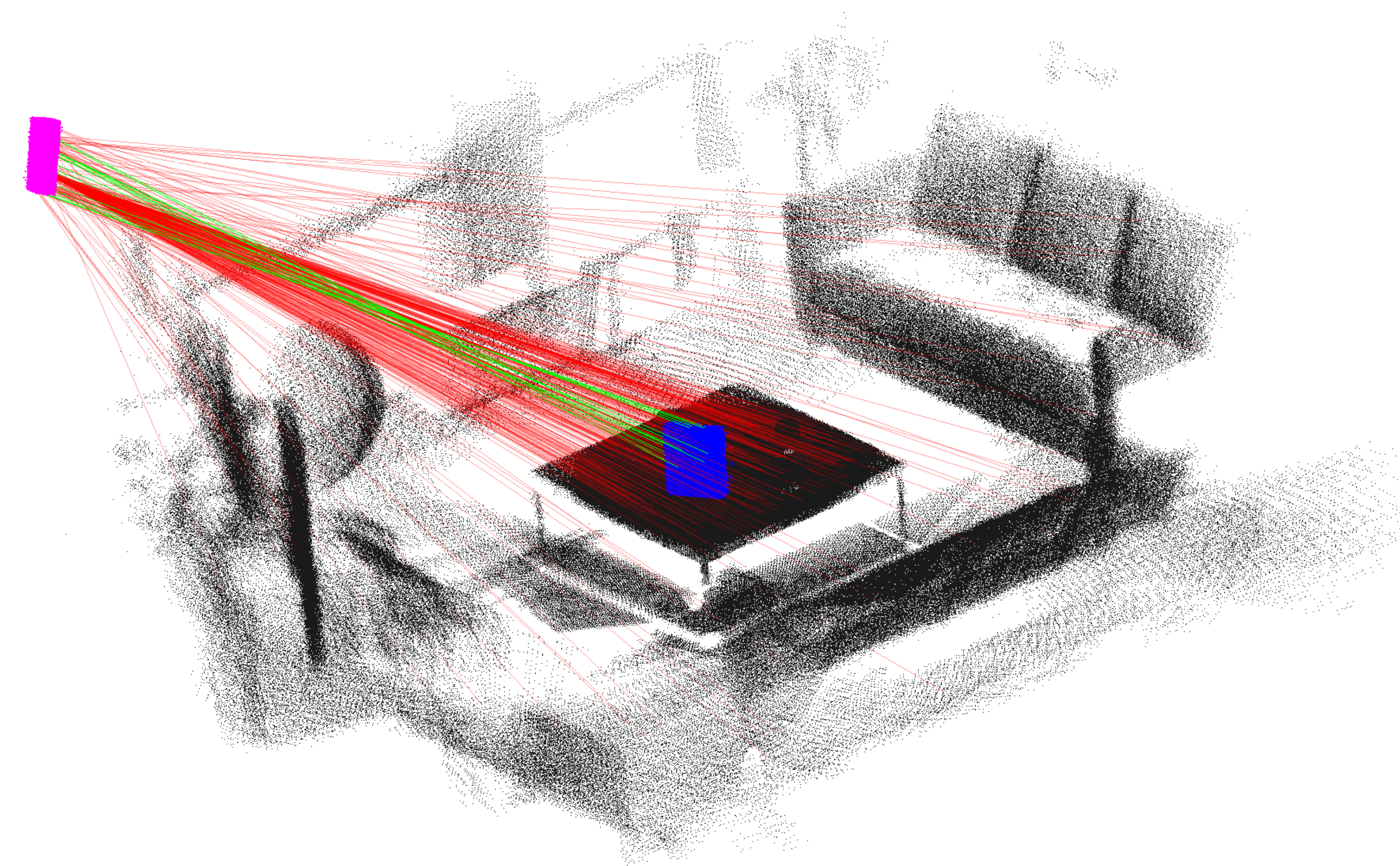}
\end{minipage}

& &

\begin{minipage}[t]{0.18\linewidth}
\centering
\includegraphics[width=1\linewidth]{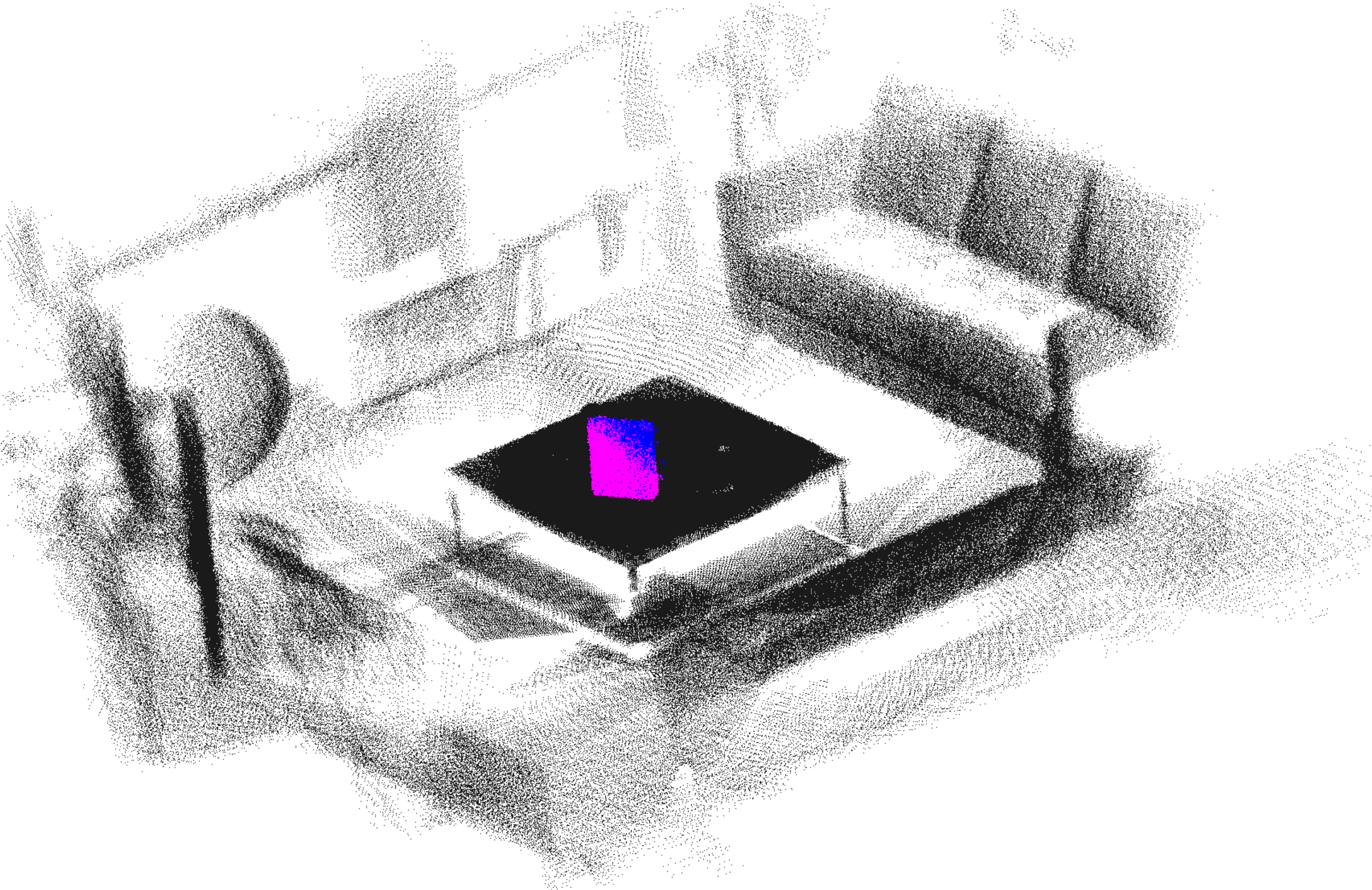}
\end{minipage}

&

\begin{minipage}[t]{0.18\linewidth}
\centering
\includegraphics[width=1\linewidth]{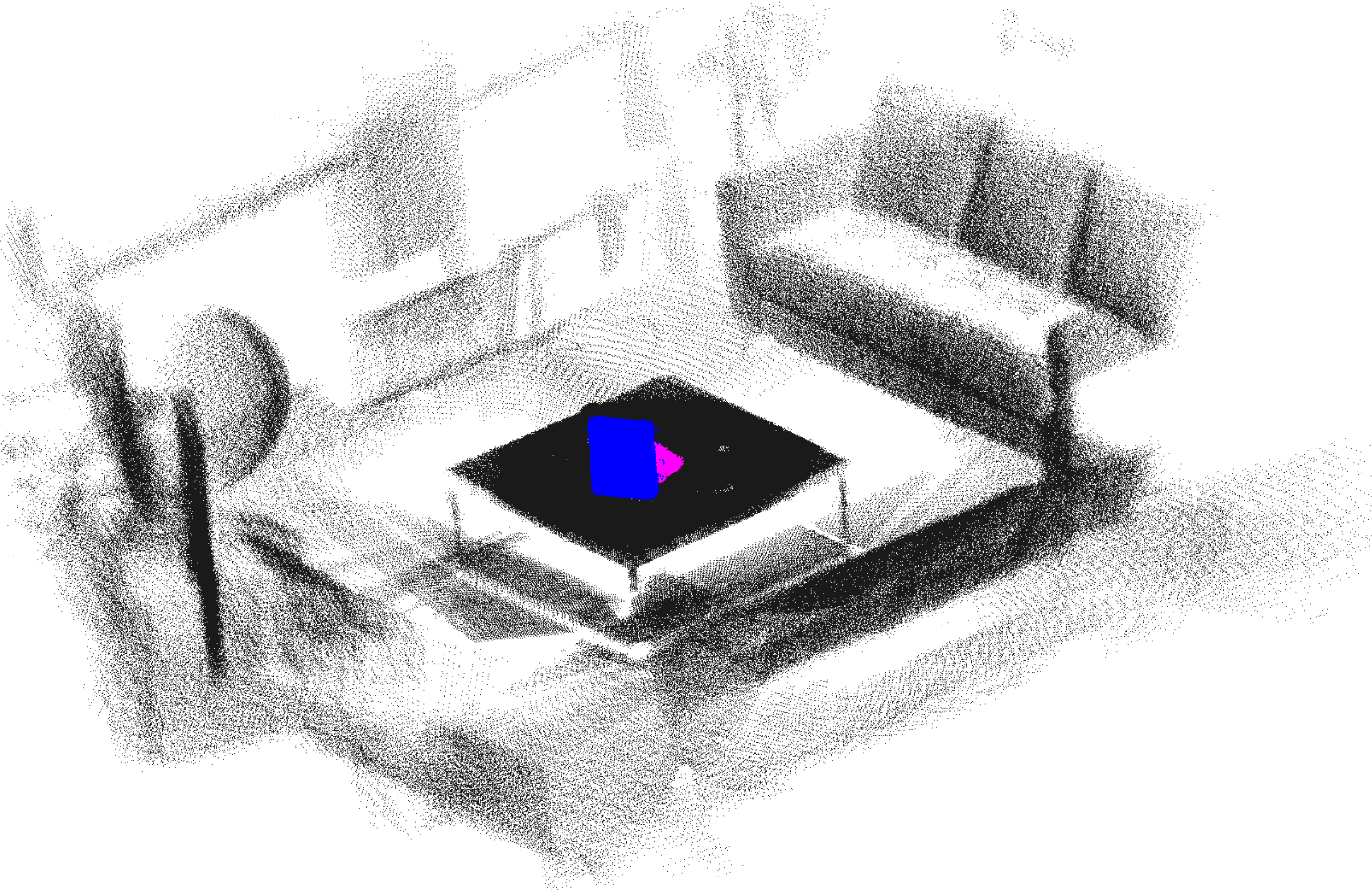}
\end{minipage}

&

\begin{minipage}[t]{0.18\linewidth}
\centering
\includegraphics[width=1\linewidth]{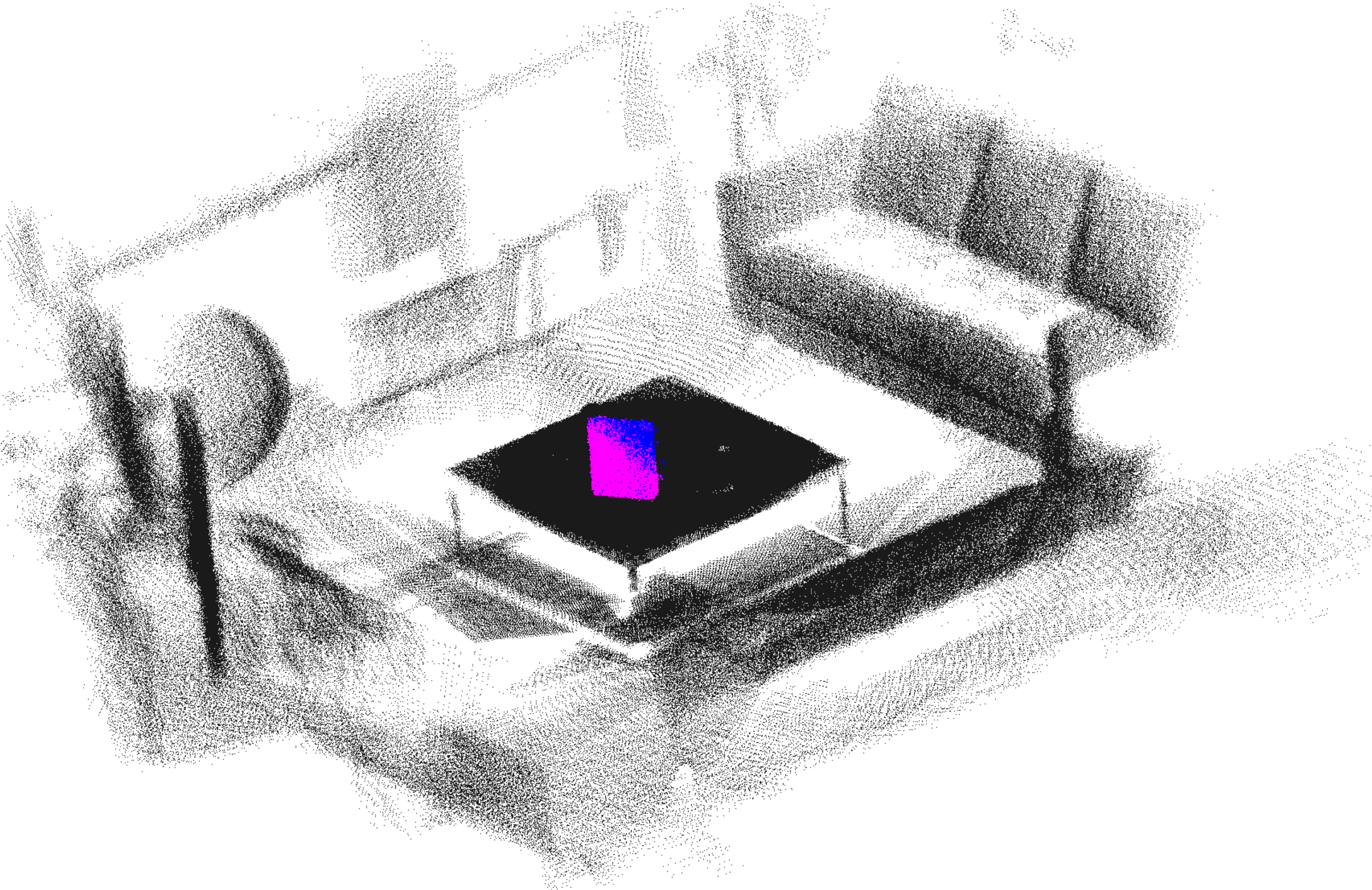}
\end{minipage}

&

\begin{minipage}[t]{0.18\linewidth}
\centering
\includegraphics[width=1\linewidth]{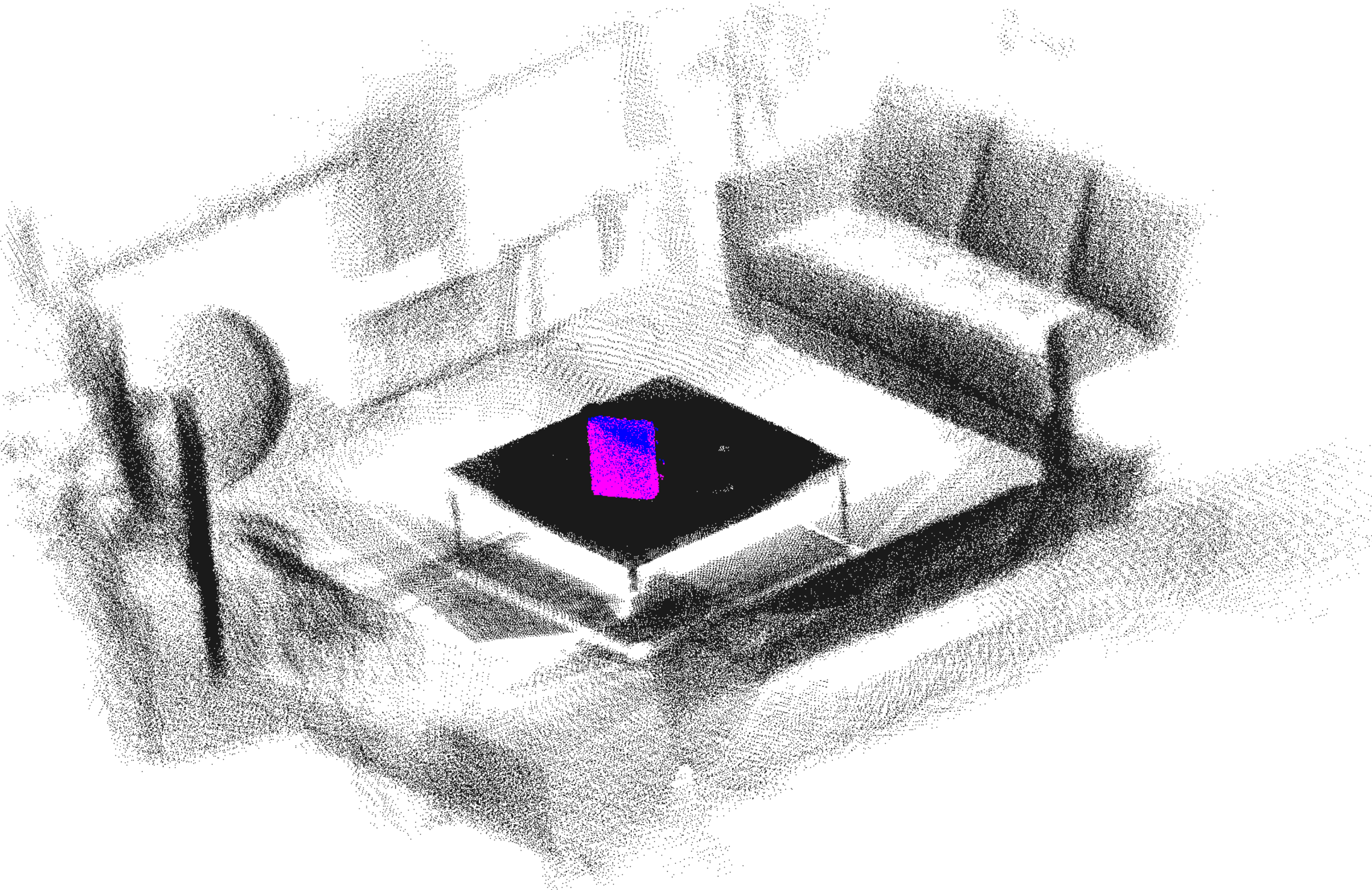}
\end{minipage}

\\

\rotatebox{90}{\,\,\footnotesize{\textit{Scene 03}}\,}\,

& &

\begin{minipage}[t]{0.18\linewidth}
\centering
\includegraphics[width=1\linewidth]{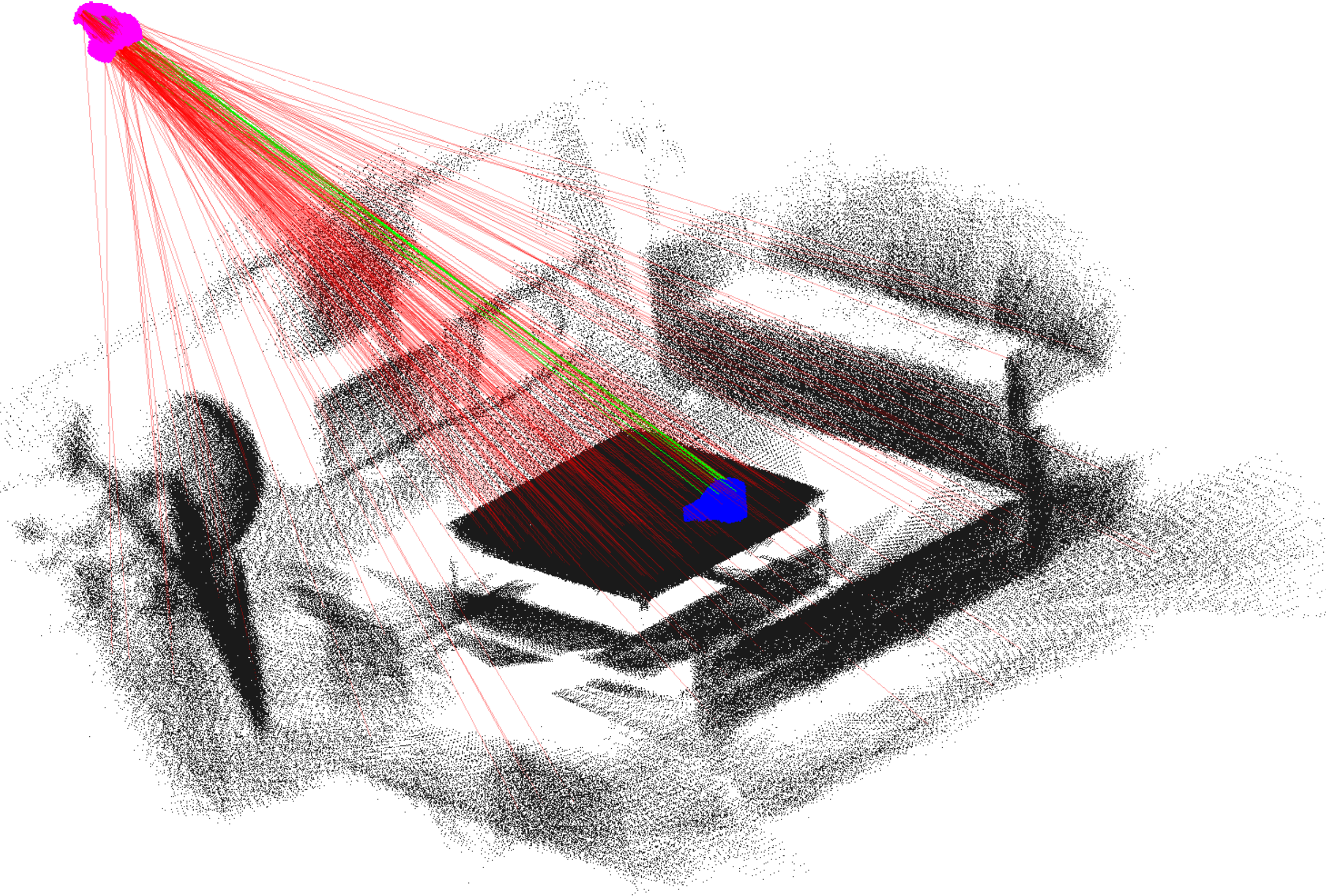}
\end{minipage}

& &

\begin{minipage}[t]{0.18\linewidth}
\centering
\includegraphics[width=1\linewidth]{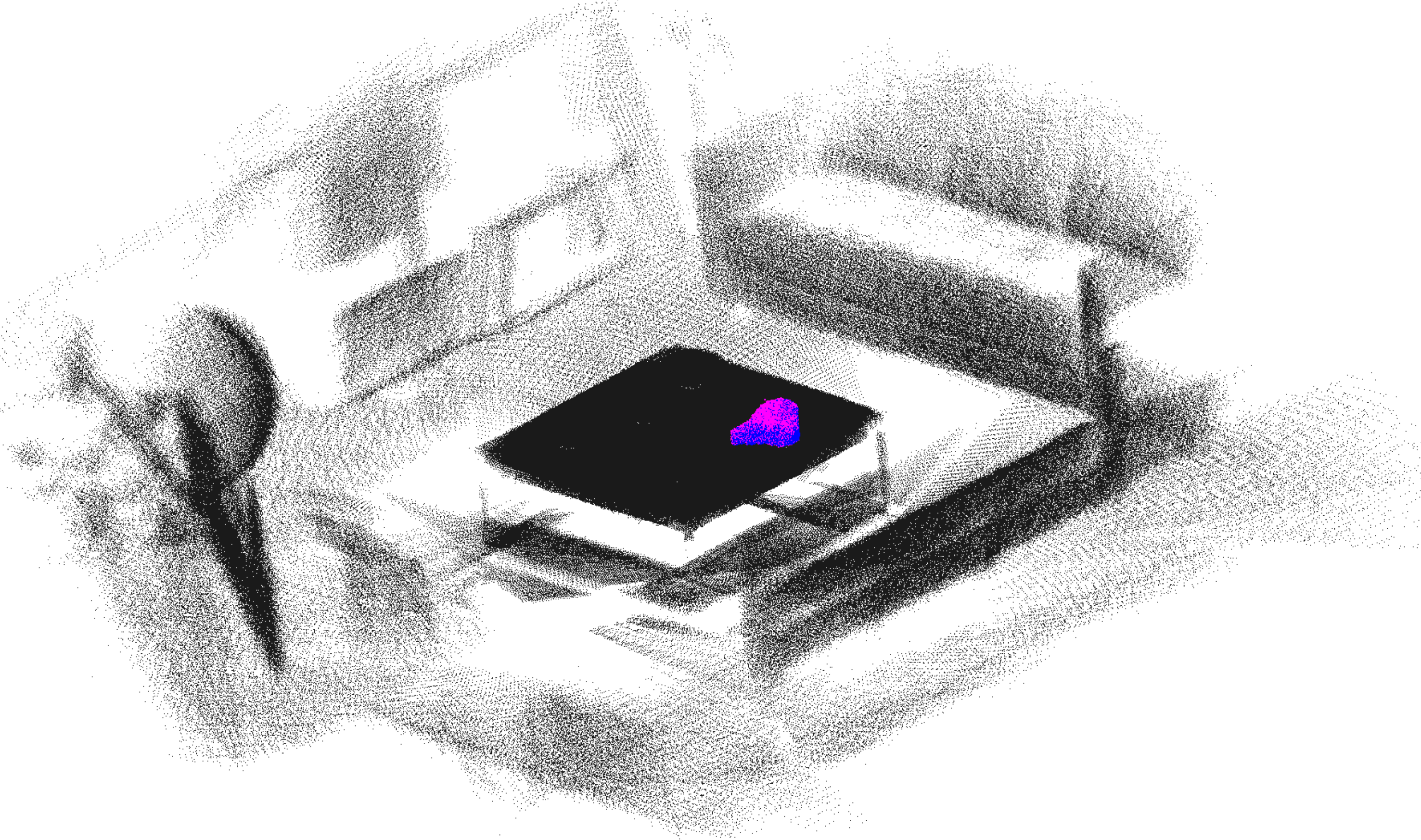}
\end{minipage}

&

\begin{minipage}[t]{0.18\linewidth}
\centering
\includegraphics[width=1\linewidth]{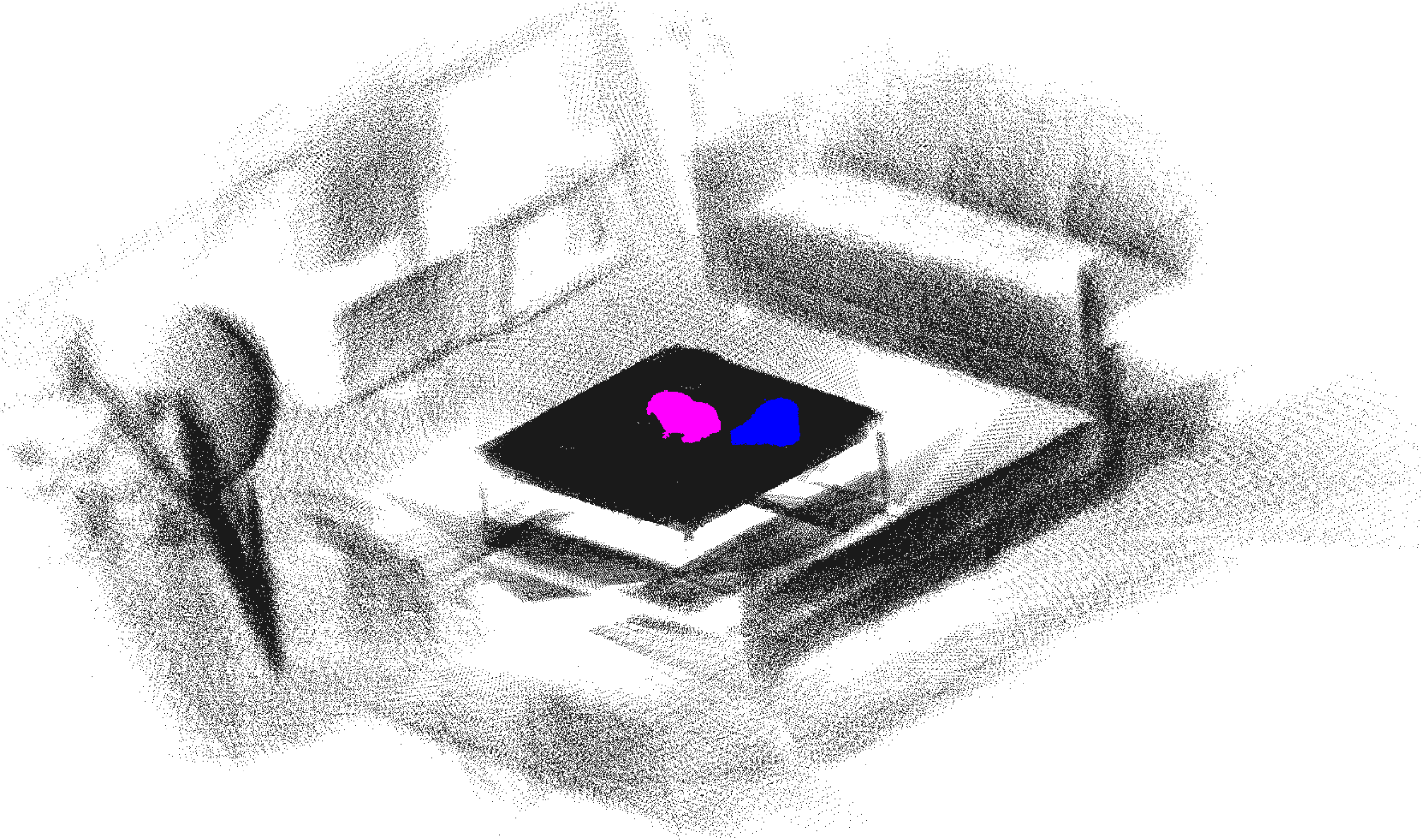}
\end{minipage}

&

\begin{minipage}[t]{0.18\linewidth}
\centering
\includegraphics[width=1\linewidth]{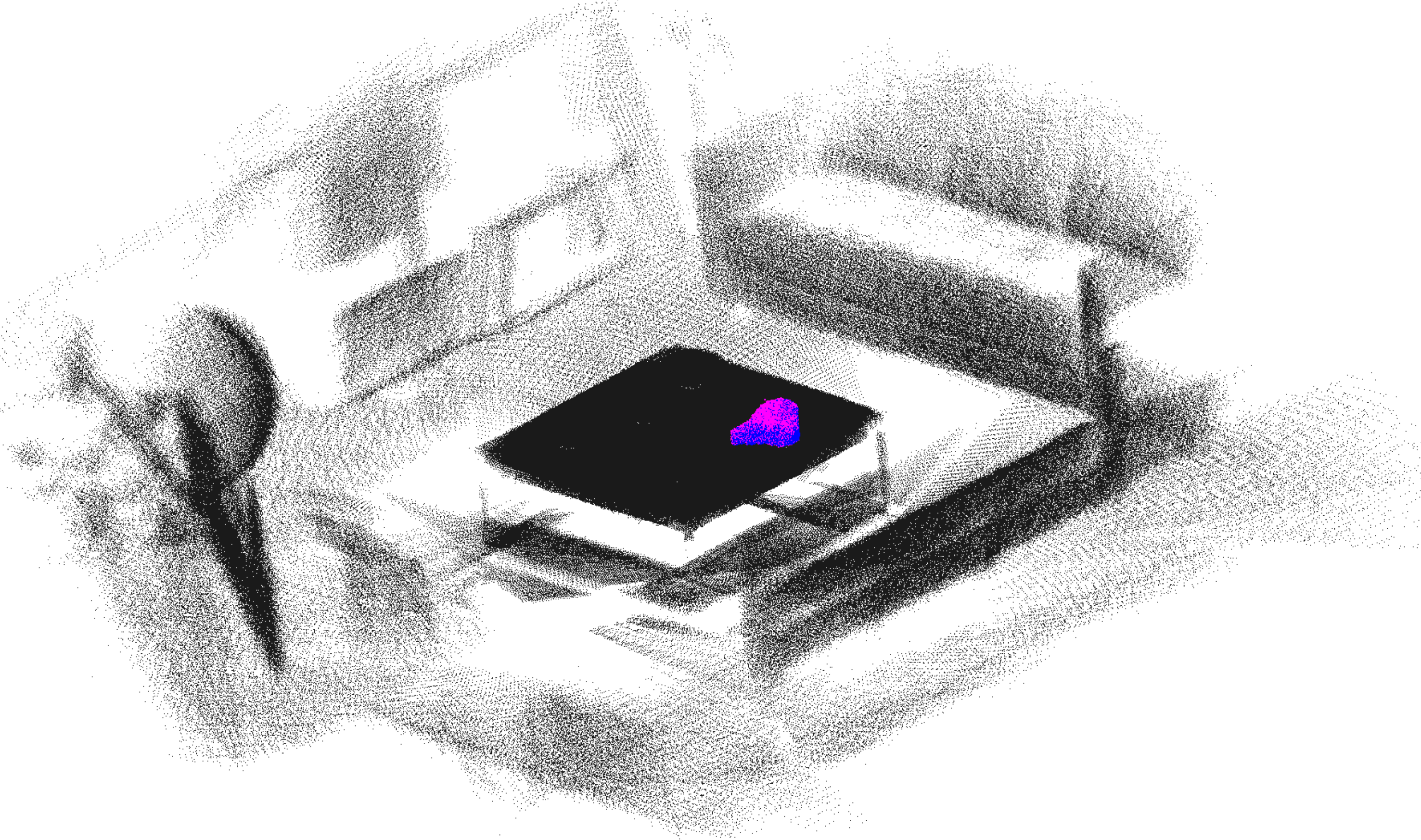}
\end{minipage}

&

\begin{minipage}[t]{0.18\linewidth}
\centering
\includegraphics[width=1\linewidth]{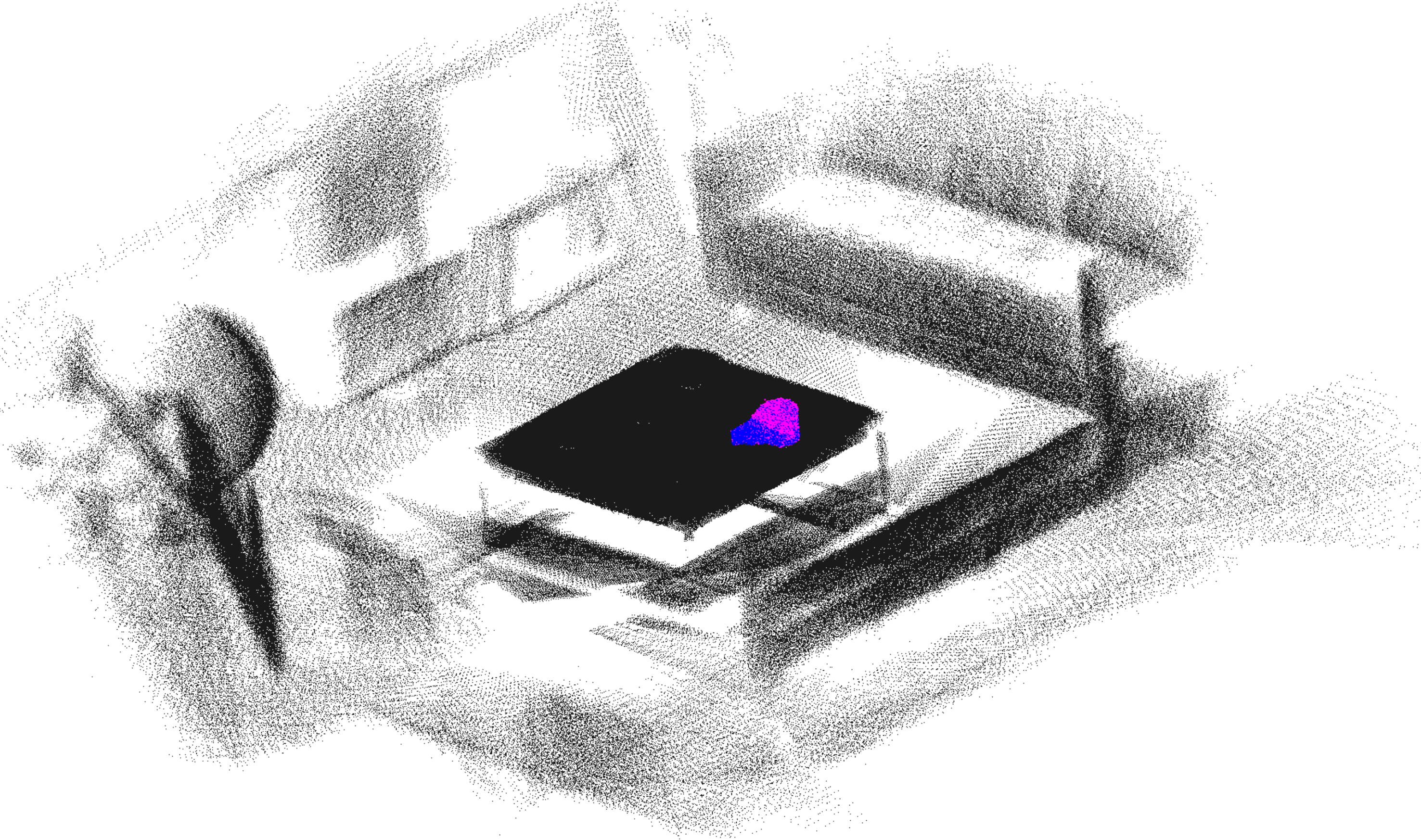}
\end{minipage}

\\

\rotatebox{90}{\,\,\footnotesize{\textit{Scene 05}}\,}\,

& &

\begin{minipage}[t]{0.18\linewidth}
\centering
\includegraphics[width=1\linewidth]{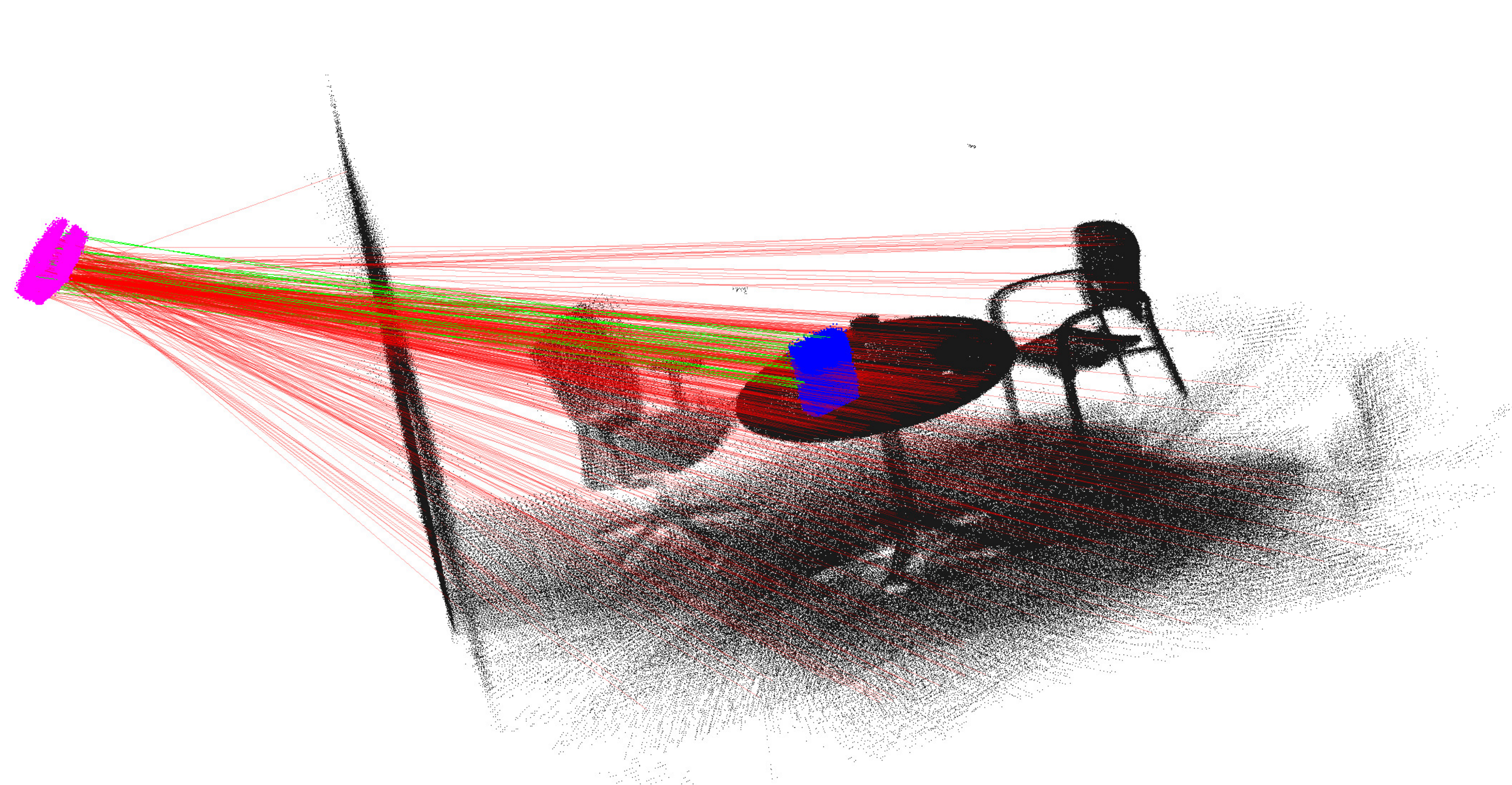}
\end{minipage}

& &

\begin{minipage}[t]{0.18\linewidth}
\centering
\includegraphics[width=1\linewidth]{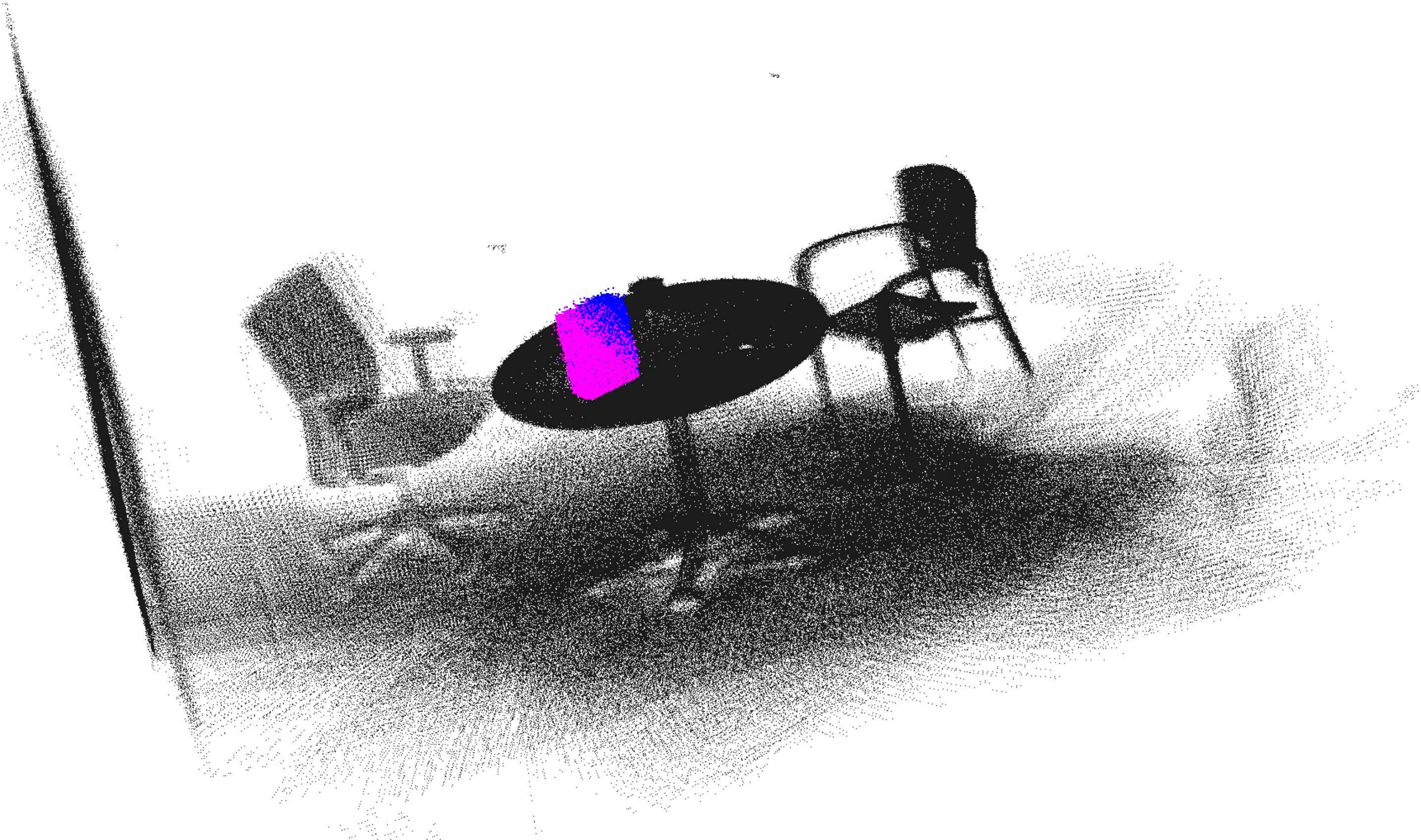}
\end{minipage}

&

\begin{minipage}[t]{0.18\linewidth}
\centering
\includegraphics[width=1\linewidth]{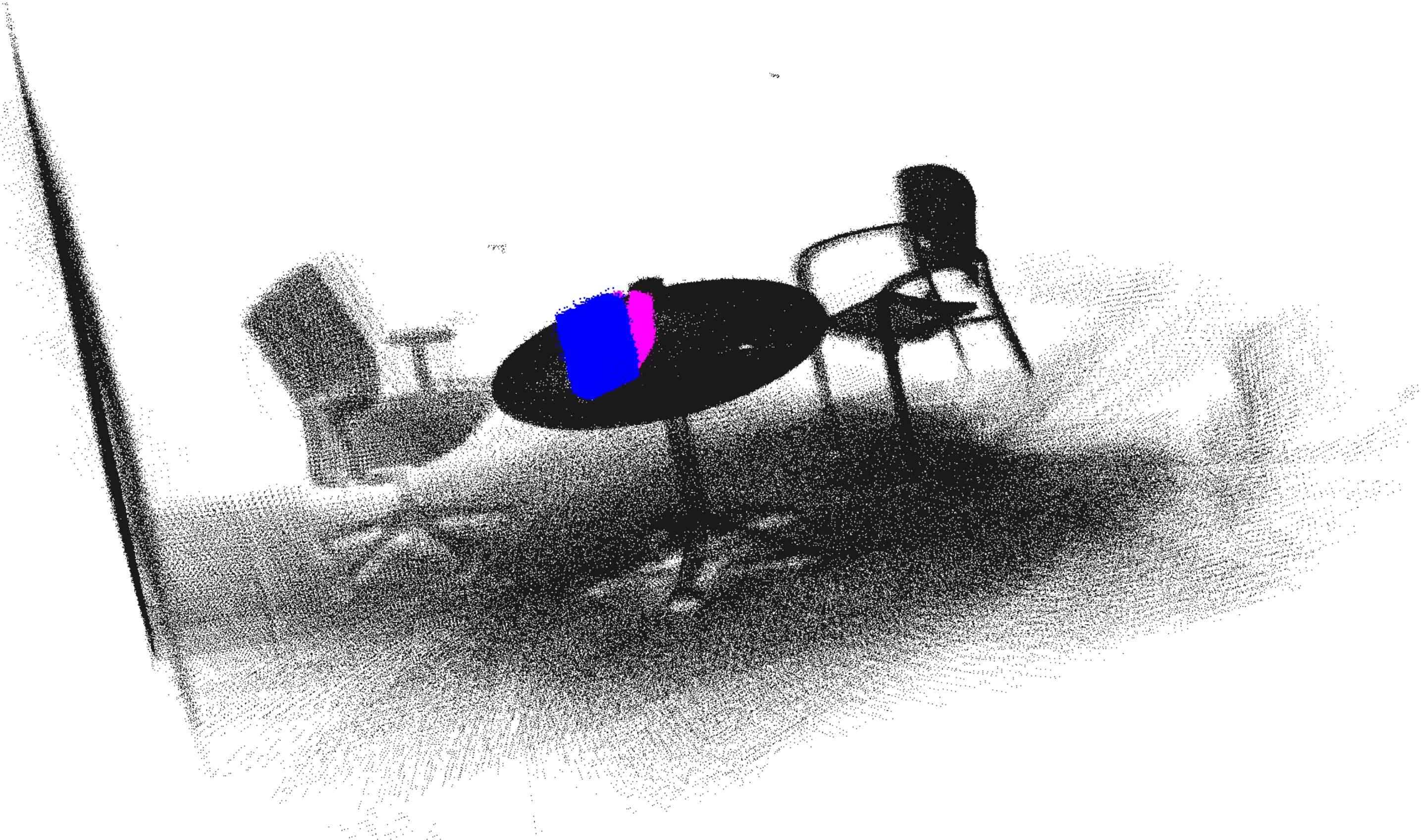}
\end{minipage}

&

\begin{minipage}[t]{0.18\linewidth}
\centering
\includegraphics[width=1\linewidth]{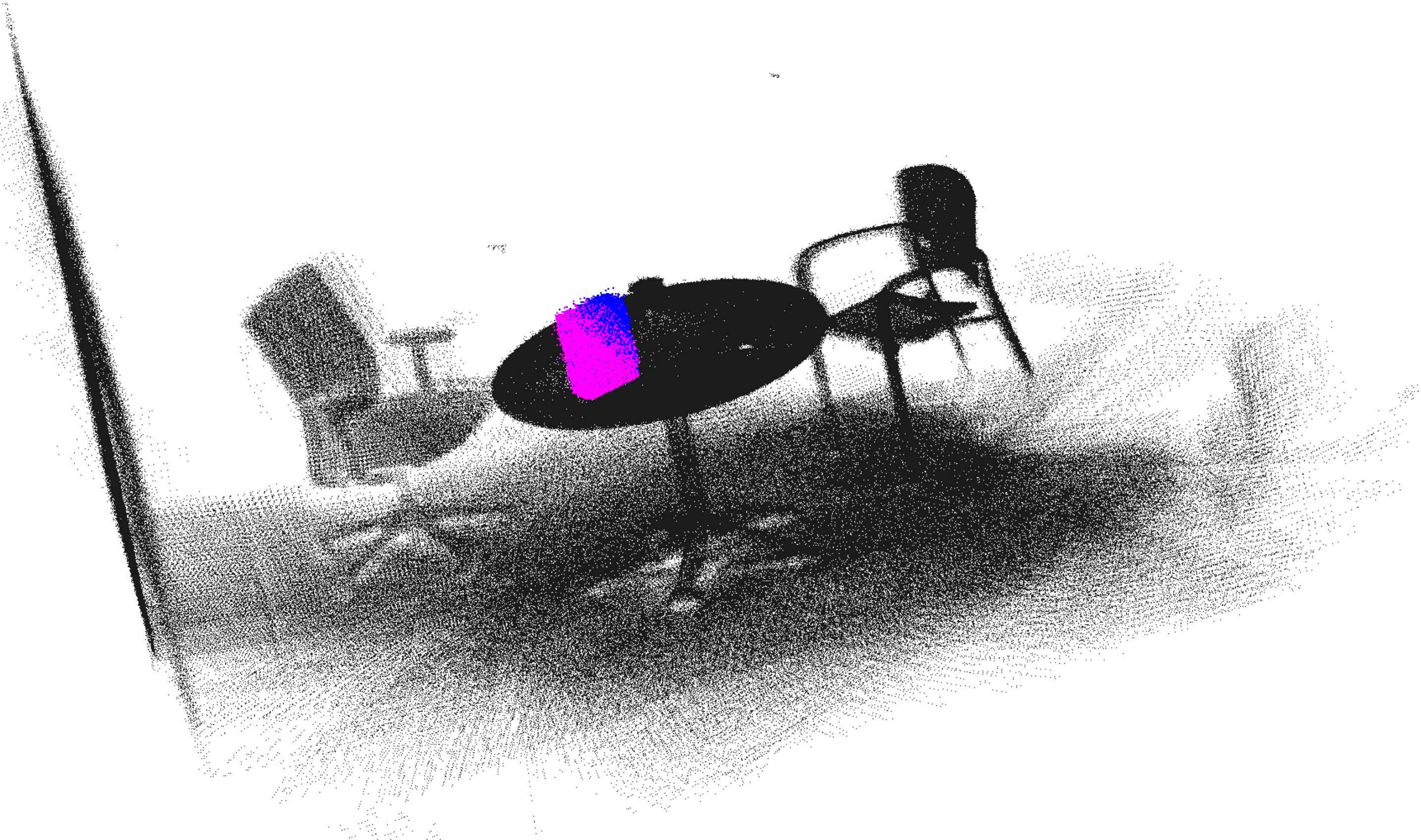}
\end{minipage}

&

\begin{minipage}[t]{0.18\linewidth}
\centering
\includegraphics[width=1\linewidth]{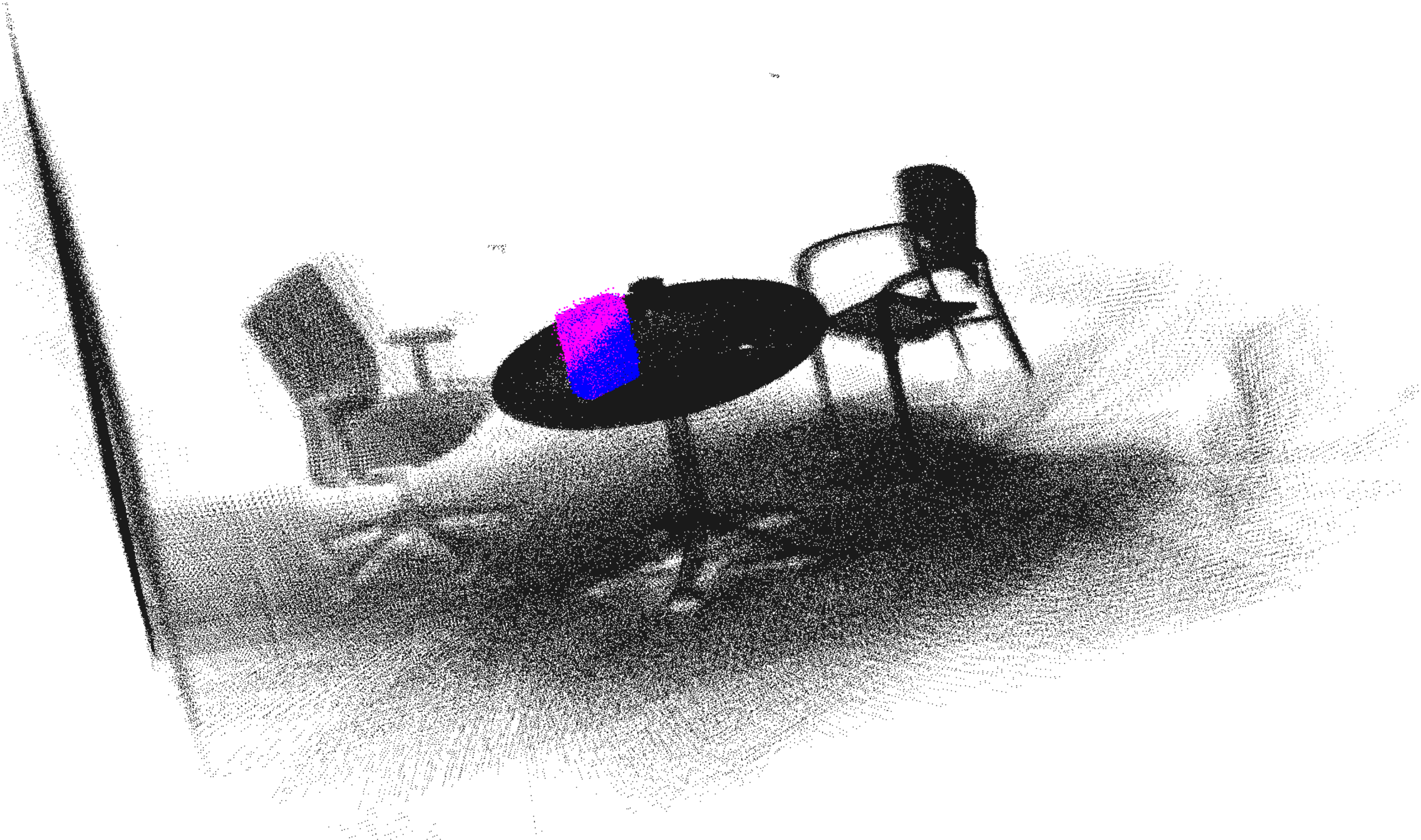}
\end{minipage}

\\

\rotatebox{90}{\,\,\footnotesize{\textit{Scene 09}}\,}\,

& &

\begin{minipage}[t]{0.18\linewidth}
\centering
\includegraphics[width=1\linewidth]{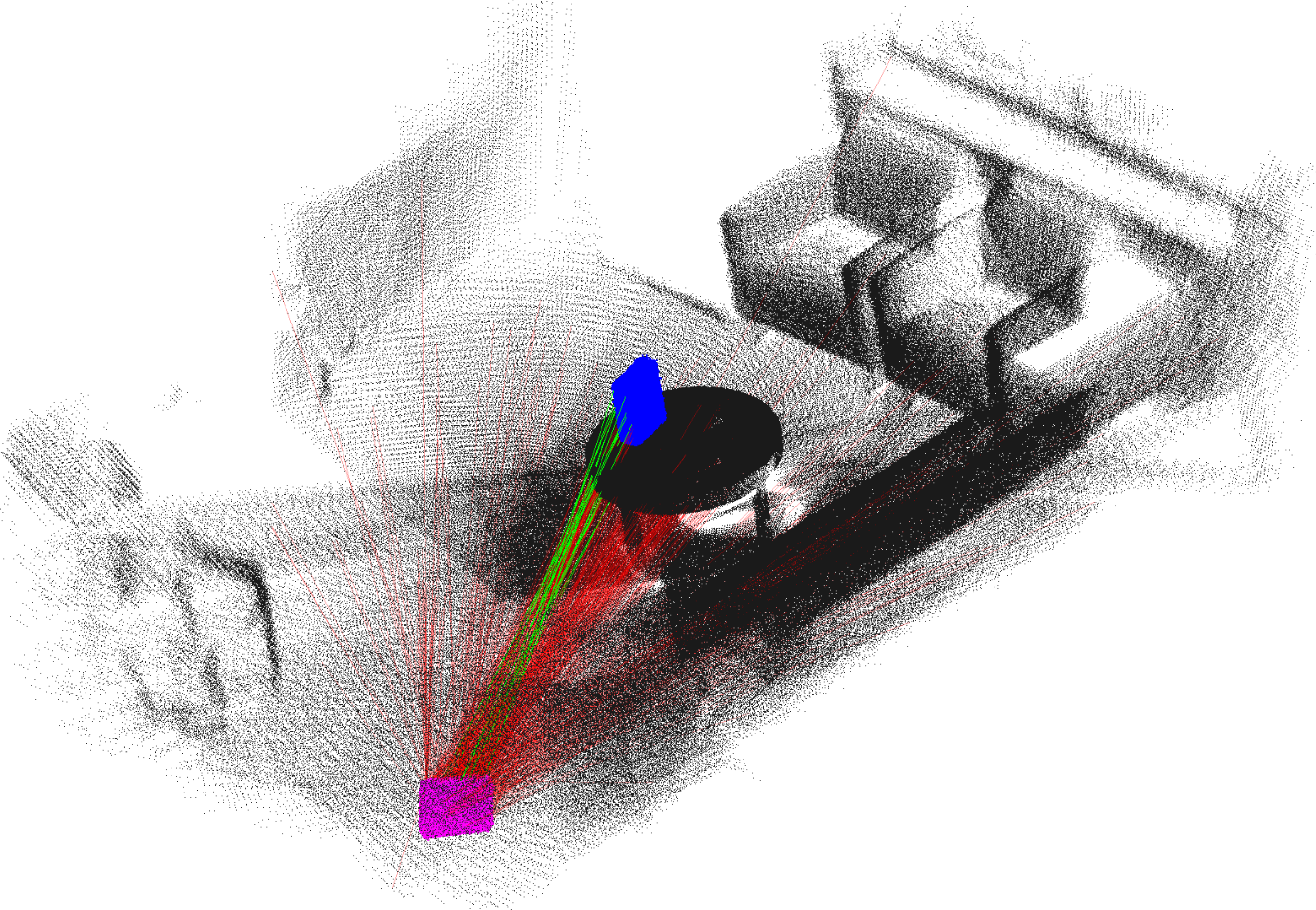}
\end{minipage}

& &

\begin{minipage}[t]{0.18\linewidth}
\centering
\includegraphics[width=1\linewidth]{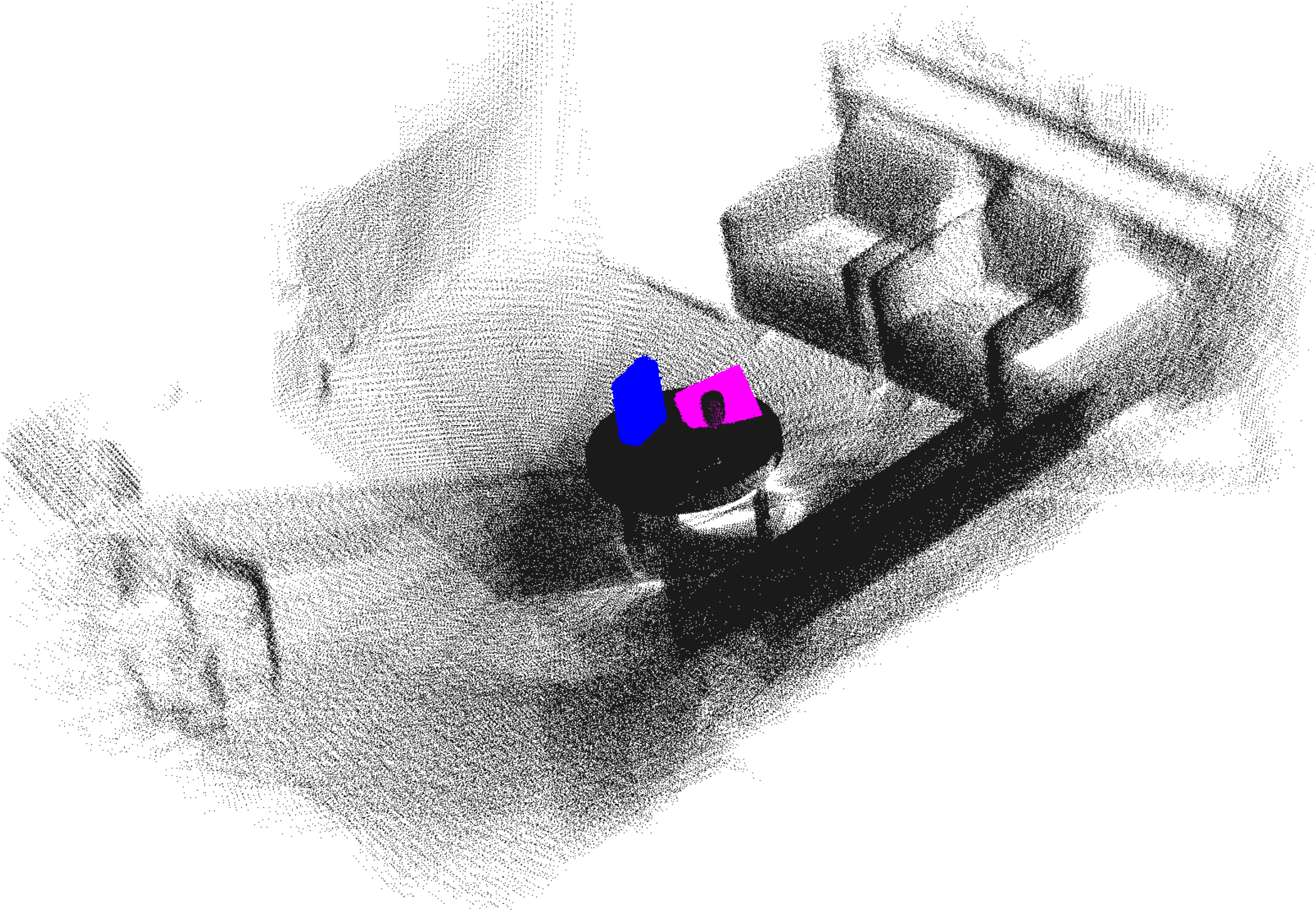}
\end{minipage}

&

\begin{minipage}[t]{0.18\linewidth}
\centering
\includegraphics[width=1\linewidth]{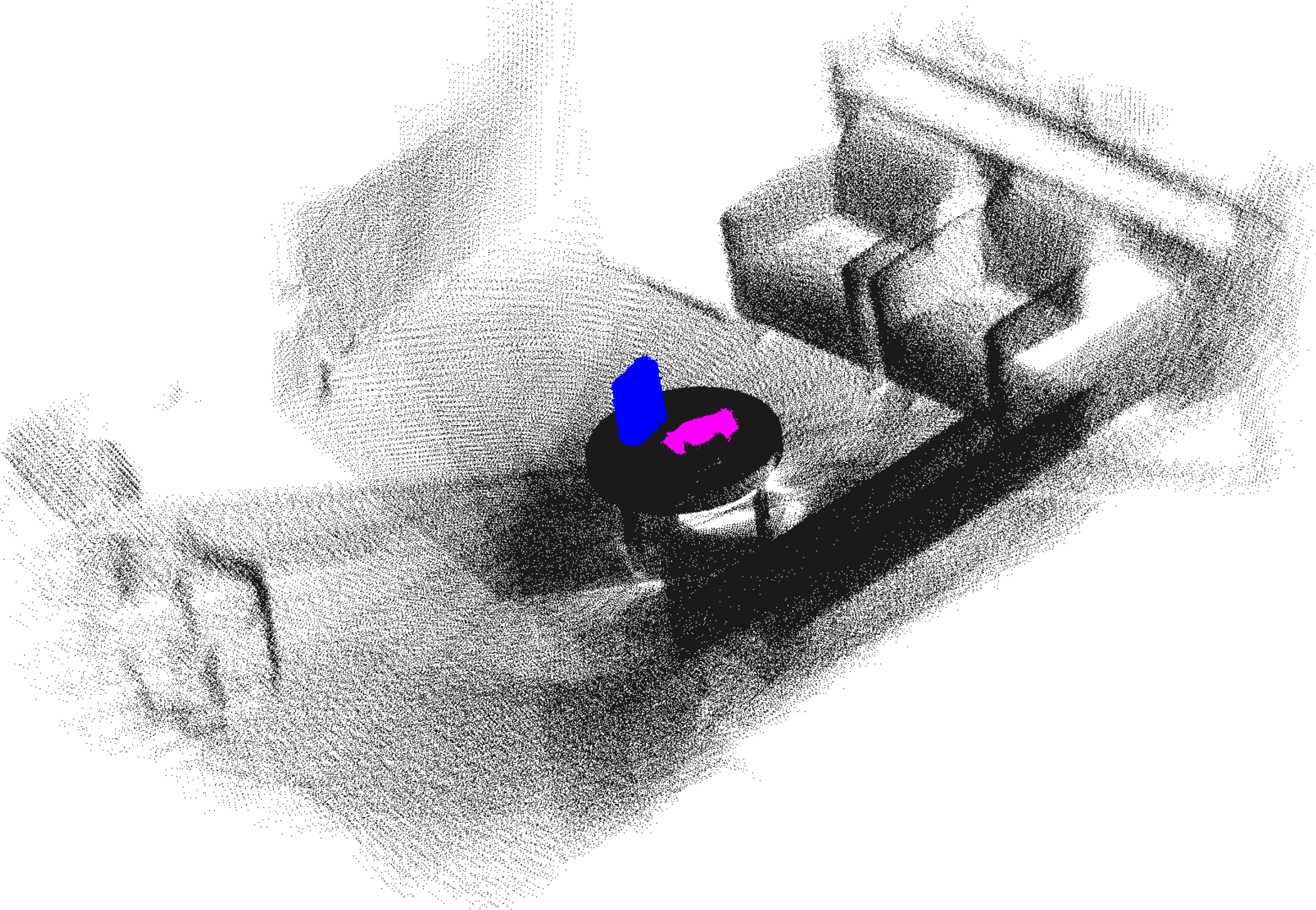}
\end{minipage}

&

\begin{minipage}[t]{0.18\linewidth}
\centering
\includegraphics[width=1\linewidth]{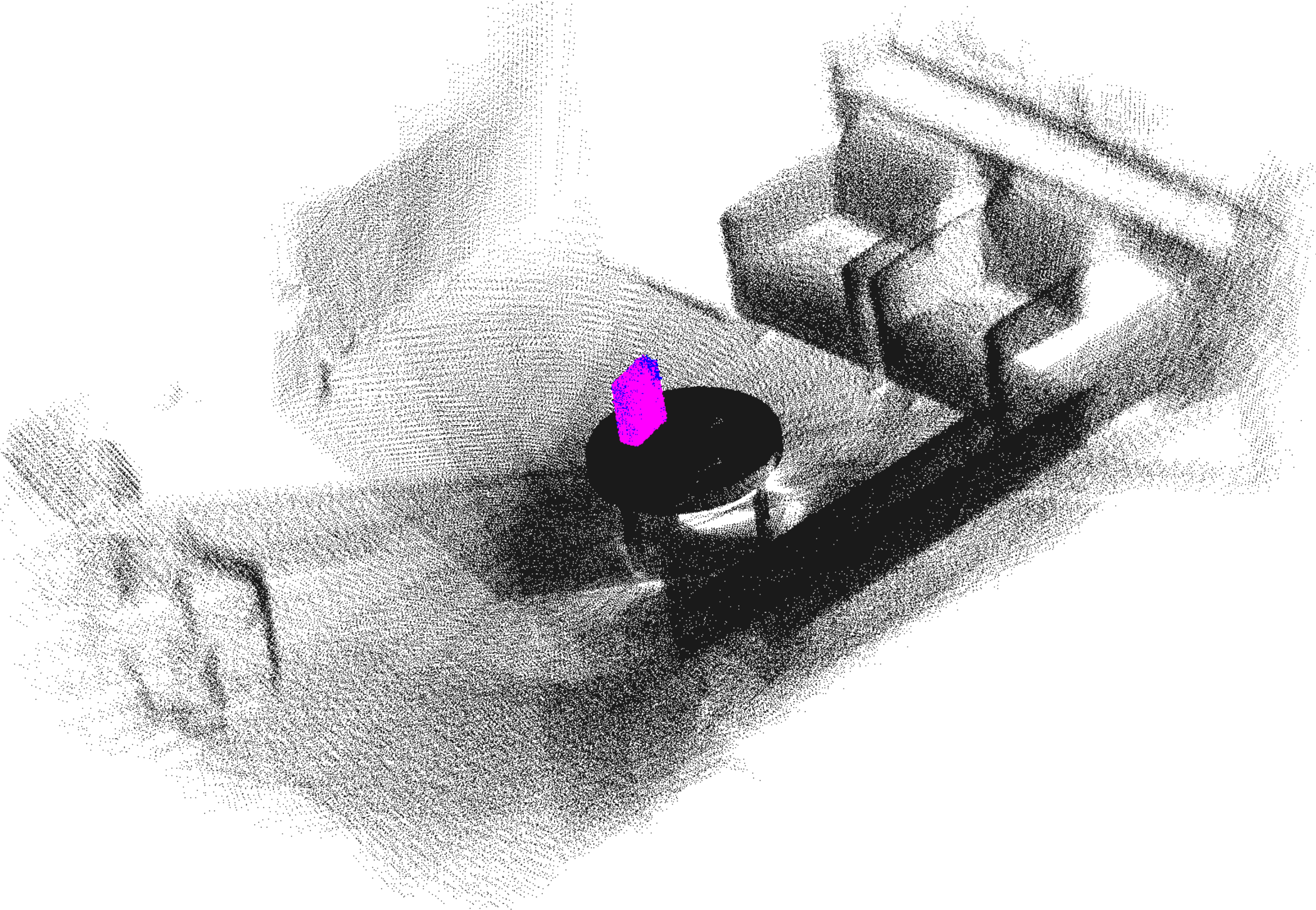}
\end{minipage}

&

\begin{minipage}[t]{0.18\linewidth}
\centering
\includegraphics[width=1\linewidth]{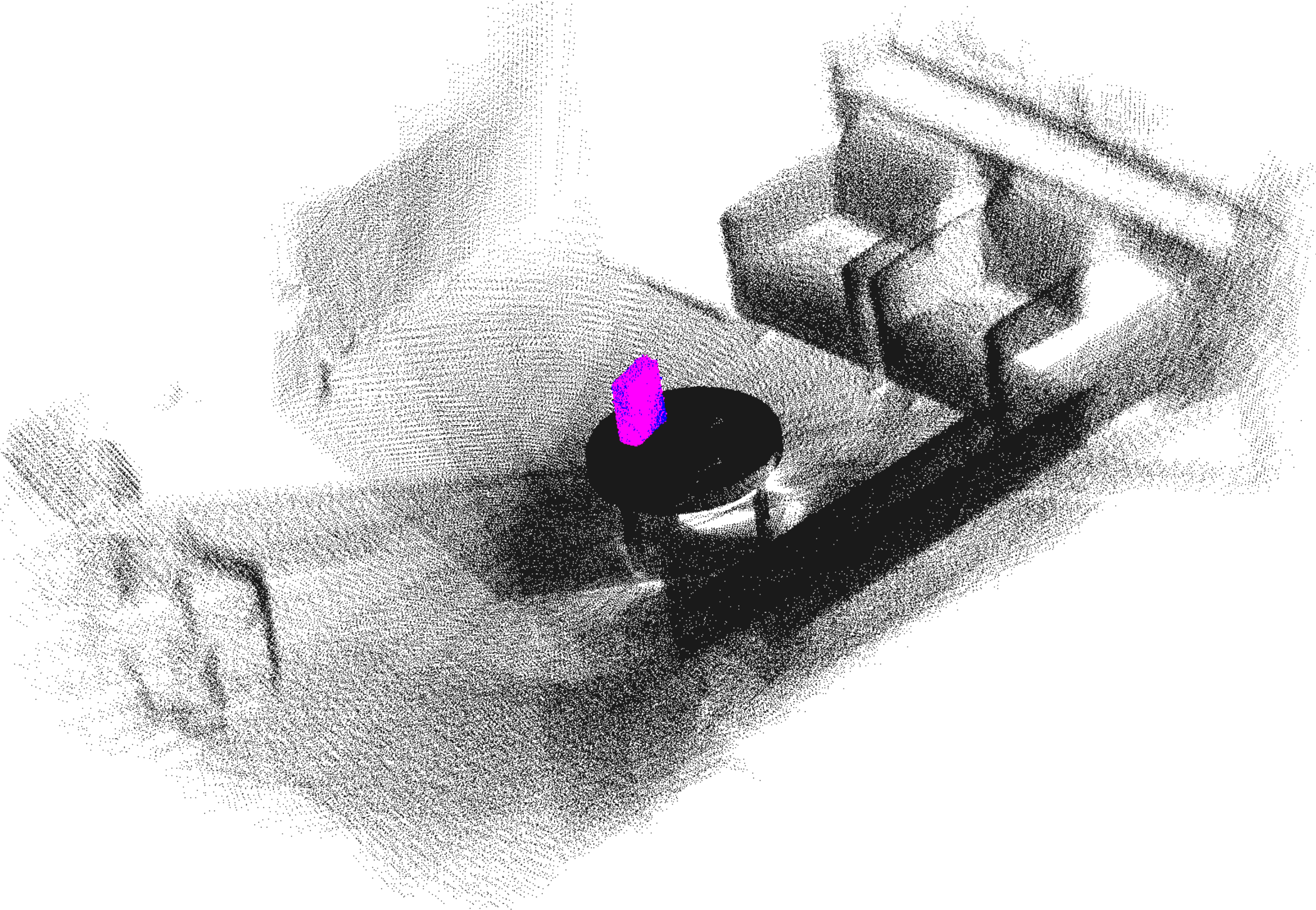}
\end{minipage}

\\

\rotatebox{90}{\,\,\footnotesize{\textit{Scene 12}}\,}\,

& &

\begin{minipage}[t]{0.18\linewidth}
\centering
\includegraphics[width=1\linewidth]{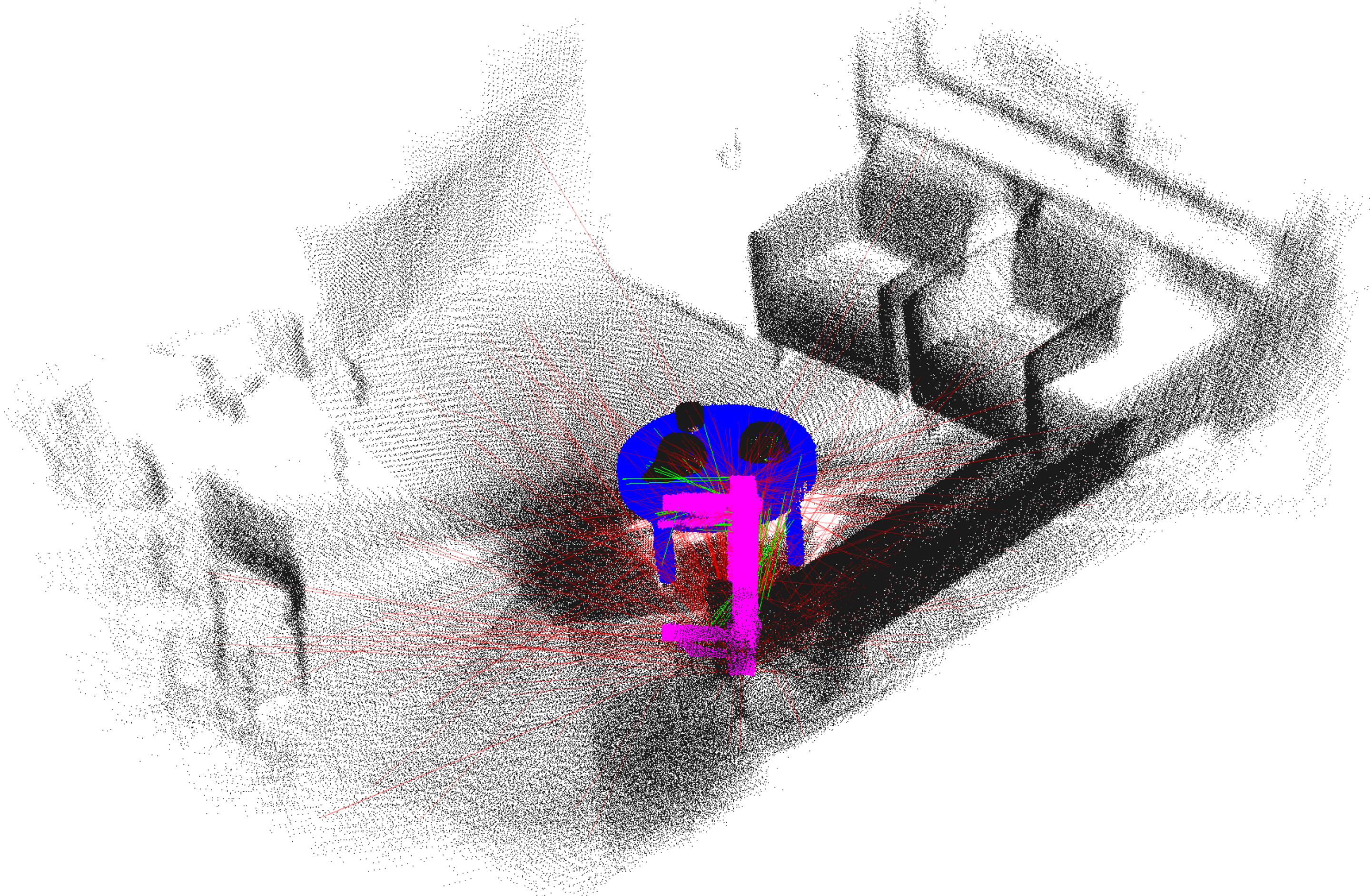}
\end{minipage}

& &

\begin{minipage}[t]{0.18\linewidth}
\centering
\includegraphics[width=1\linewidth]{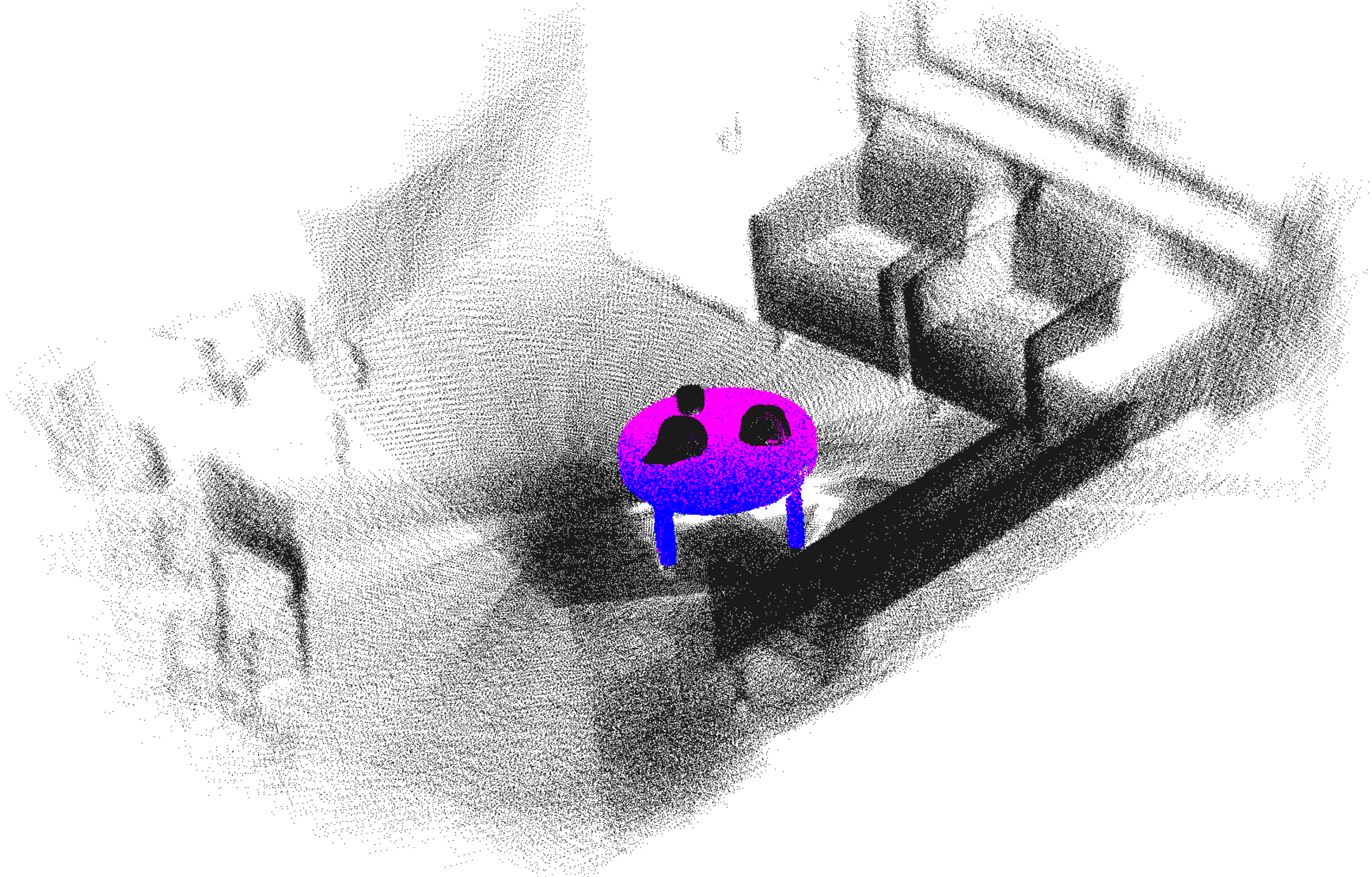}
\end{minipage}

&

\begin{minipage}[t]{0.18\linewidth}
\centering
\includegraphics[width=1\linewidth]{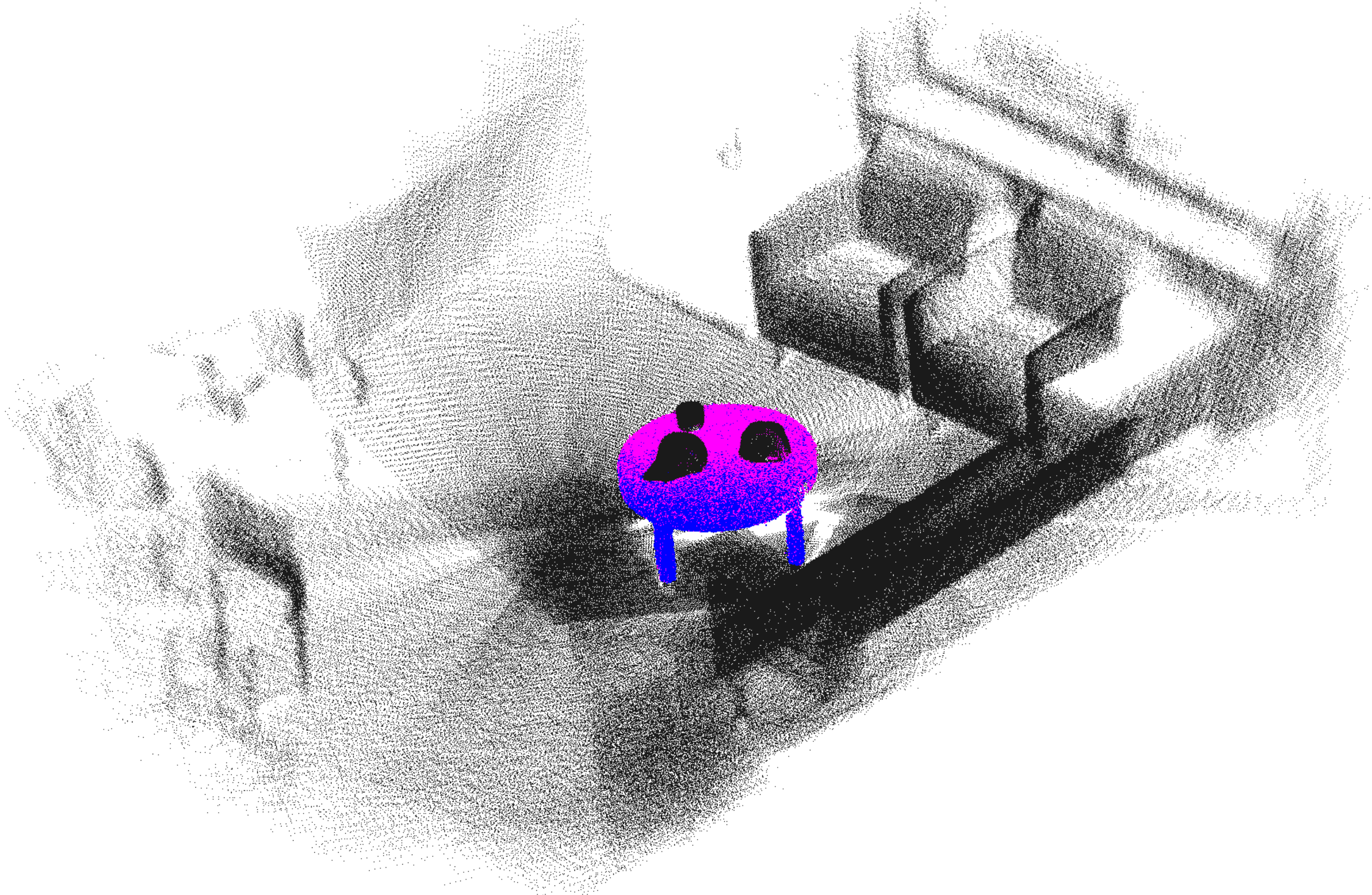}
\end{minipage}

&

\begin{minipage}[t]{0.18\linewidth}
\centering
\includegraphics[width=1\linewidth]{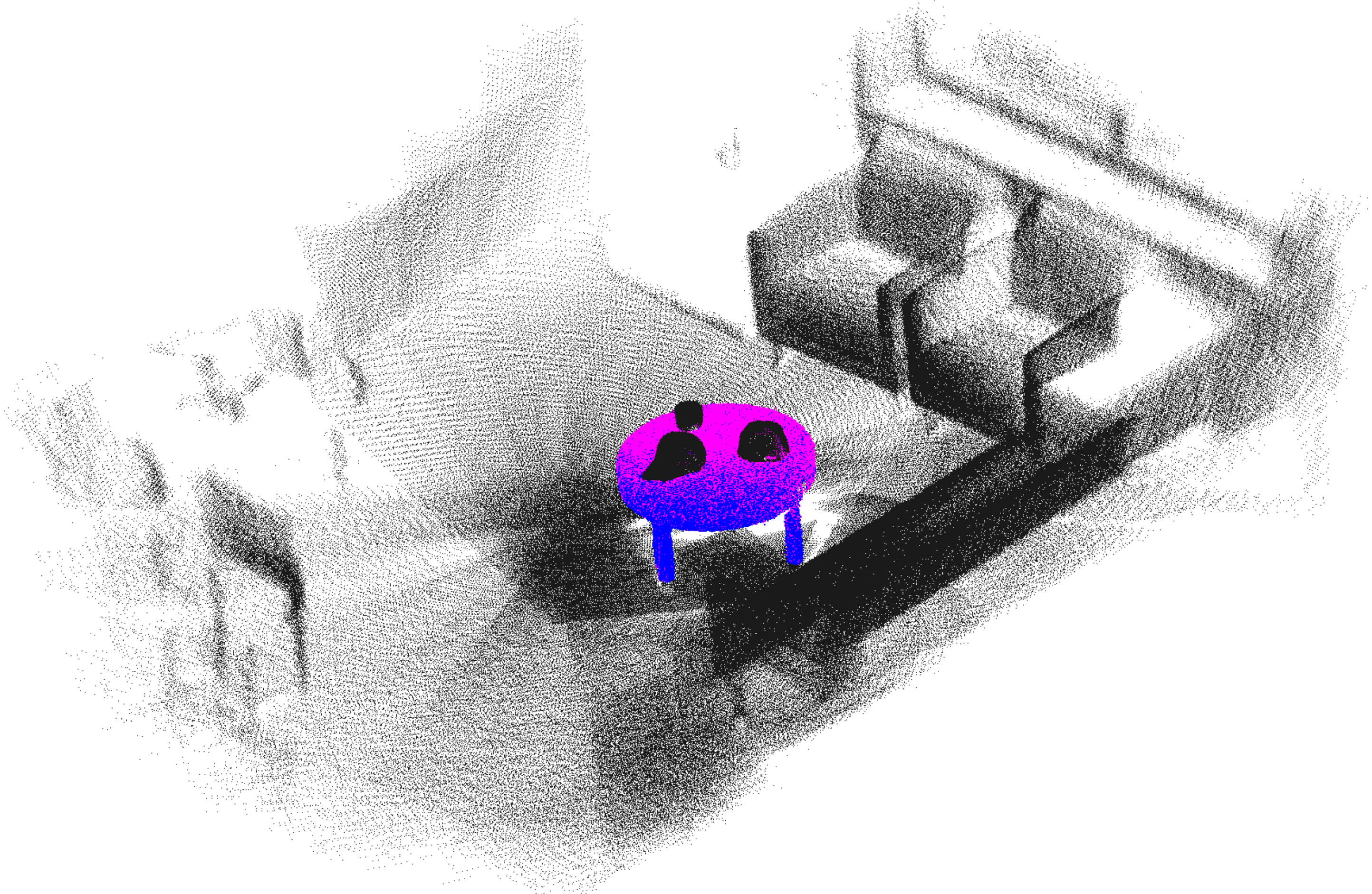}
\end{minipage}

&

\begin{minipage}[t]{0.18\linewidth}
\centering
\includegraphics[width=1\linewidth]{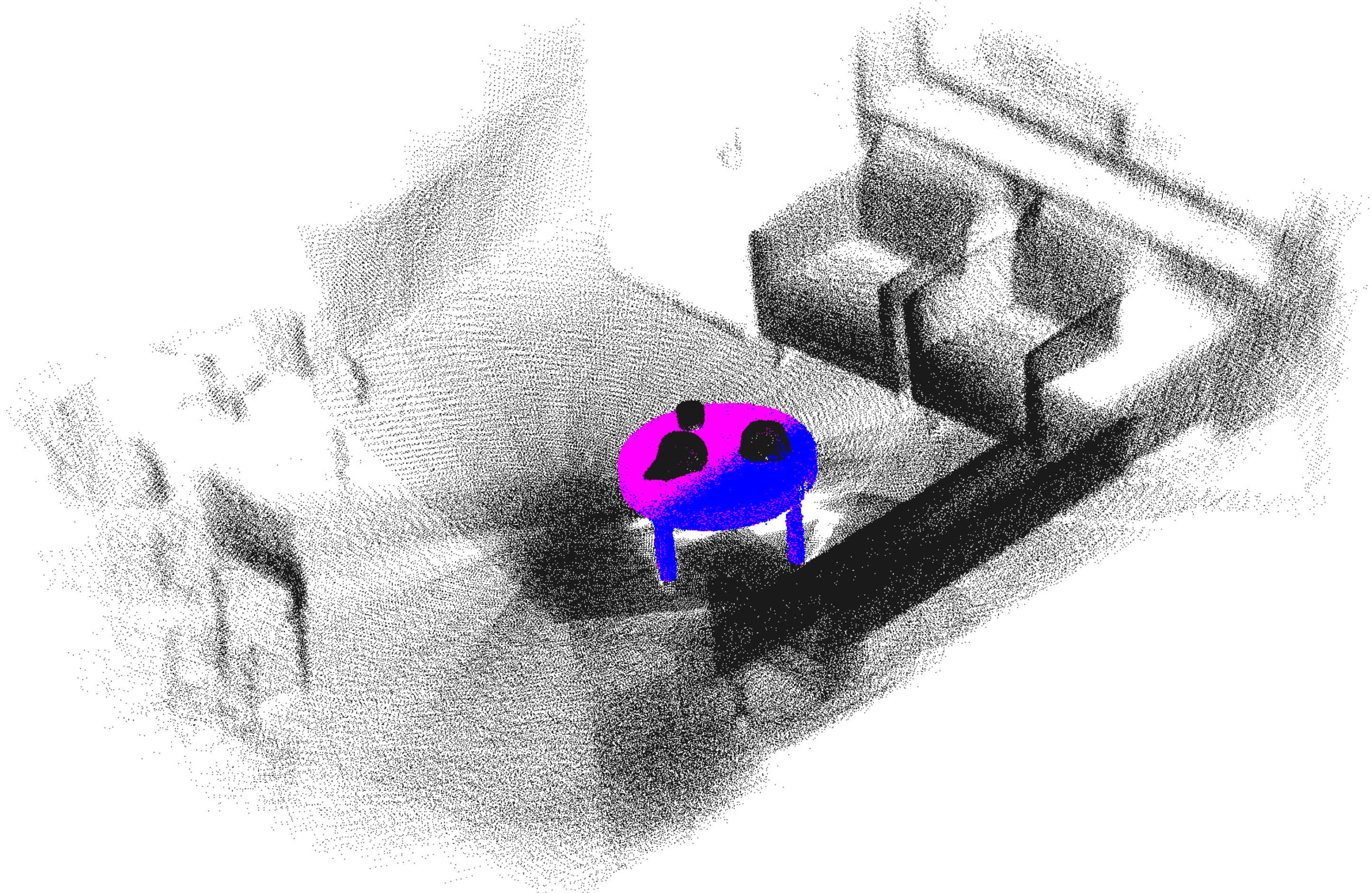}
\end{minipage}

\end{tabular}

\caption{Qualitative 3D object localization results over~\cite{lai2011large} dataset. The left-most column shows the correspondences matched by FPFH~\cite{rusu2009fast} where inliers are in green and outliers are in red. From left to right, we show qualitative reprojection results by FLO-RANSAC, GNC-TLS, GORE+RANSAC and VOCRA, respectively.}
\label{quali-obj-local}
\vspace{-3mm}
\end{figure*}

We show the putative correspondences and the qualitative registration results (reprojecting the object back to the scene with the transformation solved) in Figure~\ref{quali-obj-local} and the supplementary quantitative results (estimation errors and runtime) in Table~\ref{quan-obj-local}. It can be clearly observed that only GORE+RANSAC and VOCRA can render the correct results in all scenes, while VOCRA has the highest accuracy all the time and most often has the best efficiency, which fully reflects the practicality of VOCRA in reality.

\begin{table*}[h]

\centering

\caption{Quantitative 3D object localization results corresponding to Figure~\ref{quali-obj-local}.}
\label{quan-obj-local}
\vspace{-2mm}
\setlength\tabcolsep{2.3pt}
%\addtolength{\tabcolsep}{-0pt}

\begin{tabular}{|c|ccc|ccc|ccc|ccc|ccc|}
\hline\rule{0pt}{6pt}
\quad & \quad & \quad & \quad & \multicolumn{3}{c|}{\scriptsize{FLO-RANSAC}~\cite{lebeda2012fixing}} & \multicolumn{3}{c|}{\scriptsize{GNC-TLS}~\cite{yang2020graduated}} & \multicolumn{3}{c|}{\scriptsize{GORE+RANSAC}~\cite{bustos2017guaranteed}} & \multicolumn{3}{c|}{\scriptsize{VOCRA}} \\ \hline

\scriptsize{Scene} & \scriptsize{Outlier Rate} & \scriptsize{$N$} & \scriptsize{Object} & \scriptsize{$\boldsymbol{R}$.Err [\textit{deg}]} & \scriptsize{$\boldsymbol{t}$.Err [\textit{m}]}  & \scriptsize{Time [\textit{s}]} & \scriptsize{$\boldsymbol{R}$.Err [\textit{deg}]} & \scriptsize{$\boldsymbol{t}$.Err [\textit{m}]}  & \scriptsize{Time[\textit{s}]} &  \scriptsize{$\boldsymbol{R}$.Err [\textit{deg}]} & \scriptsize{$\boldsymbol{t}$.Err [\textit{m}]}  & \scriptsize{Time [\textit{s}]} & \scriptsize{$\boldsymbol{R}$.Err [\textit{deg}]} & \scriptsize{$\boldsymbol{t}$.Err [\textit{m}]}  & \scriptsize{Time [\textit{s}]} \\ \hline

\scriptsize{01} &\scriptsize{96.20\%} & \scriptsize{606} & \scriptsize{cap}  & \scriptsize{174.085} & \scriptsize{0.548} & \scriptsize{16.582} & \scriptsize{170.091} & \scriptsize{2.783} & \scriptsize{0.106}  & \scriptsize{0.462} & \scriptsize{0.011} & \scriptsize{1.117} & \textbf{\scriptsize{0.407}} & \textbf{\scriptsize{0.009}} & \textbf{\scriptsize{0.447}} \\ \hline

\scriptsize{01} &\scriptsize{95.44\%} & \scriptsize{373} & \scriptsize{cereal box}  & \textbf{\scriptsize{0.194}} & \textbf{\scriptsize{0.003}} & \scriptsize{12.133} & \scriptsize{169.641} & \scriptsize{2.223} & \scriptsize{0.053}  & \textbf{\scriptsize{0.194}} & \textbf{\scriptsize{0.003}} & \scriptsize{1.136} & \textbf{\scriptsize{0.194}} & \textbf{\scriptsize{0.003}} & \textbf{\scriptsize{0.203}} \\ \hline

\scriptsize{03} &\scriptsize{95.87\%} & \scriptsize{291} & \scriptsize{cap}  & \textbf{\scriptsize{0.656}} & \textbf{\scriptsize{0.006}} & \scriptsize{8.731} & \scriptsize{150.523} & \scriptsize{1.965} & \scriptsize{0.034} & \textbf{\scriptsize{0.656}} & \textbf{\scriptsize{0.006}} & \scriptsize{0.798} & \textbf{\scriptsize{0.656}} & \textbf{\scriptsize{0.006}} & \textbf{\scriptsize{0.154}} \\ \hline

\scriptsize{05} &\scriptsize{94.78\%} & \scriptsize{536} & \scriptsize{cereal box}  & \textbf{\scriptsize{0.093}} & \textbf{\scriptsize{0.002}} & \scriptsize{21.657} & \scriptsize{92.208} & \scriptsize{2.831} & \scriptsize{0.083} & \textbf{\scriptsize{0.093}} & \textbf{\scriptsize{0.002}} & \scriptsize{1.680} & \textbf{\scriptsize{0.093}} & \textbf{\scriptsize{0.002}} & \textbf{\scriptsize{0.360}} \\ \hline

\scriptsize{09} &\scriptsize{96.60\%} & \scriptsize{666} & \scriptsize{cereal box}  & {\scriptsize{177.450}} & {\scriptsize{2.402}} & \scriptsize{33.906} & \scriptsize{90.146} & \scriptsize{1.229} & \scriptsize{0.077} & \textbf{\scriptsize{0.147}} & \textbf{\scriptsize{0.003}} & \scriptsize{1.358} & \textbf{\scriptsize{0.147}} & \textbf{\scriptsize{0.003}} & \textbf{\scriptsize{0.515}} \\ \hline

\scriptsize{12} &\scriptsize{89.05\%} & \scriptsize{639} & \scriptsize{table}  & \textbf{\scriptsize{0.055}} & \textbf{\scriptsize{0.001}} & \scriptsize{6.169} & \textbf{\scriptsize{0.055}} & \textbf{\scriptsize{0.001}} & \textbf{\scriptsize{0.107}} & \textbf{\scriptsize{0.055}} & \textbf{\scriptsize{0.001}} & \scriptsize{1.049} & \textbf{\scriptsize{0.055}} & \textbf{\scriptsize{0.001}} & {\scriptsize{0.487}} \\ \hline

\end{tabular}
\end{table*}

\section{Conclusion}\label{sec-conclusion}

In this paper, a novel, fast, and robust correspondence-based point cloud registration solver, named VOCRA, is presented. First, we sort the correspondences via voting based on the TB cost function, which proves to be superior to traditional line voting. Then, we perform rapid consensus maximization by using robust single rotation averaging, circumventing the building of consensus set in each iteration. Finally, we use GNC-TB to prune outliers and find the true inlier set efficiently. 

We evaluate our VOCRA in vaired experiments on different datasets, and based on the results, we can conclude that:

(a) VOCRA is able to tolerate extremely high outlier rates (over 99\%);

(b) VOCRA is also significantly faster than other state-of-the-art solvers (e.g. RANSAC, GORE);

(c) VOCRA is applicable to, and also remains its high robustness in, the real-world application problems including scan matching and 3D object localization.

The main limitation of the proposed solver is that the time complexity of the voting process (Section~\ref{sorting-and-voting}) is $O(N^2)$, so with large problem size (correspondence number) $N$, VOCRA should require longer runtime.

We release the Matlab source code of VOCRA at: \url{https://github.com/LeiSun-98/VOCRA}.

{\small
\bibliographystyle{IEEEtran}
\bibliography{egbib}

\begin{thebibliography}{10}
\providecommand{\url}[1]{#1}
\csname url@rmstyle\endcsname
\providecommand{\newblock}{\relax}
\providecommand{\bibinfo}[2]{#2}
\providecommand\BIBentrySTDinterwordspacing{\spaceskip=0pt\relax}
\providecommand\BIBentryALTinterwordstretchfactor{4}
\providecommand\BIBentryALTinterwordspacing{\spaceskip=\fontdimen2\font plus
\BIBentryALTinterwordstretchfactor\fontdimen3\font minus
  \fontdimen4\font\relax}
\providecommand\BIBforeignlanguage[2]{{%
\expandafter\ifx\csname l@#1\endcsname\relax
\typeout{** WARNING: IEEEtran.bst: No hyphenation pattern has been}%
\typeout{** loaded for the language `#1'. Using the pattern for}%
\typeout{** the default language instead.}%
\else
\language=\csname l@#1\endcsname
\fi
#2}}

\bibitem{henry2012rgb}
P.~Henry, M.~Krainin, E.~Herbst, X.~Ren, and D.~Fox, ``Rgb-d mapping: Using
  kinect-style depth cameras for dense 3d modeling of indoor environments,''
  \emph{The International Journal of Robotics Research}, vol.~31, no.~5, pp.
  647--663, 2012.

\bibitem{choi2015robust}
S.~Choi, Q.-Y. Zhou, and V.~Koltun, ``Robust reconstruction of indoor scenes,''
  in \emph{Proceedings of the IEEE Conference on Computer Vision and Pattern
  Recognition}, 2015, pp. 5556--5565.

\bibitem{zhang2015visual}
J.~Zhang and S.~Singh, ``Visual-lidar odometry and mapping: Low-drift, robust,
  and fast,'' in \emph{2015 IEEE International Conference on Robotics and
  Automation (ICRA)}.\hskip 1em plus 0.5em minus 0.4em\relax IEEE, 2015, pp.
  2174--2181.

\bibitem{drost2010model}
B.~Drost, M.~Ulrich, N.~Navab, and S.~Ilic, ``Model globally, match locally:
  Efficient and robust 3d object recognition,'' in \emph{2010 IEEE computer
  society conference on computer vision and pattern recognition}.\hskip 1em
  plus 0.5em minus 0.4em\relax Ieee, 2010, pp. 998--1005.

\bibitem{zeng2017multi}
A.~Zeng, K.-T. Yu, S.~Song, D.~Suo, E.~Walker, A.~Rodriguez, and J.~Xiao,
  ``Multi-view self-supervised deep learning for 6d pose estimation in the
  amazon picking challenge,'' in \emph{2017 IEEE international conference on
  robotics and automation (ICRA)}.\hskip 1em plus 0.5em minus 0.4em\relax IEEE,
  2017, pp. 1386--1383.

\bibitem{zhang2014loam}
J.~Zhang and S.~Singh, ``Loam: Lidar odometry and mapping in real-time.'' in
  \emph{Robotics: Science and Systems}, vol.~2, no.~9, 2014.

\bibitem{audette2000algorithmic}
M.~A. Audette, F.~P. Ferrie, and T.~M. Peters, ``An algorithmic overview of
  surface registration techniques for medical imaging,'' \emph{Medical image
  analysis}, vol.~4, no.~3, pp. 201--217, 2000.

\bibitem{besl1992method}
P.~Besl and N.~D. McKay, ``A method for registration of 3-d shapes,''
  \emph{IEEE Transactions on Pattern Analysis and Machine Intelligence},
  vol.~14, no.~2, pp. 239--256, 1992.

\bibitem{rusu2009fast}
R.~B. Rusu, N.~Blodow, and M.~Beetz, ``Fast point feature histograms (fpfh) for
  3d registration,'' in \emph{2009 IEEE international conference on robotics
  and automation}.\hskip 1em plus 0.5em minus 0.4em\relax IEEE, 2009, pp.
  3212--3217.

\bibitem{zhong2009intrinsic}
Y.~Zhong, ``Intrinsic shape signatures: A shape descriptor for 3d object
  recognition,'' in \emph{2009 IEEE 12th International Conference on Computer
  Vision Workshops, ICCV Workshops}.\hskip 1em plus 0.5em minus 0.4em\relax
  IEEE, 2009, pp. 689--696.

\bibitem{lowe2004distinctive}
D.~G. Lowe, ``Distinctive image features from scale-invariant keypoints,''
  \emph{International journal of computer vision}, vol.~60, no.~2, pp. 91--110,
  2004.

\bibitem{bay2006surf}
H.~Bay, T.~Tuytelaars, and L.~Van~Gool, ``Surf: Speeded up robust features,''
  in \emph{European conference on computer vision}.\hskip 1em plus 0.5em minus
  0.4em\relax Springer, 2006, pp. 404--417.

\bibitem{bustos2017guaranteed}
A.~P. Bustos and T.-J. Chin, ``Guaranteed outlier removal for point cloud
  registration with correspondences,'' \emph{IEEE transactions on pattern
  analysis and machine intelligence}, vol.~40, no.~12, pp. 2868--2882, 2017.

\bibitem{fischler1981random}
M.~A. Fischler and R.~C. Bolles, ``Random sample consensus: a paradigm for
  model fitting with applications to image analysis and automated
  cartography,'' \emph{Communications of the ACM}, vol.~24, no.~6, pp.
  381--395, 1981.

\bibitem{chum2003locally}
O.~Chum, J.~Matas, and J.~Kittler, ``Locally optimized ransac,'' in \emph{Joint
  Pattern Recognition Symposium}.\hskip 1em plus 0.5em minus 0.4em\relax
  Springer, 2003, pp. 236--243.

\bibitem{parra2014fast}
A.~Parra~Bustos, T.-J. Chin, and D.~Suter, ``Fast rotation search with
  stereographic projections for 3d registration,'' in \emph{Proceedings of the
  IEEE conference on computer vision and pattern recognition}, 2014, pp.
  3930--3937.

\bibitem{yang2020graduated}
H.~Yang, P.~Antonante, V.~Tzoumas, and L.~Carlone, ``Graduated non-convexity
  for robust spatial perception: From non-minimal solvers to global outlier
  rejection,'' \emph{IEEE Robotics and Automation Letters}, vol.~5, no.~2, pp.
  1127--1134, 2020.

\bibitem{zhou2016fast}
Q.-Y. Zhou, J.~Park, and V.~Koltun, ``Fast global registration,'' in
  \emph{European Conference on Computer Vision}.\hskip 1em plus 0.5em minus
  0.4em\relax Springer, 2016, pp. 766--782.

\bibitem{yang2019polynomial}
H.~Yang and L.~Carlone, ``A polynomial-time solution for robust registration
  with extreme outlier rates,'' in \emph{Robotics: Science and Systems}, 2019.

\bibitem{yang2020teaser}
H.~Yang, J.~Shi, and L.~Carlone, ``Teaser: Fast and certifiable point cloud
  registration,'' \emph{IEEE Transactions on Robotics}, 2020.

\bibitem{horst2013global}
R.~Horst and H.~Tuy, \emph{Global optimization: Deterministic
  approaches}.\hskip 1em plus 0.5em minus 0.4em\relax Springer Science \&
  Business Media, 2013.

\bibitem{li2021practical}
J.~Li, ``A practical o (n2) outlier removal method for point cloud
  registration,'' \emph{IEEE Transactions on Pattern Analysis and Machine
  Intelligence}, 2021.

\bibitem{lee2020robust}
S.~H. Lee and J.~Civera, ``Robust single rotation averaging,'' \emph{arXiv
  preprint arXiv:2004.00732}, 2020.

\bibitem{lebeda2012fixing}
K.~Lebeda, J.~Matas, and O.~Chum, ``Fixing the locally optimized ransac--full
  experimental evaluation,'' in \emph{British machine vision conference},
  vol.~2.\hskip 1em plus 0.5em minus 0.4em\relax Citeseer, 2012.

\bibitem{chum2005matching}
O.~Chum and J.~Matas, ``Matching with prosac-progressive sample consensus,'' in
  \emph{2005 IEEE computer society conference on computer vision and pattern
  recognition (CVPR'05)}, vol.~1.\hskip 1em plus 0.5em minus 0.4em\relax IEEE,
  2005, pp. 220--226.

\bibitem{agarwal2013robust}
P.~Agarwal, G.~D. Tipaldi, L.~Spinello, C.~Stachniss, and W.~Burgard, ``Robust
  map optimization using dynamic covariance scaling,'' in \emph{2013 IEEE
  International Conference on Robotics and Automation}.\hskip 1em plus 0.5em
  minus 0.4em\relax Ieee, 2013, pp. 62--69.

\bibitem{kummerle2011g}
R.~K{\"u}mmerle, G.~Grisetti, H.~Strasdat, K.~Konolige, and W.~Burgard, ``g 2
  o: A general framework for graph optimization,'' in \emph{2011 IEEE
  International Conference on Robotics and Automation}.\hskip 1em plus 0.5em
  minus 0.4em\relax IEEE, 2011, pp. 3607--3613.

\bibitem{sunderhauf2012towards}
N.~S{\"u}nderhauf and P.~Protzel, ``Towards a robust back-end for pose graph
  slam,'' in \emph{2012 IEEE international conference on robotics and
  automation}.\hskip 1em plus 0.5em minus 0.4em\relax IEEE, 2012, pp.
  1254--1261.

\bibitem{black1996unification}
M.~J. Black and A.~Rangarajan, ``On the unification of line processes, outlier
  rejection, and robust statistics with applications in early vision,''
  \emph{International journal of computer vision}, vol.~19, no.~1, pp. 57--91,
  1996.

\bibitem{hartley2011l1}
R.~Hartley, K.~Aftab, and J.~Trumpf, ``L1 rotation averaging using the
  weiszfeld algorithm,'' in \emph{CVPR 2011}.\hskip 1em plus 0.5em minus
  0.4em\relax IEEE, 2011, pp. 3041--3048.

\bibitem{lam2007precision}
Q.~M. Lam and J.~L. Crassidis, ``Precision attitude determination using a
  multiple model adaptive estimation scheme,'' in \emph{2007 IEEE Aerospace
  Conference}.\hskip 1em plus 0.5em minus 0.4em\relax IEEE, 2007, pp. 1--20.

\bibitem{hartley2013rotation}
R.~Hartley, J.~Trumpf, Y.~Dai, and H.~Li, ``Rotation averaging,''
  \emph{International journal of computer vision}, vol. 103, no.~3, pp.
  267--305, 2013.

\bibitem{quan2020compatibility}
S.~Quan and J.~Yang, ``Compatibility-guided sampling consensus for 3-d point
  cloud registration,'' \emph{IEEE Transactions on Geoscience and Remote
  Sensing}, vol.~58, no.~10, pp. 7380--7392, 2020.

\bibitem{zach2015dynamic}
C.~Zach, A.~Penate-Sanchez, and M.-T. Pham, ``A dynamic programming approach
  for fast and robust object pose recognition from range images,'' in
  \emph{Proceedings of the IEEE conference on computer vision and pattern
  recognition}, 2015, pp. 196--203.

\bibitem{michel2017global}
F.~Michel, A.~Kirillov, E.~Brachmann, A.~Krull, S.~Gumhold, B.~Savchynskyy, and
  C.~Rother, ``Global hypothesis generation for 6d object pose estimation,'' in
  \emph{Proceedings of the IEEE Conference on Computer Vision and Pattern
  Recognition}, 2017, pp. 462--471.

\bibitem{shi2021robin}
J.~Shi, H.~Yang, and L.~Carlone, ``Robin: a graph-theoretic approach to reject
  outliers in robust estimation using invariants,'' in \emph{2021 IEEE
  International Conference on Robotics and Automation (ICRA)}.\hskip 1em plus
  0.5em minus 0.4em\relax IEEE, 2021, pp. 13\,820--13\,827.

\bibitem{barron2019general}
J.~T. Barron, ``A general and adaptive robust loss function,'' in
  \emph{Proceedings of the IEEE/CVF Conference on Computer Vision and Pattern
  Recognition}, 2019, pp. 4331--4339.

\bibitem{horn1987closed}
B.~K. Horn, ``Closed-form solution of absolute orientation using unit
  quaternions,'' \emph{Josa a}, vol.~4, no.~4, pp. 629--642, 1987.

\bibitem{curless1996volumetric}
B.~Curless and M.~Levoy, ``A volumetric method for building complex models from
  range images,'' in \emph{Proceedings of the 23rd annual conference on
  Computer graphics and interactive techniques}, 1996, pp. 303--312.

\bibitem{tzoumas2019outlier}
V.~Tzoumas, P.~Antonante, and L.~Carlone, ``Outlier-robust spatial perception:
  Hardness, general-purpose algorithms, and guarantees,'' in \emph{2019
  IEEE/RSJ International Conference on Intelligent Robots and Systems
  (IROS)}.\hskip 1em plus 0.5em minus 0.4em\relax IEEE, 2019, pp. 5383--5390.

\bibitem{arun1987least}
K.~S. Arun, T.~S. Huang, and S.~D. Blostein, ``Least-squares fitting of two 3-d
  point sets,'' \emph{IEEE Transactions on pattern analysis and machine
  intelligence}, no.~5, pp. 698--700, 1987.

\bibitem{shotton2013scene}
J.~Shotton, B.~Glocker, C.~Zach, S.~Izadi, A.~Criminisi, and A.~Fitzgibbon,
  ``Scene coordinate regression forests for camera relocalization in rgb-d
  images,'' in \emph{Proceedings of the IEEE Conference on Computer Vision and
  Pattern Recognition}, 2013, pp. 2930--2937.

\bibitem{mian2006three}
\BIBentryALTinterwordspacing
A.~S. Mian, M.~Bennamoun, and R.~Owens, ``Three-dimensional model-based object
  recognition and segmentation in cluttered scenes,'' \emph{IEEE transactions
  on pattern analysis and machine intelligence}, vol.~28, no.~10, pp.
  1584--1601, 2006. [Online]. Available:
  \url{http://vision.deis.unibo.it/keypoints3d/?page_id=2}
\BIBentrySTDinterwordspacing

\bibitem{mian2010repeatability}
A.~Mian, M.~Bennamoun, and R.~Owens, ``On the repeatability and quality of
  keypoints for local feature-based 3d object retrieval from cluttered
  scenes,'' \emph{International Journal of Computer Vision}, vol.~89, no.~2,
  pp. 348--361, 2010.

\bibitem{zeisl2013automatic}
B.~Zeisl, K.~Koser, and M.~Pollefeys, ``Automatic registration of rgb-d scans
  via salient directions,'' in \emph{Proceedings of the IEEE international
  conference on computer vision}, 2013, pp. 2808--2815.

\bibitem{lai2011large}
K.~Lai, L.~Bo, X.~Ren, and D.~Fox, ``A large-scale hierarchical multi-view
  rgb-d object dataset,'' in \emph{2011 IEEE international conference on
  robotics and automation}.\hskip 1em plus 0.5em minus 0.4em\relax IEEE, 2011,
  pp. 1817--1824.

\end{thebibliography}
}

\end{document}